\documentclass[lettersize,journal]{IEEEtran}

\usepackage{amsmath,amssymb,amsfonts,bm,setspace}
\usepackage{graphicx}
\usepackage{lipsum}
\usepackage[ruled,lined,commentsnumbered]{algorithm2e}
\usepackage{bbm}
\usepackage{tikz}
\usetikzlibrary{shapes.geometric, arrows}
\usepackage{relsize}
\usepackage{comment}
\usepackage{tabularx,ragged2e,booktabs}
\usepackage{lipsum}
\usepackage{subfiles} 
\usepackage{tikz}
\usetikzlibrary{shapes.geometric, arrows}
\usepackage{relsize}
\usepackage{array}
\usepackage{accents}
\usepackage{mathtools}
\usepackage{comment}
\usepackage{xcolor}
\usepackage{mathdots}
\usepackage{cancel}
\usepackage{amsthm}
\usepackage{amsmath}
\usepackage{stackengine}
\usepackage{lipsum, babel}
\usepackage{scalefnt}
\usepackage{caption}
  \usepackage[utf8]{inputenc}

\newcolumntype{P}[1]{>{\centering\arraybackslash}p{#1}}
\newcolumntype{M}[1]{>{\centering\arraybackslash}m{#1}}
\newcommand\myydots{\hbox to 0.7em{.\hss.\hss.}}
\newcommand*\underdot[1]{%
  \underaccent{\hspace{1.7mm}\rotatebox[origin=c]{90}{\dots}}{#1}}
\newtheorem{theorem}{Theorem}
\theoremstyle{definition}
\newtheorem{definition}{Definition}[section]
\newtheorem{corollary}{Corollary}[theorem]

\newcommand{\dimN}{N}  
\newcommand{\dimM}{M} 
\newcommand{\indexm}{m} 
\newcommand{\indexn}{n} 
\newcommand{\indexr}{r} 
\newcommand{\indexi}{i} 
\newcommand{\indexj}{j} 
\newcommand{\indexk}{k} 
\newcommand{\blocknum}{K} 
\newcommand{\TypeIIoutlier}{\mathrm{\tiny{II}}} 
\newcommand{\TypeIoutlier}{\mathrm{\tiny{I}}}
\DeclareMathSymbol{\shortminus}{\mathbin}{AMSa}{"39}
\usepackage{array}
\usepackage[caption=false,font=normalsize,labelfont=sf,textfont=sf]{subfig}
\usepackage{textcomp}
\usepackage{stfloats}
\usepackage{url}
\usepackage{verbatim}
\usepackage{graphicx}
\usepackage{cite}
\DeclareMathAlphabet{\mathcalligra}{T1}{calligra}{m}{n}
\DeclareMathAlphabet\mathbfcal{OMS}{cmsy}{b}{n}

\begin{document}

\title{Fast and Robust Sparsity-Aware Block Diagonal Representation}

\author{Aylin~Ta{\c{s}}tan$^\ast$, Michael~Muma$^\dagger$,
        and~Abdelhak~M.~Zoubir$^\ddagger$
\thanks{\noindent$^\ast$The author was with the Signal Processing Group, Technische Universität
Darmstadt, Darmstadt, Germany and is now with the Pattern Recognition Group, University of Bern, Bern, Switzerland (e-mail:  a.tastan@spg.tu-darmstadt.de; aylin.tastan@unibe.ch).\protect\\
$^\dagger$The author is with the Robust Data Science Group, Technische Universität
Darmstadt, Darmstadt, Germany (e-mail: michael.muma@tu-darmstadt.de).\protect\\
$^\ddagger$The author is with the Signal Processing Group, Technische Universität
Darmstadt, Darmstadt, Germany (e-mail:  zoubir@spg.tu-darmstadt.de).
}}

\markboth{SUBMITTED TO IEEE TRANSACTIONS ON SIGNAL PROCESSING}%
{Shell \MakeLowercase{\textit{et al.}}: A Sample Article Using IEEEtran.cls for IEEE Journals}

\IEEEpubid{}

\maketitle

\begin{abstract}
The block diagonal structure of an affinity matrix is a commonly desired property in cluster analysis because it represents clusters of feature vectors by non-zero coefficients that are concentrated in blocks. However, recovering a block diagonal affinity matrix is challenging in real-world applications, in which the data may be subject to outliers and heavy-tailed noise that obscure the hidden cluster structure. To address this issue, we first analyze the effect of different fundamental outlier types in graph-based cluster analysis.  A key idea that simplifies the analysis is to introduce a  vector that represents a block diagonal matrix as a piece-wise linear function of the similarity coefficients that form the affinity matrix. We reformulate the problem as a robust piece-wise linear fitting problem and propose a Fast and Robust Sparsity-Aware Block Diagonal Representation {(FRS-BDR)} method, which jointly estimates cluster memberships and the number of blocks. Comprehensive experiments on a variety of real-world applications demonstrate the effectiveness of FRS-BDR in terms of clustering accuracy, robustness against corrupted features, computation time and cluster enumeration performance.
\end{abstract}


\vspace{-2mm}
\section{Introduction}
\label{sec:introduction}
 \IEEEPARstart{A}{block} diagonally structured affinity matrix represents clusters of feature vectors by non-zero coefficients that are concentrated in blocks. Such a structure is an informative model to describe hidden relationships. It has numerous applications, e.g., denoising \citeform[1]-\citeform[2], recognition \cite{BDRrecognition}, semi-supervised learning \citeform[4]-\citeform[6], subspace learning  and clustering/classification \citeform[6]-\citeform[16].  

Commonly used existing block diagonal representation (BDR) methods impose a structure on the affinity matrix using regularization with block diagonal (BD) priors, e.g. based on a low-rank property \citeform[17]-\citeform[20], sparsity \citeform[21]-\citeform[23] or a known number of blocks $\blocknum$ \citeform[6]-\citeform[9]. For example the method in \cite{BDSSC},  which is one of the current benchmarks BDR methods, controls the number of connected components in the affinity matrix by imposing a rank constraint on the Laplacian matrix. An alternative popular approach \cite{BDRB}, proposes a $\blocknum$-block regularizer that is defined by the sum of the $\blocknum$ smallest eigenvalues of the Laplacian matrix to compute a BD affinity matrix. A major challenge of these methods is the need to determine appropriate BD priors which play a crucial role in achieving accurate BDR results. Due to its key role in BDR methods, the determination of sparsity/low-rank level has been intensively investigated from different viewpoints, e.g. similarity coefficients' distribution \cite{Sparcode}, connectedness \cite{Connectedness2}, geometric analysis \cite{arora2009expander} and supervised learning \citeform[27]-\citeform[28]. Recently, in \cite{EBDR}, an alternative unsupervised approach based on eigenvalues has been proposed to deduce the sparsity level in a BD matrix. The eigenvalue analysis is, however, restricted to the setting of independent blocks. 

\begin{figure}[tbp!]
  \centering
\includegraphics[trim={1.8cm 1.9cm 1cm 1.2cm},clip,width=8.3cm]{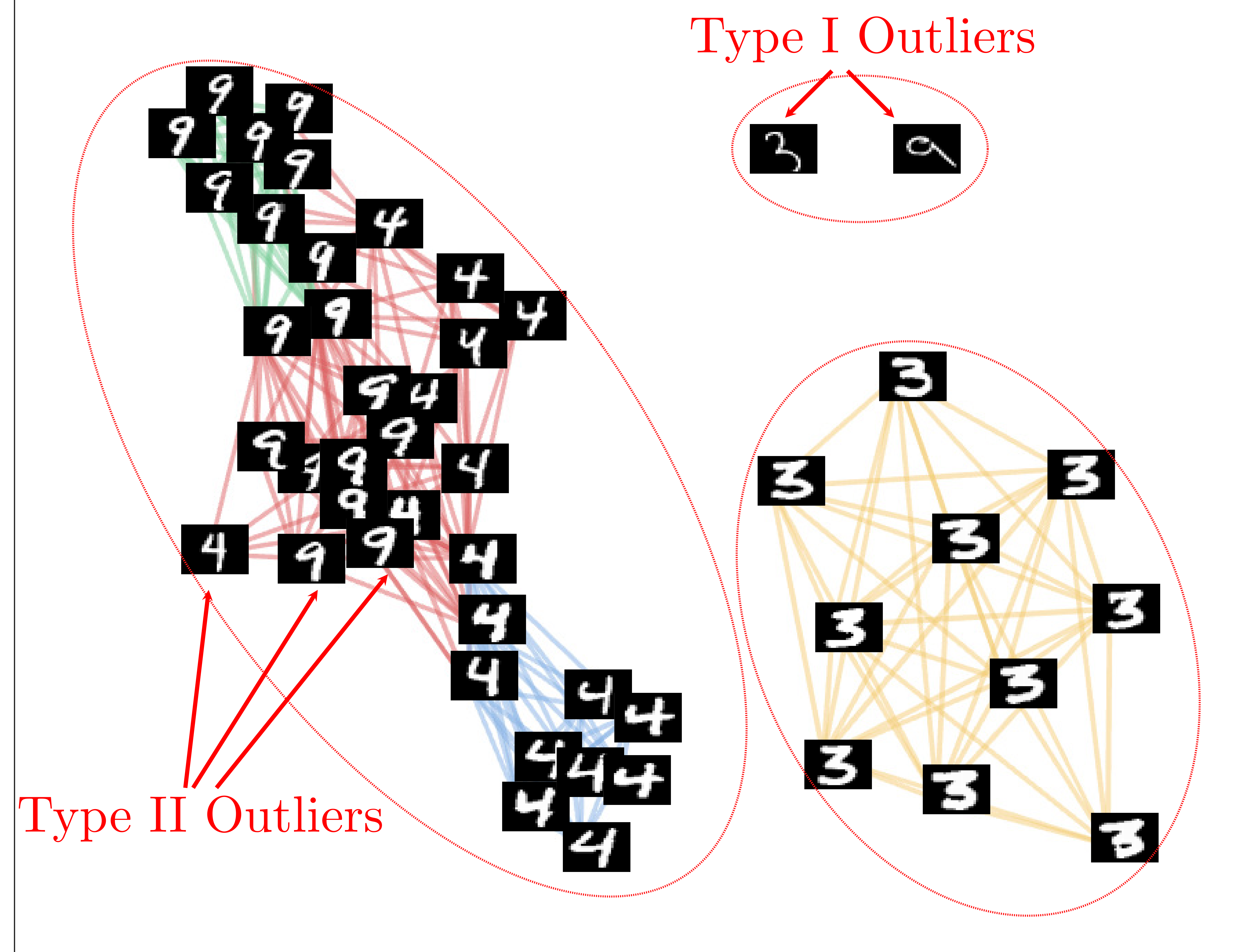}
\caption{Examplary graph partitioning digit samples from MNIST data base \cite{MNIST}.}
  \label{fig:ExamplaryPlotofcorruptedpartitioning}
\end{figure}

A further significant challenge when working with real-world data is the presence of heavy-tailed noise and outliers \citeform[29]-\citeform[31], that might obscure the eigenvalue structure in corrupted data sets. This results in a performance degradation for BDR approaches that rely on estimating eigenvalues to determine connectedness. To illustrate the necessity for robustness, a graph partitioning application is shown in Fig.~\ref{fig:ExamplaryPlotofcorruptedpartitioning} for a defined level of sparsity using the well-known handwritten digit samples from the MNIST data base \cite{MNIST}.
In the examplary graph model, the red edges represent connections to outliers while the remaining edges are the informative edges, where green, blue and yellow lines represent the within-cluster edges of digits 9, 4 and 3, respectively. The red ellipses indicate cluster assignments that are computed based on the general graph partitioning principle, in which the number of edges that cross the cut is minimized \cite{arorageometryflows}. As can be seen, unconnected outlying digit samples (`Type~I outliers') are assigned into a small cluster while a different type of outliers (`Type~II outliers') that create false positive connections between multiple clusters cause a merging of characters four and nine into one large cluster. 

In this work, we propose a method for robustly estimating an underlying BD structure, given an outlier-corrupted affinity matrix. We call this method: \emph{Fast and Robust Sparsity-Aware Block Diagonal Representation}. We build upon the definition of a vector $\mathbf{v}$ that we recently introduced in \cite{EBDR} to represent the BD affinity matrix as a piece-wise linear function.
Compared to existing popular BDR approaches, such as,
\citeform[6]-\citeform[8], the optimization is efficiently performed in a vector space instead of matrix space. Additionally and in contrast to \cite{EBDR}, the method is robust against outliers. Our main contributions are summarized as follows:

\begin{enumerate}
    \item We perform comprehensive robustness analysis that quantifies the effects of outliers. In particular, our theoretical analysis shows how the vector $\mathbf{v}$ and the eigenvalues, which carry substantial information about the BD structure, are influenced by outliers. 
    
    \item Our analysis enables the development of a BDR algorithm that is (i) robust against outliers that obscure the target BD structure and (ii) computationally efficient by re-formulating the problem as a piece-wise linear function optimization instead of a matrix-optimization. We show that our proposed method provides mathematically interpretable results in challenging settings where deriving eigenvalue information is no longer possible (i.e., in the extreme case when all blocks are connected because of corruption by outliers).
    
\end{enumerate}
 
The paper is organized as follows. Section~\ref{sec:Preliminaries} contains a summary of notations and a brief discussion on eigen-decomposition. The detailed eigenvalue analysis and outlier effects are presented in Section~\ref{sec:EigAnalysisandOutlierEffects}. The
simplification of the graph Laplacian matrix analysis by means of vector $\mathbf{v}$ and the associated outlier effect analysis are the subject of Section~\ref{sec:SimpLaplacianAnalysisandOutlierEffects}. The proposed FRS-BDR method is detailed in Section~\ref{sec:ProposedMethod} and experimental evaluations demonstrating the performance of FRS-BDR in comparison to popular BDR approaches are shown in Section~\ref{sec:ExperimentalResults}. Finally, conclusions are drawn in Section~\ref{sec:Conclusion}. {The codes that implement the FRS-BDR method are available at: \url{https://github.com/A-Tastan/FRS-BDR}}

\vspace{-2mm}
\section{PRELIMINARIES}\label{sec:Preliminaries}
\setlength{\parindent}{0pt}
\vspace{-0.5mm}
\subsection{Summary of Notation}\label{sec:notations}\vspace{-0.5mm}
Lower and upper-case bold letters denote vectors and matrices, respectively; $|x|$ denotes the absolute value of $x$; $\|\mathbf{x}\|$ denotes the norm of vector $\mathbf{x}$ while $\mathrm{med}(\mathbf{x})$ denotes its median; ${\mathrm{sign}(x)={x}/{|x|}}$; $\mathrm{diag}(x_1,\myydots,x_{\dimN})$ denotes a diagonal matrix of size $\dimN\times\dimN$ with $x_1,\myydots,x_{\dimN}$ on its diagonal; $\mathbf{I}$ denotes the identity matrix; $\mathbf{1}$ denotes the column vector of ones; $\hat{\mathbf{x}}$ denotes the estimate of vector $\mathbf{x}$; $\Tilde{\mathbf{W}}$ refers to a corrupted affinity matrix; $\indexi$, $\indexj$ and $\indexk$ are index operators for the blocks, e.g. $\Tilde{\mathbf{W}}_{\indexi}$ denotes $\indexi$th block in $\Tilde{\mathbf{W}}$; $\indexm$, $\indexn$ and $\indexr$ are index operators for the samples; finally $\TypeIoutlier$ and $\TypeIIoutlier$ denote, respectively, index operators for the Type~I and Type~II outliers. 

\vspace{-2mm}
\subsection{Eigen-decomposition of Laplacian Matrix}\label{sec:EigendecompositionLaplacian}\vspace{-0.5mm}
Let data set ${\mathbf{X}=[\mathbf{x}_1\dots\mathbf{x}_{\dimN}]\in\mathbb{R}^{\dimM\times\dimN}}$ with $\dimM$ denoting the feature dimension and $\dimN$ being the number of feature vectors, be represented as a graph ${G=\{V,E,\mathbf{W}\}}$, where $V$ denotes the vertices, $E$ represents the edges, and $\mathbf{W}\in\mathbb{R}^{\dimN\times\dimN}$ is the symmetric affinity matrix. The affinity matrix is computed from ${\mathbf{X}}$ by choosing an appropriate similarity measure, such as, the cosine similarity for which ${w_{\indexm,\indexn}=\mathbf{x}_{\indexm}^{\top}\mathbf{x}_{\indexn}}$, $\indexm\neq\indexn$ s.t. $\|\mathbf{x}_{\indexm}\|_2=1$, $\|\mathbf{x}_{\indexn}\|_2=1$. Let $\mathbf{L}\in\mathbb{R}^{\dimN\times \dimN}$ denote the nonnegative definite Laplacian matrix that is defined by the eigen-problem\vspace{-1mm}
\begin{equation}\label{eq:eigenproblem_LaplacianD1}
\mathbf{L}\mathbf{y}_{\indexm}=\lambda_{\indexm}\mathbf{y}_{\indexm}, 
\end{equation}
or in a generalized eigenvalue problem form\vspace{-0.5mm}
\begin{equation}\label{eq:eigenproblem_Laplacian}
       \mathbf{L}\mathbf{y}_{\indexm}=\lambda_{\indexm}\mathbf{D}\mathbf{y}_{\indexm},
\end{equation}
with associated eigenvalues ${0\leq\lambda_0\leq \lambda_1\leq\myydots\leq\lambda_{\dimN-1}}$ sorted in ascending order. Here, ${\mathbf{L}=\mathbf{D}-\mathbf{W}}$, where ${\mathbf{D}\in\mathbb{R}^{\dimN\times \dimN}}$ is a diagonal weight matrix with edge weights ${d_{\indexm,\indexm}=\sum_{\indexn}w_{\indexm,\indexn}}$ on the diagonal,  $\lambda_{\indexm}$ denotes the ${\indexm}$th eigenvalue and ${\mathbf{y}_{\indexm}\in\mathbb{R}^{\dimN}}$ is the eigenvector associated with $\lambda_{\indexm}$.

\vspace{-1mm}
\section{Eigenvalue Analysis and Outlier Effects}\label{sec:EigAnalysisandOutlierEffects}
\vspace{-0.5mm}
The eigen-decomposition of a Laplacian matrix has numerous applications \citeform[33]-\citeform[37] and, in particular, it plays
a crucial role in graph-based cluster analysis \cite{BDRclustering3}, \citeform[38]-\citeform[44]. However, isolated outliers and outliers that induce undesired correlations between different clusters may negatively impact the eigen-decomposition, leading to a breakdown of clustering algorithms \citeform[38]-\citeform[39]. Section~\ref{sec:ModelTargetEigenvalues} summarizes briefly our previous findings in \cite{EBDR}. A new series of solutions based on the standard eigen-decomposition in Eq.~\eqref{eq:eigenproblem_LaplacianD1} is provided in Appendix~B of the accompanying material. Then, the effect of outliers and group similarity on eigenvalues is analyzed in Section~\ref{sec:OutlierEffectonTargetEigenvalues} for both eigen-decompositions, i.e. for Eqs.~\eqref{eq:eigenproblem_LaplacianD1} and \eqref{eq:eigenproblem_Laplacian}.

\begin{figure}[tbp!]
  \centering

 \vspace{-3mm}
\subfloat[$\mathbf{L}\in\mathbb{R}^{\dimN\times\dimN}$]{\includegraphics[trim={0mm 0mm 0mm 0mm},clip,width=4.2cm]{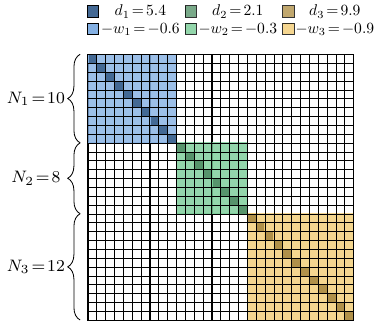}\label{fig:targetLaplacian}}
\subfloat[$\boldsymbol{\lambda}\in\mathbb{R}^{\dimN}$]{\includegraphics[trim={0mm 0mm 0mm 0mm},clip,width=5cm]{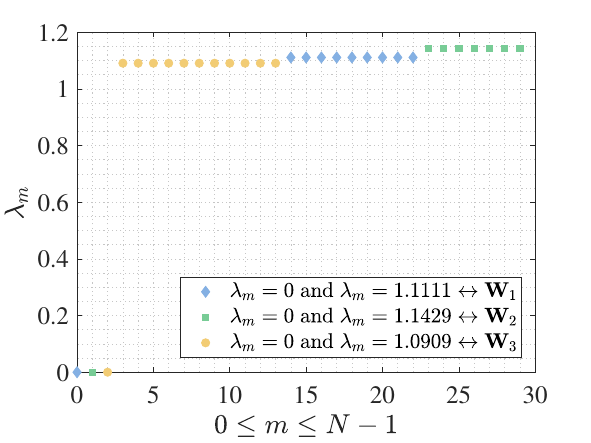}\label{fig:targeteigenvaluesvec}}
\caption{Examplary target Laplacian matrix and its eigenvalues ($\mathbf{n}=[10,8,12]^{\top}\in\mathbb{R}^{\blocknum}$, $N=30$, $\blocknum=3$).}
  \label{fig:ExamplaryPlotofTargeteigenvalues}
\end{figure}

\vspace{-2.5mm}
\subsection{Target Eigenvalues for Graph-based Clustering}\label{sec:ModelTargetEigenvalues}\vspace{-0.5mm}
As graph partitioning approaches seek to partition the set of vertices in $G$ into disjoint sets and minimizing 
the number of the edges that cross the cut \cite{arora2009expander}, \citeform[45]-\citeform[46], an ideal, i.e., target BD affinity matrix is defined in \cite{EBDR} as follows.

\begin{definition}\label{def:TargetBDR}
\textbf{\hspace{-1mm}(Target BD Affinity Matrix, \cite{EBDR})}
\textit{Let $\mathbf{W}\hspace{-0.7mm}\in\hspace{-0.7mm}\mathbb{R}^{\dimN\times \dimN}$ be a $\blocknum$ block zero-diagonal symmetric affinity matrix with blocks $\mathbf{W}_1,\mathbf{W}_2,\myydots,\mathbf{W}_{\blocknum}$ on its diagonal. Each block $\mathbf{W}_{\indexi}$, $\indexi\hspace{-0.7mm}=\hspace{-0.7mm}1,\myydots,\blocknum$ is associated with a number $\dimN_{\indexi}\hspace{-0.8mm}\in\hspace{-0.7mm}\mathbb{Z}_{+}\hspace{-0.7mm}>\hspace{-0.7mm}1$ of feature vectors and concentrated around a similarity constant $w_{\indexi}\hspace{-0.7mm}\in\hspace{-0.7mm}\mathbb{R}^{+}, \indexi\hspace{-0.7mm}=\hspace{-0.7mm}1,\myydots,\blocknum$ with negligibly small variations. $\mathbf{W}$ is called the target affinity matrix if and only if the similarity coefficients between different blocks are all zero-valued.
}\vspace{-0.5mm}
\end{definition}
Based on this definition, the corresponding ideal graph $G$ includes only edges between vertices associated with the same block. In \cite{EBDR}, using spectral analysis, we showed that if there exists a $\mathbf{W}$ as in Definition~\ref{def:TargetBDR}, the eigenvalues of the associated Laplacian matrix $\mathbf{L}\in\mathbb{R}^{\dimN\times\dimN}$ will be of the following form
\begin{align}\label{eq:lambdatarget}
  \hspace{-1.5mm}\boldsymbol{\lambda}\hspace{-1.2mm}=\hspace{-0.5mm}\mathrm{sort}\hspace{-0.1mm}\bigg(\hspace{-1mm}
   \footnotesize\underbrace{0,\myydots,0}_{\blocknum},\underbrace{\frac{\dimN_1}{\dimN_1-1},\myydots,\frac{\dimN_1}{\dimN_1-1}}_{\dimN_1-1},\myydots,\underbrace{\frac{\dimN_\blocknum}{\dimN_\blocknum-1},\myydots,\frac{\dimN_\blocknum}{\dimN_\blocknum-1}}_{\dimN_\blocknum-1}\normalsize\hspace{-1mm}\bigg),\hspace{-1mm}
\end{align}
where $\boldsymbol{\lambda}\in\mathbb{R}^{\dimN}$ denotes the vector of target eigenvalues and $\mathrm{sort}(.)$ is the sorting operation in ascending order. 

Fig.~\ref{fig:ExamplaryPlotofTargeteigenvalues} illustrates the vector of target eigenvalues $\boldsymbol{\lambda}$ associated with a Laplacian matrix of $\blocknum\hspace{-0.7mm}=\hspace{-0.7mm}3$ blocks where each block is assumed to be concentrated around a constant ${w_{\indexi}\hspace{-0.7mm}\in\hspace{-0.7mm}\mathbb{R}^{+}}$, e.g. ${\mathbf{w}=[0.6,0.3,0.9]^{\top}\in\mathbb{R}^{\blocknum}}$. Fig.~\ref{fig:targeteigenvaluesvec} confirms the findings of \cite{EBDR}, i.e., that the smallest eigenvalue is zero-valued and the remaining $ \dimN_\indexi-1$ number of eigenvalues are $\frac{\dimN_{\indexi}}{\dimN_{\indexi}-1}$ for each block $\indexi=1,\myydots,\blocknum$.

For clustered data, the target block diagonal model in Definition~\ref{def:TargetBDR} represents the optimal level of sparsity with internally dense and externally disjoint groups of vertices. If the observed data would ideally follow this model, it would not contain outliers and the sparsity level could directly be deduced from the percentage of zero-valued entries in the affinity matrix. It is evident that setting additional entries in the affinity matrix in Definition~\ref{def:TargetBDR} to zero (resulting in an overly sparse graph) will reduce the internally dense structure of a cluster and lead to the occurrence of Type~I outliers (see Definition~\ref{def:outliertype1}). In contrast, an overly dense graph, is obtained by adding undesired edges between blocks, which is consistent with the occurrence of Type~II outliers (see Definition~\ref{def:typeIIoutliers}) and its extreme case of group similarity (see Definition~\ref{def:groupsim}). The introduced theoretical analysis in the following section describes the effect of these fundamental outlier types on the optimal level of sparsity and shows how optimizing the level of sparsity based on the determined target block diagonal model prevents these fundamental outlier effects and addresses robustness and sparsity jointly.

\vspace{-2mm}
\subsection{Outlier Effects on Target Eigenvalues}\label{sec:OutlierEffectonTargetEigenvalues}
From Eq.~\eqref{eq:lambdatarget}, it follows that the non-zero components of the target eigenvalues contain the block size information. However, in practice, such a target vector is not available. Especially for outlier-corrupted affinity matrices, the blocks might be obscured (see also Fig.~\ref{fig:Examplaryplotvectorvdeviation} for an example), which results, e.g., in a performance degradation of an eigenvalue-based block size estimate. To quantify this more precisely, and subsequently derive robust BDR methods, we next define some fundamental outlier types and analyze their effects on the target eigenvalues.

\begin{definition}\label{def:outliertype1}
\textbf{(Type~I Outliers, \cite{RRLPI})}
\textit{The feature vectors corresponding to the vertices that do not share edges with any of the samples are called Type~I outliers.}
\end{definition}

\begin{figure}[tbp!]\vspace{-3mm}
  \centering

\subfloat[$\tilde{\mathbf{W}}\hspace{-0.5mm}\in\hspace{-0.5mm}\mathbb{R}^{(\dimN+2)\times(\dimN+2)}$]{\includegraphics[width=4.2cm]{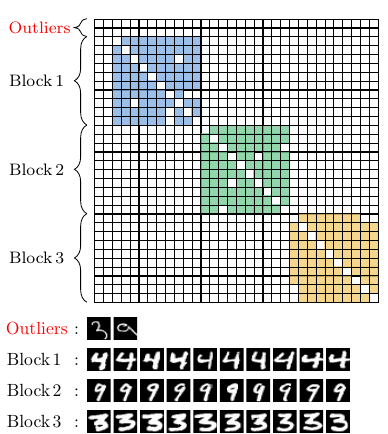}\label{fig:TypeIoutliercorruptedaffinity}}\hspace{3mm}
\subfloat[ $\tilde{G}=\{\tilde{V},\tilde{E},\tilde{\mathbf{W}}\}$ ]{\includegraphics[width=4.2cm]{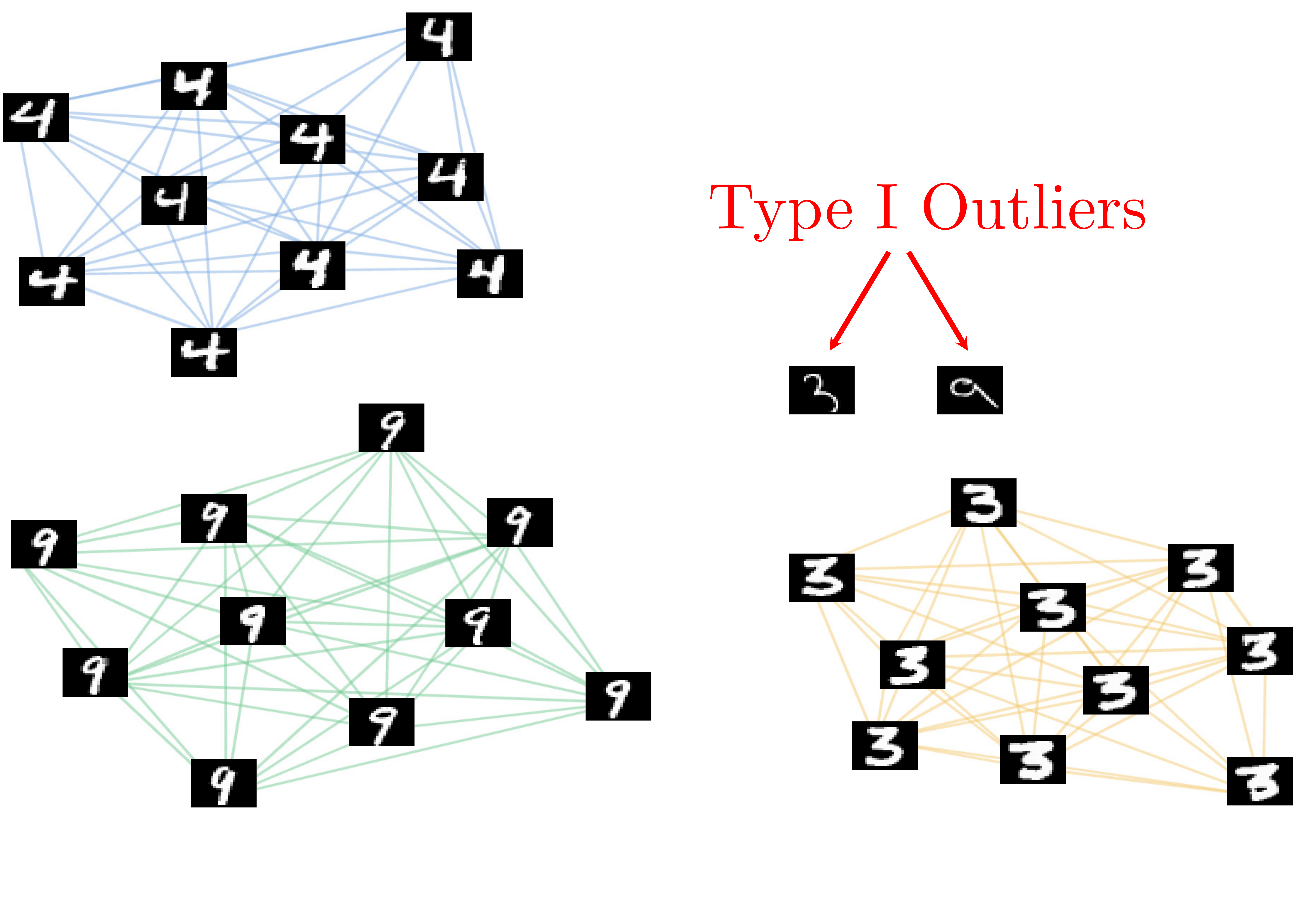}\label{fig:TypeIoutliercorruptedgraphmodel}}
\caption{Illustration of Type~I outliers. The colored cells in the corrupted BD affinity matrix $\tilde{\mathbf{W}}$ represent non-zero edge weights in graph $\tilde{G}$.}
  \label{fig:ExamplaryPlotofTypeIOutliers}
\end{figure}

Definition~\ref{def:outliertype1} is illustrated in 
Fig.~\ref{fig:ExamplaryPlotofTypeIOutliers} in which the unconnected vertices in $\tilde{G}$ are Type~I outliers. Since the multiplicity of the zero-valued eigenvalues of $\mathbf{L}$ equals the number of connected components \cite{LuxburgSC}, this means that $\dimN_{\mathrm{I}}$ Type~I outliers lead to $\dimN_{\mathrm{I}}$ additional zero-valued eigenvalues \cite{RRLPI}.

In real-world scenarios the number of Type~I outliers varies  and their occurrence, generally speaking, is affected by multiple factors: One significant delimiter for the number of Type~I outliers is the data structure. For example, a simple similarity measure, i.e. $\mathbf{W}=\mathbf{X}^{\top}\mathbf{X}$ will produce a sparse affinity matrix only when the feature vectors are sparse. In practice, using images or medical observations as feature vectors usually generates non-sparse affinity matrices for a simple similarity measure (e.g. $\mathbf{W}=\mathbf{X}^{\top}\mathbf{X}$) while using, e.g. a term-document matrix as data matrix may result in a sparse matrix and consequently to the occurrence of Type~I outliers. The second important delimiter is the affinity matrix construction. In more details, for a sparse affinity matrix construction method increasing sparsity  produces Type~I outliers. An example illustrating the link between Type~I outliers and sparse affinity matrix construction is shown in Fig.~\ref{fig:ExamplaryPlotofSparsityLevel} and in Appendix~E.1 of the accompanying material for the MNIST data base. 

Next, we study the effect of Type~II outliers, defined as follows:

\begin{definition}\vspace{-0.75mm}\label{def:typeIIoutliers}
\textbf{(Type~II Outliers, \cite{RRLPI})}
\textit{The feature vectors corresponding to the vertices that share edges with more than one group of feature vectors are called Type~II outliers.}\vspace{-0.75mm}
\end{definition}

\begin{figure}[tbp!]
  \centering

\subfloat[$\tilde{\mathbf{W}}\hspace{-0.5mm}\in\hspace{-0.5mm}\mathbb{R}^{(\dimN+3)\times(\dimN+3)}$]{\includegraphics[width=4.1cm]{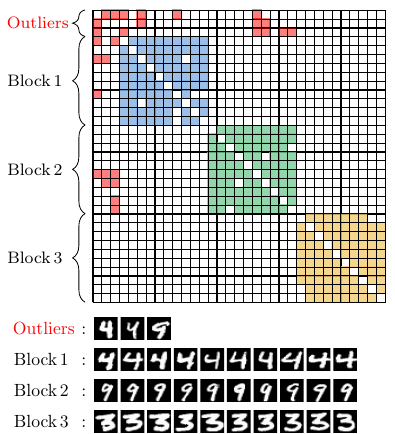}\label{fig:TypeIIoutliercorruptedaffinity}}\hspace{3mm}
\subfloat[ $\tilde{G}=\{\tilde{V},\tilde{E},\tilde{\mathbf{W}}\}$ ]{\includegraphics[width=4.3cm]{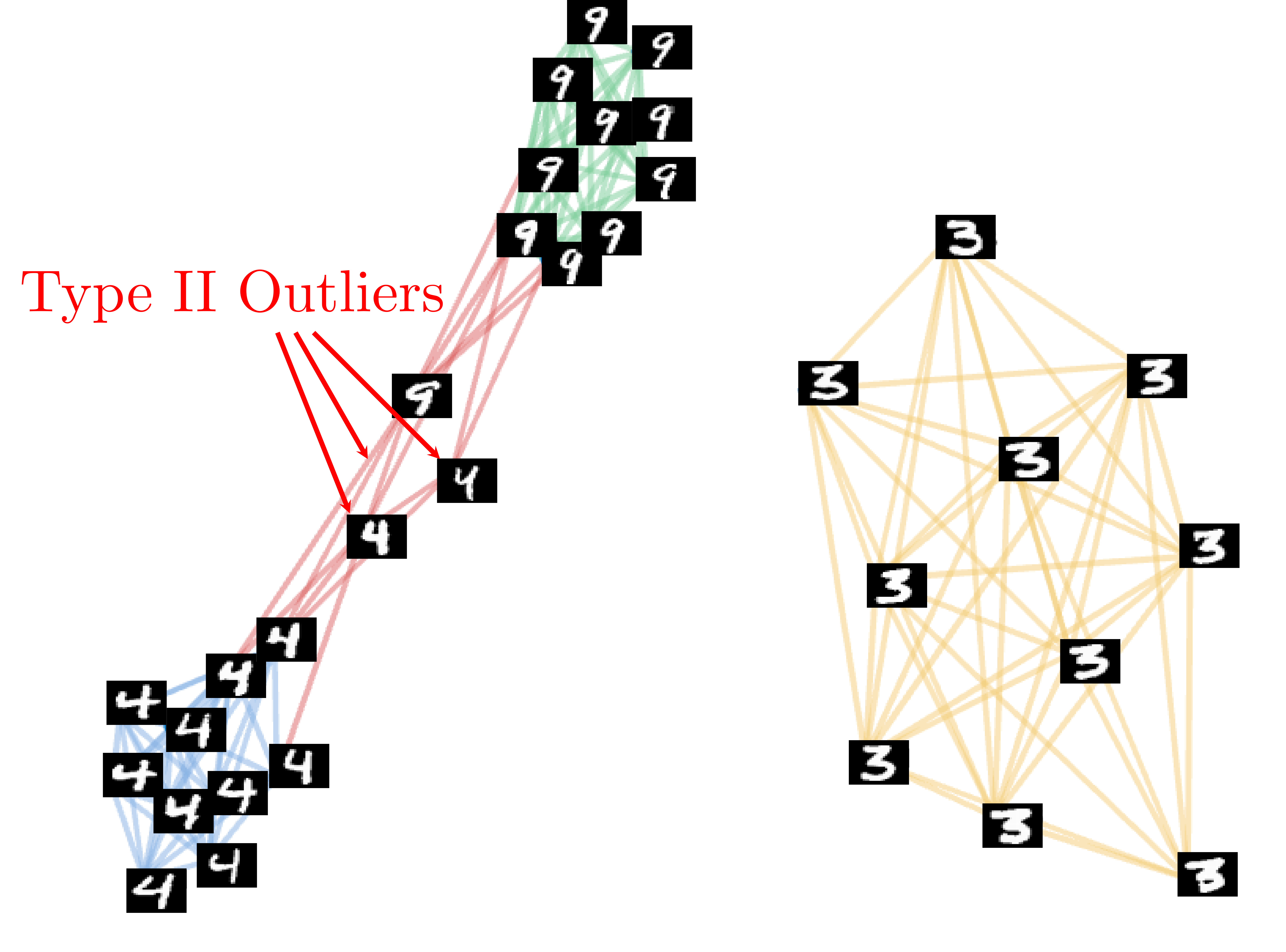}\label{fig:TypeIIoutliercorruptedgraphmodel}}
\caption{Illustration of Type~II outliers. The red colored cells in $\tilde{\mathbf{W}}$ correspond to edges of Type~II outliers.}
  \label{fig:ExamplaryPlotofTypeIIOutliers}
\end{figure}

Definition~\ref{def:typeIIoutliers} is illustrated in Fig.~\ref{fig:ExamplaryPlotofTypeIIOutliers}, which shows that the connectedness of Type~II outliers to multiple groups of feature vectors obscures the target group structure and poses a challenge to BDR methods.

In contrast to Type~I outliers studied in \cite{RRLPI}, the effect of Type~II outliers on eigenvalues is still an open problem. Therefore, an analysis of the Type~II outliers' effect on the eigenvalues of the Laplacian matrix is provided for the generalized eigen-decomposition in Eq.~\eqref{eq:eigenproblem_Laplacian} as follows.\footnote{For an analysis based on the standard eigen-decomposition in Eq.~\eqref{eq:eigenproblem_LaplacianD1}, see Appendix~B of the accompanying material.}

\begin{theorem}\label{theorem:theoremtype2outliereffectEq2}\vspace{-0.75mm}
Let $\Tilde{\mathbf{W}}\hspace{-0.7mm}\in\hspace{-0.7mm}\mathbb{R}^{(\dimN+1)\times (\dimN+1)}$ define a symmetric affinity matrix, that is equal to $\mathbf{W}$, except for an additional Type~II outlier that shares similarity coefficients with $\blocknum$ blocks where $\Tilde{w}_{\TypeIIoutlier,\blocknum}\hspace{-0.7mm}>\hspace{-0.7mm}0$ denotes the similarity coefficient between the Type~II outlier and the $\blocknum$th block. Then, for the associated corrupted Laplacian matrix $\Tilde{\mathbf{L}}\hspace{-0.7mm}\in\hspace{-0.7mm}\mathbb{R}^{(\dimN+1)\times (\dimN+1)}$ with eigenvalues $\Tilde{\boldsymbol{\lambda}}\in\mathbb{R}^{\dimN+1}$, it holds that
\begin{align*}
\small
\begin{cases}
\begin{split}
    &\dimN_1-1\hspace{1mm}\mathrm{elements\hspace{1mm}of}\hspace{1mm} \Tilde{\boldsymbol{\lambda}}\mathrm{\hspace{1mm}are\hspace{1mm}equal\hspace{1mm}to\hspace{2mm}} \frac{\dimN_1w_1+\Tilde{w}_{\TypeIIoutlier,1}}{\Tilde{d}_1},\\\vspace{1mm}
    &\dimN_2-1\hspace{1mm}\mathrm{elements\hspace{1mm}of\hspace{1mm}} \Tilde{\boldsymbol{\lambda}}\underdot{\hspace{1mm}\mathrm{are}\hspace{1mm}\mathrm{equal\hspace{1mm}to}\hspace{2mm}}\frac{\dimN_2w_2+\Tilde{w}_{\TypeIIoutlier,2}}{\Tilde{d}_2},\\
    &\\\vspace{-3mm}
    &\dimN_{\blocknum}-1\mathrm{\hspace{1mm}elements\hspace{1mm}of}\hspace{1mm}
    \Tilde{\boldsymbol{\lambda}}\mathrm{\hspace{1mm}are\hspace{1mm}equal\hspace{1mm}to\hspace{2mm}}\frac{\dimN_{\blocknum}w_{\blocknum}+\Tilde{w}_{\TypeIIoutlier,\blocknum}}{\Tilde{d}_\blocknum},\\\vspace{1mm}
    &\mathrm{the\hspace{1mm}smallest}\mathrm{\hspace{1mm}element\hspace{1mm}of}\hspace{1mm}
    \Tilde{\boldsymbol{\lambda}}\mathrm{\hspace{1mm}is\hspace{1mm}equal\hspace{1mm}to\hspace{1mm}zero,}
     \end{split}
\end{cases}
\end{align*}
and the remaining $\blocknum$ eigenvalues are the roots of
\begin{align*}\small
\prod_{\indexj=1}^{\blocknum}(\Tilde{w}_{\TypeIIoutlier,\indexj}-\Tilde{\lambda}\Tilde{d}_{\indexj})\Bigg(-\sum_{\indexj=1}^{\blocknum}\frac{\dimN_{\indexj}\Tilde{w}_{\TypeIIoutlier,\indexj}\Tilde{d}_{\indexj}}{\Tilde{w}_{\TypeIIoutlier,\indexj}-\Tilde{\lambda}\Tilde{d}_{\indexj}}-\Tilde{d}_{\TypeIIoutlier} \Bigg)=0,
\end{align*}
where $\Tilde{d}_{\TypeIIoutlier}=\sum\limits_{\indexj=1}^{\blocknum} \dimN_{\indexj}\Tilde{w}_{\TypeIIoutlier,\indexj}$ and $\Tilde{d}_{\indexj}=(\dimN_{\indexj}-1)w_{\indexj}+\Tilde{w}_{\TypeIIoutlier,\indexj}$.
\end{theorem}

\begin{proof}\hspace{-2.3mm}
See Appendix~A.1 of the accompanying material.\hspace{-4.5mm}
\end{proof}

\vspace{1mm}
We next introduce an extreme case of Type~II outliers based on the following definition.

\begin{definition}
\textbf{(Group Similarity)}\label{def:groupsim}\vspace{-1mm}
\textit{If an entire group of vertices shares edges with another group of vertices we call this, group similarity.}\vspace{-0.75mm}
\end{definition}

The Laplacian matrix of Definition~\ref{def:groupsim} can be considered as a single connected component which means that the number of zero-valued eigenvalues equals to one \cite{LuxburgSC}. In contrast to this simple interpretation, the remaining eigenvalues can be formulated as a function of intra-blocks and inter-blocks similarity coefficients where inter-blocks similarity coefficients are generally smaller-valued than those of intra-blocks in real-world scenarios. To provide a mathematical understanding of this,  the following theorem quantifies the effect of group similarity on the target eigenvalues.

\begin{theorem}\label{theorem:corollarykgroupcorrelationeffectongeneralizedeigendecomposition}\textit{Let $\Tilde{\mathbf{W}}\in\mathbb{R}^{\dimN\times \dimN}$ define an affinity matrix that is equal to $\mathbf{W}$, except that block $\indexi$ has similarity with the remaining $\blocknum\hspace{-0.2mm}-\hspace{-0.2mm}1$ blocks with $\Tilde{w}_{\indexi,\indexj}\hspace{-0.7mm}=\hspace{-0.7mm}\Tilde{w}_{\indexj,\indexi}\hspace{-0.7mm}>\hspace{-0.7mm}0$ denoting the value around which the similarity coefficients between blocks $\indexi$ and $\indexj$ are concentrated for $\indexj=1,\myydots,\blocknum$ and $\indexi\neq\indexj$. Then, 
the eigenvalues 
$\Tilde{\boldsymbol{\lambda}}\in\mathbb{R}^{\dimN}$ of $\Tilde{\mathbf{L}}\in\mathbb{R}^{\dimN\times \dimN}$ are as follows:\vspace{-1mm}
\begin{align*}
\begin{cases}
\small
\begin{split}
    &\dimN_{\indexi}-1\hspace{1mm}\mathrm{elements\hspace{1mm}of}\hspace{1mm} \Tilde{\boldsymbol{\lambda}}\hspace{1mm}\mathrm{are}\hspace{1mm}\mathrm{equal\hspace{1mm}to\hspace{2mm}}\frac{\dimN_{\indexi}w_{\indexi}+\sum\limits_{\substack{\indexj=1,\\ \indexj\neq \indexi}}^{\blocknum}\dimN_{\indexj}\Tilde{w}_{\indexi,\indexj}}{\Tilde{d}_{\indexi}},\\
    &\dimN_{\indexj}-1\hspace{1mm}\mathrm{elements\hspace{1mm}of}\hspace{1mm} \Tilde{\boldsymbol{\lambda}}\hspace{1mm}\underdot{\mathrm{are}\hspace{1mm}\mathrm{equal\hspace{1mm}to\hspace{2mm}}}\frac{\dimN_{\indexj}w_{\indexj}+\dimN_{\indexi}\Tilde{w}_{\indexi,\indexj}}{\Tilde{d}_{\indexj}},\\
    &\\\vspace{-2mm}
    &\dimN_{\blocknum}-1\hspace{1mm}\mathrm{elements\hspace{1mm}of}\hspace{1mm} \Tilde{\boldsymbol{\lambda}}\hspace{1mm}\mathrm{are}\hspace{1mm}\mathrm{equal\hspace{1mm}to\hspace{2mm}}\frac{\dimN_{\blocknum}w_{\blocknum}+\dimN_{\indexi}\tilde{w}_{\indexi,\blocknum}}{\Tilde{d}_{\blocknum}},\\
    &\mathrm{the\hspace{1mm} smallest}\hspace{1mm}\mathrm{element\hspace{1mm}of}\hspace{1mm} \Tilde{\boldsymbol{\lambda}}\hspace{1mm}\mathrm{is}\hspace{1mm}\mathrm{equal\hspace{1mm}to\hspace{1mm}zero,}
    \end{split}
\end{cases}
\end{align*}
and the remaining $\blocknum\hspace{-0.45mm}-\hspace{-0.45mm}1$ eigenvalues in ${\Tilde{\boldsymbol{\lambda}}}$ are the roots of
\begin{align*}\small
\prod\limits_{\substack{\indexj=1 \\ \indexj\neq \indexi}}^{\blocknum}(\dimN_{\indexi}\Tilde{w}_{\indexi,\indexj}-\Tilde{\lambda}\Tilde{d}_{\indexj})\Bigg(-\sum\limits_{\substack{\indexj=1 \\ \indexj\neq \indexi}}^{\blocknum}\frac{\Tilde{d}_{\indexj}\dimN_{\indexj}\Tilde{w}_{\indexi,\indexj}}{\dimN_{\indexi}\Tilde{w}_{\indexi,\indexj}-\Tilde{\lambda}\Tilde{d}_{\indexj}}-\Tilde{d}_{\indexi}\Bigg)=0,
\end{align*}}where ${\Tilde{d}_{\indexj}\hspace{-0.8mm}=\hspace{-0.8mm}(\dimN_{\indexj}\hspace{-0.5mm}-\hspace{-0.5mm}1)w_{\indexj}\hspace{-0.6mm}+\hspace{-0.6mm}\dimN_{\indexi}\Tilde{w}_{\indexi,\indexj}}$, ${\Tilde{d}_{\indexi}\hspace{-0.8mm}=\hspace{-0.8mm}(\dimN_{\indexi}\hspace{-0.5mm}-\hspace{-0.5mm}1)w_{\indexi}\hspace{-0.6mm}+\hspace{-1.1mm}\sum\limits_{\substack{\indexj=1 \\ \indexj\neq \indexi}}^{\blocknum}\hspace{-0.5mm}\dimN_{\indexj}\Tilde{w}_{\indexi,\indexj}}$.
\end{theorem}

\begin{proof}\hspace{-2.3mm}\vspace{-0.8mm}
See Appendix~A.2 of the accompanying material.\hspace{-3.7mm}
\end{proof}

\begin{figure}[tbp!]
  \centering

 \vspace{-3mm}
\subfloat[$\tilde{\mathbf{L}}\in\mathbb{R}^{(\dimN+2)\times(\dimN+2)}$]{\includegraphics[trim={0mm 0mm 0mm 0mm},clip,width=4.2cm]{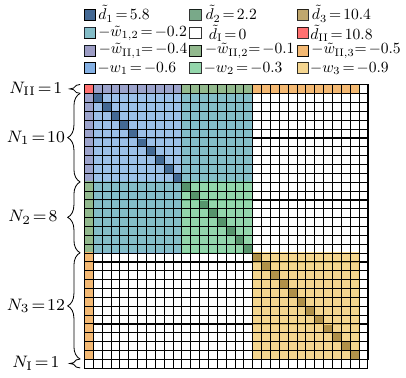}\label{fig:corruptedLaplacian}}\hspace{-3mm}
\subfloat[$\tilde{\boldsymbol{\lambda}}\in\mathbb{R}^{\dimN+2}$]{\includegraphics[trim={0mm 1mm 2mm 3mm},clip,width=4.8cm]{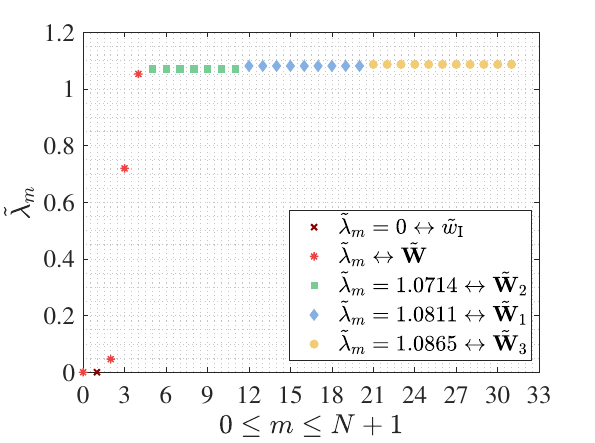}\label{fig:corruptedeigenvaluvec}}
\caption{Examplary corrupted Laplacian matrix and its eigenvalues ($\mathbf{n}=[10,8,12]^{\top}\hspace{-0.7mm}\in\hspace{-0.5mm}\mathbb{R}^{\blocknum}$, $N=30$, $\blocknum=3$).}
  \label{fig:ExamplaryPlotofcorruptedeigenvalues}\vspace{-1mm}
\end{figure}

A Laplacian matrix $\tilde{\mathbf{L}}$ that is corrupted with all discussed outlier types is displayed in Fig.~\ref{fig:corruptedLaplacian}, while the above derived outlier effects on the eigenvalues are visually summarized in Fig.~\ref{fig:corruptedeigenvaluvec}.

\textbf{Remark 1.} To derive the theoretical results, simplifying assumptions\footnote{For our further analysis about loosening assumptions based on eigenvectors, see Theorem~1 in \cite{atastanEUSIPCO23}.}, such as, concentration of the similarity coefficients within a block around a mean value are required. In practice, these assumptions may only be approximately fulfilled. However, will see later in Section~\ref{sec:ExperimentalResults} that the numerical performance gain obtained by suppressing outliers' effects outweigh the model mismatch in the considered benchmark data sets. Further, existing theoretical works on the spectrum of BDR methods, e.g. \citeform[49]-\citeform[52], make even stricter assumptions, because the spectrum of an adjacency matrix can be found in closed form for simple models \cite{esagrapherror}. According to the most recent generic analyses \cite{PPMeigenvalueanalysis} and \cite{esagrapherror}, the spectrum of a BDR has been computed for planted partition model (PPM) with equal community sizes which is a special case of stochastic block model (SBM) assuming that the probability of having an edge within the cluster is constant and equal for all clusters while probability of having an edge to a different cluster is also constant and the same for all clusters. There are also some other researches on eigenvalues to determine the limiting distribution of the edge eigenvalues \cite{tang2018eigenvalues} and that of the outlier eigenvalues \cite{athreya2021eigenvalues}. Even though these previously introduced spectral properties of random matrices are interesting to understand the complex graph structures, the available information about the spectrum of a BD matrix is limited to the considered simple models, i.e. PPM,    and the available information about the eigenvalues is limited to eigenvalues of non-weighted graphs.

\vspace{-1mm}
\section{Simplified Laplacian Matrix Analysis and Outlier Effects}\label{sec:SimpLaplacianAnalysisandOutlierEffects}\vspace{-0.5mm}
In the preceeding sections, outlier effects have been analyzed for $\dimN\hspace{-0.7mm}\times\hspace{-0.7mm} \dimN$ Laplacian matrices, which may lead to computationally heavy methods for large graphs. In this section, we therefore re-formulate the problem in $\dimN\times1$ vector space. In particular, assuming that $\mathbf{W}$ is symmetric and BD\footnote{A sparse matrix can be transformed into a BD form using the Reverse Cuthill-McKee (RCM) algorithm \cite{RCM}.}, the analysis is simplified by defining the vector $\mathbf{v}\in\mathbb{R}^{\dimN}$ as follows\vspace{-1.5mm} \cite{EBDR}
\begin{align}\label{eq:vectorvcomputation}
    v_{\indexm}=\sum_{\indexn=\indexm}^{\dimN}l_{\indexm,\indexn},
\end{align}
where $v_{\indexm}$ and $l_{\indexm,\indexn}$, respectively, denote the $\indexm$th  and $(\indexm,\indexn)$th components of $\mathbf{v}$ and $\mathbf{L}$.

\begin{figure}[tbp!]
  \centering

 \vspace{-3mm}
\subfloat[$\mathbf{L}\in\mathbb{R}^{\dimN\times\dimN}$]{\includegraphics[trim={0mm 0mm 0mm 0mm},clip,width=4.2cm]{Images_Journal/Target_Laplacian_TikZ.pdf}\label{fig:targetLaplacian}}
\subfloat[$\mathbf{v}\in\mathbb{R}^{\dimN}$]{\includegraphics[trim={3mm 0mm 5mm 3mm},clip,width=4.8cm]{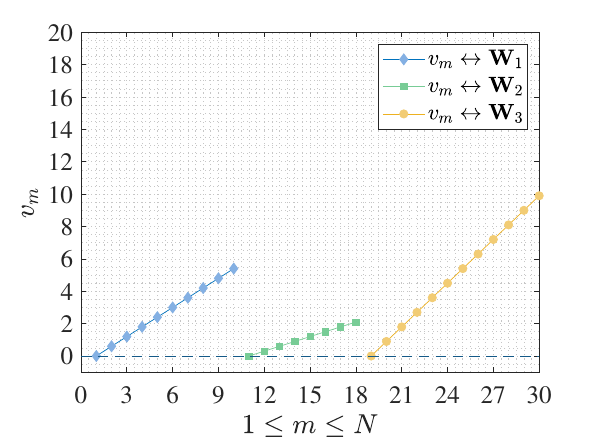}\label{fig:targetvectorv}}\vspace{-1mm}
\caption{Examplary target Laplacian matrix and corresponding vector $\mathbf{v}$ ($\mathbf{n}=[10,8,12]^{\top}\in\mathbb{R}^{\blocknum}$, $N=30$, $\blocknum=3$).}
  \label{fig:ExamplaryPlotofTargetVectorv}
\end{figure}

\vspace{-2mm}
\subsection{Target Vector $\mathbf{v}$ for  Graph-based Clustering}\label{sec:ModelTargetVectorv}\vspace{-0.5mm}
In \cite{EBDR}, we have shown that the target vector $\mathbf{v}$ is a piece-wise linear function of the following form.

\begin{definition}\label{def:targetvectorv}\textbf{(Target Vector $\mathbf{v}$, \cite{EBDR})}\textit{
The target vector $\mathbf{v}$ is a piece-wise linear function of the following form
\begin{align*}\small
v_{\indexm}=f(\indexm)=\begin{cases}(\indexm-\ell_1)w_1   &\underdot{\mathrm{if}}\hspace{3mm}\ell_1\leq \indexm\leq u_1\\\vspace{2mm}
(\indexm-\ell_{\blocknum})w_{\blocknum}&\mathrm{if}\hspace{3mm}\ell_{\blocknum}\leq \indexm\leq u_{\blocknum},
\end{cases}
\end{align*}
where $\ell_1\hspace{-1mm}=\hspace{-1mm}1$, $u_1\hspace{-1mm}=\hspace{-1mm}\dimN_1$, $\ell_{\indexi}\hspace{-1mm}=\hspace{-1.6mm}\sum\limits_{\indexk=1}^{\indexi-1}\hspace{-1mm}\dimN_{\indexk}\hspace{-0.4mm}+\hspace{-0.4mm}1$ $\mathrm{and}$ ${u_{\indexi}\hspace{-1mm}=\hspace{-1.6mm}\sum\limits_{\indexk=1}^{\indexi}\hspace{-1mm}\dimN_{\indexk}}$ for $\indexi\hspace{-1mm}=\hspace{-1mm}2,\hspace{-0.3mm}\myydots,\hspace{-0.3mm}\blocknum$.}
\end{definition}

An illustration is provided in Fig~\ref{fig:ExamplaryPlotofTargetVectorv} for a $\blocknum=3$ block Laplacian matrix. As can be seen, the changepoints of the piece-wise linear function provide information about the block size. To arrive at robust methods, we next determine the outlier effects on $\mathbf{v}$.
\vspace{-2mm}
\subsection{Outlier Effects on Target Vector $\mathbf{v}$}\label{sec:OutlierEffectsonTargetVectorv}\vspace{-0.5mm}



For a Type~I outlier-corrupted affinity matrix $\tilde{\mathbf{W}}\in\mathbb{R}^{(\dimN+1)\times{(\dimN+1)}}$ that is identical to $\mathbf{W}$, except for a single Type~I outlier $\mathrm{o}_{\TypeIoutlier}$, the overall edge weight associated with $\mathrm{o}_{\TypeIoutlier}$ is zero-valued, i.e. $\tilde{d}_{\TypeIoutlier}=0$. Based on Def.~\ref{def:targetvectorv}, it is straightforward to show that 
the component in the associated corrupted
vector $\Tilde{\mathbf{v}}\in\mathbb{R}^{\dimN+1}$ that is
associated with Type~I outliers is zero valued, i.e., $\Tilde{v}_{\TypeIoutlier}=0$.  The Type II outliers' effect on $\mathbf{v}$ is as follows.

\begin{theorem}\vspace{-0.75mm}\label{theorem:CorollaryTypeIIOutlierEffectonTargetVectorv}\textit{
Let $\Tilde{\mathbf{W}}\in\mathbb{R}^{(\dimN+1)\times (\dimN+1)}$ define a Type~II outlier-corrupted BD affinity matrix that is identical to $\mathbf{W}\in\mathbb{R}^{\dimN\times\dimN}$ except for a single Type~II outlier that has non-zero similarity coefficients with all blocks. Assuming that the similarity coefficients associated with the outlier $\mathrm{o}_{\TypeIIoutlier}$ and the blocks $\indexj\in\{1,\myydots,\blocknum\}$ are concentrated around $\Tilde{w}_{\TypeIIoutlier,\indexj}$, the components, whose indexes are valued between the outlier index and the largest index of the $\indexj$th block, such that ${\indexm}_{\TypeIIoutlier}<\indexm\leq u_{\indexj}$, increase by $\Tilde{w}_{\TypeIIoutlier,\indexj}$ in the corrupted vector $\tilde{\mathbf{v}}\in\mathbb{R}^{\dimN+1}$. Further, the component associated with the Type~II outlier is given by
\begin{align*}\small
\Tilde{v}_{\TypeIIoutlier}\hspace{-1mm}=\hspace{-0.8mm}\begin{cases}
   0\hspace{0.1cm}&\mathrm{if}\hspace{2mm}0<\hspace{-0.5mm}{\indexm}_{\TypeIIoutlier}\hspace{-0.5mm}<\ell_1\\
   ({\indexm}_{\TypeIIoutlier}\hspace{-0.5mm}-\hspace{-0.5mm}\ell_1)\tilde{w}_{\TypeIIoutlier,1}&\underdot{\mathrm{if}}\hspace{2mm}\ell_1\hspace{-0.5mm}<\hspace{-0.5mm}{\indexm}_{\TypeIIoutlier}\hspace{-0.5mm}<\ell_2\\
   \sum\limits_{\indexj=1}^{\blocknum-1}\hspace{-1mm}\dimN_{\indexj}\tilde{w}_{\TypeIIoutlier,\indexj}\hspace{-0.5mm}+\hspace{-0.5mm}({\indexm}_{\TypeIIoutlier}\hspace{-0.5mm}-\hspace{-0.5mm}\ell_{\blocknum})\tilde{w}_{\TypeIIoutlier,\blocknum}&\mathrm{if}\hspace{2mm}\ell_{\blocknum}\hspace{-0.9mm}<\hspace{-0.5mm}{\indexm}_{\TypeIIoutlier}\hspace{-0.5mm}\leq \hspace{-0.5mm}\dimN\hspace{-0.7mm}+\hspace{-0.7mm}1
   \end{cases}\hspace{-1mm},
\end{align*}
where $\ell_{\indexj}$ denotes the lowest index of the $\indexj$th block.
}
\end{theorem}

\begin{proof}\vspace{-0.5mm}\hspace{-2.5mm}
See Appendix~C.1 of the accompanying material.\hspace{-3.3mm}
\vspace{-0.5mm}
\end{proof}

\begin{figure}[tbp!]
  \centering
 \vspace{-3mm}
\subfloat[$\tilde{\mathbf{L}}\in\mathbb{R}^{(\dimN+2)\times(\dimN+2)}$]{\includegraphics[trim={0mm 0mm 0mm 0mm},clip,width=4.2cm]{Images_Journal/Example_Corrupted_Laplacian.pdf}\label{fig:corruptedLaplaciantwo}}\hspace{-4mm}
\subfloat[$\tilde{\boldsymbol{\lambda}}\in\mathbb{R}^{\dimN+2}$]{\includegraphics[trim={5mm 4mm 9mm 5.5mm},clip,width=4.9cm]{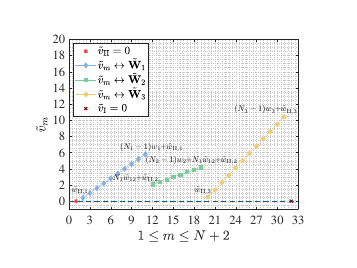}\label{fig:corruptedvectorv}}
\caption{Examplary corrupted Laplacian matrix and corresponding $\tilde{\mathbf{v}}$ ($\mathbf{n}\hspace{-0.7mm}=\hspace{-0.7mm}[10,8,12]^{\top}\hspace{-0.7mm}\in\hspace{-0.5mm}\mathbb{R}^{\blocknum}$, $N\hspace{-0.7mm}=\hspace{-0.7mm}30$, $\blocknum\hspace{-0.7mm}=\hspace{-0.7mm}3$).}
  \label{fig:ExamplaryPlotofcorruptedvectorv}
\end{figure}
The effect of group similarity on $\mathbf{v}$ is as follows.

\begin{theorem}\vspace{-0.75mm}\label{theorem:groupsimilarityeffectonvksimilarity}\textit{
Let $\Tilde{\mathbf{W}}\in\mathbb{R}^{\dimN\times \dimN}$ define a corrupted affinity matrix that is identical to $\mathbf{W}\in\mathbb{R}^{\dimN\times \dimN}$, except that block $\indexi$ has non-zero similarity coefficients with the remaining $\blocknum-1$ blocks with ${\Tilde{w}_{\indexi,\indexj}=\Tilde{w}_{\indexj,\indexi}>0}$ denoting the similarity coefficients around which, blocks $\indexi$ and $\indexj$ are concentrated. These similarities result in an increase by $\dimN_{\indexi}\Tilde{w}_{\indexi,\indexj}$ in the components associated with the blocks ${\indexj=\indexi+1,\myydots,\blocknum}$ of $\tilde{\mathbf{v}}\in\mathbb{R}^{\dimN}$ while the components of $\indexj<\indexi$ remain the same. Further, the components associated with block $\indexi$ remain the same for $\indexi=1$ and increase by $\sum\limits_{\indexj=1}^{\indexi-1}\dimN_{\indexj}\Tilde{w}_{\indexi,\indexj}$ for $2\leq \indexi\leq \blocknum$.}
\end{theorem}

\begin{proof}\vspace{-0.75mm}\hspace{-2.5mm}
See Appendix~C.2 of the accompanying material.\hspace{-3.3mm}
\end{proof}
In the sequel, we analyze the worst case of group similarity, i.e., similarity of all blocks.  Note that, in this case, eigenvalues can not be formulated as a function of similarity coefficients due to the impossibility of simplifying determinants of full matrices via Gaussian elimination. However, recovering the structure of $\mathbf{W}$ based on $\mathbf{v}$ is possible based on the following result.

\begin{figure}[tbp!]
  \centering
\subfloat[Iris]{\includegraphics[trim={3mm 0mm 3mm 2mm},clip,width=4.5cm]{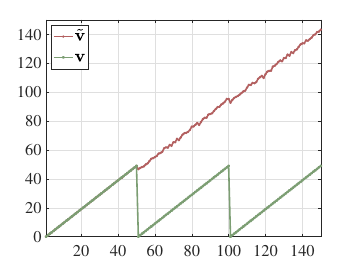}\label{fig:irisvectorvdeviation}}
\subfloat[Person Identification]{\includegraphics[trim={3mm 0mm 3mm 2mm},clip,width=4.5cm]{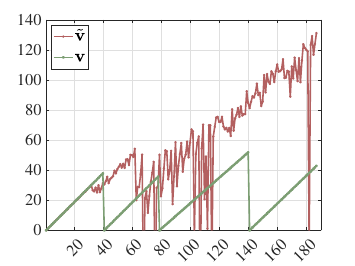}\label{fig:personidvectorvdeviation}}
\caption{Examplary deviations from the target vector $\mathbf{v}$. The affinity matrix is defined by $\mathbf{W}=\mathbf{X}^{\top}\mathbf{X}$.}
  \label{fig:Examplaryplotvectorvdeviation}
\end{figure}

\begin{figure}[tbp!]
  \centering
\subfloat[Iris]{\includegraphics[trim={3mm 0mm 3mm 2mm},clip,width=4.5cm]{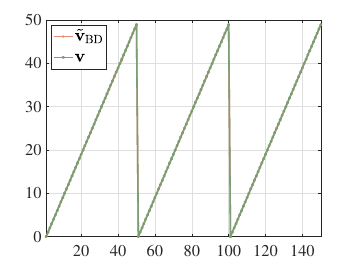}\label{fig:irisvectorvBDdeviation}}
\subfloat[Person Identification]{\includegraphics[trim={3mm 0mm 3mm 2mm},clip,width=4.5cm]{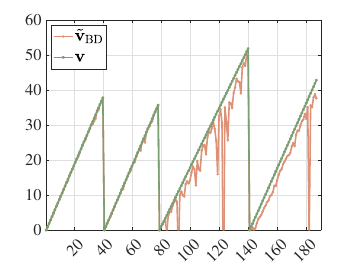}\label{fig:personidvectorvBDdeviation}}
\caption{Examplary deviations from the target vector $\mathbf{v}$. The BD affinity matrix is defined by removing the undesired similarity coefficients between different blocks of $\mathbf{W}=\mathbf{X}^{\top}\mathbf{X}$.}
  \label{fig:ExamplaryplotvectorvBDdeviation}
\end{figure}


\begin{corollary}\vspace{-0.75mm}\label{corollary:groupsimilarityeffectonvallsimilar}\textit{
Let $\Tilde{\mathbf{W}}\in\mathbb{R}^{\dimN\times \dimN}$ define a corrupted affinity matrix that is identical to $\mathbf{W}\in\mathbb{R}^{\dimN\times \dimN}$, except that each block $\indexi=1,\myydots,\blocknum$ has non-zero similarity coefficients with the remaining $\blocknum-1$ blocks with ${\Tilde{w}_{\indexi,\indexj}=\Tilde{w}_{\indexj,\indexi}>0}$ denoting the similarity coefficients around which, blocks $\indexi$ and $\indexj$ are concentrated for $\indexj=1,\myydots,\blocknum$ and $\indexi\neq\indexj$. This leads to a piece-wise linear function given by
\begin{align*}
\small
\tilde{v}_{\indexm}\hspace{-0.7mm}=\hspace{-1mm}\begin{cases}(\indexm-\ell_1)w_1   &\hspace{-1.2mm}\mathrm{if}\hspace{1.5mm}\ell_1\leq \indexm\leq u_1\\
(u_1-\ell_1+1)\Tilde{w}_{1,2}\hspace{-0.7mm}+\hspace{-0.7mm}(\indexm-\ell_2)w_2 &\hspace{-1.2mm}\underdot{\mathrm{if}}\hspace{1.5mm}\ell_2\leq \indexm\leq u_2
\\\vspace{2mm}
\sum\limits_{\indexi=1}^{\blocknum-1}(u_{\indexi}-\ell_{\indexi}+1)\Tilde{w}_{\indexi,\blocknum}\hspace{-0.7mm}+\hspace{-0.7mm}(\indexm-\ell_{\blocknum})w_{\blocknum}&\hspace{-1.2mm}\mathrm{if}\hspace{1.5mm}\ell_{\blocknum}\leq \indexm\leq u_{\blocknum}
\end{cases}
\end{align*}
where $\ell_1=1$, $u_1=\dimN_1$, $\ell_{\indexi}=\sum\limits_{\indexk=1}^{\indexi-1}\dimN_{\indexk}+1$ and ${u_{\indexi}=\sum\limits_{\indexk=1}^{\indexi}\dimN_{\indexk}}$ for $\indexi=2,\myydots,\blocknum$.
}
\end{corollary}

\begin{proof}\hspace{-2.5mm}
See Appendix~C.2 of the accompanying material.\hspace{-3.5mm}
\end{proof}

An examplary corrupted Laplacian matrix $\tilde{\mathbf{L}}$ and corresponding $\tilde{\mathbf{v}}$ illustrating our theoretical findings are shown in Fig.~\ref{fig:corruptedLaplaciantwo} and Fig.~\ref{fig:corruptedvectorv}, respectively. Consistent with Section~\ref{sec:OutlierEffectonTargetEigenvalues}, outliers of Type~I result in zeros in $\tilde{\mathbf{v}}$. Additionally, Type~II outliers and group similarity lead to an increase in the target vector $\mathbf{v}$ as quantified in Theorems~\ref{theorem:CorollaryTypeIIOutlierEffectonTargetVectorv} and \ref{theorem:groupsimilarityeffectonvksimilarity}, respectively. 

 
\begin{figure}[tbp!]
\includegraphics[trim={2mm 0mm 2mm 0mm},clip,width=\linewidth]{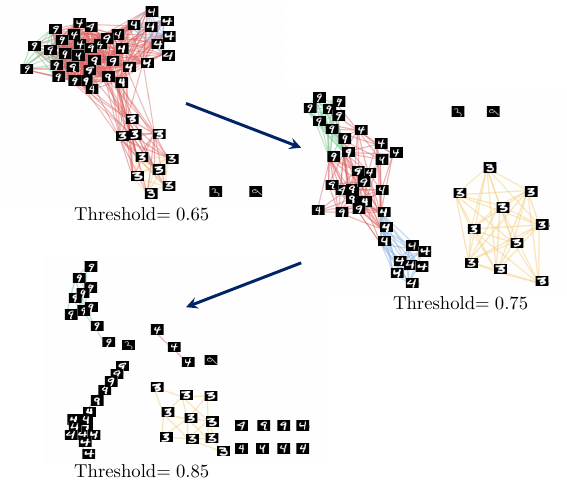}
\caption{Example graphs for increasing sparsity. An initial affinity matrix is defined by $\mathbf{W}=\mathbf{X}^{\top}\mathbf{X}$ and the example graphs are obtained by removing the edges whose corresponding edge weight is smaller than the defined threshold value. }
  \label{fig:ExamplaryPlotofSparsityLevel}
\end{figure}

To demonstrate the degree of model mismatch in the considered real-world data sets due to the simplifying assumptions
that were necessary to derive our theory, examplary vectors associated with the corrupted affinity matrices that are subject to group similarity as in Corollary~\ref{corollary:groupsimilarityeffectonvallsimilar} and the corresponding target vector $\mathbf{v}$'s are illustrated in Figs.~\ref{fig:irisvectorvdeviation} and~\ref{fig:personidvectorvdeviation} , respectively, for the Iris \cite{Iris} and Person Identification \cite{PersonIdentification} data sets.  As can be seen, undesired similarity coefficients between different blocks result in shifts from the target piece-wise linear functions starting from the second linear pieces, consistent with our theory in Corollary~\ref{corollary:groupsimilarityeffectonvallsimilar}. In particular, assumptions and findings of Corollary~\ref{corollary:groupsimilarityeffectonvallsimilar} hold well in real-world data sets, especially, when the data sets include densely connected clusters of points, e.g. Ceramic \cite{Ceramic} and Iris \cite{Iris}.\footnote{For further real-world data examples, see Appendix~E.2 of the accompanying material.} Additionally, corrupted data sets, e.g. Person Identification \cite{PersonIdentification} whose corresponding affinity matrix is subject to Type~I outliers and group similarity results in large deviations from the target piece-wise linear function with group similarity shifts and small-valued $\Tilde{\mathbf{v}}$ components corresponding to Type~I outliers as it has been theoretically shown in previous. A further analysis illustrating the degree of model mismatch between the target BD model and a BD model with varying similarity coefficients within the blocks is shown in Figs.~\ref{fig:irisvectorvBDdeviation} and~\ref{fig:personidvectorvBDdeviation}, respectively, for the Iris and Person Identification data sets. Even though highly corrupted data sets generate large deviations from the assumed models in real-world scenarios, an appropriate BDR suppresses these outlier effects by providing an optimal level of sparsity which is a major motivation of our proposed algorithm that will be detailed in the sequel.

\begin{figure*}[!tbp]
  \centering
\includegraphics[trim={1mm 0mm 1mm 1mm},clip,width=\linewidth]{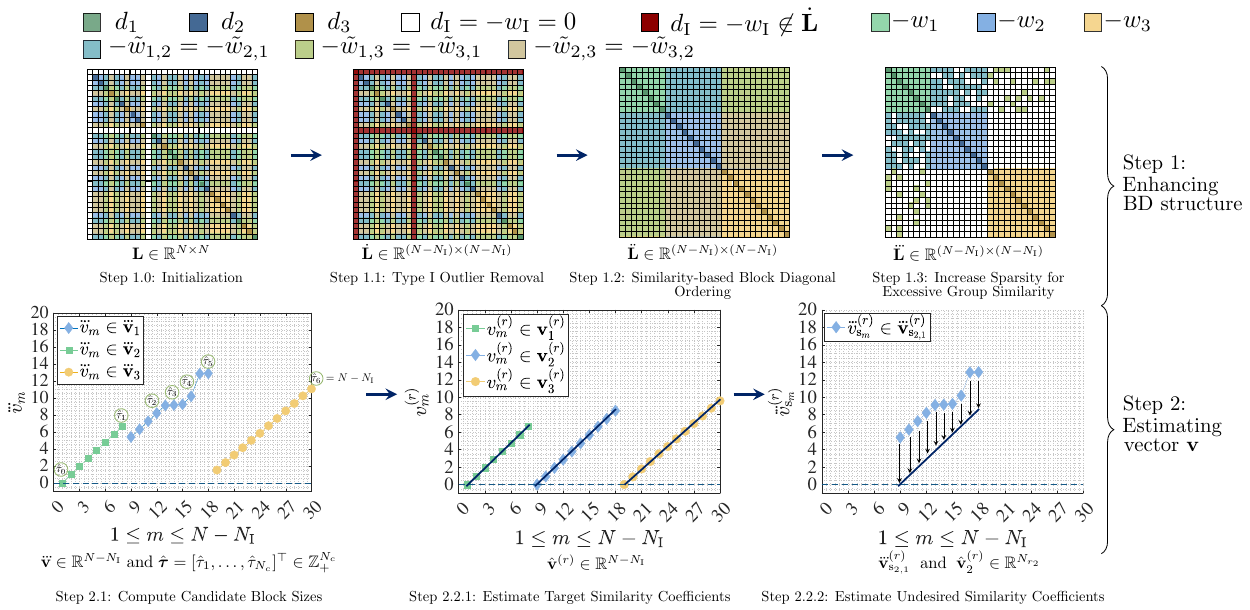}
\caption{High-level flow diagram illustrating the key steps of FRS-BDR using a generic example with $\blocknum=3$ clusters.}
 \label{fig:ExamplaryPlotofRvBDR}
\end{figure*}

\section{The Proposed Method}\label{sec:ProposedMethod}
In Section~\ref{sec:outline}, we briefly discuss the key ideas of the proposed method. Following this, a step-by-step detailed mathematical explanation is provided in Section~\ref{sec:FRS-BDR Algorithm}. We then analyze the computational complexity in Section~\ref{sec:ComputationalAnalysis}. Additionally, a comprehensive visual summary is provided in Appendix~F.1 of the accompanying material and a pseudo-code algorithm of FRS-BDR is given in Algorithm~\ref{alg:FRS-BDR}.

\vspace{-1mm}
\subsection{Problem Statement: Jointly Addressing Robustness and Sparsity}\label{sec:ProblemStatement}

With the results of Sections~\ref{sec:EigAnalysisandOutlierEffects} and~\ref{sec:SimpLaplacianAnalysisandOutlierEffects} in place, we are ready to understand the relationship between the level of sparsity and the previously defined outlier types to highlight the importance of jointly addressing robustness and sparsity. In a generic example, Fig.~\ref{fig:ExamplaryPlotofSparsityLevel} shows that a dense graph (top) contains high amounts of group similarity while increasing sparsity reduces the number of Type~II outliers (middle). Finally, further increasing sparsity generates Type~I outliers until at some point the underlying true cluster structure is completely lost. This means that an inaccurate determination of the sparsity level leads to the above discussed outlier effects for existing approaches, such as, e.g. \citeform[21]-\citeform[23]. In this section, we therefore propose a new method that addresses robustness and sparsity \emph{jointly}.

More precisely, let a given data set of feature vectors $\mathbf{X}\in\mathbb{R}^{\dimM\times\dimN}$ be represented as a weighted graph ${G=\{V,E,\mathbf{W}\}}$, i.e., ${\mathbf{W}=\mathbf{X}^{\top}\mathbf{X}}$ and ${\|\mathbf{x}_m\|=1}$, ${m=1,\myydots,N}$. Further, let ${\mathbf{D}}$ and ${\mathbf{L}\in\mathbb{R}^{\dimN\times\dimN}}$ denote, respectively, the overall edge weight and the Laplacian matrices associated with ${\mathbf{W}}$. Then, the goal of this work is to robustly estimate 
a $\blocknum$ block zero-diagonal symmetric affinity matrix $\mathbf{W}\in\mathbb{R}^{\dimN\times\dimN}$ using the available information about the vector $\mathbf{v}$ and an eigen-decomposition. The number of blocks $\blocknum$ is assumed to be unknown and ${\mathbf{X}}$ may be subject to heavy-tailed noise and outliers which results in undesired effects, such as group similarity. The number of outliers is assumed to be unknown. Computational efficiency is also of fundamental interest. \emph{Thus, in brief, the overall aim is to develop a fast and sparsity aware BDR method that is robust against outliers and group similarity}.

\subsection{Main Ideas and Outline of Proposed Method}\label{sec:outline}
This section summarizes the main ideas of our proposed \emph{Fast and Robust Sparsity-Aware Block Diagonal Representation (FRS-BDR)} method. The full details of each step are given in Section~\ref{sec:FRS-BDR Algorithm}.

To provide a general understanding, a high-level flow diagram illustrating the key steps of FRS-BDR is provided in Fig.~\ref{fig:ExamplaryPlotofRvBDR}. As shown in the figure, the method consists of two general steps, i.e., enhancing BD structure (Step 1) and estimating vector $\mathbf{v}$ (Step 2). The computation step starts with a given Type~I outlier-corrupted and non-sparse Laplacian matrix $\mathbf{L}$ (Step 1.0: Initialization in Fig.~\ref{fig:ExamplaryPlotofRvBDR}). According to the explicit Definition~\ref{def:outliertype1} of Type~I outliers, the method first removes the similarity coefficients associated with Type~I outliers, which are represented in red color, from $\mathbf{L}$ (Step 1.1: Type I Outlier Removal in Fig.~\ref{fig:ExamplaryPlotofRvBDR}). Then, the next step is to structure the resulting matrix $\dot{\mathbf{L}}$ in a BD form $\ddot{\mathbf{L}}$ with a similarity-based BD ordering that we present in the sequel (Step 1.2: Similarity-based Block Diagonal Ordering in Fig.~\ref{fig:ExamplaryPlotofRvBDR}). The last part of Step~1 is, to obtain vector $\mathbf{v}$ in form of $\blocknum$ discrete linear segments by computing an ordered sparse Laplacian matrix $\dddot{\mathbf{L}}$ (Step 1.3: Sparsity for Excessive Group Similarity in Fig.~\ref{fig:ExamplaryPlotofRvBDR}). Then, the estimation step starts with a changepoint detection that we propose, to compute the possible block sizes (Step 2.1: Compute Candidate Block Sizes in Fig.~\ref{fig:ExamplaryPlotofRvBDR}). For each possible block size vector, i.e., $\mathbf{n}_r=[8,10,12]^{\top}\hspace{-0.7mm}\in\hspace{-0.7mm}\mathbb{Z}_{+}^{\blocknum}$ in this illustrating example, the method computes a target vector ${\mathbf{v}^{(r)}}$ and a corresponding estimate ${\hat{\mathbf{v}}^{(r)}}$ as a function of the estimated target similarity coefficients (Step 2.2.1: Estimate Target Similarity Coefficients in Fig.~\ref{fig:ExamplaryPlotofRvBDR}). 
Further, for every undesired similarity coefficient around which the blocks are concentrated, the shifted vectors (see Corollary~\ref{corollary:groupsimilarityeffectonvallsimilar}) are computed separately and the undesired similarity coefficients are estimated (Step 2.2.2: Estimate Undesired Similarity Coefficients in Fig.~\ref{fig:ExamplaryPlotofRvBDR}). Finally, the estimate ${\hat{\mathbf{v}}\hspace{-0.7mm}\in\hspace{-0.7mm}\mathbb{R}^{\dimN-\dimN_{\TypeIoutlier}}}$ is computed for the block size vector which provides the best fit to the computed vector $\dddot{\mathbf{v}}$.

\vspace{-2mm}
\subsection{FRS-BDR Algorithm}\label{sec:FRS-BDR Algorithm}
\subsubsection{Step 1 : Enhancing BD Structure}\label{sec:Computingvectorv}
The key requirement for computing vector $\mathbf{v}$ based on Eq.~\eqref{eq:vectorvcomputation} is recovering an approximately BD structured Laplacian matrix. Assuming that ${\mathbf{W}}$ (and the associated ${\mathbf{L}}$) are symmetric and sparse matrices, they can be ordered in a BD form \cite{RCM} based on which vector $\mathbf{v}$ can be directly computed. However, in general, similarity measures may not produce sparse affinity matrices. We therefore discuss the most challenging scenario, i.e., that ${\mathbf{W}}$ is subject to Type~I outliers and all blocks exhibit similarity. Considering the Type~I outliers' effect on the target vector $\mathbf{v}$ (see, Section~\ref{sec:OutlierEffectsonTargetVectorv}), the proposed vector $\mathbf{v}$ computation starts with Type~I outlier detection (Step~1.1). Then, a new BD ordering based on the similarity coefficients is proposed to generate a BD ordered Laplacian matrix (Step~1.2). Lastly, a sparse Laplacian matrix design is detailed for the case of excessive group similarity (Step~1.3).

\begin{figure}[!tbp]
  \centering
\includegraphics[trim={0mm 14mm 0mm 13mm},clip,width=\linewidth]{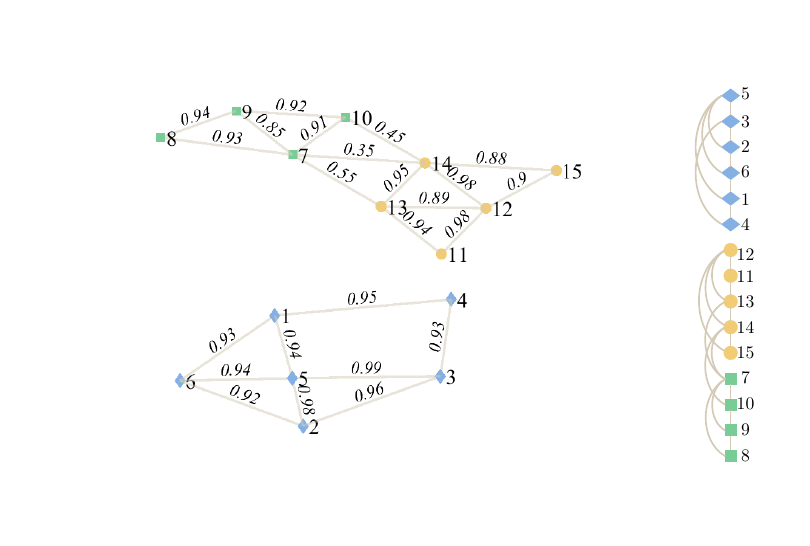}
\caption{Examplary plot of the sBDO algorithm.}
 \label{fig:ExamplaryPlotofsBDO}
\end{figure}

\vspace{1.5mm}
\paragraph{Step 1.1 : Type~I Outlier Removal}\label{sec:preprocessing}
Type~I outliers are detected according to\vspace{1mm}
\begin{align}\label{eq:preprocessing}
    \mathbf{x}_m\in\mathbf{O}_{\mathrm{I}} \hspace{3mm}\mathrm{if}\hspace{3mm}w_{\indexm,\indexn}=0\hspace{1mm}\mathrm{for}\hspace{1mm}\forall \indexn=1,\myydots,\dimN\hspace{1mm}\mathrm{and}\hspace{1mm}\indexm\neq\indexn,\vspace{2mm}
\end{align}
where $\mathbf{O}_{\mathrm{I}}\in\mathbb{R}^{\dimM\times\dimN_{\mathrm{I}}}$ denotes the matrix of Type~I outliers, $\mathbf{x}_m\in\mathbb{R}^{\dimM}$ is the 
$m$th feature vector for ${m=1,\myydots,\dimN}$, $w_{\indexm,\indexn}$ is the $\indexm,\indexn$th similarity coefficient corresponding to $\mathbf{x}_m$ (due to the symmetry of $\mathbf{W}$, $w_{\indexm,\indexn}=w_{\indexn,\indexm}$). 

Type~I outlier removal based on Eq.~\eqref{eq:preprocessing} directly follows Definition~\ref{def:outliertype1} which means that the operation does not require a determination of the number of outliers. It is an important preliminary step since the presence of Type~I outliers may lead to an inaccurate sparsity increase in Step~1.3  due to their effects on the eigenvalues or an incorrect candidate block size estimation in Step~2.1  based on their effects on the vector $\mathbf{v}$.

\vspace{1.5mm}
\paragraph{Step 1.2 : Similarity-based BD Ordering (sBDO)}\label{sec:sBDOordering}

Let ${\dot{\mathbf{X}}\in\mathbb{R}^{\dimM\times(\dimN\shortminus\dimN_{\mathrm{I}})}}$, ${\dot{\mathbf{W}},\dot{\mathbf{D}}}$ and ${\dot{\mathbf{L}}\in\mathbb{R}^{(\dimN\shortminus\dimN_{\mathrm{I}})\times(\dimN\shortminus\dimN_{\mathrm{I}})}}$ be the resulting matrices after Step~1.1. The vector of the BD order, i.e., $\hat{\mathbf{b}}\in\mathbb{Z}_{+}^{\dimN\shortminus\dimN_{\mathrm{I}}}$ is determined based on the following steps.\\
\noindent\textit{Step 1.2.1: Initialization:} The BD order vector $\hat{\mathbf{b}}^{(1)}$ is comprised of the node index of maximum overall edge weight (i.e., $\dot{d}_{\mathrm{max}}$).\\
\noindent\textit{Step 1.2.2: Adding the most similar neighbor to $\hat{\mathbf{b}}^{(s)}$:} Let $\hat{\mathbf{b}}^{(s)}\hspace{-0.7mm}=\hspace{-0.7mm}[\hat{b}_1,\myydots,\hat{b}_{s-1}]^{\top}\hspace{-0.7mm}\in\hspace{-0.7mm}\mathbb{Z}_{+}^{s-1}$, with $s=2,...,\dimN-\dimN_{\mathrm{I}}$, denote the BD order vector at the $s$th stage. Assuming that the neighbors set is non-empty\footnote{If it is empty the method simply stacks the node index of maximum overall edge weight into $\hat{\mathbf{b}}^{(s)}$.}, the most similar neighbor to $\hat{\mathbf{b}}^{(s)}$ at the $s$th stage is determined by
\begin{equation}\label{eq:selectneighbor}
\hat{b}_{s}=\underset{\indexm\in\{1,2,\myydots,{\dimN\shortminus\dimN_{\mathrm{I}}}\}}{\arg\max} \Big\{ \sum_{\indexn=1}^{s-1}\dot{w}_{\indexm,\hat{b}_\indexn}\Big\},
\end{equation}
where $\indexm\in\mathbb{Z}_{+}$ such that $1\leq m \leq \dimN\shortminus\dimN_{\mathrm{I}}$ denotes a neighbor node.

An example of the sBDO algorithm is illustrated in Fig.~\ref{fig:ExamplaryPlotofsBDO} and technically summarized in Algorithm~1. As can be seen from Fig.\ref{fig:ExamplaryPlotofsBDO}, starting from node five, whose overall edge weight is largest valued, the method selects the neighbors based on their edge weights that represent the similarity to previously selected nodes. After selecting all neighbors, the method jumps to the node that has the maximum overall edge weight among the remaining nodes and determines the ordering of the associated neighbors.

\begin{algorithm}[tbp!]\small
\setstretch{1}
\DontPrintSemicolon
\KwIn{
$\dot{\mathbf{W}},\dot{\mathbf{D}}\in \mathbb{R}^{(\dimN-\dimN_{\mathrm{I}})\times(\dimN-\dimN_{\mathrm{I}})}$}
\textit{Initialization:}\\
Find the node of maximum overall edge weight $\dot{d}_{\mathrm{max}}\hspace{-0.5mm}$\\
\vspace{0.2mm}
\SetInd{0.5em}{0.25em}
\For{$s=2,\myydots,(\dimN-\dimN_{\mathrm{I}})$}{
\SetInd{0.5em}{0.25em}
\vspace{0.2mm}
\textit{Adding the most similar neighbor to $\hat{\mathbf{b}}^{(s)}$:}\\
\vspace{0.2mm}
\eIf{at least one neighbor exists}{Estimate $\hat{b}_{s}$ using Eq.~\eqref{eq:selectneighbor} and stack into $\hat{\mathbf{b}}^{(s)}$}
{Find the node with maximum overall edge weight$\hspace{-5mm}$\\ among unselected nodes and stack $\hat{b}_{s}$ into $\hat{\mathbf{b}}^{(m)}\hspace{-2cm}$\\
}\vspace{0.2mm}
}
\caption{sBDO}
\label{alg:sBDO}
\KwOut{Estimated order vector $\hat{\mathbf{b}}^{(s)}\hspace{-0.7mm}\in\hspace{-0.7mm}\mathbb{Z}_{+}^{(\dimN-\dimN_{\mathrm{I}})}\hspace{-7mm}$ }
\end{algorithm}

Different from the Reverse Cuthill-McKee (RCM) \cite{RCM}, which is a well-known block diagonal ordering method, the proposed sBDO algorithm incorporates useful information from the similarity coefficients. By doing this, the sBDO ordering method does not require making specific assumptions\footnote{For examples of similarity coefficients' empirical distributions, see Appendix~E.3 of the accompanying material.}. on the similarity coefficients or a sparse matrix structure that is necessary in RCM algorithm. In challenging scenarios, for example, starting the ordering with a Type~II outlier the sBDO algorithm continues selecting vertices from the most similar cluster therewith quickly suppressing the effect of Type~II outlier's similarity coefficients.\footnote{For the analysis of sBDO performance, see Appendix~F.4.5 of the accompanying material.}

\vspace{1.5mm}
\paragraph{Step 1.3: Increase Sparsity for Excessive Group Similarity}\label{sec:IncreaseSparsity}
Let $\ddot{\mathbf{W}}$, $\ddot{\mathbf{D}}$ and $\ddot{\mathbf{L}}\in\mathbb{R}^{(\dimN-\dimN_{\mathrm{I}})\times(\dimN-\dimN_{\mathrm{I}})}$ be the matrices resulting from Step~1.2. A sparsity improved Laplacian matrix ${\dddot{\mathbf{L}}\in\mathbb{R}^{(\dimN\shortminus\dimN_{\mathrm{I}})\times(\dimN\shortminus\dimN_{\mathrm{I}})}}$ is designed\footnote{For the examplary sparse Laplacian matrix design algorithms, see Appendix~F.3 of the accompanying material.} by increasing sparsity as long as, at least, the two smallest eigenvalues are close to zero\footnote{For the definition of \emph{close to zero}, see Appendix~F.2 in of the accompanying material.}. After computing ${\dddot{\mathbf{L}}}$, the vector ${\dddot{\mathbf{v}}\in\mathbb{R}^{\dimN\shortminus\dimN_{\mathrm{I}}}}$ is obtained using Eq.~\eqref{eq:vectorvcomputation}\footnote{The vector $\mathbf{v}$ can alternatively be computed using ${\ddot{\mathbf{L}}\in\mathbb{R}^{(\dimN\shortminus\dimN_{\mathrm{I}})\times(\dimN\shortminus\dimN_{\mathrm{I}})}}$ after executing Steps 1.1 and 1.2 if, at least, the two smallest eigenvalues of ${\ddot{\mathbf{L}}}$ are close to zero.}. 

 Increasing sparsity for excessive group similarity is an optional step for the FRS-BDR algorithm. In particular, it is designed to 
obtain a sparsity improved Laplacian matrix ${\dddot{\mathbf{L}}}$ whose associated vector $\dddot{\mathbf{v}}$ provides distinct changepoints that can be easily computed in Step~2.1. In this way, the negative impact of excessive group similarity, which obscures the piece-wise linear functions (for details, see Corollary~\ref{corollary:groupsimilarityeffectonvallsimilar}), is suppressed and changepoints become more visible. However, this operation does not enforce a block diagonal affinity matrix since it eliminates only a small portion of the undesired similarity coefficients. Therefore, the following steps estimate the vector $\mathbf{v}$ as a function of desired similarity coefficients and that of undesired that will be removed to obtain a BDR.

\begin{figure}[!tbp]\vspace{-3mm}
  \centering
\includegraphics[trim={1mm 3mm 2mm 3mm},clip,width=\linewidth]{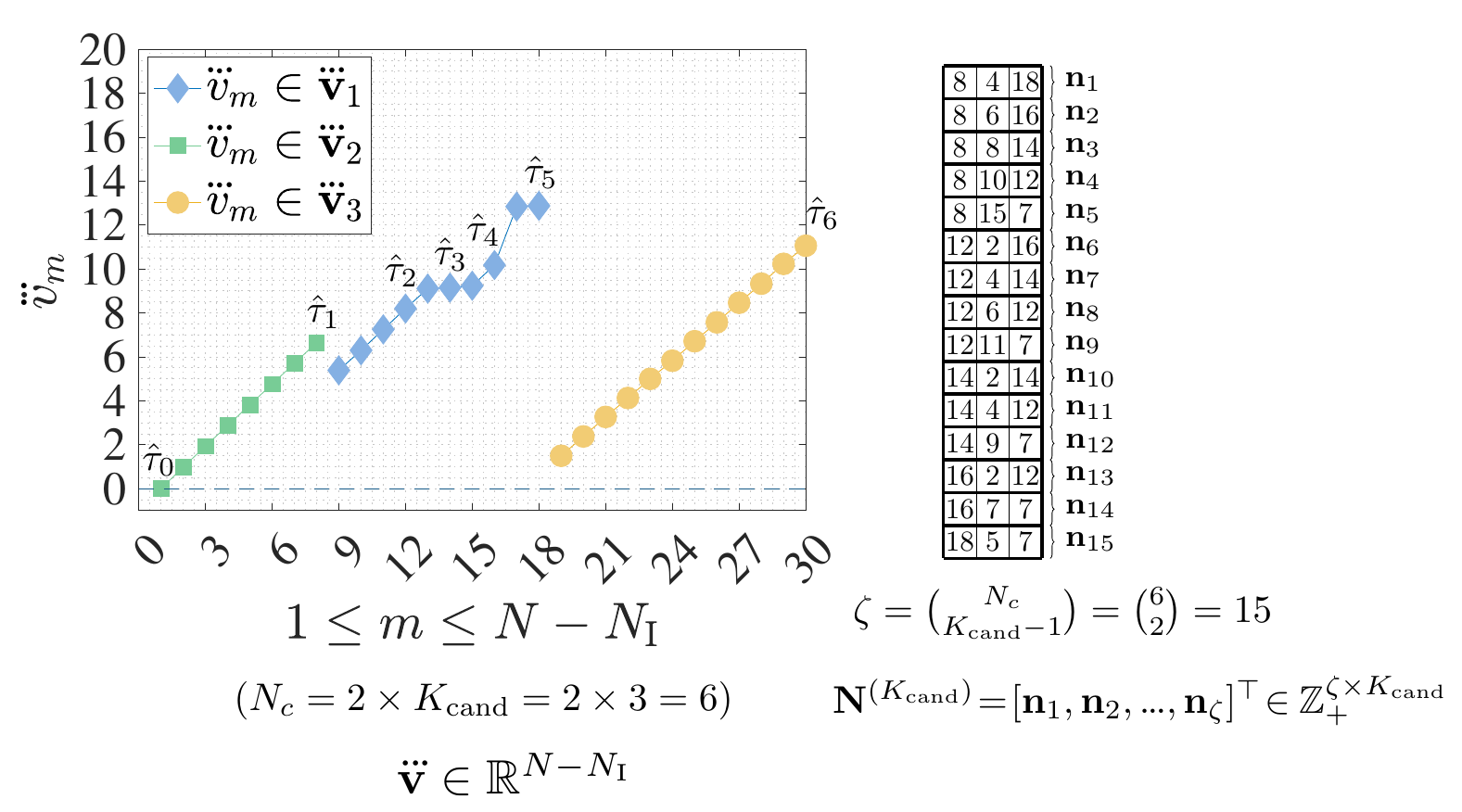}
\caption{Examplary plot of computing candidate block sizes.}
\label{fig:ExamplaryPlotofPossibleBlockSizes}
\end{figure}

\subsubsection{Step 2: Estimating Vector ${\mathbf{v}}$}\label{sec:EstimatingVectorv}
This step models ${\dddot{\mathbf{v}}}$ as a $K$-piece linear function of similarity coefficients around which the blocks are assumed to be concentrated (for details, see Corollary~\ref{corollary:groupsimilarityeffectonvallsimilar}.), i.e.,
\begin{equation}\label{eq:modellingvectorv}
\dddot{\mathbf{v}}_{\indexi}=\mathbf{v}_\indexi+\mathbf{1}\sum\limits_{\substack{\indexj=1 \\ \indexj\neq \indexi}}^{\indexi-1}\dimN_{\indexj}\Tilde{w}_{\indexi,\indexj},\hspace{5mm}\indexi=1,\myydots,\blocknum,
\end{equation}
where\vspace{-1mm}\label{eq:modellingtargetvectorv}
\begin{equation}
\mathbf{v}_\indexi=[0,w_{\indexi},\myydots,(\dimN_{\indexi}-1)w_{\indexi}]^{\top}\in\mathbb{R}^{\dimN_{\indexi}}
\end{equation}
denotes the $\indexi$th linear segment of the target vector $\mathbf{v}_\indexi$, $w_{\indexi}$ is the similarity coefficient around which the block $\indexi$ is concentrated and $\Tilde{w}_{\indexi,\indexj}$ is the undesired similarity coefficient between blocks $\indexi$ and $\indexj$ around which they are concentrated, $\mathbf{1}\in\mathbb{R}^{\dimN_{\indexi}}$ is the column vector of ones, $\dimN_{\indexi}$ and $\dimN_{\indexj}$ are, respectively, the size of block $\indexi$ and $\indexj$.

\paragraph{Step 2.1: Computing Candidate Block Sizes}\label{sec:PossibleBlockSizes}
Let  $\dimN_c\in\mathbb{Z}_{+}$ denote the number of changepoints, let ${\boldsymbol{\tau}=[\tau_1,\tau_2,\myydots,\tau_{\dimN_c}]^{\top}\in\mathbb{Z}^{\dimN_c}_{+}}$ be the vector containing the corresponding locations in ${\dddot{\mathbf{v}}}$, and let ${\tau_0=0}$ and ${\tau_{\dimN_c+1}=\dimN}$. Then, to estimate the model for vector $\dddot{\mathbf{v}}$ based on Eq.~\eqref{eq:modellingvectorv}, our first step is to detect the changepoints in ${\dddot{\mathbf{v}}}$ by minimizing the following penalized least-squares function as in \cite{changepoints}
\begin{align}\label{eq:generatecandidates}
   \sum_{\indexi=1}^{N_{c}+1}\sum_{\indexm=\tau_{\indexi-1}+1}^{\tau_{\indexi}}(\dddot{v}_{\indexm}-\hat{v}_{\indexm})^2+\gamma N_{c},
\end{align}
where $\dddot{v}_{\indexm}$ and $\hat{v}_{\indexm}$ denote, respectively, the ${\indexm}$th point in the $\indexi$th linear segment of ${\dddot{\mathbf{v}}}$ and the corresponding least-squares linear fit. $\gamma$ is the penalty parameter that controls the number of changepoints $\dimN_c$. In particular, Eq.~\eqref{eq:generatecandidates} considers all possible changepoints for $\gamma=0$ and it rejects including additional changepoints if the residual error is smaller than the determined penalty parameter $\gamma$. Different from determining $\gamma$ directly, this step increases the value of $\gamma$ gradually as long as the function finds a lower number of changepoints than a predefined maximum number of changepoints $N_{c_{\mathrm{max}}}\in\mathbb{Z}_{+}$ which is a reasonably small number satisfying ${\blocknum-1\leq N_{c_{\mathrm{max}}}}$. Then, for a candidate number of blocks from a given vector, i.e., ${\blocknum_{\mathrm{cand}}\in[\blocknum_{\mathrm{min}},\myydots,\blocknum_{\mathrm{max}}]^{\top}\in\mathbb{Z}_{+}^{\dimN_{\blocknum}}}$, the
resulting number of changepoints $\dimN_c$ and corresponding locations ${\boldsymbol{\tau}}$ in Eq.~\eqref{eq:generatecandidates} are used to compute the candidate size vectors ${\mathbf{n}_{r}=[\dimN_{r_{1}},\dimN_{r_{2}},\myydots,\dimN_{r_{\blocknum_{\mathrm{cand}}}}]^{\top}\in\mathbb{Z}_{+}^{\blocknum_{\mathrm{cand}}}}$, ${r=1,\myydots,\zeta}$ that are designed by combination of all possible size vectors with $\zeta=\binom{\dimN_c}{\blocknum_{\mathrm{cand}}-1}$. Lastly, the block-size matrix associated with a candidate number of blocks, i.e., 
\begin{equation}\label{eq:computepossibleblocksize}
\mathbf{N}^{(\blocknum_{\mathrm{cand}})}=[\mathbf{n}_{1},\mathbf{n}_{2},\myydots,\mathbf{n}_{\zeta}]^{\top}\in\mathbb{Z}^{\zeta\times\blocknum_{\mathrm{cand}}}_{+},
\end{equation}
is formed.\footnote{In practice, the candidate size vectors including the block sizes that are smaller than a predefined minimum number of nodes in the blocks $\dimN_{\mathrm{min}}$ can be removed from ${\mathbf{N}^{(\blocknum_{\mathrm{cand}})}}$.}

The computation of candidate block sizes illustrated in Fig.~\ref{fig:ExamplaryPlotofPossibleBlockSizes} for a candidate block number $\blocknum_{\mathrm{cand}}\hspace{-0.7mm}=\hspace{-0.7mm}3$.  After estimating the changepoints using Eq.~\eqref{eq:generatecandidates}, a possible block size matrix, i.e. ${\mathbf{N}^{(\blocknum_{\mathrm{cand}})}}\in\mathbb{Z}^{\zeta\times\blocknum_{\mathrm{cand}}}_{+}$, with $\zeta=15$ is computed for all possible block size combinations. 

In this step, the changepoint locations are determined based on a piece-wise linear fit of the vector $\dddot{\mathbf{v}}$ using Eq.~\eqref{eq:generatecandidates}. This is a fundamental step to compute the candidate block sizes. However, the obtained information from Eq.~\eqref{eq:generatecandidates} does not provide the target and undesired similarity coefficients which are needed to structure the affinity matrix in a block diagonal form. In other words, the estimated piece-wise linear fit is a combination of these similarity coefficients as it has been illustrated in Fig.~\ref{fig:corruptedvectorv}. Therefore, Step~2.2, i.e. estimating the target and undesired similarity coefficients individually is a necessary step to obtain information about all similarity coefficients. A more detailed explanation of similarity coefficients' estimation is provided in the following. 

\vspace{3mm}
\paragraph{\hspace{-0.5mm}Step 2.2:\hspace{-0.5mm} Estimating Matrix of Similarity Coefficients}$\hspace{5mm}$\label{sec:Estimatingv}\\
\vspace{1mm}
    $\hspace{-1.3mm}$\textit{b.1) Step 2.2.1: Estimate Target Similarity Coefficients}\\
    Suppose that $\dimN_{r_\indexi}$ denotes the size of the $\indexi$th linear segment from a candidate size vector $\mathbf{n}_{r}$, as defined in Eq.~\eqref{eq:computepossibleblocksize}. Further, let ${\mathbf{v}^{(r)}\in\mathbb{R}^{(\dimN\shortminus\dimN_{\mathrm{I}})}}$ denote the target vector $\mathbf{v}$ associated with ${\mathbf{n}_{r}}$ defined by\vspace{-1mm}
\begin{equation}\label{eq:computetargetvectorvforvdddot}
v_{\indexm}^{(r)}=\sum_{\indexn=\indexm}^{u_{r_{\indexi}}}\dddot{l}_{\indexm,\indexn}\hspace{3mm}\mathrm{s.t.}\hspace{3mm}
\begin{aligned}
 &\ell_{r_{\indexi}}\leq\indexm\leq u_{r_{\indexi}}\\
 &\indexi\hspace{-0.7mm}=\hspace{-0.7mm}1,\myydots,\blocknum_{\mathrm{cand}}
 \end{aligned},
\end{equation}
where the $\indexm$th and $(\indexm,\indexn)$th components of ${\mathbf{v}^{(r)}}$ and 
${\dddot{\mathbf{L}}}$ are denoted, respectively, by $v_{\indexm}^{(r)}$ and $\dddot{l}_{\indexm,\indexn}$, ${\ell_{r_{1}}=1}$, ${u_{r_{1}}\hspace{-0.7mm}=\hspace{-0.7mm}\dimN_{r_1}}$, ${\ell_{r_{\indexi}}\hspace{-0.7mm}=\hspace{-0.7mm}\sum_{\indexk=1}^{\indexi-1}\dimN_{r_\indexk}+1}$ and ${u_{r_{\indexi}}\hspace{-0.7mm}=\hspace{-0.7mm}\sum_{\indexk=1}^{\indexi}\dimN_{r_\indexk}}$ for $\indexi=2,\myydots,\blocknum_{\mathrm{cand}}$.

After computing ${\mathbf{v}^{(r)}}$ using Eq.~\eqref{eq:computetargetvectorvforvdddot}, with Definition~\ref{def:targetvectorv}, we model it as a $\blocknum$-piece linear function of the target similarity coefficients. The model parameters are estimated in the FRS-BDR algorithm by applying the algorithm from \cite{PlrPC} that determines a plane-based piece-wise linear fit. In more details, for every linear segment $\indexi=1,\myydots,\blocknum_{\mathrm{cand}}$ associated with $\blocknum_{\mathrm{cand}}$, the method 
first estimates the parameters of the linear fit. Then, 
it estimates the target similarity coefficients $w_\indexi,\myydots,w_{\blocknum_{\mathrm{cand}}}$ based on the slope of piece-wise linear fit estimates. A step-by-step detailed description of the plane-based piece-wise linear fit algorithm to determine ${\mathbf{v}^{(r)}}$ is given in Section~IX.A of the supplementary material.

\vspace{2mm}
\textit{b.2) Step 2.2.2: Estimate Undesired Similarity Coefficients }\\
In this step, the shifted vectors of ${\mathbf{v}^{(r)}}$ are computed as follows
\begin{equation}\label{eq:computeshiftedvectorv}
\dddot{\mathbf{v}}_{\mathrm{s}_{\indexi,\indexj}}^{(r)}=\mathbf{v}^{(r)}_{\indexi}+\dddot{\mathbf{v}}^{(r)}_{\indexi,\indexj},\hspace{5mm}
\begin{aligned}
 &\indexi=2,\myydots,\blocknum_{\mathrm{cand}},\\
 &\indexj=1,\myydots,\indexi-1
 \end{aligned}\vspace{-0.5mm}
\end{equation}
where ${\dddot{\mathbf{v}}^{(r)}_{\indexi,\indexj}\in\mathbb{R}^{\dimN_{r_i}}}$ denotes the vector of increase, associated with the undesired group similarity between block $\indexi$ and $\indexj$, and $\dddot{\mathbf{v}}_{\mathrm{s}_{\indexi,\indexj}}^{(r)}$ is the associated shifted target vector.\footnote{For details, see Section~IX.B of the supplementary material.} Then, combining the results from Eq.~\eqref{eq:modellingvectorv}, Eq.~\eqref{eq:computetargetvectorvforvdddot} and Eq.~\eqref{eq:computeshiftedvectorv},  the undesired similarity coefficients between different blocks can be estimated as
\begin{equation}\label{eq:estimateundesiredsimcoeff}
\hat{w}_{\indexi,\indexj}^{(r)}=\frac{\mathrm{med}(\dddot{\mathbf{v}}^{(r)}_{\mathrm{s}_{\indexi,\indexj}}-\hat{\mathbf{v}}^{(r)}_{\indexi})}{\dimN_{r_\indexj}}\hspace{5mm}
\begin{aligned}
 &\indexi=2,\myydots,\blocknum_{\mathrm{cand}}\\
 &\indexj=1,\myydots,\indexi-1
 \end{aligned},
\end{equation}
where $\mathrm{med}(.)$ denotes the median operator, $\dimN_{r_\indexj}$ is defined in Eq.~\eqref{eq:computepossibleblocksize}, and $\hat{w}_{\indexi,\indexj}^{(r)}$ is the undesired similarity coefficient estimate between $\indexi$ and $\indexj$. 

\textbf{Remark 2.} Alternative to using the median operator as an estimator in Eq.~\eqref{eq:estimateundesiredsimcoeff}, one could consider using the sample mean estimator based on the theory in Section~\ref{sec:EigAnalysisandOutlierEffects} and~\ref{sec:SimpLaplacianAnalysisandOutlierEffects}. However, for the sample mean, a single outlying component has an unbounded effect on estimating undesired similarity coefficient, while the median operator provide robustness with the highest possible breakdown value of $50\%$ (for a detailed discussion about robustness comparisons, see Section~1.3 in \cite{Robuststatistics}). This property of the median provides robustness even in real-world cases where our theoretical assumptions are not fully fulfilled.

\begin{figure}[!tbp]
  \centering
\includegraphics[trim={0mm 0mm 0mm 2.3mm},clip,width=\linewidth]{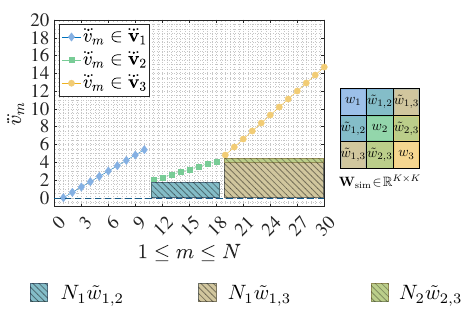}
\caption{Examplary plot of $\dddot{\mathbf{v}}$ and $\mathbf{W}_{\mathrm{sim}}$ with ${\blocknum_\mathrm{cand}\hspace{-0.7mm}=\hspace{-0.7mm} \blocknum}$, ${\mathbf{n}=[10,8,12]^{\top}\hspace{-0.7mm}\in\hspace{-0.7mm}\mathbb{R}^{\blocknum}}$, ${\mathrm{diag}(\mathbf{W}_{\mathrm{sim}})=[0.6,0.3,0.9]^{\top}\hspace{-0.7mm}\in\hspace{-0.7mm}\mathbb{R}^{\blocknum}}$, ${\tilde{w}_{1,2}=0.2}$, ${\tilde{w}_{1,3}=0.4}$, amd ${\tilde{w}_{2,3}=0.1}$.}
\label{fig:ExamplaryPlotofVectorvEstimation}
\end{figure}

To clarify Steps~2.2.1 and 2.2.2, an example with ${\blocknum_{\mathrm{cand}}=\blocknum}$ illustrating the computation of vector $\dddot{\mathbf{v}}$ and a matrix $\mathbf{W}_{\mathrm{sim}}\in\mathbb{R}^{\blocknum\times\blocknum}$ is shown in Fig~\ref{fig:ExamplaryPlotofVectorvEstimation}. As can be seen, the target similarity coefficients, which are the diagonal elements of $\mathbf{W}_{\mathrm{sim}}$, i.e., ${{\mathrm{diag}(\mathbf{W}_{\mathrm{sim}})=[w_1,w_2,\myydots,w_{\blocknum}]}^{\top}\in\mathbb{R}^{\blocknum}}$, represent an estimate of the slopes of the ${\blocknum_{\mathrm{cand}}=\blocknum}$ linear segments in $\dddot{\mathbf{v}}$. Further, off-diagonal elements of $\mathbf{W}_{\mathrm{sim}}$ represent undesired similarity coefficients between different blocks and are calculated by computing the undesired shifts that have been highlighted as shaded areas in Fig~\ref{fig:ExamplaryPlotofVectorvEstimation}.
\\
\textit{b.3) Step 2.3: Estimating vector $\dddot{\mathbf{v}}$ and $\mathbf{W}_{\mathrm{sim}}$ }\\
From the computed estimates ${\hat{\mathbf{W}}_{\mathrm{sim}}^{(r)}}\hspace{-0.5mm}\in\hspace{-0.5mm}\mathbb{R}^{\blocknum_{\mathrm{cand}}\times\blocknum_{\mathrm{cand}}}$ and ${\hat{\mathbf{v}}^{(r)}\hspace{-0.5mm}\in\hspace{-0.5mm}\mathbb{R}^{(\dimN\shortminus\dimN_{\mathrm{I}})}}$, the vector 
$\widehat{\dddot{\mathbf{v}}}_{\indexi}^{(r)}$ is computed by plugging in the associated intermediate estimates for all ${r\hspace{-0.7mm}=\hspace{-0.7mm}1,\myydots,\zeta}$ and ${\blocknum_{\mathrm{cand}}\hspace{-0.7mm}=\hspace{-0.7mm}\blocknum_{\mathrm{min}},\myydots,\blocknum_{\mathrm{max}}}$ into Eq.~\eqref{eq:modellingvectorv} and determining the final estimate as
\begin{equation}\label{eq:vectorvestimation}
\widehat{\dddot{\mathbf{v}}}=\underset{\mathbf{n}_r\in\mathbf{N}^{(\blocknum_{\mathrm{cand}})}}{\mathrm{argmin}}\|\dddot{\mathbf{v}}-\widehat{\dddot{\mathbf{v}}}^{(r)}\|_2
\end{equation}
where ${\forall\hat{w}^{(r)}_i\in\mathrm{diag}(\hat{\mathbf{W}}_{\mathrm{sim}}^{(r)})}$, ${\hat{w}^{(r)}_i>\hat{w}^{(r)}_{\indexi,\indexj}}$ holds for ${\indexi=1,\myydots,\blocknum_{\mathrm{cand}}}$,
${\indexj=1,\myydots,\blocknum_{\mathrm{cand}}}$ and ${\indexi\neq\indexj}$. 

Since the target block diagonal model with internally dense externally disjoint clusters represents the optimum level of sparsity, the closeness of the estimate of $\dddot{\mathbf{v}}$ to the target piece-wise linear function directly provides information of how well the algorithm was able to remove the undesired edges and therewith determine the sparsity level. In particular, the estimate of vector $\dddot{\mathbf{v}}$ provides fundamental information about the number of blocks, the number of elements for every block, desired and undesired similarity coefficients associated with each block that have been collected in the matrix $\mathbf{W}_{\mathrm{sim}}$. To design a BDR that provides a good balance with internally dense externally sparse clusters, the desired similarity coefficients, the proposed strategy preserves the similarity coefficients corresponding to the diagonal entries of $\mathbf{W}_{\mathrm{sim}}$ while removing that of undesired similarity coefficients corresponding to the off-diagonal entries of $\mathbf{W}_{\mathrm{sim}}$.\footnote{For examples that analyze the mismatch between the target and estimated BD structure, see Appendix~E.2 of the accompanying material.}

The proposed FRS-BDR is summarized in Algorithm~\ref{alg:FRS-BDR}. {The codes are provided at: \url{https://github.com/A-Tastan/FRS-BDR}}

\begin{algorithm}[tbp!]\small
\setstretch{1}
\DontPrintSemicolon
\KwIn{ 
$\hspace{-0.5mm}\mathbf{X}\hspace{-0.5mm}\in\hspace{-0.5mm}\mathbb{R}^{\dimM\times\dimN}$, $\blocknum_{\mathrm{min}}$, $\blocknum_{\mathrm{max}}$, $N_{c_\mathrm{max}}$, $\dimN_{\mathrm{min}}$(opt.)\hspace{-1cm}}
\vspace{0.2mm}
Compute $\mathbf{W}\in\mathbb{R}^{\dimN\times \dimN}$ i.e. $\mathbf{W}=\mathbf{X}^{\top}\mathbf{X}$ for ${\forall\mathbf{x}_{\indexm}\in\mathbf{X}, \|\mathbf{x}_{\indexm}\|\hspace{-1mm}=\hspace{-1mm}1}\hspace{-1cm}$\\
\vspace{0.28mm}
\textbf{Step 1: Enhancing BD Structure}\\
\vspace{0.28mm}
\textit{Step 1.1: Type~I Outlier Removal}\\
\vspace{0.2mm}
Compute $\dot{\mathbf{W}}$, $\dot{\mathbf{D}}$ and ${\dot{\mathbf{L}}\hspace{-0.7mm}\in\hspace{-0.7mm}\mathbb{R}^{(\dimN-\dimN_{\mathrm{I}})\times(\dimN-\dimN_{\mathrm{I}})}}$ via Eq.~\eqref{eq:preprocessing}$\hspace{-5mm}$\\
\vspace{0.28mm}
\textit{Step 1.2: Similarity-based Block Diagonal Ordering}\\
\vspace{0.2mm}
Perform Algorithm~\ref{alg:sBDO} to achieve 
$\hat{\mathbf{b}}^{(s)}\in\mathbb{Z}_{+}^{(\dimN-\dimN_{\mathrm{I}})}$\\
\vspace{0.2mm}
Obtain $\ddot{\mathbf{W}}$, $\ddot{\mathbf{D}}$ and ${\ddot{\mathbf{L}}\hspace{-0.7mm}\in\hspace{-0.7mm}\mathbb{R}^{(\dimN-\dimN_{\mathrm{I}})\times(\dimN-\dimN_{\mathrm{I}})}}$ using $\hat{\mathbf{b}}^{(s)}\hspace{-3mm}$\\
\vspace{0.28mm}
\textit{Step 1.3 (opt.): Sparsity for Excessive Group Similarity$\hspace{-4mm}$}\\
\vspace{0.2mm}
Design ${\dddot{\mathbf{L}}\in\mathbb{R}^{(\dimN-\dimN_{\mathrm{I}})\times(\dimN-\dimN_{\mathrm{I}})}}$ for the desired method,$\hspace{-5mm}$\\i.e. Algorithm~3 or~4 of the accompanying material.\\
\vspace{0.2mm}
Compute ${\dddot{\mathbf{v}}\hspace{-0.7mm}\in\hspace{-0.7mm}\mathbb{R}^{(\dimN-\dimN_{\mathrm{I}})\times 1}}$ corresponding to ${\dddot{\mathbf{L}}}$ using Eq.~\eqref{eq:vectorvcomputation}$\hspace{-6mm}$\\(or alternatively ${\ddot{\mathbf{v}}\in\mathbb{R}^{(\dimN-\dimN_{\mathrm{I}})\times 1}}$ corresponding to ${\ddot{\mathbf{L}}}$)\\
\vspace{0.28mm}
\textbf{Step 2: Estimating Vector $\mathbf{v}$}\\
\vspace{0.2mm}
\SetInd{0.5em}{0.25em}
\For{$\blocknum_{\mathrm{cand}}=\blocknum_{\mathrm{min}},\myydots,\blocknum_{\mathrm{max}}$}{
\vspace{0.28mm}
\textit{Step 2.1: Computing Candidate Block Sizes}\\
\vspace{0.2mm}
Compute ${\mathbf{N}^{(\blocknum_{\mathrm{cand}})}\in\mathbb{Z}_{+}^{\zeta\times\blocknum_{\mathrm{cand}}}}$ using Eqs.~\eqref{eq:generatecandidates}-\eqref{eq:computepossibleblocksize}\\
\vspace{0.2mm}
\SetInd{0.5em}{0.25em}
\textit{Step 2.2: Estimating $\mathbf{W}_{\mathrm{sim}}$}\\
\For{$\mathbf{n}_{r}=\mathbf{n}_{1},\myydots,\mathbf{n}_{\zeta}$}{
\vspace{0.28mm}
\textit{Step 2.2.1: Estimating Target Similarity Coefficients$\hspace{-5mm}$}\\
\vspace{0.2mm}
Compute ${\mathbf{v}^{(r)}\in\mathbb{R}^{(\dimN-\dimN_{\mathrm{I}})}}$ using Eq.~\eqref{eq:computetargetvectorvforvdddot}\\
\vspace{0.2mm}
\For{$\indexi=1,\myydots,\blocknum_{\mathrm{cand}}$}{
Calculate $\boldsymbol{\Sigma}_{\indexi}^{(r)}\hspace{-0.7mm}\in\hspace{-0.7mm}\mathbb{R}^{2\times2}$ and $\boldsymbol{\mu}_{\indexi}^{(r)}\hspace{-0.7mm}\in\hspace{-0.7mm}\mathbb{R}^{2}$ for $\boldsymbol{\Upsilon}_{\indexi}^{(r)}\hspace{-5mm}$\\
\vspace{0.2mm}
Find $\hat{\boldsymbol{\vartheta}}^{(r)}_{\indexi}\hspace{-0.7mm}\in\hspace{-0.7mm}\mathbb{R}^{2}$ and $\hat{\mathrm{b}}^{(r)}_{\indexi}\in\mathbb{R}$\\
\vspace{0.2mm}
Find ${\hat{\mathbf{v}}_{\indexi}^{(r)}\hspace{-0.9mm}\in\hspace{-0.7mm}\mathbb{R}^{\dimN_{r_\indexi}}}$ and compute $\hat{w}_{\indexi}$\\
\vspace{0.2mm}
}
Form ${\mathrm{diag}(\hat{\mathbf{W}}^{(r)}_{\mathrm{sim}})\hspace{-0.7mm}=\hspace{-0.7mm}[\hat{w}^{(r)}_{1},\hat{w}^{(r)}_{2},\myydots,\hat{w}^{(r)}_{\blocknum_{\mathrm{cand}}}]^{\top}\hspace{-1mm}\in\hspace{-0.7mm}\mathbb{R}^{\blocknum_{\mathrm{cand}}}}\hspace{-1.2cm}$ 
\\\vspace{0.2mm}
and $\hat{\mathbf{v}}^{(r)}\hspace{-1mm}=\hspace{-0.7mm}[(\hat{\mathbf{v}}^{(r)}_{1})^{\top}\hspace{-0.7mm},(\hat{\mathbf{v}}^{(r)}_{2})^{\top}\hspace{-0.7mm},\myydots,(\hat{\mathbf{v}}^{(r)}_{\blocknum_{\mathrm{cand}}})^{\top}]^{\top}\hspace{-1.5mm}\in\hspace{-0.7mm}\mathbb{R}^{(\dimN-\dimN_{\mathrm{I}})}\hspace{-3cm}$\\\vspace{0.28mm}
\textit{Step 2.2.2: Estimating Undesired Similarity Coefficients$\hspace{-1cm}$}\\
\vspace{0.2mm}
\For{$\indexi=2,\myydots,\blocknum_{\mathrm{cand}}$}{
\For{$\indexj=1,\myydots,\indexi-1$}{
Compute $\dddot{\mathbf{v}}_{\mathrm{s}_{\indexi,\indexj}}^{(r)}\in\mathbb{R}^{(\dimN-\dimN_{\mathrm{I}})}$ using Eqs.~\eqref{eq:computeshiftedvectorv}$\hspace{-5mm}$\\
\vspace{0.2mm}
Compute $\hat{w}_{\indexi,\indexj}^{(r)}$ using Eq.~\eqref{eq:estimateundesiredsimcoeff} and stack $\hat{\mathbf{W}}^{(r)}_{\mathrm{sim}}\hspace{-5mm}$\\
\vspace{0.2mm}
}}
Estimate $\dddot{\mathbf{v}}^{(r)}$ using Eq.~\eqref{eq:modellingvectorv}\\
\vspace{0.2mm}
Update $\widehat{\dddot{\mathbf{v}}}$ based on Eq.~\eqref{eq:vectorvestimation}\\
\vspace{0.2mm}
}
}
\caption{FRS-BDR}
\label{alg:FRS-BDR}
\KwOut{ $\widehat{\dddot{\mathbf{v}}}$, $\hat{\mathbf{W}}_{\mathrm{sim}}$, $\hat{\mathbf{n}}$}
\vspace{-0.5mm}
\end{algorithm}

\vspace{-3mm}

\subsection{Computational Analysis of FRS-BDR}\label{sec:ComputationalAnalysis}
A comprehensive computational analysis is computed in Section~X of the supplementary material by determining the number of fladd, flmlt, fldiv and flam. The Landau’s big $O$ symbol is used for the cases when the complexity is not specified as above operations. For a detailed information, see \citeform[59]-\citeform[60].  Our analysis showed that the complexity of FRS-BDR strongly depends on the initial structure of the affinity matrix and the number of blocks $\blocknum$.  In addition to the numeric analysis, the complexity is analyzed experimentally in the following sections.

\vspace{-2.25mm}
\section{Experimental Results}\label{sec:ExperimentalResults}
This section benchmarks the proposed FRS-BDR method in a broad range of real data experiments, including cluster enumeration and handwritten digit, object and face clustering.

\paragraph*{\textit{Data Sets}} The performance is analyzed using the well-known data sets for handwritten digit clustering \cite{MNIST,USPS}, for object clustering \cite{COIL20}, for face clustering \citeform[63]-\citeform[65] and for cluster enumeration \citeform[66]-\citeform[70]. The detailed information about the data sets is given in the following sections. 
\paragraph*{\textit{Baselines}} For the task of subspace clustering, FRS-BDR is benchmarked against seven state-of-the-art BDR approaches \citeform[6]-\citeform[10], two low-rank representation methods \citeform[17]-\citeform[18], a sparse representation method (SSC) \cite{SSC}, a robust principal component analysis method (FRPCAG) \cite{FRPCAG}, a robust spectral clustering method (RSC) \cite{RSC} and the initial affinity matrix that is defined by $\mathbf{W}=\mathbf{X}^{\top}\mathbf{X}$. For cluster enumeration\footnote{For the numerical cluster enumeration results, see Appendix~F.4.4.2 of the accompanying material.}, the method is benchmarked against seven popular community detection methods, i.e. \citeform[73]-\citeform[78] and our previously proposed method that is called SPARCODE in \cite{Sparcode}.

\paragraph*{\textit{Parameter Setting}} In all experiments, the parameters are optimally tuned for the competitor approaches, while FRS-BDR is computed with the default parameters that are detailed in Section~XI of the supplementary material.
\paragraph*{\textit{Evaluation Metrics}} The computation time $(t)$ and average clustering accuracy $(\Bar{c}_{\mathrm{acc}})$ are used for the subspace clustering performance analysis. In cluster enumeration, the empirical probability of detection $(p_{\mathrm{det}})$, modularity ($\mathrm{mod}$) and conductance $(\mathrm{cond})$ are used in addition to $t$. The evaluation metrics are comprehensively explained in Section~XI of the supplementary material.
\vspace{-2.5mm}
\subsection{Handwritten Digit Clustering}\label{sec:clusterhandwrittendigit}
The effectiveness of FRS-BDR in handwritten digit clustering is shown based on the following popular real-world data sets:
\paragraph*{\textit{MNIST Data Set}} The data base includes 60,000 training and 10,000 test images corresponding to 10 digits. For a varying number of subjects $\blocknum=\{2,3,5,8,10\}$, the data 
matrix $\mathbf{X}$ is generated using 100 randomly selected images from the test set for every subject where the images are used as feature vectors and normalized. As in \cite{BDRB}, $\mathbf{X}$ of size $784\times100\blocknum$ is produced for the images of size $28\times28$.
\paragraph*{\textit{USPS Data Set}} 7291 training and 2007 test images of size $16\times16$ are contained in the data set. The data matrix $\mathbf{X}$ is computed by following the same procedure, except for using 50 randomly selected images from the test set for every subject. As a result, for the images of size $16\times16$, the data matrix $\mathbf{X}$ of size $256\times50\blocknum$ corresponding to a number of subjects $\blocknum=\{2,3,5,8,10\}$, is obtained.

In contrast to object and face applications that we will detail in the following sections, the data matrix $\mathbf{X}$ of high dimensional feature vectors is directly used in initial affinity matrix design. The initial affinity matrix, i.e. $\mathbf{W}=\mathbf{X}^{\top}\mathbf{X}$ is used as an input to BDR approaches \citeform[6]-\citeform[10], low-rank representation methods \citeform[17]-\citeform[18] and the sparse representation method in \cite{SSC} to design affinity matrices in a desired form. Then, spectral clustering\footnote{For the details about spectral clustering, see Section~XI of the supplementary material.} is applied to the resulting affinity matrices of different methods. Different from affinity matrix construction methods, FRPCAG \cite{FRPCAG} and RSC \cite{RSC} algorithms use the data matrix $\mathbf{X}$ as an input. These methods determine the affinity matrices based on their default construction where the number of neighbors is defined by gradually decreasing the number of all neighbors until the methods do not fail.

\begin{figure*}[!tbp]
  \centering
\includegraphics[trim={0mm 0mm 0.2mm 0mm},clip,width=\linewidth]{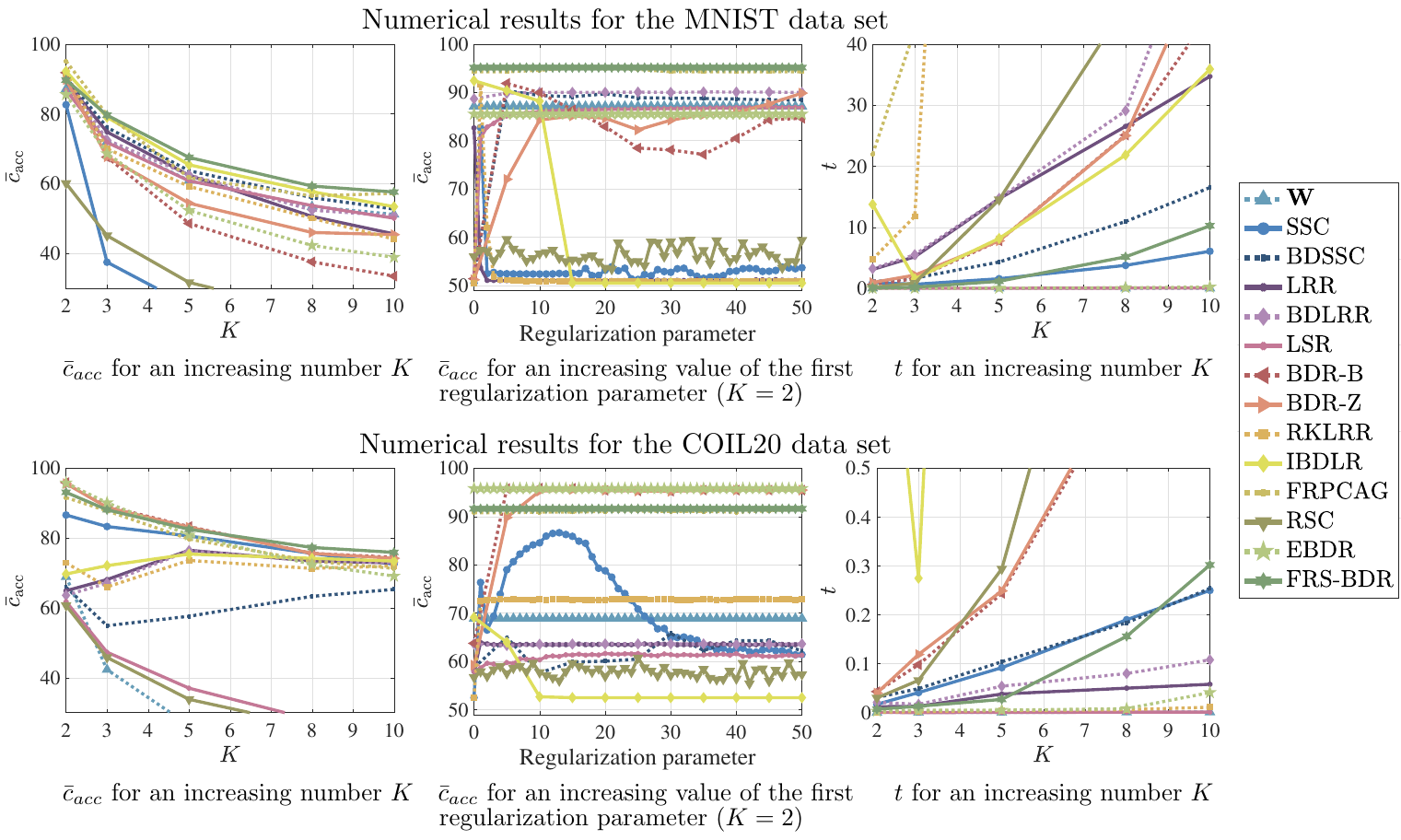}
\caption{Numerical results for the COIL20 and MNIST data sets. The regularization parameters of the competing methods are tuned for optimal performance in all settings while the proposed method determines the parameters using Algorithms~\ref{alg:sBDO} and~\ref{alg:FRS-BDR}. In  the regularization parameter performance analysis, for all competing methods that use two parameters, the second one is tuned optimally while varying the first parameter.}
\label{fig:NumericalAnalysisMNISTandCOIL20}
\end{figure*}

An example of digit clustering results is shown in Fig.~\ref{fig:NumericalAnalysisMNISTandCOIL20} for the MNIST data base. A broad set of analyses including MNIST and USPS data bases is provided in Appendix~F.4.1 of the accompanying material. Even though the performance of SSC \cite{SSC}, BDSSC \cite{BDSSC}, BDLRR \cite{BDSSC}, BDR-B \cite{BDRB}, BDR-Z \cite{BDRB}, IBDLR \cite{IBDLR}, LSR \cite{LSR}, LRR \cite{LRR}, RKLRR \cite{RKLRR}, FRPCAG \cite{FRPCAG} and RSC \cite{RSC} is reported for an optimal tuning of the parameters, which is not feasible in practice, the FRS-BDR achieves the highest clustering accuracy results in almost all cases. Further, the regularization parameter effect analysis in Fig.~\ref{fig:NumericalAnalysisMNISTandCOIL20} shows that BDR-B and BDR-Z performances are sensitive to the choice of the first regularization parameter, even when tuning the second one optimally. Based on the computation time analysis, the main drawback of competitor approaches is that they are sensitive to the dimension of the feature space whereas FRS-BDR is an efficient algorithm for the data sets including high dimensional feature vectors.

To quantify the performance of different BDR approaches in terms of the sparsity, an additional 
set of experiments analyzing modularity (mod) and conductance (cond) scores, which are the commonly used quality metrics for this analysis, are introduced in Appendix~F.4 of the accompanying material. The numerical analysis demonstrates that the proposed FRS-BDR algorithm provides a ``good balance'' in sparsity with large-valued modularity scores and small-valued conductance scores in most
of the cases.\footnote{The modularity and conductance performance of the proposed FRS-BDR algorithm could be further improved by enforcing the estimated blocks to be distinct but such a step may result in a performance degradation in clustering accuracy which is more important in these clustering applications.} The analysis confirms the results of the clustering accuracy performance analysis that FRS-BDR algorithm shows a good performance compared to the optimally tuned BDR approaches while providing considerably better performance than optimally tuned low-rank representation methods. Different from structuring all clusters based on a single determined sparsity parameter (which may be
difficult to tune in practice), our approach allows for treating every block differently, depending on the occurrence of the outliers’ effect within each block and this makes the proposed FRS-BDR method advantageous in terms of balancing the sparsity.

\vspace{-2.5mm}
\subsection{Object Clustering}\vspace{-0.5mm}
This section introduces a set of experiments that are performed on the COIL20 \cite{COIL20} data base of 20 objects. In COIL20, each object has 72 images where the images are taken by rotating the object on a turntable in five degree intervals. In our experiments, the processed COIL20 data set in \cite{DengCaiCOIL20} containing images of size $32\times32$ pixels is used. Then, the data set $\mathbf{X}$ of size $1024\times400$ is generated by selecting 20 images randomly for every object. The feature space is reduced to 10 based on the PCA performance, which is provided in Appendix~F.4.2.1 of the accompanying material.

As in \cite{BDRB}, a performance analysis of every application is conducted for an increasing value of $\blocknum$, i.e., ${\blocknum=\{2,3,5,8,10\}}$ using $100$ randomly selected subject combinations. To obtain the affinity matrices for the competing methods, the regularization parameters are manually tuned on a grid of 50 values. Finally, spectral clustering \cite{LuxburgSC} is applied and the results in Fig.~\ref{fig:NumericalAnalysisMNISTandCOIL20}, for an increasing value of $\blocknum$, are obtained analogously to \cite{BDRB} (see Appendix~F.4.2 of the accompanying material for further details). The average clustering accuracy $\Bar{c}_{\mathrm{acc}}$ results show that FRS-BDR performs best while EBDR is an efficient method for small values of $\blocknum$. In terms of $t$, the main competitors BDR-B and BDR-Z show poor performance whereas FRS-BDR performs relatively good even for large values of $\blocknum$.
This computational advantage of the proposed method can be explained with its simple nature, i.e. finding a piece-wise linear function robustly, which is easy to solve in comparison to analyzing the graph structure in a matrix space as in the existing BDR methods.

\begin{figure*}[!tbp]
  \centering
\includegraphics[trim={0mm 0mm 0.2mm 0mm},clip,width=\linewidth]{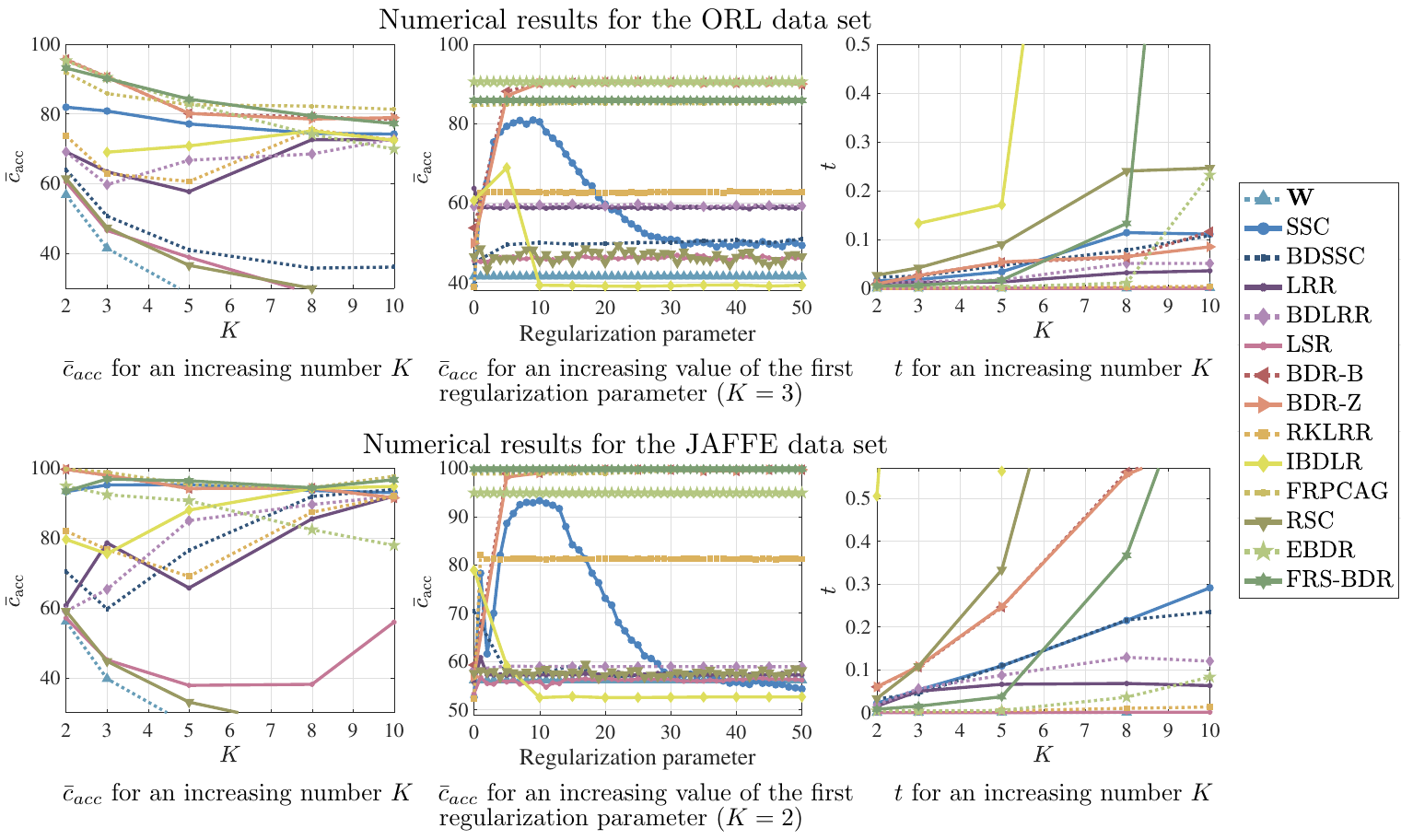}
\caption{Numerical results for the ORL and JAFFE data sets. The regularization parameters of the competing methods are tuned for optimal performance in all settings while the proposed method determines the parameters using Algorithms~\ref{alg:sBDO} and~\ref{alg:FRS-BDR}. In  the regularization parameter performance analysis, for all competing methods that use two parameters, the second one is tuned optimally while varying the first parameter.}
\label{fig:NumericalAnalysisORLandJAFFE}
\end{figure*}

The BDR-B and BDR-Z methods show poor performance for small-valued regularization parameters even though the second regularization parameter is optimally tuned. An important point is that these approaches reach their best results lately in comparison to experiments on face clustering data sets that are explained in the following section.

\vspace{-2mm}
\subsection{Face Clustering}
In this section, the subspace clustering performances of different methods are benchmarked in terms of their $\Bar{c}_{\mathrm{acc}}$ and $t$ by using the following application details:
\paragraph*{\textit{ORL Data Set}} The data set includes 10 images of 40 different subjects that are taken at different times by varying the lighting, facial expressions and details. As in \cite{IBDLR}, we resize all images to $32\times32$ to obtain a data matrix $\mathbf{X}$ of size $1024\times400$ using normalized features. The feature space dimension is reduced to nine using Principal Component Analysis (PCA) in order to reduce the computation time\footnote{For the PCA analysis of the ORL data set, see Appendix~F.4.3.1 of the accompanying material.}.
\paragraph*{\textit{JAFFE Data Set}} The JAFFE data set comprises 213 images of seven facial expressions from 10 Japanese female models. As in \cite{IBDLR}, the images are resized to $64\times64$ pixels and the data set $\mathbf{X}$ of size $4096\times213$ is computed using resized images as normalized feature vectors before applying PCA to reduce the dimensionality to 14 features\footnote{For the PCA analysis of the JAFFE data set, see Appendix~F.4.3.2 of the accompanying material.}.
\paragraph*{\textit{Yale Data Set}} 165 grayscale images of 15 different individuals. For every subject, the data set contains 11 images that capture different facial expressions. The data matrix $\mathbf{X}$ of size $1024\times165$ is constructed as in the ORL Data Set\footnote{For the PCA analysis of the Yale data set, see Appendix~F.4.3.3 of the accompanying material}.


\begin{table*}[tbp!]
\centering
\resizebox{\linewidth}{!}{%
\begin{tabular}{p{4cm}M{1.5cm}M{1.5cm}M{1.5cm}M{1.5cm}M{1.5cm}M{1.5cm}M{1.5cm}M{1.5cm}M{1.5cm}M{1.5cm}M{1.5cm}M{1.5cm}M{1.5cm}M{1.8cm}}
\hline\hline
\\[-3mm]
 & \multicolumn{14}{c}{Subspace Clustering Performances  for Different Block Diagonal Representation Methods}\\\\[-3mm]
\cline{2-15}\\[-3mm]
& &\multicolumn{11}{c}{Minimum-Maximum Clustering Accuracy $(c_\mathrm{accmin}-c_\mathrm{accmax})$ for Different Regularization Parameters} &  &\\\\[-3mm]
\cline{3-13}\\[-3mm]
Data Set & $\mathbf{W}$ & SSC & BD-SSC & LRR & BD-LRR & LSR & BDR-B & BDR-Z & RKLRR & IBDLR &FRPCAG & RSC & EBDR & \textbf{FRS-BDR} \\
\midrule
Breast Cancer\cite{BreastCancer},& 88.2 & 51.0-74.7 & 50.3-88.2 & 54.3-90.3 & 88.0-90.0 & 73.5-88.2 & 62.4-90.0 & 52.9-90.2 & 62.6-91.7 & 60.3-90.0 &60.5-88.2&50.1-58.5& 85.2 & 90.1\\
Ceramic \cite{Ceramic},&98.9 & 51.1-98.9 & 51.1-100 & 95.5-98.9 & 95.5-98.9 & 54.5-98.9 & 51.1-100 & 51.1-98.9 & 51.1-95.5 & 51.1-98.9 &50.0-100&50.0-69.3& 98.9 & 98.9 \\
Vertebral Column \cite{VertebralColumn},&73.2 & 50.0-77.7 & 50.3-74.8 & 53.9-72.6 & 72.6-72.6 & 62.6-75.8 & 67.4-76.8 & 71.9-76.8 & 67.4-71.3 & 67.4-76.1 &51.0-75.8&50.0-69.7& 74.8 & 75.8 \\
Iris \cite{Iris},&78.0& 34.7-82.7 & 34.0-83.3 & 38.7-80.7 & 80.0-98.0 & 78.0-82.7 & 34.0-96.7 & 65.3-96.7 & 34.0-80.0 & 34.7-84.0 &34.0-89.3&35.3-50.0& 98.0 & 96.7\\
Human Gait \cite{HumanGait},& 77.3 & 20.3-77.4 & 20.1-77.5 & 26.1-83.9 & 78.9-83.5 & 55.4-75.9 & 20.3-84.8 & 26.4-84.5 & 20.5-85.5 & 20.4-81.6 &50.8-77.8&22.1-26.0& 81.1 & 77.1 \\
Ovarian Cancer \cite{OvarianCancer},& 61.7 & 51.4-73.6 & 50.9-71.3 & 52.3-76.4 & 54.2-76.4 & 51.9-66.2 & 53.7-75.9 & 51.9-74.1 & 55.6-88.4 & 55.6-75.5 &77.8-89.3&50.0-69.4& 77.8 & 77.3\\
Person Identification \cite{PersonIdentification},& x & 33.7-96.8 & 31.6-95.7 & 49.7-94.7 & 71.1-94.7 & 33.2-64.2 & 31.6-96.3 & 59.4-95.7 & 34.2-94.1 & 33.7-95.7 &29.4-92.5&28.9-41.2& 97.3 & 96.8\\
Parkinson \cite{Parkinson},& 61.3 & 50.4-58.8 & 50.0-61.3 & 50.4-54.2 & 50.4-61.3 & 57.9-61.3 & 50.4-61.3 & 50.0-61.3 & 50.4-61.7 & 50.4-61.3 &50.0-67.5&50.4-72.1& 56.7 & 58.2\\
\hline\\[-3mm]
Average & 76.9 & 42.8-80.1 & 42.3-81.5 & 52.6-81.4 & 73.8-84.4 & 58.4-76.6 & 46.4-85.2 & 53.6-84.8 & 47.0-83.5 & 46.7-82.9 &50.4-85.1&42.1-57.0& 83.7 & 83.9\\
\hline\hline
\end{tabular}}
\caption{Subspace clustering performance of different block diagonal representation approaches on well-known clustering data sets. $\mathbf{W}$ represents the subspace clustering results that are obtained by using the initial affinity matrix $\mathbf{W}=\mathbf{X}^{\top}\mathbf{X}$ as an input to spectral clustering algorithm. The remaining columns show affinity matrix construction methods that are using $\mathbf{W}$ as input and performing spectral clustering on the sparse affinity matrix estimates. The performances are summarized in terms of $\bar{c}_\mathrm{acc}$ for parameter-free approaches including $\mathbf{W}$, EBDR and FRS-BDR while the remaining methods are shown for $c_\mathrm{accmin}-c_\mathrm{accmax}$.  `x' denotes the failed results due to the complex-valued eigenvectors.}
\label{tab:subspaceclusteringadditionaldatasetsaccuracy}
\end{table*}




After determining the number of PCA features, the same procedure as in object clustering is performed and the performance is reported for a different number of subjects $\blocknum\hspace{-0.7mm}=\hspace{-0.7mm}\{2,3,5,8,10\}$ in Fig.~\ref{fig:NumericalAnalysisORLandJAFFE}. For a detailed performance analysis, see Appendix~F.4.3 of the accompanying material.

The average clustering accuracy $\Bar{c}_{\mathrm{acc}}$ and computation time $t$ for the ORL and the JAFFE data sets are provided in Fig.~\ref{fig:NumericalAnalysisORLandJAFFE}.
Consistent with the previous experiments, FRS-BDR shows the best clustering accuracy performance among all approaches in almost all cases. In terms of $t$, FRS-BDR shows a reasonably good performance until the number of subjects reaches $\blocknum=8$. A reduction for a large value of $\blocknum$ can be obtained by adjusting $\dimN_{c_{\mathrm{max}}}$. Extensive further numerical experiments are reported in Appendices E.5.3.1, E.5.3.2, and E.5.3.3. of the accompanying material.

\vspace{-3mm}
\subsection{Subspace Clustering on Well-Known Clustering Data Sets}\label{sec:SubspaceClustering}\vspace{-1mm}
This section investigates the subspace clustering performance of different approaches in terms of their average clustering accuracy using the following popular clustering data sets: Breast Cancer Wisconsin (Breast Cancer) \cite{BreastCancer}, Chemical Composition of Ceramic (Ceramic) \cite{Ceramic}, Vertebral Column \cite{VertebralColumn}, Fisher’s iris (Iris) \cite{Iris}, Radar-based Human Gait (Human Gait) \cite{HumanGait}, Ovarian Cancer \cite{OvarianCancer}, Person Identification \cite{PersonIdentification} and Parkinson \cite{Parkinson}. To analyze subspace clustering performances on popular clustering data sets, subspace clustering is first performed on the initial affinity matrix that is defined by $\mathbf{W}=\mathbf{X}^{\top}\mathbf{X}$. Analogous to the handwritten digit clustering application in Section~\ref{sec:clusterhandwrittendigit}, the data matrix is used as an input to the FRPCAG \cite{FRPCAG} and RSC \cite{RSC} methods while state-of-the-art BDR methods use the initial affinity matrix that is defined by $\mathbf{W}=\mathbf{X}^{\top}\mathbf{X}$ as an input to design BD structured affinity matrices. Then, spectral clustering as detailed in Section~XI of the supplementary material is performed on the BD affinity matrix estimates. For the FRPCAG \cite{FRPCAG} and RSC \cite{RSC} methods, spectral clustering is performed based their eigenvector estimates. As in previous experiments, the competitor approaches' results are shown for optimally tuned parameters while the proposed FRS-BDR is performed with the default parameters.

The clustering accuracy performances of different block-diagonal representation approaches are detailed in terms of their average clustering accuracy in Table~\ref{tab:subspaceclusteringadditionaldatasetsaccuracy}. As can be seen from Table~\ref{tab:subspaceclusteringadditionaldatasetsaccuracy}, FRS-BDR provides a similar performance as the maximum clustering accuracy of its strongest 
competitors (BDR-B, BDR-Z, BD-LRR) while it outperforms all other block diagonal representation approaches. The method is also computationally efficient in comparison to most of the competitors based on the additional experiments that are given in Appendix~F.4.4 of the accompanying material.

\begin{figure}[!tbp]
  \centering
\includegraphics[trim={0mm 0mm 0mm 0mm},clip,width=\linewidth]{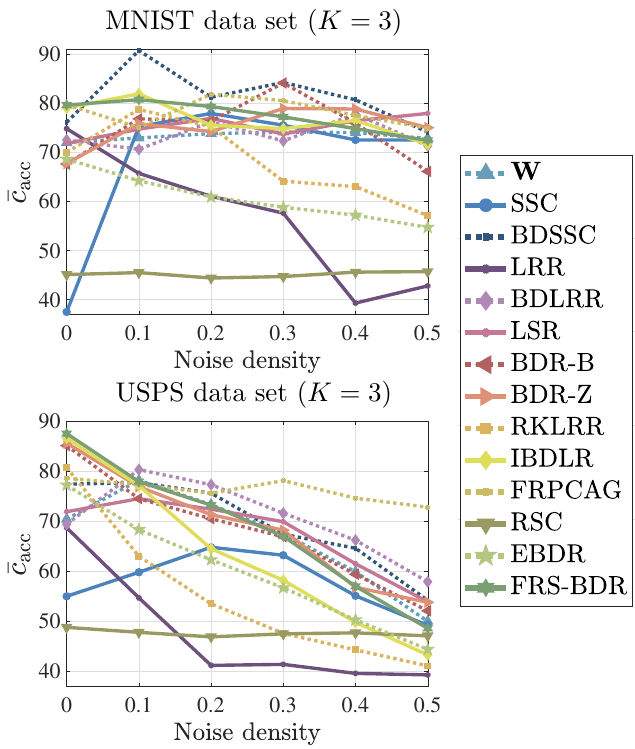}
\caption{Robustness analysis results for the MNIST and USPS data sets. The regularization parameters of the competing methods are tuned for optimal performance in all settings while the proposed method determines the parameters using Algorithms~\ref{alg:sBDO} and~\ref{alg:FRS-BDR}. $\bar{c}_\mathrm{acc}$ performances are shown for increasing density value of the salt and pepper noise.}
\label{fig:RobustnessanalysisMNISTandUSPS}
\end{figure}

\subsection{Robustness Analysis}
A further analysis evaluating robustness of the proposed FRS-BDR method in noisy scenarios with corruptions in data/feature space, is reported in this section. To analyze robustness, against outliers, digit samples from MNIST \cite{MNIST} and USPS \cite{USPS} data sets are corrupted with salt and pepper noise and Poisson noise. Object and face recognition data sets are not included to robustness analysis due to the performance degradation of the (non-robust) PCA that is part of the feature generation.

The robustness analysis results of different methods are shown in Fig.~\ref{fig:RobustnessanalysisMNISTandUSPS} for the MNIST and USPS data sets that are corrupted with salt and pepper noise for an increasing percentage of outlier contamination. As in previously analyzed scenarios, the proposed FRS-BDR shows relatively good performance compared to the optimally tuned approaches for both data sets. This is because the proposed method leverages the derived
theory on how an ideal block diagonal structure is disturbed by outliers and this allows to precisely
remove the effects of Type I and Type II outliers as well as the group similarity. Many block diagonal affinity
matrix construction methods that we compare against are not robust against outliers and it is well-known
that performance of non-robust methods can severely be degraded in presence of outliers \cite{Robuststatistics}. 

\section{Conclusion}\label{sec:Conclusion}
 A robust method to recover a block diagonal affinity matrix in challenging scenarios has been presented. The proposed Fast and Robust Sparsity-Aware Block Diagonal Representation (FRS-BDR) method jointly estimates cluster memberships and the number of blocks. It builds upon our presented theoretical results that describe the effect of different fundamental outlier types in cluster analysis, allowing a reformulation of the problem as a robust piece-wise linear fitting problem. Comprehensive experiments, including a variety of real-world applications demonstrate the effectiveness of FRS-BDR compared to optimally tuned benchmark methods in terms of clustering accuracy, computation time and cluster enumeration performance. {Since all codes are made available, the FRS-BDR method can also easily be benchmarked on other larger-scale data sets, e.g. \citeform[80]-\citeform[82].}

\section*{Acknowledgments}
The work of A. Taştan is supported by the Republic of Turkey Ministry of National Education. The work of M. Muma has been funded by the LOEWE initiative (Hesse, Germany) within the emergenCITY centre and is supported by the ERC Starting Grant ScReeningData (Project Number: 101042407).

\end{document}


\title{Supplementary Information: Fast and Robust Sparsity-Aware Block Diagonal Representation}

\author{Aylin~Ta{\c{s}}tan$^\ast$, Michael~Muma$^\dagger$,
        and~Abdelhak~M.~Zoubir$^\ddagger$
\thanks{\noindent$^\ast$The author was with the Signal Processing Group, Technische Universität
Darmstadt, Darmstadt, Germany and is now with the Pattern Recognition Group, University of Bern, Bern, Switzerland (e-mail:  a.tastan@spg.tu-darmstadt.de; aylin.tastan@unibe.ch).\protect\\
$^\dagger$The author is with the Robust Data Science Group, Technische Universität
Darmstadt, Darmstadt, Germany (e-mail: michael.muma@tu-darmstadt.de).\protect\\
$^\ddagger$The author is with the Signal Processing Group, Technische Universität
Darmstadt, Darmstadt, Germany (e-mail:  zoubir@spg.tu-darmstadt.de).
}}

\markboth{SUBMITTED TO IEEE TRANSACTIONS ON SIGNAL PROCESSING}%
{Shell \MakeLowercase{\textit{et al.}}: A Sample Article Using IEEEtran.cls for IEEE Journals}

\IEEEpubid{}

\maketitle
\section*{Structure}
\vspace{-1.5mm}
\label{sec:introduction}
\setlength{\parindent}{0pt}
The Supplementary Information for the paper Fast and Robust Sparsity-Aware Block Diagonal Representation (FRS-BDR) is organized as follows: In Section~\ref{sec:exsolutionpaths} examplary solution paths are shown for eigenvalue and simplified Laplacian matrix analysis, respectively. Section~\ref{sec:FRS-BDRAlgorithmDetails} comprises the additional information for FRS-BDR algorithm. A step-by-step analysis of computational complexity is presented in Section~\ref{sec:comcompanalysis}. Finally, the details about experimental setting are introduced in Section~\ref{sec:experimentalsetting}. 

A comprehensive supplementary file is given separately. The file is named by ``Accompanying Material: Fast and Robust Sparsity-Aware Block Diagonal Representation" and it is organized as follows:\\\vspace{-3.5mm}

\textit{Appendix A: The Generalized Eigen-decomposition based Eigenvalue Analysis}\\
The theorems (and corresponding corollaries) by reference to outlier effects on eigenvalues are proved based on the generalized eigen-decomposition and they are visually exemplified.\\\vspace{-3.5mm}

 \textit{Appendix B: The Standard Eigen-decomposition based Eigenvalue Analysis}\\
It includes analysis of outlier effects on eigenvalues theorems (and corresponding corollaries) based on the standard eigen-decomposition and their visual examples.\\\vspace{-3.5mm}

\textit{Appendix C: Simplified Laplacian Matrix Analysis and Outlier Effects}\\
The proofs associated with simplified Laplacian matrix analysis and outlier effects are presented with visual illustrations in Appendix C.\\\vspace{-3.5mm}

\textit{Appendix D: The Generalized Matrix Determinant Lemma}\\
The appendix explains the generalization of matrix determinant lemma.\\\vspace{-3.5mm}

\textit{Appendix E: Real-world Data Examples}\\
The real-world data examples that are analyzing the model mismatch are introduced.\\\vspace{-3.5mm}

\textit{Appendix F: Additional Information for FRS-BDR}\\
It contains a step by step detailed visual summary of FRS-BDR, an analysis for sparse Laplacian matrix design, additional algorithms for sparse Laplacian matrix design,  and additional experimental results for all applications, respectively.

\vspace{-2.5mm}
\setcounter{section}{7}
\section{Examplary Solution Paths} \label{sec:exsolutionpaths}
\subsection{Eigenvalue Analysis and Outlier Effects}
This  section provides a high-level description of the solution path for the eigenvalue analysis and outlier effects detailed in Theorem~1 and 2 in Section~III of the paper Fast and Robust Sparsity-Aware Block Diagonal Representation (FRS-BDR). The below diagram shows the steps involved from the input, which is an affinity matrix subject to outliers or group similarity, to the output, which is the set of eigenvalues. The detailed solutions are given in Appendix~A of the accompanying material in accompanying material.
\begin{figure}[h!]\vspace{-2mm}
  \centering
  \captionsetup{justification=centering}
\includegraphics[trim={0cm 0cm 0cm 0cm},clip,width=6.3cm]{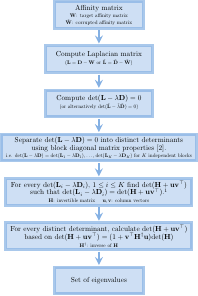}
  \label{fig:FlowChart_EigenvalueAnalysis}
\vspace{-6mm}
\end{figure}
\blfootnote{$^1$If the matrix $\mathrm{det}(\mathbf{L}_i-\lambda\mathbf{D}_i)$ can not be represented in terms of $\mathbf{H}$, $\mathbf{u}$ and $\mathbf{v}$, the generalized matrix determinant lemma can be used. For a detailed information, see Appendix~D in accompanying material.}
\subsection{Simplified Laplacian Matrix Analysis and Outlier Effects}
This  section provides a high-level description of the solution path to obtain vector $\mathbf{v}$ for the Theorems 3-4 and Corollary~4.1 in Section~IV of the paper Fast and Robust Sparsity-Aware Block Diagonal Representation. The detailed solutions are given in Appendix~C of the accompanying material in accompanying material.

\begin{figure}[h!]
  \centering
  \captionsetup{justification=centering}
\includegraphics[trim={0cm 0cm 0cm 0cm},clip,width=6.5cm]{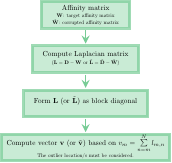}
  \label{fig:FlowChart_VectorvAnalysis}
\vspace{-5mm}
\end{figure}

\section{FRS-BDR Algorithm Details}\label{sec:FRS-BDRAlgorithmDetails}\vspace{-1mm}
\subsection{Plane-based Piece-wise Linear Fit of $\mathbf{v}^{(r)}$}
Assuming that ${\boldsymbol{\Upsilon}^{(r)}_{\indexi}=[\boldsymbol{\upsilon}^{(r)}_{\indexi_1},\boldsymbol{\upsilon}^{(r)}_{\indexi_2},\myydots,\boldsymbol{\upsilon}^{(r)}_{\indexi_{N_{r_i}}}]^{\top}\in\mathbb{R}^{N_{r_i}\times 2}}$
denotes a sample matrix associated with the $\indexi$th linear segment such that ${\boldsymbol{\upsilon}^{(r)}_{\indexi_\indexm}\hspace{-0.5mm}=\hspace{-0.5mm}[\indexm,v_{\indexm}^{(r)}]^{\top}\hspace{-0.7mm}\in\hspace{-0.7mm}\mathbb{R}^{2}}$, ${\indexm=1,\myydots,\dimN_{r_{\indexi}}}$, our goal is to approximate ${\mathbf{v}^{(r)}\in\mathbb{R}^{(\dimN\shortminus\dimN_{\mathrm{I}})}}$,  using a piece-wise linear function that is determined by estimating $\blocknum_{\mathrm{cand}}$ planes, i.e.\vspace{1mm}
\begin{align}
\hspace{-0.8mm}\hat{\mathcal{P}}^{(r)}_{\indexi}\hspace{-0.7mm}=\hspace{-0.7mm}\{\boldsymbol{\upsilon}^{(r)}_{\indexi_\indexm}| \boldsymbol{\upsilon}^{(r)}_{\indexi_\indexm}\in\mathbb{R}^2, (\hat{\boldsymbol{\vartheta}}^{(r)}_{\indexi})^{\top}\boldsymbol{\upsilon}^{(r)}_{\indexi_\indexm}+\hat{\mathrm{b}}^{(r)}_{\indexi}\hspace{-0.7mm}=\hspace{-0.7mm}0\},\hspace{-1mm}\vspace{1mm}
\end{align}
where $\hat{\boldsymbol{\vartheta}}^{(r)}_{\indexi}\hspace{-0.7mm}\in\hspace{-0.7mm}\mathbb{R}^2$ and $\hat{\mathrm{b}}^{(r)}_{\indexi}\hspace{-0.7mm}\in\hspace{-0.7mm}\mathbb{R}$ denote, respectively, the normal vector and the bias associated with the estimated $\indexi$th plane $\hat{\mathcal{P}}^{(r)}_{\indexi}$ for $\indexi\hspace{-0.7mm}=\hspace{-0.7mm}1,\myydots,\blocknum_{\mathrm{cand}}$. The estimation can be performed by solving the $\blocknum_{\mathrm{cand}}$ individual ordinary eigenvalue problems as in \cite{PlrPC}\vspace{1mm}
\begin{align} \label{eq:estimatenormalvector}
    \boldsymbol{\Sigma}^{(r)}_{\indexi}\hat{\boldsymbol{\vartheta}}^{(r)}_{\indexi}=\Lambda_{\indexi}^{(r)}\hat{\boldsymbol{\vartheta}}^{(r)}_{\indexi}\hspace{5mm}\indexi=1,\myydots,\blocknum_{\mathrm{cand}}\vspace{1mm}
\end{align}
and
\begin{align}\label{eq:estimatebias}\vspace{1mm}
   \hat{\mathrm{b}}^{(r)}_{\indexi}=-(\hat{\boldsymbol{\vartheta}}^{(r)}_{\indexi})^{\top}\boldsymbol{\mu}^{(r)}_{\indexi}\hspace{5mm}\indexi=1,\myydots,\blocknum_{\mathrm{cand}},\vspace{1mm}
\end{align}
where $\Lambda^{(r)}_{\indexi}\in\mathbb{R}$ is the smallest eigenvalue associated with the $\indexi$th plane, and $\boldsymbol{\Sigma}^{(r)}_{\indexi}\in\mathbb{R}^{2\times 2}$ and $\boldsymbol{\mu}^{(r)}_{\indexi}\in\mathbb{R}^2$ are, respectively, the covariance and the mean matrices of $\boldsymbol{\upsilon}^{(r)}_{\indexi}$.
Then, using the estimated parameters of the $\blocknum_{\mathrm{cand}}$ planes, each segment in the vector ${\mathbf{v}^{(r)}}$ is estimated as follows\vspace{1mm}
\begin{align}\label{eq:estimatethefit}
    (\hat{\boldsymbol{\vartheta}}^{(r)}_{\indexi})^{\top}
    \begin{bmatrix}
    \indexm\\
    \hat{v}^{(r)}_{i_\indexm}
    \end{bmatrix}+\hat{\mathrm{b}}^{(r)}_{\indexi}=0,\hspace{4mm}
    \begin{aligned}
    &\indexi=1,\myydots,\blocknum_{\mathrm{cand}}\\
    &\indexm=1,\myydots,\dimN_{r_{\indexi}}
    \end{aligned},\vspace{1mm}
\end{align}
where $\hat{v}^{(r)}_{i_\indexm}$ denotes the $\indexm$th estimated point correspond to the $\indexi$th segment. Based on this linear function, the target similarity coefficients $\mathrm{diag}(\mathbf{W}^{(r)}_{\mathrm{sim}})$ can be computed as\vspace{1mm}
\begin{equation}\label{eq:targetsimcoeffestimate}
    \hat{w}^{(r)}_{\indexi}=-\frac{\hat{\vartheta}^{(r)}_{\indexi_1}}{\hat{\vartheta}^{(r)}_{\indexi_2}},\hspace{5mm}\indexi=1,\myydots,\blocknum_{\mathrm{cand}}\vspace{1mm}
\end{equation}
where $\hat{\vartheta}^{(r)}_{\indexi_1}$ and $\hat{\vartheta}^{(r)}_{\indexi_2}$ are denote, respectively, the first and the second elements in $\hat{\boldsymbol{\vartheta}}^{(r)}_{\indexi}$, and $\hat{w}^{(r)}_{\indexi}$ is the target similarity coefficient estimate associated with $\indexi$th block.

\subsection{Computing vector of increases ${\dddot{\mathbf{v}}^{(r)}_{\mathrm{u}_{\indexi\indexj}}}$}\label{sec:Vectorofincrease}
The vector of increase ${\dddot{\mathbf{v}}^{(r)}_{\indexi,\indexj}\in\mathbb{R}^{\dimN_{r_i}}}$ associated with group similarity between block $\indexi$ and $\indexj$, is computed as follows\vspace{1mm}
\begin{equation}\label{eq:computeundesiredsimij}
\dddot{\mathbf{v}}^{(r)}_{\indexi,\indexj}=\Bigg[\sum\limits_{\indexn=\ell_{r_j}}^{u_{r_\indexj}}\dddot{l}_{\ell_{r_i},\indexn},\myydots,-\sum\limits_{\indexn=\ell_{r_j}}^{u_{r_\indexj}}\dddot{l}_{u_{r_i},\indexn}\Bigg].\vspace{1mm}
\end{equation}

\vspace{-3mm}
\section{Computational Complexity Analysis} \label{sec:comcompanalysis}\vspace{-0.5mm}
Due to its essential role in graph analysis, the computational complexity of the proposed FRS-BDR method is analyzed in terms
of its main operations. The computational analysis is detailed using the following terms \cite{GStewartMatrixI,GStewartMatrixII}:
\begin{itemize}
    \item [] fladd: an operation that consists of a floating-point addition
    
    \item [] flmlt: an operation that consists of a floating-point multiplication
    
    \item [] fldiv: an operation that consists of a floating-point division
    
    \item [] flam: a compound operation that consists of one addition and one multiplication
\end{itemize}
Additionally, the Landau’s big $O$ symbol is used when the complexity is not specified as above terms. In the sequel, the computational complexity of the proposed approach is detailed for the fundamental steps.
\vspace{-2.7mm}
\subsection{Initial Graph Construction}
As in \cite{RLPI}, the pairwise cosine similarity which takes $\frac{1}{2}\dimN^2\dimM+2\dimN\dimM$ flam  can be used for constructing an initial graph $G$.

\vspace{-2.7mm}
\subsection{Step 1: Enhancing BD Structure}
\subsubsection{Step 1.1 (optional): Type I Outlier Removal}
Since removing unconnected vertices associated with Type~I outliers results in negligibly smaller cost in comparison to remaining estimation steps, the complexity of preprocessing is ignored in the complexity analysis.
\subsubsection{Step 1.2: sBDO Algorithm}
In this step, the complexity of sBDO algorithm is detailed for two main operations as follows.

To determine the starting nodes, the overall edge weights must be sorted which is of complexity $O(\dimN\mathrm{log}\dimN)$ and there are computationally efficient alternatives such as \cite{efficientsorting} for which the complexity is reduced to $O(\dimN\sqrt{\mathrm{log}\dimN})$.
    
The second main operation is adding the most similar node to the vector of previously estimated nodes. For a vector of previously estimated nodes $\hat{\mathbf{b}}^{(s)}\in\hspace{-0.7mm}\mathbb{Z}_{+}^{s-1}$ at the $s$th stage, the method sums up the similarity coefficients that takes 
$s-2$ additions for every neighbor. For $\dimN^{(s)}_{\mathrm{neigh}}$ number of neighbors, it follows that 
\begin{align*}
   \sum\limits_{s=2}^{\dimN}(s-2)\dimN^{(s)}_{\mathrm{neigh}}
\end{align*}
fladd. Even though the complexity is directly linked to number of neighbors, it can be explicitly calculated if every node is connected to remaining $\dimN-1$ number of nodes, i.e.
   \begin{align*}
   \sum\limits_{s=2}^{\dimN}(s-2)(\dimN-s+1)
   \end{align*}
fladd. In addition to similarity computation, the method sorts vector of neighbor similarities to find the most similar node. This sorting operation results in minimum  $O(\dimN^{(s)}_{\mathrm{neigh}}\sqrt{\mathrm{log}\dimN^{(s)}_{\mathrm{neigh}}})$ complexity at every stage ${s=2,\myydots,\dimN}$.

\subsubsection{Step 1.3 (optional): Increase Sparsity for Excessive Group Similarity}
This step evaluates computational complexity of sparse Laplacian matrix design for the two examplary algorithms which have been provided in Appendix~E.3 accompanying material.

\vspace{3mm}
\textbf{Sparse Laplacian Matrix Design based on Adaptive Thresholding:} As eigen-decomposition and sorting operations are computationally demanding in comparison to thresholding operation, the complexity is detailed in terms of these two main processes. The proposed sparse Laplacian matrix design computes eigenvalues for each iteration in which e.g. MATLAB uses a Krylov Schur decomposition \cite{KrylovLargeEigenproblems}. The decomposition is built upon two main phases that are known as expansion and contraction. The computational cost of decomposition mainly depends on these phases when $\dimN$ is larger than $\dimN_{\mathrm{Lan}}$, where $\dimN_{\mathrm{Lan}}$ denotes the number of Lanczos basis vectors (preferably chosen as $\dimN_{\mathrm{Lan}}\geq2\dimN_{\mathrm{eig}}$ for $\dimN_{\mathrm{eig}}$ eigenvectors). In more details, the expansion phase requires between $\dimN(\dimN_{\mathrm{Lan}}^2-\dimN_{\mathrm{eig}}^2)$ flam and $2\dimN(\dimN_{\mathrm{Lan}}^2-\dimN_{\mathrm{eig}}^2)$ flam while the contraction phase requires $\dimN\dimN_{\mathrm{Lan}}\dimN_{\mathrm{eig}}$ flam \cite{GStewartMatrixII}. To find the second smallest eigenvalues, the computed eigenvalues must be sorted which is of complexity $O(\dimN\sqrt{\mathrm{log}\dimN})$ in \cite{efficientsorting}.
When $\dimN_{\mathrm{Lan}}\rightarrow \dimN$, the cost of eigen-decomposition dominates the sorting operation. Thus, sparse Laplacian matrix design using adaptive thresholding with respect to flam yields minimally
\begin{align*}
    \dimN_{\mathrm{iter}}(\dimN(\dimN_{\mathrm{Lan}}^2-\dimN_{\mathrm{eig}}^2)+\dimN \dimN_{\mathrm{Lan}}\dimN_{\mathrm{eig}}),
\end{align*}
where $\dimN_{\mathrm{iter}}$ denotes the number of performed iterations to achieve a sparse Laplacian matrix.\\

\textbf{Sparse Laplacian Matrix Design based on $p$-Nearest Neighbor Graph:}
In contrast to adaptive thresholding-based graph construction, computing a $p$-nearest neighbor graph results in considerable cost. In addition to eigen-decomposition and sorting operations, the algorithm finds $p$-nearest neighbors which is of complexity $\dimN^2\mathrm{log}\dimN$ flam \cite{RLPI} in each iteration. Therefore, the overall computational cost of sparse Laplacian matrix design using $p$-nearest graph is written in terms of flam as follows
\begin{align*}
    \dimN_{p_\mathrm{iter}}(\dimN(\dimN_{\mathrm{Lan}}^2-\dimN_{\mathrm{eig}}^2)+\dimN \dimN_{\mathrm{Lan}}\dimN_{\mathrm{eig}}+\dimN^2\mathrm{log}\dimN).
\end{align*}
 
\subsection{Step 2: Estimating Vector $\mathbf{v}$}
\subsubsection{Step 2.1: Computing Candidate Block Sizes}
The estimation of candidate block size matrix can mainly be attributed to changepoint detection which is a widely researched topic in the literature and there are variety of different alternatives for changepoint detection, e.g. binary segmentation (BS) or optimal partitioning (OP) approaches. In \cite{PELT}, the computational cost of BS and OP are indicated as $O(\dimN\mathrm{log}\dimN)$ and $O(\dimN^2)$, respectively. Then, a computationally efficient the pruned exact linear time (PELT) method has been provided. The complexity of the PELT method is $O(\dimN)$ (under certain conditions) which can be reach $O(\dimN^2)$ in the worst-case \cite{PELT}.
\vspace{3mm}
\subsubsection{Step 2.2: Estimating Matrix of Similarity Coefficients}

\textit{Step 2.2.1: Estimating Target Similarity Coefficients}\\
Since the eigen-decomposition of  $2\times2$ matrix does not require a considerable time, the computational complexity of plane-based piece wise linear fitting algorithm in \cite{PlrPC} mainly depends on covariance matrix and the mean vector estimation. 
    
First, let consider the covariance matrix computation. For two random vectors, the covariance function includes $\dimN$ executions where each execution includes two addition and one multiplication. Therefore, it can be said that calculation of covariance for two random vectors requires $2\dimN \mathrm{fladd}+\dimN\mathrm{flmlt}+1\mathrm{fldiv}$. As vector $\dddot{\mathbf{v}}$ fitting generates covariance matrix of dimension $2\times2$, our covariance matrix computation necessitates $6\dimN \mathrm{fladd}+3\dimN\mathrm{flmlt}+3\mathrm{fldiv}$.
    
After computing covariance matrix, the mean operator including $\dimN-1$ additions and a division is executed two times for two vectors. Thus, mean vector computation mainly requires $2(\dimN-1)$ fladd $+2$ fldiv and total piece-wise linear fitting complexity results in $8\dimN-2$ fladd $+3\dimN$ flmlt $+5$ fldiv.\\ 

\textit{Step 2.2.2: Estimating Undesired Similarity Coefficients}\\
The undesired similarity coefficients' estimation consists of two consecutive steps. For every candidate size vector $\mathbf{n}_r=[\dimN_{r_1},\dimN_{r_2},\myydots,\dimN_{r_{\blocknum_{\mathrm{cand}}}}]\in\mathbb{Z}_{+}^{\blocknum_{\mathrm{cand}}}$ the proposed method first computes shifted vector whose complexity principally depends on vector of increase computation. To compute a vector of increase  associated with group similarity between block $\indexi$ and $\indexj$, the method executes $\dimN_{r_\indexi}$ times where each execution includes $\dimN_{r_\indexj}-1$ addition. This means that the vector of increase  associated with group similarity between block $\indexi$ and $\indexj$ results in $\dimN_{r_\indexi}(\dimN_{r_\indexj}-1)$ fladd. Further, the algorithm finds median of computed vector which is of dimension $\dimN_{r_\indexi}\times1$. To compute the median, the vector can be sorted in $O(\dimN_{r_\indexi}\sqrt{\mathrm{log}\dimN_{r_\indexi}})$ complexity using \cite{efficientsorting}. When size of the $\indexj$th block $\dimN_{r_\indexj}$ is sufficiently large valued, the vector of increase computation dominates median operation. Therefore, computing an undesired similarity coefficient corresponds to group similarity between block $\indexi$ and $\indexj$ takes minimally $\dimN_{r_\indexi}(\dimN_{r_\indexj}-1)$ fladd and that can be written for 
$\indexi=2,\myydots,\blocknum_{\mathrm{cand}}$ and $\indexj=1,\myydots,\indexi-1$ as follows\vspace{-1mm}
\begin{align*}
\sum\limits_{\indexi=2}^{\blocknum_{\mathrm{cand}}}\sum\limits_{\indexj=1}^{_{\indexi-1}}\dimN_{r_\indexi}(\dimN_{r_\indexj}-1)\hspace{2mm}\mathrm{fladd}.
\end{align*}

To summarize, estimating the similarity coefficients matrix $\hat{\mathbf{W}}^{(r)}_{\mathrm{sim}}$ and the vector $\hat{\mathbf{v}}^{(r)}\in\mathbb{R}^{\dimN}$ 
associated with a candidate block size vector $\mathbf{n}_r\in\mathbb{Z}_{+}^{\blocknum_{\mathrm{cand}}}$ requires minimally\vspace{-1mm}
\begin{align*}
    &\blocknum_{\mathrm{cand}}(8\dimN-2\hspace{2mm}\mathrm{fladd}+3\dimN\hspace{2mm}\mathrm{flmlt} +5\hspace{2mm}\mathrm{fldiv})\\&+   \sum\limits_{\indexi=2}^{\blocknum_{\mathrm{cand}}}\sum\limits_{\indexj=1}^{_{\indexi-1}}\dimN_{r_\indexi}(\dimN_{r_\indexj}-1)\hspace{2mm}\mathrm{fladd}.
\end{align*}

To compute all possible $\hat{\mathbf{W}}^{(r)}_{\mathrm{sim}}\in\mathbb{R}^{\blocknum_{\mathrm{cand}}\times\blocknum_{\mathrm{cand}}}$ and $\widehat{\dddot{\mathbf{v}}}^{(r)}\in\mathbb{R}^{\dimN}$ correspond to ${\mathbf{\dimN}^{(\mathrm{\blocknum}_{\mathrm{cand}})}=[\mathbf{n}_1,\mathbf{n}_2,\myydots,\mathbf{n}_{\zeta}]^{\top}\in\mathbb{Z}_{+}^{\zeta\times{\blocknum_{\mathrm{cand}}}}}$ for ${\blocknum_{\mathrm{cand}}=\blocknum_{\mathrm{min}},\myydots,\blocknum_{\mathrm{max}}}$, the proposed method requires
\begin{align*}\vspace{-1mm}
    &\zeta\Big(\blocknum_{\mathrm{cand}}(8\dimN-2\hspace{2mm}\mathrm{fladd}+3\dimN\hspace{2mm}\mathrm{flmlt} +5\hspace{2mm}\mathrm{fldiv}) \\&+   \sum\limits_{\indexi=2}^{\blocknum_{\mathrm{cand}}}\sum\limits_{\indexj=1}^{_{\indexi-1}}\dimN_{r_\indexi}(\dimN_{r_\indexj}-1)\hspace{2mm}\mathrm{fladd}\Big).
\end{align*}
When the number of candidate block size vectors increases, the complexity of these numerous operations is significantly larger than changepoint detection using PELT method. Thus, its computational cost can be ignored in vector $\mathbf{v}$ estimation.

\section{Experimental Setting} \label{sec:experimentalsetting}
\subsection{Parameter Definition}
\subsubsection{FRS-BDR for Unknown Number of Blocks}\label{sec:UnKnownKSubspaceClustering}
\begin{itemize}[leftmargin=*]
    \item []Eigen-decomposition function: the generalized eigen-decomposition
    \item []Minimum number of blocks ($\blocknum_{\mathrm{min}}$): 2
    \item []Maximum number of blocks ($\blocknum_{\mathrm{max}}$): $2\times\blocknum$
    \item []Minimum number of nodes in the blocks ($\dimN_{\mathrm{min}}$): $\frac{\dimN}{\blocknum_{\mathrm{max}}}$
    \item []Maximum number of changepoints ($\dimN_{c_\mathrm{max}}$): $2\times(\blocknum_{\mathrm{max}}\hspace{-0.4mm}-\hspace{-0.4mm}1)$
    \item []Sparse Laplacian matrix design algorithm: $p$-nearest graph (For details, see Algorithm~4 in accompanying material.)
\end{itemize}
\subsubsection{FRS-BDR for Known Number of Blocks}\label{sec:KnownKSubspaceClustering}
\begin{itemize}[leftmargin=*]
    \item []Eigen-decomposition function : the generalized eigen-decomposition
    \item []Minimum number of nodes in the blocks ($\dimN_{\mathrm{min}}$) : $\frac{\dimN}{2\times\blocknum}$
    \item []Maximum number of changepoints $\dimN_{c_\mathrm{max}}$ : $2\times(\blocknum-1)$
    \item []Sparse Laplacian matrix design algorithm : $p$-nearest graph (For details, see Algorithm~4 in accompanying material.)
\end{itemize}
\subsubsection{Spectral Clustering}\label{sec:SpectralClustering}
\begin{itemize}[leftmargin=*]
    \item[]Laplacian matrix : $\mathbf{L}=\mathbf{D}-\mathbf{W}$
    \item[]Partitioning method : $\blocknum$-means
    \item[]Eigen-decomposition function : the generalized eigen-decomposition
\end{itemize}
\vspace{-4mm}
\subsection{Evaluation Metrics}
In the following, the different clustering performance evaluation metrics are detailed.
\begin{itemize}[leftmargin=0.5cm]
    \item [$\small{\blacktriangleright}$]Average clustering accuracy $(\bar{c}_{\mathrm{acc}})$ is measured by
    \begin{equation*}
\Bar{c}_{\mathrm{acc}}=\frac{1}{\dimN\dimN_E}\sum_{\indexm=1}^{\dimN_E}\sum_{\indexn=1}^{\dimN}\mathbbm{1}_{\{\hat{c}_{\indexn}=c_{\indexn}\}},
\end{equation*}
where 
\begin{equation*}
\mathbbm{1}_{\{\hat{c}_{\indexn}=c_{\indexn}\}}=\begin{cases}
     1 ,        & \text{if } \hat{c}_{\indexn}=c_{\indexn}\\
     0 ,           & \text{otherwise}
\end{cases},
\end{equation*}
$\dimN$ is the number of observations, $\dimN_E$ is the total number of experiments, $c_{\indexn}$ and  $\hat{c}_{\indexn}$ are the estimated and ground truth labels for the $\indexn$th observation, respectively.

\item [$\small{\blacktriangleright}$]The empirical probability of detection $p_{\mathrm{det}}$ is used to analyze cluster enumeration performance as follows
\begin{equation*}
p_{\mathrm{det}}=\frac{1}{\dimN_E}\sum_{\indexm=1}^{\dimN_E}\mathbbm{1}_{\{\hat{\blocknum}=\blocknum\}},
\end{equation*}
where $\hat{\blocknum}$ denotes the number of clusters estimate and $\mathbbm{1}_{\{\hat{\blocknum}=\blocknum\}}$ is the indicator function.

\item [$\small{\blacktriangleright}$]Modularity ($\mathrm{mod}$) is a metric that evaluates the quality of partition. The modularity score tends to one if a vertex has more edges within the assigned community. As in  
\cite{Newman2002} and \cite{Newman2006}, $\mathrm{mod}$ is calculated by
\begin{equation*}
\mathrm{mod}=\frac{1}{2g}\sum_{\indexm,\indexn}^{\dimN}\Big[{w}_{\indexm,\indexn}-\frac{d_{\indexm}d_{\indexn}}{2g}\Big]\delta(c_{\indexm},c_{\indexn})
\end{equation*}
Here, $d_{\indexm}=\sum_{\indexn=1}^{\dimN}w_{\indexm,\indexn}$ is the overall edge weight attached to vertex $\indexm$,  $g=\frac{1}{2}\sum_{\indexm,\indexn}w_{\indexm,\indexn}$ and the function $\delta(c_{\indexm},c_{\indexn})$ equals one if $c_{\indexm}=c_{\indexn}$ and is zero otherwise.

\item [$\small{\blacktriangleright}$]Conductance ($\mathrm{cond}$) evaluates quality of partition in terms of the fraction of edges that points outside the community. Therefore, a small-valued conductance score means good partitioning performance and it can be calculated as \cite{conductance}
\begin{equation*}
\mathrm{cond}=\frac{\sum_{\indexm,\indexn}^{\dimN}w_{\indexm,\indexn}\mathbbm{1}_{\{c_{\indexm}\neq c_{\indexn}\}}}{\sum_{\indexm,\indexn}^{\dimN}w_{\indexm,\indexn}},
\end{equation*}
where $w_{\indexm,\indexn}$ denotes the weight of the edge between the $\indexm$th and the $\indexn$th feature vector of $\mathbf{X}$, the function $\mathbbm{1}_{\{c_{\indexm}\neq c_{\indexn}\}}$ equals one if $c_{\indexm}\neq c_{\indexn}$ and is zero otherwise.

\end{itemize}


\title{Accompanying Material: Fast and Robust Sparsity-Aware Block Diagonal Representation}

\author{Aylin~Ta{\c{s}}tan$^\ast$, Michael~Muma$^\dagger$,
        and~Abdelhak~M.~Zoubir$^\ddagger$
\thanks{\noindent$^\ast$The author was with the Signal Processing Group, Technische Universität
Darmstadt, Darmstadt, Germany and is now with the Pattern Recognition Group, University of Bern, Bern, Switzerland (e-mail:  a.tastan@spg.tu-darmstadt.de; aylin.tastan@unibe.ch).\protect\\
$^\dagger$The author is with the Robust Data Science Group, Technische Universität
Darmstadt, Darmstadt, Germany (e-mail: michael.muma@tu-darmstadt.de).\protect\\
$^\ddagger$The author is with the Signal Processing Group, Technische Universität
Darmstadt, Darmstadt, Germany (e-mail:  zoubir@spg.tu-darmstadt.de).
}}

\markboth{SUBMITTED TO IEEE TRANSACTIONS ON SIGNAL PROCESSING}%
{Shell \MakeLowercase{\textit{et al.}}: A Sample Article Using IEEEtran.cls for IEEE Journals}

\IEEEpubid{}

\maketitle
\section*{Structure}
\label{sec:introduction}
\setlength{\parindent}{0pt}
The Accompanying Material for the paper Fast and Robust Sparsity-Aware Block Diagonal Representation (FRS-BDR) is organized as follows: In Appendix A and B, the theorems by reference to outlier effects on eigenvalues are proved based on the generalized and the standard eigen-decompositions, respectively. The theoretical analysis of the outlier effects on vector $\mathbf{v}$ is the subject of Appendix C. Appendix D contains the auxiliary information that is used to analyse the effects of outliers on the eigenvalues and in Appendix~E real-world data examples analyzing deviations from the theoretical analysis are given. Lastly, additional information on FRS-BDR, including visual summary, sparse Laplacian matrix analysis, algorithms and  detailed experimental results, is provided. 
\vspace{5mm}
\setcounter{secnumdepth}{0}
\section{Appendix A: The Generalized Eigen-decomposition based Eigenvalue Analysis}
\vspace{1mm}
\subsection{A.1~Outlier Effects on Target Eigenvalues: Proof of Theorem 1}\label{sec:app:ProofTheorem3}
\begin{proof}[\unskip\nopunct]
Let $\tilde{\mathbf{W}}\in\mathbb{R}^{(\dimN+1)\times(\dimN+1)}$ and $\tilde{\mathbf{L}}\in\mathbb{R}^{(\dimN+1)\times(\dimN+1)}$, respectively, denote a block zero-diagonal symmetric affinity matrix and associated Laplacian matrix for $\blocknum$ blocks with an additional Type II outlier that is correlated with all blocks, i.e.,
\begin{align*}
\Tilde{\mathbf{W}}=
\begin{bmatrix}
\begin{smallmatrix}
0&\Tilde{w}_{\TypeIIoutlier,1}&\Tilde{w}_{\TypeIIoutlier,1} &\myydots&\Tilde{w}_{\TypeIIoutlier,1}&\Tilde{w}_{\TypeIIoutlier,2}&\Tilde{w}_{\TypeIIoutlier,2} &\myydots&\Tilde{w}_{\TypeIIoutlier,2} &\myydots& \Tilde{w}_{\TypeIIoutlier,\blocknum}& \Tilde{w}_{\TypeIIoutlier,\blocknum}&\myydots&\Tilde{w}_{\TypeIIoutlier,\blocknum}\\
\Tilde{w}_{\TypeIIoutlier,1}&0& w_1&\myydots&w_1&& & && && & &\\
\Tilde{w}_{\TypeIIoutlier,1}&w_1&0&\myydots&w_1& & & & && & & &\\
\vdots&\vdots&\vdots&\ddots&&& & && & & & &\\
\Tilde{w}_{\TypeIIoutlier,1}&w_1 &w_1 &\myydots & 0 & && &&& & & &\\
\Tilde{w}_{\TypeIIoutlier,2}&& & & &0& w_2&\myydots&w_2& & & & &\\
\Tilde{w}_{\TypeIIoutlier,2}& & & &  &w_2&0&\myydots&w_2& & & & &\\
\vdots&& & &  &\vdots&\vdots&\ddots&& & & & &\\
\Tilde{w}_{\TypeIIoutlier,2}& & & & &w_2 &w_2 &\myydots & 0 & & & & &\\
\vdots&& & & && & & &\ddots & &\vdots& & \vdots\\
\Tilde{w}_{\TypeIIoutlier,\blocknum}&&& & &&&&&&0&w_{\blocknum}&\myydots&w_{\blocknum}\\
\Tilde{w}_{\TypeIIoutlier,\blocknum}&& & & &&&&&&w_{\blocknum}&0& \myydots& w_{\blocknum} \\
\vdots&& & &&&&&&&\vdots&\vdots&\ddots\\
\Tilde{w}_{\TypeIIoutlier,\blocknum}&& & & &&&&&\myydots&w_{\blocknum}& w_{\blocknum} &\myydots&0\\
\end{smallmatrix}
\end{bmatrix}
\end{align*}
and
\begin{align*}
\Tilde{\mathbf{L}}=
\begin{bmatrix}
\begin{smallmatrix}
\Tilde{d}_{\TypeIIoutlier}&-\Tilde{w}_{\TypeIIoutlier,1}&-\Tilde{w}_{\TypeIIoutlier,1} &\myydots&-\Tilde{w}_{\TypeIIoutlier,1}&-\Tilde{w}_{\TypeIIoutlier,2}&-\Tilde{w}_{\TypeIIoutlier,2} &\myydots&-\Tilde{w}_{\TypeIIoutlier,2} &\myydots& -\Tilde{w}_{\TypeIIoutlier,\blocknum}& -\Tilde{w}_{\TypeIIoutlier,\blocknum}&\myydots&-\Tilde{w}_{\TypeIIoutlier,\blocknum}\\
-\Tilde{w}_{\TypeIIoutlier,1}&\Tilde{d}_1& -w_1&\myydots&-w_1&& & && && & &\\
-\Tilde{w}_{\TypeIIoutlier,1}&-w_1&\Tilde{d}_1&\myydots&-w_1& & & & && & & &\\
\vdots&\vdots&\vdots&\ddots&&& & && & & & &\\
-\Tilde{w}_{\TypeIIoutlier,1}&-w_1 &-w_1 &\myydots & \Tilde{d}_1 & && &&& & & &\\
-\Tilde{w}_{\TypeIIoutlier,2}&& & & &\Tilde{d}_2& -w_2&\myydots&-w_2& & & & &\\
-\Tilde{w}_{\TypeIIoutlier,2}& & & &  &-w_2&\Tilde{d}_2&\myydots&-w_2& & & & &\\
\vdots&& & &  &\vdots&\vdots&\ddots&& & & & &\\
-\Tilde{w}_{\TypeIIoutlier,2}& & & & &-w_2 &-w_2 &\myydots & \Tilde{d}_2 & & & & &\\
\vdots&& & & && & & &\ddots & &\vdots& & \vdots\\
-\Tilde{w}_{\TypeIIoutlier,\blocknum}&&& & &&&&&&\Tilde{d}_{\blocknum}&-w_{\blocknum}&\myydots&-w_{\blocknum}\\
-\Tilde{w}_{\TypeIIoutlier,\blocknum}&& & & &&&&&&-w_{\blocknum}&\Tilde{d}_{\blocknum}& \myydots& -w_{\blocknum} \\
\vdots&& & &&&&&&&\vdots&\vdots&\ddots\\
-\Tilde{w}_{\TypeIIoutlier,\blocknum}&& & & &&&&&\myydots&-w_{\blocknum}& -w_{\blocknum} &\myydots&\Tilde{d}_{\blocknum}\\
\end{smallmatrix}
\end{bmatrix}
\end{align*}
where $\Tilde{d}_{\TypeIIoutlier}=\sum_{\indexj=1}^{\blocknum}\dimN_{\indexj}\Tilde{w}_{\TypeIIoutlier,\indexj}$ and $\Tilde{d}_{\indexj}=(\dimN_{\indexj}-1)w_{\indexj}+\Tilde{w}_{\TypeIIoutlier,\indexj}$ such that $\indexj=1,\myydots,\blocknum$. To compute the eigenvalues of the Laplacian matrix $\Tilde{\mathbf{L}}$, $\mathrm{det}(\Tilde{\mathbf{L}}-\Tilde{\lambda}\Tilde{\mathbf{D}})=0$ is considered which can equivalently be written in matrix form as follows
\begin{align*}
\mathrm{det}(\Tilde{\mathbf{L}}-\Tilde{\lambda}\Tilde{\mathbf{D}})=
\begin{vmatrix}
\begin{smallmatrix}
\Tilde{d}_{\TypeIIoutlier}-\Tilde{\lambda}\Tilde{d}_{\TypeIIoutlier}&-\Tilde{w}_{\TypeIIoutlier,1}&-\Tilde{w}_{\TypeIIoutlier,1} &\myydots&-\Tilde{w}_{\TypeIIoutlier,1}&-\Tilde{w}_{\TypeIIoutlier,2}&-\Tilde{w}_{\TypeIIoutlier,2} &\myydots&-\Tilde{w}_{\TypeIIoutlier,2} &\myydots& -\Tilde{w}_{\TypeIIoutlier,\blocknum}& -\Tilde{w}_{\TypeIIoutlier,\blocknum}&\myydots&-\Tilde{w}_{\TypeIIoutlier,\blocknum}\\
-\Tilde{w}_{\TypeIIoutlier,1}&\Tilde{d}_1-\Tilde{\lambda}\Tilde{d}_1& -w_1&\myydots&-w_1&& & && && & &\\
-\Tilde{w}_{\TypeIIoutlier,1}&-w_1&\Tilde{d}_1-\Tilde{\lambda}\Tilde{d}_1&\myydots&-w_1& & & & && & & &\\
\vdots&\vdots&\vdots&\ddots&&& & && & & & &\\
-\Tilde{w}_{\TypeIIoutlier,1}&-w_1 &-w_1 &\myydots & \Tilde{d}_1-\Tilde{\lambda}\Tilde{d}_1 & && &&& & & &\\
-\Tilde{w}_{\TypeIIoutlier,2}&& & & &\Tilde{d}_2-\Tilde{\lambda}\Tilde{d}_2& -w_2&\myydots&-w_2& & & & &\\
-\Tilde{w}_{\TypeIIoutlier,2}& & & &  &-w_2&\Tilde{d}_2-\Tilde{\lambda}\Tilde{d}_2&\myydots&-w_2& & & & &\\
\vdots&& & &  &\vdots&\vdots&\ddots&& & & & &\\
-\Tilde{w}_{\TypeIIoutlier,2}& & & & &-w_2 &-w_2 &\myydots & \Tilde{d}_2-\Tilde{\lambda}\Tilde{d}_2& & & & &\\
\vdots&& & & && & & &\ddots & &\vdots& & \vdots\\
-\Tilde{w}_{\TypeIIoutlier,\blocknum}&&& & &&&&&&\Tilde{d}_{\blocknum}-\Tilde{\lambda}\Tilde{d}_{\blocknum}&-w_{\blocknum}&\myydots&-w_{\blocknum}\\
-\Tilde{w}_{\TypeIIoutlier,\blocknum}&& & & &&&&&&-w_{\blocknum}&\Tilde{d}_{\blocknum}-\Tilde{\lambda}\Tilde{d}_{\blocknum}& \myydots& -w_{\blocknum} \\
\vdots&& & &&&&&&&\vdots&\vdots&\ddots\\
-\Tilde{w}_{\TypeIIoutlier,\blocknum}&& & & &&&&&\myydots&-w_{\blocknum}& -w_{\blocknum} &\myydots&\Tilde{d}_{\blocknum}-\Tilde{\lambda}\Tilde{d}_{\blocknum}\\
\end{smallmatrix}
\end{vmatrix}=0.
\end{align*}
To simplify this determinant, the matrix determinant lemma \cite{matrixdeterminantlemma} can be generalized as follows\footnote{For a detailed information about generalization of the matrix determinant lemma, see Appendix~D.}
\begin{align*}
\mathrm{det}(\mathbf{H}+\mathbf{U}\mathbf{V}^{\top})=\mathrm{det}(\mathbf{H})\mathrm{det}(\mathbf{I}+\mathbf{V}^{\top}\mathbf{H}^{\dagger}\mathbf{U}) 
\end{align*}
where $\mathbf{H}\in\mathbb{R}^{(\dimN+1)\times(\dimN+1)}$ denotes an invertible matrix, $\mathbf{I}$ is the identity matrix and $\mathbf{U},\mathbf{V}\in\mathbb{R}^{(\dimN+1)\times(\dimN+1)}$. Then, for $\mathrm{det}(\tilde{\mathbf{L}}-\tilde{\lambda}\tilde{\mathbf{D}})=\mathrm{det}(\mathbf{H}+\mathbf{U}\mathbf{V}^{\top})=0$, it follows that\vspace{5mm}\\
\resizebox{\linewidth}{!}{
  \begin{minipage}{\linewidth}
\begin{align*}
\begin{split}
\hspace{-0.7mm}0\hspace{-0.7mm}=&
\mathrm{det}\begin{pmatrix}
\begin{bmatrix}
\begin{smallmatrix}
z_{\TypeIIoutlier}&\myydots &&&& & & & & & &\myydots&0\\\vspace{-1mm}
&\hspace{1.5mm}z_1& &&&& & & & & & &\vdots\\\vspace{-0.7mm}
&&\hspace{1.5mm}\ddots&&&& & && & & & \\\vspace{-0.7mm}
&& & &z_1 & & & & & & & & & \\\vspace{-1mm}
&&&& &\hspace{-1mm}z_2& & & & & & &\\\vspace{-0.6mm}
&&&& &&\hspace{-1mm}\ddots& && & & & \\\vspace{-0.6mm}
&& & & && & &\hspace{-2mm}z_2  & & & & & \\\vspace{-1mm}
&& & & && & &  &\ddots& & & & \\\vspace{-1mm}
&&&& && & & &\hspace{3mm}z_{\blocknum}& & & \\\vspace{-1mm}
\vdots&&&& &&& & & &\hspace{1mm}\ddots& & \\\vspace{-1mm}
0&\myydots && & && & && & & &\hspace{-4mm}z_{\blocknum}  &  \\\vspace{0.6mm}
\end{smallmatrix}
\end{bmatrix}+
\begin{bmatrix}
\begin{smallmatrix}
1&\hspace{-1mm}\bovermat{$\blocknum$}{0&\myydots& 
&}&&\hspace{-2mm}&&0\hspace{1mm}&\bovermat{$\dimN-\blocknum$}{0&\myydots&} &\myydots &0\\
0&\hspace{-1.3mm}1&0&\myydots 
&&&\hspace{-2mm}&&0\hspace{1mm}&0&\myydots& &\myydots &0\\
\vdots&\vdots&\vdots&&&&\hspace{-2mm}&&\vdots&\vdots&\ddots& &\iddots &\vdots\\
0&1&0& & & &\hspace{-2mm} &&&&& & &\\
&0&1& &&&&\hspace{-2mm}&&&& & &\\
&\vdots&\vdots & &&\hspace{-2mm}&  &&&&& & &\\
& &1 &&\hspace{-2mm}& & &&&&& & &\\
& &0 && &\hspace{-2mm} & &&0&&& & &\\
& &&&&\hspace{-2mm} & &&1&&& & &\\
&\vdots&\vdots & &&& \hspace{-2mm} &&\vdots&\vdots&\iddots& &\ddots &\vdots\\
0&0 &0 &\myydots& &\hspace{-2mm}&& & 1&0&\myydots& &\myydots &0\\
\end{smallmatrix}
\end{bmatrix}
\begin{bmatrix}
\begin{smallmatrix}
0&\bovermat{$\dimN_1$}{-\Tilde{w}_{\TypeIIoutlier,1}&\myydots &}-\Tilde{w}_{\TypeIIoutlier,1}&\bovermat{$\dimN_2$}{-\Tilde{w}_{\TypeIIoutlier,2}&\myydots&}-\Tilde{w}_{\TypeIIoutlier,2} &\myydots&\bovermat{$\dimN_{\blocknum}$}{-\Tilde{w}_{\TypeIIoutlier,\blocknum} &\myydots &}-\Tilde{w}_{\TypeIIoutlier,\blocknum}\\
-\Tilde{w}_{\TypeIIoutlier,1}&-w_1&\myydots&-w_1&0&\myydots && &&\myydots &0 \\
-\Tilde{w}_{\TypeIIoutlier,2}&0&\myydots&0&-w_2&\myydots &-w_2& &&\myydots &0 \\
\vdots&& & & & & & & & &\vdots &\\
-\Tilde{w}_{\TypeIIoutlier,\blocknum}&0&\myydots &&&& & &-w_{\blocknum}&\myydots &-w_{\blocknum} \\
0&\myydots&& &&& & & &\myydots &0\\
\vdots&\ddots&& &&&& &&\iddots&\vdots  \\\vspace{-2mm}
& & & && & && & & \\
\vdots&\iddots & && && & &  &\ddots&\vdots & \\
0&\myydots&&& && & & &\myydots&  0\\
\end{smallmatrix}
\end{bmatrix}
\end{pmatrix}\\\\
\hspace{-0.7mm}0\hspace{-0.7mm}=&\mathrm{det}\hspace{-1mm}\begin{pmatrix}\hspace{-1mm}
\mathbf{I}\hspace{-1mm}+\hspace{-2mm}
\begin{bmatrix}
\begin{smallmatrix}
0&-\Tilde{w}_{\TypeIIoutlier,1}z_1^{-1}&\myydots &-\Tilde{w}_{\TypeIIoutlier,1}z_1^{-1}&-\Tilde{w}_{\TypeIIoutlier,2}z_2^{-1}&\myydots&-\Tilde{w}_{\TypeIIoutlier,2}z_2^{-1} &\myydots&-\Tilde{w}_{\TypeIIoutlier,\blocknum}z_{\blocknum}^{-1} &\myydots &-\Tilde{w}_{\TypeIIoutlier,\blocknum}z_{\blocknum}^{-1}\\
-\Tilde{w}_{\TypeIIoutlier,1}z_{\TypeIIoutlier}^{-1}&-w_1z_1^{-1}&\myydots&-w_1z_1^{-1}&0&\myydots && &&\myydots &0 \\
-\Tilde{w}_{\TypeIIoutlier,2}z_{\TypeIIoutlier}^{-1}&0&\myydots&0&-w_2z_2^{-1}&\myydots &-w_2z_2^{-1}& &&\myydots &0 \\
\vdots&& & & & & & & & &\vdots &\\
-\Tilde{w}_{\TypeIIoutlier,\blocknum}z_{\TypeIIoutlier}^{-1}&0&\myydots &&&& & &-w_{\blocknum}z_{\blocknum}^{-1}&\myydots &-w_{\blocknum}z_{\blocknum}^{-1} \\
0&\myydots&& &&& & & &\myydots &0\\
\vdots&\ddots&& &&&& &&\iddots&\vdots  \\\vspace{-2mm}
& & & && & && & & \\
\vdots&\iddots & && && & &  &\ddots&\vdots & \\
0&\myydots&&& && & & &\myydots&  0\\
\end{smallmatrix}
\end{bmatrix}\hspace{-2mm}
\begin{bmatrix}
\begin{smallmatrix}
1&\hspace{-1mm}0&\myydots& 
&&&\hspace{-2mm}&&0&0&\myydots&&\myydots &0\\\vspace{0.2mm}
0&\hspace{-1.3mm}1&0&\myydots 
&&&\hspace{-2mm}&\hspace{-2mm}&0&0&\myydots& &\myydots &0\\\vspace{0.2mm}
\vdots&\vdots&\vdots&&&&\hspace{-2mm}&&\vdots&\vdots&\ddots& &\iddots &\vdots\\\vspace{0.2mm}
0&1&0& & & &\hspace{-2mm} &&&&& & &\\\vspace{0.2mm}
&0&1& &&&&\hspace{-2mm}&&&& & &\\\vspace{0.2mm}
&\vdots&\vdots & &&\hspace{-2mm}&  &&&&& & &\\\vspace{0.2mm}
& &1 &&\hspace{-2mm}& & &&&&& & &\\\vspace{0.2mm}
& &0 && &\hspace{-2mm} & &&0&&& & &\\\vspace{0.2mm}
& &&&&\hspace{-2mm} & &&1&&& & &\\\vspace{0.2mm}
&\vdots&\vdots & &&& \hspace{-2mm} &&\vdots&\vdots&\iddots& &\ddots &\vdots\\\vspace{0.2mm}
0&0 &0 &\myydots& &\hspace{-2mm}&& & 1&0&\myydots& &\myydots &0\\
\end{smallmatrix}
\end{bmatrix}
\end{pmatrix}\mathrm{det}(\mathbf{H})\\
\vspace{3mm}
\hspace{-0.7mm}0\hspace{-0.7mm}=&
\begin{vmatrix}
\begin{smallmatrix}
1&-\dimN_1\Tilde{w}_{\TypeIIoutlier,1}z_1^{-1}&-\dimN_2\Tilde{w}_{\TypeIIoutlier,2}z_2^{-1} &\myydots&-\dimN_{\blocknum}\Tilde{w}_{\TypeIIoutlier,\blocknum}z_{\blocknum}^{-1}&0&\myydots&\hspace{1cm}& &\myydots &0\\
-\Tilde{w}_{\TypeIIoutlier,1}z_{\TypeIIoutlier}^{-1}&-\dimN_1w_1z_1^{-1}+1&0&\myydots&0&\vdots&\ddots& &&\iddots &\vdots \\
-\Tilde{w}_{\TypeIIoutlier,2}z_{\TypeIIoutlier}^{-1}&0&-\dimN_2w_2z_2^{-1}+1&\myydots&0& && && & \\
\vdots&& & &\vdots &\vdots &\iddots& & &\ddots &\vdots &\\
-\Tilde{w}_{\TypeIIoutlier,\blocknum}z_{\TypeIIoutlier}^{-1}&0&\myydots &&-\dimN_{\blocknum}w_{\blocknum}z_{\blocknum}^{-1}+1&0&\myydots & &&\myydots &0 \\
0&\myydots&& &0&1 &\myydots & & &\myydots &0\\
\vdots&\ddots&&\iddots &\vdots&\vdots&\ddots& &&\iddots&\vdots  \\\vspace{-2mm}
& & & && & && & & \\
\vdots&\iddots & &\ddots&\vdots &\vdots&\iddots & &  &\ddots&\vdots & \\
0&\myydots&&\myydots&0 &0&\myydots & & &\myydots&  1\\
\end{smallmatrix}
\end{vmatrix}\mathrm{det}(\mathbf{H})
\end{split}
\end{align*}
\end{minipage}}
\\\vspace{2mm}
where $z_{\TypeIIoutlier}=\sum_{\indexj=1}^{\blocknum}\dimN_{\indexj}\Tilde{w}_{\TypeIIoutlier,\indexj}-\tilde{\lambda}\sum_{\indexj=1}^{\blocknum}\dimN_{\indexj}\Tilde{w}_{\TypeIIoutlier,\indexj}$ and $z_{\indexj}=\dimN_{\indexj}w_{\indexj}+\Tilde{w}_{\TypeIIoutlier,\indexj}-\tilde{\lambda}\big((\dimN_{\indexj}-1)w_{\indexj}+\Tilde{w}_{\TypeIIoutlier,\indexj}\big)$ for $\indexj=1,\myydots,\blocknum$. Using determinant properties of block matrices \cite{detblockmatrices}, it holds that\vspace{1mm}
\begin{align*}
    \begin{split}
       0=&
\begin{vmatrix}
\begin{smallmatrix}
1&-\dimN_1\Tilde{w}_{\TypeIIoutlier,1}z_1^{-1}&-\dimN_2\Tilde{w}_{\TypeIIoutlier,2}z_2^{-1} &\myydots&-\dimN_{\blocknum}\Tilde{w}_{\TypeIIoutlier,\blocknum}z_{\blocknum}^{-1}\\
-\Tilde{w}_{\TypeIIoutlier,1}z_{\TypeIIoutlier}^{-1}&-\dimN_1w_1z_1^{-1}+1&0&\myydots&0\\
-\Tilde{w}_{\TypeIIoutlier,2}z_{\TypeIIoutlier}^{-1}&0&-\dimN_2w_2z_2^{-1}+1&\myydots&0\\
\vdots&& & &\vdots\\
-\Tilde{w}_{\TypeIIoutlier,\blocknum}z_{\TypeIIoutlier}^{-1}&0&\myydots &&-\dimN_{\blocknum}w_{\blocknum}z_{\blocknum}^{-1}+1
\end{smallmatrix}
\end{vmatrix}\mathrm{det}(\mathbf{H}).
    \end{split}
\end{align*}
\newpage
\begin{figure*}[tbp!]
  \centering
  \captionsetup{justification=centering}
\subfloat[ $\tilde{G}=\{\tilde{V},\tilde{E},\tilde{\mathbf{W}}\}$ ]{\includegraphics[trim={0cm 0cm 0cm 0cm},clip,width=4.25cm]{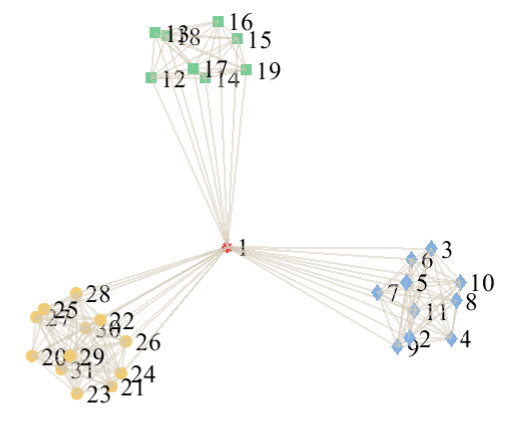}}
\subfloat[$\tilde{\mathbf{W}}\in\mathbb{R}^{(\dimN+1)\times (\dimN+1)}$]{\includegraphics[trim={0mm 0mm 0mm 0mm},clip,width=3.9cm]{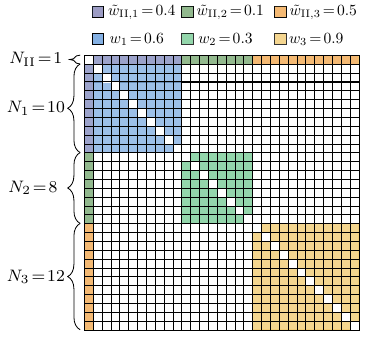}\label{fig:TypeIIOutlieraffinity}}\hspace{1mm}
\subfloat[$\tilde{\mathbf{L}}\in\mathbb{R}^{(\dimN+1)\times (\dimN+1)}$]{\includegraphics[trim={0mm 0mm 0mm 0mm},clip,width=4.95cm]{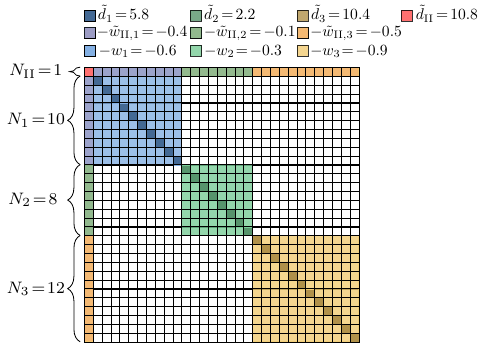}}
\subfloat[$\tilde{\boldsymbol{\lambda}}\in\mathbb{R}^{\dimN+1}$]{\includegraphics[trim={0mm 0mm 0mm 0mm},clip,width=4.75cm]{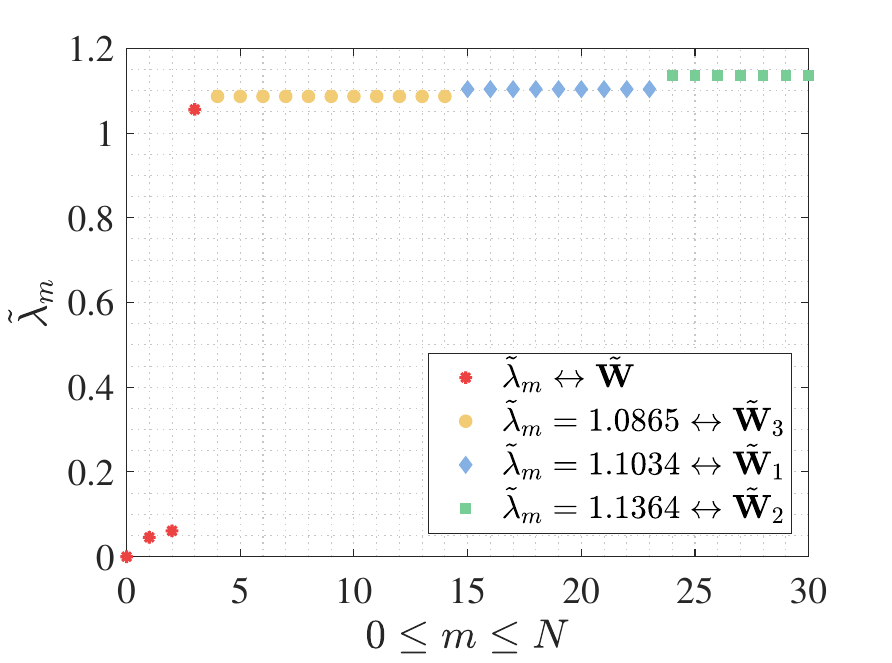}}
\caption{Examplary plot of Theorem~1 ($\mathbf{n}=[1,10,8,12]^{\top}\in\mathbb{R}^{\blocknum+1}$, $\dimN+1=31$, $\blocknum=3$).}
  \label{fig:ExamplaryPlotofTheorem3}
\end{figure*}
To simplify the determinant of the first matrix, it transformed into a lower diagonal matrix by applying the following Gaussian elimination steps
\begin{align*}
\small
\begin{split}
    \frac{\dimN_1\Tilde{w}_{\TypeIIoutlier,1}z_1^{-1}}{-\dimN_1w_1z_1^{-1}+1}R_2+R_1 &\rightarrow R_1\\
    \frac{\dimN_2\Tilde{w}_{\TypeIIoutlier,2}z_2^{-1}}{-\dimN_2w_2z_2^{-1}+1}R_3+R_1 &\rightarrow R_1\\
    &\vdots\\
    \frac{\dimN_{\blocknum}\Tilde{w}_{\TypeIIoutlier,\blocknum}z_{\blocknum}^{-1}}{-\dimN_{\blocknum}w_{\blocknum}z_{\blocknum}^{-1}+1}R_{\blocknum+1}+R_1 &\rightarrow R_1
\end{split}
\end{align*}
where $R_{\blocknum}$ denotes $\blocknum$th row.
Then, the simplified determinant yields
\begin{align*}
\begin{split}
    0=&c_{\TypeIIoutlier}(-\dimN_1w_1z_1^{-1}+1)(-\dimN_2w_2z_2^{-1}+1)\myydots(-\dimN_{\blocknum}w_{\blocknum}z_{\blocknum}^{-1}+1)z_{\TypeIIoutlier} z_1^{\dimN_1} z_2^{\dimN_2} \myydots z_{\blocknum}^{\dimN_{\blocknum}}
\end{split}
\end{align*}
where
\begin{align*}
\begin{split}
    c_{\TypeIIoutlier}=&\Bigg(1-\frac{\dimN_1\Tilde{w}_{\TypeIIoutlier,1}^2z_1^{-1}z_{\TypeIIoutlier}^{-1}}{-\dimN_1w_1z_1^{-1}+1}-\frac{\dimN_2\Tilde{w}_{\TypeIIoutlier,2}^2z_2^{-1}z_{\TypeIIoutlier}^{-1}}{-\dimN_2w_2z_2^{-1}+1}-\myydots-\frac{\dimN_{\blocknum}\Tilde{w}_{\TypeIIoutlier,\blocknum}^2z_{\blocknum}^{-1}z_{\TypeIIoutlier}^{-1}}{-\dimN_{\blocknum}w_{\blocknum}z_{\blocknum}^{-1}+1}   \Bigg).
\end{split}
\end{align*}
For $z_{\TypeIIoutlier}=\sum_{\indexj=1}^{\blocknum}\dimN_{\indexj}\Tilde{w}_{\TypeIIoutlier,\indexj}-\tilde{\lambda}\sum_{\indexj=1}^{\blocknum}\dimN_{\indexj}\Tilde{w}_{\TypeIIoutlier,\indexj}$ and $z_{\indexj}=\dimN_{\indexj}w_{\indexj}+\Tilde{w}_{\TypeIIoutlier,\indexj}-\tilde{\lambda}\big((\dimN_{\indexj}-1)w_{\indexj}+\Tilde{w}_{\TypeIIoutlier,\indexj}\big)$ such that $\indexj=1,\myydots,\blocknum$ the determinant yields\\
\resizebox{\linewidth}{!}{
  \begin{minipage}{\linewidth}
\begin{align*}
    \begin{split}
        0=&(-\dimN_1w_1z_1^{-1}+1)(-\dimN_2w_2z_2^{-1}+1)\myydots(-\dimN_{\blocknum}w_{\blocknum}z_{\blocknum}^{-1}+1)\cancel{z_{\TypeIIoutlier}} z_1^{\dimN_1} z_2^{\dimN_2} \myydots z_{\blocknum}^{\dimN_{\blocknum}}\cancel{z_{\TypeIIoutlier}^{-1}}\\&\Bigg(z_{\TypeIIoutlier}-\frac{\dimN_1\Tilde{w}_{\TypeIIoutlier,1}^2z_1^{-1}}{-\dimN_1w_1z_1^{-1}+1}-\frac{\dimN_2\Tilde{w}_{\TypeIIoutlier,2}^2z_2^{-1}}{-\dimN_2w_2z_2^{-1}+1}-\myydots-\frac{\dimN_{\blocknum}\Tilde{w}_{\TypeIIoutlier,\blocknum}^2z_{\blocknum}^{-1}}{-\dimN_{\blocknum}w_{\blocknum}z_{\blocknum}^{-1}+1}\Bigg)\\
        0=&(\Tilde{w}_{\TypeIIoutlier,1}-\tilde{\lambda}\Tilde{d}_1)(\Tilde{w}_{\TypeIIoutlier,2}-\tilde{\lambda}\Tilde{d}_2)\myydots(\Tilde{w}_{\TypeIIoutlier,\blocknum}-\tilde{\lambda}\Tilde{d}_{\blocknum})
        (\dimN_1w_1+\Tilde{w}_{\TypeIIoutlier,1}-\tilde{\lambda}\Tilde{d}_1)^{\dimN_1-1}(\dimN_2w_2+\Tilde{w}_{\TypeIIoutlier,2}-\tilde{\lambda}\Tilde{d}_2)^{\dimN_2-1}\myydots(\dimN_{\blocknum}w_{\blocknum}+\Tilde{w}_{\TypeIIoutlier,\blocknum}-\tilde{\lambda}\Tilde{d}_{\blocknum})^{\dimN_{\blocknum}-1}\\&\Bigg(\sum_{\indexj=1}^{\blocknum}\dimN_{\indexj}\Tilde{w}_{\TypeIIoutlier,\indexj}-\tilde{\lambda}\Tilde{d}_{\TypeIIoutlier}-\frac{\dimN_1\Tilde{w}_{\TypeIIoutlier,1}^2}{\Tilde{w}_{\TypeIIoutlier,1}-\tilde{\lambda}\Tilde{d}_1}-\frac{\dimN_2\Tilde{w}_{\TypeIIoutlier,2}^2}{\Tilde{w}_{\TypeIIoutlier,2}-\tilde{\lambda}\Tilde{d}_2}\myydots-\frac{\dimN_{\blocknum}\Tilde{w}_{\TypeIIoutlier,\blocknum}^2}{\Tilde{w}_{\TypeIIoutlier,\blocknum}-\tilde{\lambda}\Tilde{d}_{\blocknum}}\Bigg)\\
        0=&(\Tilde{w}_{\TypeIIoutlier,1}-\tilde{\lambda}\Tilde{d}_1)(\Tilde{w}_{\TypeIIoutlier,2}-\tilde{\lambda}\Tilde{d}_2)\myydots(\Tilde{w}_{\TypeIIoutlier,\blocknum}-\tilde{\lambda}\Tilde{d}_{\blocknum})
        (\dimN_1w_1+\Tilde{w}_{\TypeIIoutlier,1}-\tilde{\lambda}\Tilde{d}_1)^{\dimN_1-1}(\dimN_2w_2+\Tilde{w}_{\TypeIIoutlier,2}-\tilde{\lambda}\Tilde{d}_2)^{\dimN_2-1}\myydots(\dimN_{\blocknum}w_{\blocknum}+\Tilde{w}_{\TypeIIoutlier,\blocknum}-\tilde{\lambda}\Tilde{d}_{\blocknum})^{\dimN_{\blocknum}-1}\\&\Bigg(\dimN_1\Tilde{w}_{\TypeIIoutlier,1}-\frac{\dimN_1\Tilde{w}_{\TypeIIoutlier,1}^2}{\Tilde{w}_{\TypeIIoutlier,1}-\tilde{\lambda}\Tilde{d}_1}+\dimN_2\Tilde{w}_{\TypeIIoutlier,2}-\frac{\dimN_2\Tilde{w}_{\TypeIIoutlier,2}^2}{\Tilde{w}_{\TypeIIoutlier,2}-\tilde{\lambda}\Tilde{d}_2}\myydots+ \dimN_{\blocknum}\Tilde{w}_{\TypeIIoutlier,\blocknum}-\frac{\dimN_{\blocknum}\Tilde{w}_{\TypeIIoutlier,\blocknum}^2}{\Tilde{w}_{\TypeIIoutlier,\blocknum}-\tilde{\lambda}\Tilde{d}_{\blocknum}}-\tilde{\lambda}\Tilde{d}_{\TypeIIoutlier}\Bigg)\\
        0=&(\Tilde{w}_{\TypeIIoutlier,1}-\tilde{\lambda}\Tilde{d}_1)(\Tilde{w}_{\TypeIIoutlier,2}-\tilde{\lambda}\Tilde{d}_2)\myydots(\Tilde{w}_{\TypeIIoutlier,\blocknum}-\tilde{\lambda}\Tilde{d}_{\blocknum})
        (\dimN_1w_1+\Tilde{w}_{\TypeIIoutlier,1}-\tilde{\lambda}\Tilde{d}_1)^{\dimN_1-1}(\dimN_2w_2+\Tilde{w}_{\TypeIIoutlier,2}-\tilde{\lambda}\Tilde{d}_2)^{\dimN_2-1}\myydots(\dimN_{\blocknum}w_{\blocknum}+\Tilde{w}_{\TypeIIoutlier,\blocknum}-\tilde{\lambda}\Tilde{d}_{\blocknum})^{\dimN_{\blocknum}-1}\\&\Bigg(-\frac{\dimN_1\Tilde{w}_{\TypeIIoutlier,1}\tilde{\lambda}\Tilde{d}_1}{\Tilde{w}_{\TypeIIoutlier,1}-\tilde{\lambda}\Tilde{d}_1}-\frac{\dimN_2\Tilde{w}_{\TypeIIoutlier,2}\tilde{\lambda}\Tilde{d}_2}{\Tilde{w}_{\TypeIIoutlier,2}-\tilde{\lambda}\Tilde{d}_2}\myydots-\frac{\dimN_{\blocknum}\Tilde{w}_{\TypeIIoutlier,\blocknum}\tilde{\lambda}\Tilde{d}_{\blocknum}}{\Tilde{w}_{\TypeIIoutlier,\blocknum}-\tilde{\lambda}\Tilde{d}_{\blocknum}}-\tilde{\lambda}\Tilde{d}_{\TypeIIoutlier}\Bigg)\\
        0=&\tilde{\lambda}
        (\dimN_1w_1+\Tilde{w}_{\TypeIIoutlier,1}-\tilde{\lambda}\Tilde{d}_1)^{\dimN_1-1}(\dimN_2w_2+\Tilde{w}_{\TypeIIoutlier,2}-\tilde{\lambda}\Tilde{d}_2)^{\dimN_2-1}\myydots(\dimN_{\blocknum}w_{\blocknum}+\Tilde{w}_{\TypeIIoutlier,\blocknum}-\tilde{\lambda}\Tilde{d}_{\blocknum})^{\dimN_{\blocknum}-1}(\Tilde{w}_{\TypeIIoutlier,1}-\tilde{\lambda}\Tilde{d}_1)(\Tilde{w}_{\TypeIIoutlier,2}-\tilde{\lambda}\Tilde{d}_2)\myydots(\Tilde{w}_{\TypeIIoutlier,\blocknum}-\tilde{\lambda}\Tilde{d}_{\blocknum})\\&\Bigg(-\frac{\dimN_1\Tilde{w}_{\TypeIIoutlier,1}\Tilde{d}_1}{\Tilde{w}_{\TypeIIoutlier,1}-\tilde{\lambda}\Tilde{d}_1}-\frac{\dimN_2\Tilde{w}_{\TypeIIoutlier,2}\Tilde{d}_2}{\Tilde{w}_{\TypeIIoutlier,2}-\tilde{\lambda}\Tilde{d}_2}\myydots-\frac{\dimN_{\blocknum}\Tilde{w}_{\TypeIIoutlier,\blocknum}\Tilde{d}_{\blocknum}}{\Tilde{w}_{\TypeIIoutlier,\blocknum}-\tilde{\lambda}\Tilde{d}_{\blocknum}}-\Tilde{d}_{\TypeIIoutlier}\Bigg)\\
    \end{split}\\
\end{align*}
\end{minipage}}

Now, $\dimN+1-\blocknum$ number of eigenvalues can be written as
\begin{align*}
\small
\begin{cases}
\begin{split}
    &\dimN_1-1\hspace{1mm}\mathrm{elements\hspace{1mm}of}\hspace{1mm} \Tilde{\boldsymbol{\lambda}}\mathrm{\hspace{1mm}equal\hspace{1mm}to\hspace{2mm}} \frac{\dimN_1w_1+\Tilde{w}_{\TypeIIoutlier,1}}{\Tilde{d}_1}\\
    &\dimN_2-1\hspace{1mm}\mathrm{elements\hspace{1mm}of\hspace{1mm}} \Tilde{\boldsymbol{\lambda}}\underdot{\hspace{1mm}\mathrm{equal\hspace{1mm}to}\hspace{2mm}}\frac{\dimN_2w_2+\Tilde{w}_{\TypeIIoutlier,2}}{\Tilde{d}_2}\\
    &\\\vspace{-3mm}
    &\dimN_{\blocknum}-1\mathrm{\hspace{1mm}elements\hspace{1mm}of}\hspace{1mm}
    \Tilde{\boldsymbol{\lambda}}\mathrm{\hspace{1mm}equal\hspace{1mm}to\hspace{2mm}}\frac{\dimN_{\blocknum}w_{\blocknum}+\Tilde{w}_{\TypeIIoutlier,\blocknum}}{\Tilde{d}_\blocknum}\\
    &\mathrm{the\hspace{1mm}smallest}\mathrm{\hspace{1mm}element\hspace{1mm}of}\hspace{1mm}
    \Tilde{\boldsymbol{\lambda}}\mathrm{\hspace{1mm}equal\hspace{1mm}to\hspace{1mm}zero}
     \end{split}
\end{cases}
\end{align*}
and the remaining $\blocknum$ number of eigenvalues are roots of the equation
\begin{align*}
\begin{split}
0=&(\Tilde{w}_{\TypeIIoutlier,1}-\tilde{\lambda}\Tilde{d}_1)(\Tilde{w}_{\TypeIIoutlier,2}-\tilde{\lambda}\Tilde{d}_2)\myydots(\Tilde{w}_{\TypeIIoutlier,\blocknum}-\tilde{\lambda}\Tilde{d}_{\blocknum})\Bigg(-\frac{\dimN_1\Tilde{w}_{\TypeIIoutlier,1}\Tilde{d}_1}{\Tilde{w}_{\TypeIIoutlier,1}-\tilde{\lambda}\Tilde{d}_1}-\frac{\dimN_2\Tilde{w}_{\TypeIIoutlier,2}\Tilde{d}_2}{\Tilde{w}_{\TypeIIoutlier,2}-\tilde{\lambda}\Tilde{d}_2}\myydots-\frac{\dimN_{\blocknum}\Tilde{w}_{\TypeIIoutlier,\blocknum}\Tilde{d}_{\blocknum}}{\Tilde{w}_{\TypeIIoutlier,\blocknum}-\tilde{\lambda}\Tilde{d}_{\blocknum}}-\Tilde{d}_{\TypeIIoutlier}\Bigg)\\
0=&\prod_{\indexj=1}^{\blocknum}(\Tilde{w}_{\TypeIIoutlier,\indexj}-\Tilde{\lambda}\Tilde{d}_{\indexj})\Bigg(-\sum_{\indexj=1}^{\blocknum}\frac{\dimN_{\indexj}\Tilde{w}_{\TypeIIoutlier,\indexj}\Tilde{d}_{\indexj}}{\Tilde{w}_{\TypeIIoutlier,\indexj}-\Tilde{\lambda}\Tilde{d}_{\indexj}}-\Tilde{d}_{\TypeIIoutlier} \Bigg).
\end{split}
\end{align*}
\end{proof}

\subsection{A.2~Outlier Effects on Target Eigenvalues: Proof of Theorem 2}
\begin{proof}[\unskip\nopunct]
Let $\Tilde{\mathbf{W}}\in\mathbb{R}^{\dimN\times\dimN}$ and $\Tilde{\mathbf{L}}\in\mathbb{R}^{\dimN\times\dimN}$ denote $\blocknum$ block zero diagonal affinity matrix and associated Laplacian in which $\indexi$th block has similarity with remaining $\blocknum-1$ number of blocks. For simplicity, let $\indexi=1$, i.e. 
\vspace{2mm}
\begin{align*}
\Tilde{\mathbf{W}}=
\begin{bmatrix}
\begin{smallmatrix}
0& w_1&\myydots&w_1&\tilde{w}_{1,2}&\myydots & &\tilde{w}_{1,2} &\myydots & \tilde{w}_{1,\blocknum}& \myydots& &\tilde{w}_{1,\blocknum}\\
w_1&0&\myydots&w_1&\tilde{w}_{1,2} & & & \tilde{w}_{1,2} &\myydots & \tilde{w}_{1,\blocknum}& \myydots& &\tilde{w}_{1,\blocknum}\\
\vdots&\vdots&\ddots&&\vdots& & \ddots& \vdots & & & &\ddots &\\
w_1 &w_1 &\myydots & 0 &\tilde{w}_{1,2} &\myydots & &\tilde{w}_{1,2}&\myydots & \tilde{w}_{1,\blocknum}& \myydots& &\tilde{w}_{1,\blocknum}\\
\tilde{w}_{1,2}&\myydots & &\tilde{w}_{1,2} &0& w_2&\myydots&w_2& & & & &\\
\tilde{w}_{1,2} & & & \tilde{w}_{1,2} &w_2&0&\myydots&w_2& & & & &\\
\vdots& & \ddots& \vdots &\vdots&\vdots&\ddots&& & & & &\\
\tilde{w}_{1,2} &\myydots & &\tilde{w}_{1,2} &w_2 &w_2 &\myydots & 0 & & & & &\\
\vdots& & &\vdots &\vdots& \vdots& & &\ddots & &\vdots& & \vdots\\
\tilde{w}_{1,\blocknum}&\myydots & &\tilde{w}_{1,\blocknum} &&&&&&0&w_{\blocknum}&\myydots&w_{\blocknum}\\
\tilde{w}_{1,\blocknum}& & &\tilde{w}_{1,\blocknum} &&&&&\myydots&w_{\blocknum}&0& \myydots& w_{\blocknum} \\
\vdots& & &\vdots&&&&&&\vdots&\vdots&\ddots\\
\tilde{w}_{1,\blocknum}& & &\tilde{w}_{1,\blocknum} &&&&&\myydots&w_{\blocknum}& w_{\blocknum} &\myydots&0\\
\end{smallmatrix}
\end{bmatrix}\vspace{2mm}
\end{align*}
and
\begin{align*}
\Tilde{\mathbf{L}}=
\begin{bmatrix}
\begin{smallmatrix}
\Tilde{d}_1& -w_1&\myydots&-w_1&-\tilde{w}_{1,2}&\myydots & &-\tilde{w}_{1,2} &\myydots & -\tilde{w}_{1,\blocknum}& \myydots& &-\tilde{w}_{1,\blocknum}\\
-w_1&\Tilde{d}_1&\myydots&-w_1&-\tilde{w}_{1,2} & & & -\tilde{w}_{1,2} &\myydots & -\tilde{w}_{1,\blocknum}& \myydots& &-\tilde{w}_{1,\blocknum}\\
\vdots&\vdots&\ddots&&\vdots& & \ddots& \vdots & & & &\ddots &\\
-w_1 &-w_1 &\myydots & \Tilde{d}_1 &-\tilde{w}_{1,2} &\myydots & &-\tilde{w}_{1,2}&\myydots & -\tilde{w}_{1,\blocknum}& \myydots& &-\tilde{w}_{1,\blocknum}\\
-\tilde{w}_{1,2}&\myydots & &-\tilde{w}_{1,2} &\Tilde{d}_2& -w_2&\myydots&-w_2& & & & &\\
-\tilde{w}_{1,2} & & & -\tilde{w}_{1,2} &-w_2&\Tilde{d}_2&\myydots&-w_2& & & & &\\
\vdots& & \ddots& \vdots &\vdots&\vdots&\ddots&& & & & &\\
-\tilde{w}_{1,2} &\myydots & &-\tilde{w}_{1,2} &-w_2 &-w_2 &\myydots & \Tilde{d}_2 & & & & &\\
\vdots& & &\vdots &\vdots& \vdots& & &\ddots & &\vdots& & \vdots\\
-\tilde{w}_{1,\blocknum}&\myydots & &-\tilde{w}_{1,\blocknum} &&&&&&\Tilde{d}_{\blocknum}&-w_{\blocknum}&\myydots&-w_{\blocknum}\\
-\tilde{w}_{1,\blocknum}& & &-\tilde{w}_{1,\blocknum} &&&&&\myydots&-w_{\blocknum}&\Tilde{d}_{\blocknum}& \myydots&-w_{\blocknum} \\
\vdots& & &\vdots&&&&&&\vdots&\vdots&\ddots\\
-\tilde{w}_{1,\blocknum}& & &-\tilde{w}_{1,\blocknum} &&&&&\myydots&-w_{\blocknum}&-w_{\blocknum} &\myydots&\Tilde{d}_{\blocknum}\\
\end{smallmatrix}
\end{bmatrix}\vspace{2mm}
\end{align*}
where $\Tilde{d}_1=(\dimN_1-1)w_1+\sum_{\indexj=2}^{\blocknum}\dimN_{\indexj}\tilde{w}_{1,\indexj}$ and $\Tilde{d}_{\indexj}=(\dimN_{\indexj}-1)w_{\indexj}+\dimN_1\tilde{w}_{1,\indexj}, \indexj=2,\myydots,\blocknum$.
To estimate the eigenvalues of the Laplacian matrix $\Tilde{\mathbf{L}}$, $\mathrm{det}(\Tilde{\mathbf{L}}-\Tilde{\lambda}\Tilde{\mathbf{D}})=0$ is considered which can equivalently be written in matrix form as follows
\begin{align*}
\mathrm{det}(\Tilde{\mathbf{L}}-\Tilde{\lambda}\Tilde{\mathbf{D}})=
\begin{vmatrix}
\begin{smallmatrix}
\Tilde{d}_1-\Tilde{\lambda}\Tilde{d}_1& -w_1&\myydots&-w_1&-\tilde{w}_{1,2}&\myydots & &-\tilde{w}_{1,2} &\myydots & -\tilde{w}_{1,\blocknum}& \myydots& &-\tilde{w}_{1,\blocknum}\\
-w_1&\Tilde{d}_1-\Tilde{\lambda}\Tilde{d}_1&\myydots&-w_1&-\tilde{w}_{1,2} & & & -\tilde{w}_{1,2} &\myydots & -\tilde{w}_{1,\blocknum}& \myydots& &-\tilde{w}_{1,\blocknum}\\
\vdots&\vdots&\ddots&&\vdots& & \ddots& \vdots & & & &\ddots &\\
-w_1 &-w_1 &\myydots & \Tilde{d}_1-\Tilde{\lambda}\Tilde{d}_1 &-\tilde{w}_{1,2} &\myydots & &-\tilde{w}_{1,2}&\myydots & -\tilde{w}_{1,\blocknum}& \myydots& &-\tilde{w}_{1,\blocknum}\\
-\tilde{w}_{1,2}&\myydots & &-\tilde{w}_{1,2} &\Tilde{d}_2-\Tilde{\lambda}\Tilde{d}_2& -w_2&\myydots&-w_2& & & & &\\
-\tilde{w}_{1,2} & & & -\tilde{w}_{1,2} &-w_2&\Tilde{d}_2-\Tilde{\lambda}\Tilde{d}_2&\myydots&-w_2& & & & &\\
\vdots& & \ddots& \vdots &\vdots&\vdots&\ddots&& & & & &\\
-\tilde{w}_{1,2} &\myydots & &-\tilde{w}_{1,2} &-w_2 &-w_2 &\myydots & \Tilde{d}_2-\Tilde{\lambda}\Tilde{d}_2 & & & & &\\
\vdots& & &\vdots &\vdots& \vdots& & &\ddots & &\vdots& & \vdots\\
-\tilde{w}_{1,\blocknum}&\myydots & &-\tilde{w}_{1,\blocknum} &&&&&&\Tilde{d}_{\blocknum}-\Tilde{\lambda}\Tilde{d}_{\blocknum}&-w_{\blocknum}&\myydots&-w_{\blocknum}\\
-\tilde{w}_{1,\blocknum}& & &-\tilde{w}_{1,\blocknum} &&&&&\myydots&-w_{\blocknum}&\Tilde{d}_{\blocknum}-\Tilde{\lambda}\Tilde{d}_{\blocknum}& \myydots&-w_{\blocknum} \\
\vdots& & &\vdots&&&&&&\vdots&\vdots&\ddots\\
-\tilde{w}_{1,\blocknum}& & &-\tilde{w}_{1,\blocknum} &&&&&\myydots&-w_{\blocknum}&-w_{\blocknum} &\myydots&\Tilde{d}_{\blocknum}-\Tilde{\lambda}\Tilde{d}_{\blocknum}\\
\end{smallmatrix}
\end{vmatrix}=0
\end{align*}
To simplify this determinant, the matrix determinant lemma \cite{matrixdeterminantlemma} can be generalized as follows\footnote{For a detailed information about generalization of the matrix determinant lemma, see Appendix~D.}
\begin{align*}
\mathrm{det}(\mathbf{H}+\mathbf{U}\mathbf{V}^{\top})=\mathrm{det}(\mathbf{H})\mathrm{det}(\mathbf{I}+\mathbf{V}^{\top}\mathbf{H}^{\dagger}\mathbf{U}) 
\end{align*}
where $\mathbf{H}\in\mathbb{R}^{(\dimN+1)\times(\dimN+1)}$ denotes an invertible matrix, $\mathbf{I}$ is the identity matrix and $\mathbf{U},\mathbf{V}\in\mathbb{R}^{(\dimN+1)\times(\dimN+1)}$. Then, for $\mathrm{det}(\tilde{\mathbf{L}}-\tilde{\lambda}\tilde{\mathbf{D}})=\mathrm{det}(\mathbf{H}+\mathbf{U}\mathbf{V}^{\top})=0$, it follows that\vspace{5mm}\\
\resizebox{\linewidth}{!}{
  \begin{minipage}{\linewidth}
\begin{align*}
\begin{split}
\hspace{-0.7mm}0\hspace{-0.7mm}&=
\mathrm{det}\begin{pmatrix}
\begin{bmatrix}
\begin{smallmatrix}
z_1&\myydots &&&& & & & & &\myydots&\hspace{-1mm}0\\\vspace{-1mm}
&\hspace{3mm}\ddots&&&& & && & & & \hspace{-1mm}\vdots\\\vspace{0.2mm}
& & &\hspace{-1.5mm}z_1 & & & & & & & & & \\\vspace{-0.6mm}
&&& &\hspace{-2mm}z_2& & & & & & &\\\vspace{-0.6mm}
&&& &&\hspace{-0.5mm}\ddots& && & & & \\\vspace{-0.6mm}
& & & && & &\hspace{-1mm}z_2  & & & & & \\\vspace{-1mm}
& & & && & &  &\hspace{-1mm}\ddots& & & & \\\vspace{-1mm}
&&& && & & &\hspace{3mm}z_{\blocknum}& & & \\\vspace{-1mm}
\vdots&&& &&& & & &\hspace{-1mm}\ddots& & \\\vspace{-1mm}
0&\myydots & & && & && & & &\hspace{-2mm}z_{\blocknum}  &  \\\vspace{0.6mm}
\end{smallmatrix}
\end{bmatrix}+
\begin{bmatrix}
\begin{smallmatrix}
\bovermat{$\blocknum$}{\hspace{-1.3mm}1&0&\myydots } &&&\hspace{-2mm}&&0\hspace{1mm}&\bovermat{$\dimN-\blocknum$}{0&\myydots&} &\myydots &0\\
\vdots&\vdots&&&&\hspace{-2mm}&&\vdots&\vdots&\ddots& &\iddots &\vdots\\
1&0& & & &\hspace{-2mm} &&&&& & &\\
0&1& &&&&\hspace{-2mm}&&&& & &\\
\vdots&\vdots & &&\hspace{-2mm}&  &&&&& & &\\
 &1 &&\hspace{-2mm}& & &&&&& & &\\
 &0 && &\hspace{-2mm} & &&0&&& & &\\
 &&&&\hspace{-2mm} & &&1&&& & &\\
\vdots&\vdots & &&& \hspace{-2mm} &&\vdots&\vdots&\iddots& &\ddots &\vdots\\
0 &0 &\myydots& &\hspace{-2mm}&& & 1&0&\myydots& &\myydots &0\\
\end{smallmatrix}
\end{bmatrix}
\begin{bmatrix}
\begin{smallmatrix}
\bovermat{$\dimN_1$}{-w_1&\myydots &}-w_1&\bovermat{$\dimN_2$}{-\tilde{w}_{1,2}&\myydots&}-\tilde{w}_{1,2} &\myydots&\bovermat{$\dimN_{\blocknum}$}{-\tilde{w}_{1,\blocknum} &\myydots &}-\tilde{w}_{1,\blocknum}\\
-\tilde{w}_{1,2}&\myydots&-\tilde{w}_{1,2}&-w_2&\myydots &-w_2& &&\myydots &0 \\
\vdots& & & & & & & & &\vdots &\\
-\tilde{w}_{1,\blocknum}&\myydots &-\tilde{w}_{1,\blocknum}&0&\myydots& & &-w_{\blocknum}&\myydots &-w_{\blocknum} \\
0&\myydots&& && & & &\myydots &0\\
\vdots&\ddots&& &&& &&\iddots&\vdots  \\\vspace{-2mm}
& & & && & && & & \\
\vdots&\iddots & & && & &  &\ddots&\vdots & \\
0&\myydots&& && & & &\myydots&  0\\
\end{smallmatrix}
\end{bmatrix}
\end{pmatrix}\\
\\
\hspace{-0.7mm}0\hspace{-0.7mm}&=\hspace{-0.7mm}\mathrm{det}(\mathbf{H})
\mathrm{det}\hspace{-1.1mm}\begin{pmatrix}
\hspace{-1.5mm}\mathbf{I}\hspace{-0.7mm}+\hspace{-1.7mm}
\begin{bmatrix}
\begin{smallmatrix}
-w_1z_1^{-1}&\myydots &-w_1z_1^{-1}&-\tilde{w}_{1,2}z_2^{-1}&\myydots&-\tilde{w}_{1,2}z_2^{-1} &\myydots&-\tilde{w}_{1,\blocknum}z_{\blocknum}^{-1} &\myydots &-\tilde{w}_{1,\blocknum}z_{\blocknum}^{-1}\\
-\tilde{w}_{1,2}z_1^{-1}&\myydots&-\tilde{w}_{1,2}z_1^{-1}&-w_2z_2^{-1}&\myydots &-w_2z_2^{-1}& &&\myydots &0 \\
\vdots& & & & & & & & &\vdots &\\
-\tilde{w}_{1,\blocknum}z_1^{-1}&\myydots &-\tilde{w}_{1,\blocknum}z_1^{-1}&0&\myydots& & &-w_{\blocknum}z_{\blocknum}^{-1}&\myydots &-w_{\blocknum}z_{\blocknum}^{-1} \\
0&\myydots&& && & & &\myydots &0\\
\vdots&\ddots&& &&& &&\iddots &\vdots \\\vspace{-3mm}
& & & && & && & & \\
\vdots&\iddots & & && & &  &\ddots&\vdots & \\
0&\myydots&& && & & &\myydots&0  \\
\end{smallmatrix}
\end{bmatrix}
\hspace{-1.5mm}
\begin{bmatrix}
\begin{smallmatrix}
1&0&\myydots&&&\hspace{-1mm}&&0&0&\myydots& &\myydots &0\\
\vdots&\vdots&&&&\hspace{-1mm}&&\vdots&\vdots&\ddots& &\iddots &\vdots\\
1&0& & & &\hspace{-1mm} &&&&& & &\\
0&1& &&&&\hspace{-1mm}&&&& & &\\
\vdots&\vdots & &&\hspace{-1mm}&  &&&&& & &\\
 &1 &&\hspace{-1mm}& & &&&&& & &\\
 &0 && &\hspace{-1mm} & &&0&&& & &\\
 &&&&\hspace{-1mm} & &&1&&& & &\\
\vdots&\vdots & &&& \hspace{-1mm} &&\vdots&\vdots&\iddots& &\ddots &\vdots\\
0 &0 &\myydots& &\hspace{-1mm}&& & 1&0&\myydots& &\myydots &0\\
\end{smallmatrix}
\end{bmatrix}
\hspace{-1mm}
\end{pmatrix}\\
\\
0&=\mathrm{det}(\mathbf{H})
\begin{vmatrix}
\begin{smallmatrix}
-\dimN_1w_1z_1^{-1}+1 & -\dimN_2\tilde{w}_{1,2}z_2^{-1} & -\dimN_3\tilde{w}_{1,3}z_3^{-1} & \myydots & -\dimN_{\blocknum}\tilde{w}_{1,\blocknum}z_{\blocknum}^{-1}& 0 &\myydots &\hspace{1.5cm} &\myydots & 0\\
-\dimN_1\tilde{w}_{1,2}z_1^{-1} & -\dimN_2w_2z_2^{-1}+1 & 0 & \myydots & 0& \vdots &\ddots &\hspace{1.5cm} &\iddots & \vdots\\
-\dimN_1\tilde{w}_{1,3}z_1^{-1} & 0 &-\dimN_3w_3z_3^{-1}+1 & \myydots & 0 &  &&\hspace{1.5cm} & & \\
\vdots & & & & \vdots&\vdots &\iddots &\hspace{1.5cm} &\ddots & \vdots\\
-\dimN_1\tilde{w}_{1,\blocknum}z_1^{-1} &0 &\myydots & &-\dimN_{\blocknum}w_{\blocknum}z_{\blocknum}^{-1}+1& 0 &\myydots &\hspace{1.5cm} &\myydots & 0\\
0 &\myydots & &\myydots &0& 1 & &\hspace{1.5cm} &\myydots & 0\\
\vdots &\ddots & &\iddots &\vdots&  & &\hspace{0.25cm}\ddots\hspace{1.25cm} & & \vdots\\
 & & & &&  & &\hspace{1.5cm} && \\
\vdots &\iddots & &\ddots &\vdots&  \vdots& &\hspace{1.5cm} &\hspace{-0.25cm}\ddots\hspace{0.25cm} & \vdots\\
0 &\myydots & &\myydots &0& 0 &\myydots&\hspace{1.5cm} & & 1\\
\end{smallmatrix}
\end{vmatrix}
\end{split}
\end{align*}
\end{minipage}}\\
where $z_1=\dimN_1w_1+\sum_{\indexj=2}^{\blocknum}\dimN_{\indexj}\tilde{w}_{1,\indexj}-\Tilde{\lambda}\big((\dimN_1-1)w_1+\sum_{\indexj=2}^{\blocknum}\dimN_{\indexj}\tilde{w}_{1,\indexj}\big)$ and $z_{\indexj}=\dimN_{\indexj}w_{\indexj}+\dimN_1\tilde{w}_{1,\indexj}-\Tilde{\lambda}\big((\dimN_{\indexj}-1)w_{\indexj}+\dimN_1\tilde{w}_{1,\indexj} \big)$ with $\indexj=2,\myydots,\blocknum$. Using
determinant properties of block matrices \cite{detblockmatrices}, it holds that\vspace{2mm}
\begin{align*}
    \begin{split}
        0&=\mathrm{det}(\mathbf{H})
\begin{vmatrix}
\begin{smallmatrix}
-\dimN_1w_1z_1^{-1}+1 & -\dimN_2\tilde{w}_{1,2}z_2^{-1} & -\dimN_3\tilde{w}_{1,3}z_3^{-1} & \myydots & -\dimN_{\blocknum}\tilde{w}_{1,\blocknum}z_{\blocknum}^{-1}\\
-\dimN_1\tilde{w}_{1,2}z_1^{-1} & -\dimN_2w_2z_2^{-1}+1 & 0 & \myydots & 0\\
-\dimN_1\tilde{w}_{1,3}z_1^{-1} & 0 &-\dimN_3w_3z_3^{-1}+1 & \myydots & 0 \\
\vdots & & & & \vdots\\
-\dimN_1\tilde{w}_{1,\blocknum}z_1^{-1}&0 &\myydots & &-\dimN_{\blocknum}w_{\blocknum}z_{\blocknum}^{-1}+1\\
\end{smallmatrix}
\end{vmatrix}
    \end{split}.
\end{align*}
\newpage
To simplify the determinant of the first matrix, it transformed into a lower diagonal matrix by applying the following Gaussian elimination steps
\begin{align*}
\small
\begin{split}
    \frac{\dimN_2\tilde{w}_{1,2}z_2^{-1}}{-\dimN_2w_2z_2^{-1}+1}R_2+R_1 &\rightarrow R_1\\
    \frac{\dimN_3\tilde{w}_{1,3}z_3^{-1}}{-\dimN_3w_3z_3^{-1}+1}R_3+R_1 &\rightarrow R_1\\
    &\vdots\\
    \frac{\dimN_{\blocknum}\tilde{w}_{1,\blocknum}z_{\blocknum}^{-1}}{-\dimN_{\blocknum}w_{\blocknum}z_{\blocknum}^{-1}+1}R_{\blocknum}+R_1 &\rightarrow R_1
\end{split}
\end{align*}
where $R_{\blocknum}$ denotes $\blocknum$th row.
Then, the simplified determinant yields
\begin{align*}
    \begin{split}
        0&=\mathrm{det}(\mathbf{H})
\begin{vmatrix}
\begin{smallmatrix}
c_1 & 0 & 0 & \myydots & 0 \\
-\dimN_1\tilde{w}_{1,2}z_1^{-1} & -\dimN_2w_2z_2^{-1}+1 & 0 & \myydots & 0\\
-\dimN_1\tilde{w}_{1,3}z_1^{-1} & 0 &-\dimN_3w_3z_3^{-1}+1 & \myydots & 0 \\
\vdots & & & & \vdots\\
-\dimN_1\tilde{w}_{1,\blocknum}z_1^{-1} &0 &\myydots & &-\dimN_{\blocknum}w_{\blocknum}z_{\blocknum}^{-1}+1\\
\end{smallmatrix}
\end{vmatrix}
    \end{split}
\end{align*}
where $c_1$ equals to\vspace{-2mm}
\begin{align*}
c_1=-\dimN_1w_1z_1^{-1}+1- \frac{\dimN_2\tilde{w}_{1,2}z_2^{-1}\dimN_1\tilde{w}_{1,2}z_1^{-1}}{-\dimN_2w_2z_2^{-1}+1}-\myydots-\frac{\dimN_{\blocknum}\tilde{w}_{1,\blocknum}z_{\blocknum}^{-1}\dimN_1\tilde{w}_{1,\blocknum}z_1^{-1}}{-\dimN_{\blocknum}w_{\blocknum}z_{\blocknum}^{-1}+1}. 
\end{align*}
For $z_1=\dimN_1w_1+\sum_{\indexj=2}^{\blocknum}\dimN_{\indexj}\tilde{w}_{1,\indexj}-\Tilde{\lambda}\big((\dimN_1-1)w_1+\sum_{\indexj=2}^{\blocknum}\dimN_{\indexj}\tilde{w}_{1,\indexj}\big)$ and $z_{\indexj}=\dimN_{\indexj}w_{\indexj}+\dimN_1\tilde{w}_{1,\indexj}-\Tilde{\lambda}\big((\dimN_{\indexj}-1)w_{\indexj}+\dimN_1\tilde{w}_{1,\indexj} \big)$ with $\indexj=2,\myydots,\blocknum$, the determinant $\mathrm{det}(\Tilde{\mathbf{L}}-\Tilde{\lambda}\Tilde{\mathbf{D}})=0$ yields
\begin{align*}
\begin{split}
   0=&z_1^{\dimN_1}z_2^{\dimN_2}\myydots z_{\blocknum}^{\dimN_{\blocknum}}(-\dimN_2w_2z_2^{-1}+1)\myydots(-\dimN_{\blocknum}w_{\blocknum}z_{\blocknum}^{-1}+1)c_1\\
   0=&z_1^{\dimN_1}z_2^{\dimN_2-1}\myydots z_{\blocknum}^{\dimN_{\blocknum}-1}(-\dimN_2w_2+z_2)\myydots(-\dimN_{\blocknum}w_{\blocknum}+z_{\blocknum})c_1\\
   0=&z_1^{\dimN_1}z_2^{\dimN_2-1}\myydots z_{\blocknum}^{\dimN_{\blocknum}-1}(-\dimN_2w_2+z_2)\myydots(-\dimN_{\blocknum}w_{\blocknum}+z_{\blocknum})z_1^{-1}\Bigg(-\dimN_1w_1+z_1- \frac{\dimN_1\dimN_2\tilde{w}_{1,2}^2z_2^{-1}}{-\dimN_2w_2z_2^{-1}+1}-\myydots-\frac{\dimN_1\dimN_{\blocknum}\tilde{w}_{1,\blocknum}^2z_{\blocknum}^{-1}}{-\dimN_{\blocknum}w_{\blocknum}z_{\blocknum}^{-1}+1}\Bigg)\\
   0=&z_1^{\dimN_1-1}z_2^{\dimN_2-1}\myydots z_{\blocknum}^{\dimN_{\blocknum}-1}(-\dimN_2w_2+z_2)\myydots(-\dimN_{\blocknum}w_{\blocknum}+z_{\blocknum})\Bigg(-\dimN_1w_1+z_1- \frac{\dimN_1\dimN_2\tilde{w}_{1,2}^2}{z_2-\dimN_2w_2}-\myydots-\frac{\dimN_1\dimN_{\blocknum}\tilde{w}_{1,\blocknum}^2}{z_{\blocknum}-\dimN_{\blocknum}w_{\blocknum}}\Bigg)\\
   0=&\Big(\dimN_1w_1+\sum_{\indexj=2}^{\blocknum}\dimN_{\indexj}\tilde{w}_{1,\indexj}-\Tilde{\lambda}\Tilde{d}_1\Big)^{\dimN_1-1}\Big(\dimN_2w_2+\dimN_1\tilde{w}_{1,2}-\Tilde{\lambda}\Tilde{d}_2\Big)^{\dimN_2-1}\myydots\Big(\dimN_{\blocknum}w_{\blocknum}+\dimN_1\tilde{w}_{1,\blocknum}-\Tilde{\lambda}\Tilde{d}_{\blocknum}\Big)^{\dimN_{\blocknum}-1}\\&(\dimN_1\tilde{w}_{1,2}-\Tilde{\lambda}\Tilde{d}_2)\myydots(\dimN_1\tilde{w}_{1,\blocknum}-\Tilde{\lambda}\Tilde{d}_{\blocknum})\Bigg(\sum_{\indexj=2}^{\blocknum}\dimN_{\indexj}\tilde{w}_{1,\indexj}-\tilde{\lambda}\Tilde{d}_1-\frac{\dimN_1\dimN_2\tilde{w}^2_{1,2}}{\dimN_1\tilde{w}_{1,2}-\Tilde{\lambda}\Tilde{d}_2}-\myydots-\frac{\dimN_1\dimN_{\blocknum}\tilde{w}^2_{1,\blocknum}}{\dimN_1\tilde{w}_{1,\blocknum}-\Tilde{\lambda}\Tilde{d}_{\blocknum}}\Bigg)\\
      0=&\Big(\dimN_1w_1+\sum_{\indexj=2}^{\blocknum}\dimN_{\indexj}\tilde{w}_{1,\indexj}-\Tilde{\lambda}\Tilde{d}_1\Big)^{\dimN_1-1}\Big(\dimN_2w_2+\dimN_1\tilde{w}_{1,2}-\Tilde{\lambda}\Tilde{d}_2\Big)^{\dimN_2-1}\myydots\Big(\dimN_{\blocknum}w_{\blocknum}+\dimN_1\tilde{w}_{1,\blocknum}-\Tilde{\lambda}\Tilde{d}_{\blocknum}\Big)^{\dimN_{\blocknum}-1}\\&(\dimN_1\tilde{w}_{1,2}-\Tilde{\lambda}\Tilde{d}_2)\myydots(\dimN_1\tilde{w}_{1,\blocknum}-\Tilde{\lambda}\Tilde{d}_{\blocknum})\Bigg(\dimN_2\tilde{w}_{1,2}- \frac{\dimN_1\dimN_2\tilde{w}^2_{1,2}}{\dimN_1\tilde{w}_{1,2}-\Tilde{\lambda}\Tilde{d}_2}+\myydots+\dimN_{\blocknum}\tilde{w}_{1,\blocknum}-\frac{\dimN_1\dimN_{\blocknum}\tilde{w}^2_{1,\blocknum}}{\dimN_1\tilde{w}_{1,\blocknum}-\Tilde{\lambda}\Tilde{d}_{\blocknum}}-\tilde{\lambda}\Tilde{d}_1\Bigg)\\
     0=&\Big(\dimN_1w_1+\sum_{\indexj=2}^{\blocknum}\dimN_{\indexj}\tilde{w}_{1,\indexj}-\Tilde{\lambda}\Tilde{d}_1\Big)^{\dimN_1-1}\Big(\dimN_2w_2+\dimN_1\tilde{w}_{1,2}-\Tilde{\lambda}\Tilde{d}_2\Big)^{\dimN_2-1}\myydots\Big(\dimN_{\blocknum}w_{\blocknum}+\dimN_1\tilde{w}_{1,\blocknum}-\Tilde{\lambda}\Tilde{d}_{\blocknum}\Big)^{\dimN_{\blocknum}-1}\\&(\dimN_1\tilde{w}_{1,2}-\Tilde{\lambda}\Tilde{d}_2)\myydots(\dimN_1\tilde{w}_{1,\blocknum}-\Tilde{\lambda}\Tilde{d}_{\blocknum})\Bigg(-\frac{\tilde{\lambda}\Tilde{d}_2\dimN_2\tilde{w}_{1,2}}{\dimN_1\tilde{w}_{1,2}-\Tilde{\lambda}\Tilde{d}_2}\myydots-\frac{\tilde{\lambda}\Tilde{d}_{\blocknum}\dimN_{\blocknum}\tilde{w}_{1,\blocknum}}{\dimN_1\tilde{w}_{1,\blocknum}-\Tilde{\lambda}\Tilde{d}_{\blocknum}}-\tilde{\lambda}\Tilde{d}_1\Bigg)\\
      0=&\tilde{\lambda}\Big(\dimN_1w_1+\sum_{\indexj=2}^{\blocknum}\dimN_{\indexj}\tilde{w}_{1,\indexj}-\Tilde{\lambda}\Tilde{d}_1\Big)^{\dimN_1-1}\Big(\dimN_2w_2+\dimN_1\tilde{w}_{1,2}-\Tilde{\lambda}\Tilde{d}_2\Big)^{\dimN_2-1}\myydots\Big(\dimN_{\blocknum}w_{\blocknum}+\dimN_1\tilde{w}_{1,\blocknum}-\Tilde{\lambda}\Tilde{d}_{\blocknum}\Big)^{\dimN_{\blocknum}-1}\\&(\dimN_1\tilde{w}_{1,2}-\Tilde{\lambda}\Tilde{d}_2)\myydots(\dimN_1\tilde{w}_{1,\blocknum}-\Tilde{\lambda}\Tilde{d}_{\blocknum})\Bigg(-\frac{\Tilde{d}_2\dimN_2\tilde{w}_{1,2}}{\dimN_1\tilde{w}_{1,2}-\Tilde{\lambda}\Tilde{d}_2}\myydots-\frac{\Tilde{d}_{\blocknum}\dimN_{\blocknum}\tilde{w}_{1,\blocknum}}{\dimN_1\tilde{w}_{1,\blocknum}-\Tilde{\lambda}\Tilde{d}_{\blocknum}}-\Tilde{d}_1\Bigg)\\
\end{split}
\end{align*}

Based on this, $\dimN+1-\blocknum$ number of eigenvalues are
\begin{align*}
\begin{cases}
\small
\begin{split}
    &\dimN_{\indexi}-1\hspace{1mm}\mathrm{elements\hspace{1mm}of}\hspace{1mm} \Tilde{\boldsymbol{\lambda}}\hspace{1mm}\mathrm{equal\hspace{1mm}to\hspace{2mm}}\frac{\dimN_{\indexi}w_{\indexi}+\sum\limits_{\substack{\indexj=1 \\ \indexj\neq \indexi}}^{\blocknum}\dimN_{\indexj}\tilde{w}_{\indexi,\indexj}}{\Tilde{d}_{\indexi}}\\
    &\dimN_{\indexj}-1\hspace{1mm}\mathrm{elements\hspace{1mm}of}\hspace{1mm} \Tilde{\boldsymbol{\lambda}}\hspace{1mm}\underdot{\mathrm{equal\hspace{1mm}to\hspace{2mm}}}\frac{\dimN_{\indexj}w_{\indexj}+\dimN_{\indexi}\tilde{w}_{\indexi,\indexj}}{\Tilde{d}_{\indexj}}\\
    &\\\vspace{-2mm}
    &\dimN_{\blocknum}-1\hspace{1mm}\mathrm{elements\hspace{1mm}of}\hspace{1mm} \Tilde{\boldsymbol{\lambda}}\hspace{1mm}\mathrm{equal\hspace{1mm}to\hspace{2mm}}\frac{\dimN_{\blocknum}w_{\blocknum}+\dimN_{\indexi}\tilde{w}_{\indexi,\blocknum}}{\Tilde{d}_{\blocknum}}\\
    &\mathrm{the\hspace{1mm} smallest}\hspace{1mm}\mathrm{element\hspace{1mm}of}\hspace{1mm} \Tilde{\boldsymbol{\lambda}}\hspace{1mm}\mathrm{equal\hspace{1mm}to\hspace{1mm}zero}
    \end{split}
\end{cases}
\end{align*}
and the remaining $\blocknum\hspace{-0.45mm}-\hspace{-0.45mm}1$ eigenvalues are roots of the equation\vspace{-1mm}
\begin{align*}
\begin{split}
0=&(\dimN_1\tilde{w}_{1,2}-\Tilde{\lambda}\Tilde{d}_2)\myydots(\dimN_1\tilde{w}_{1,\blocknum}-\Tilde{\lambda}\Tilde{d}_{\blocknum})\Bigg(-\frac{\Tilde{d}_2\dimN_2\tilde{w}_{1,2}}{\dimN_1\tilde{w}_{1,2}-\Tilde{\lambda}\Tilde{d}_2}\myydots-\frac{\Tilde{d}_{\blocknum}\dimN_{\blocknum}\tilde{w}_{1,\blocknum}}{\dimN_1\tilde{w}_{1,\blocknum}-\Tilde{\lambda}\Tilde{d}_{\blocknum}}-\Tilde{d}_1\Bigg)\\
0=&\prod\limits_{\substack{\indexj=1 \\ \indexj\neq \indexi}}^{\blocknum}(\dimN_{\indexi}\tilde{w}_{\indexi,\indexj}-\Tilde{\lambda}\Tilde{d}_{\indexj})\Bigg(-\sum\limits_{\substack{\indexj=1 \\ \indexj\neq \indexi}}^{\blocknum}\frac{\Tilde{d}_{\indexj}\dimN_{\indexj}\tilde{w}_{\indexi,\indexj}}{\dimN_{\indexi}\tilde{w}_{\indexi,\indexj}-\Tilde{\lambda}\Tilde{d}_{\indexj}}-\Tilde{d}_{\indexi}\Bigg)
\end{split}\vspace{-3mm}
\end{align*}
where $\indexi=1$ and $\indexj=2\myydots,\blocknum$.
\end{proof}
\begin{figure*}[tbp!]
\vspace{-2mm}
  \centering
  \captionsetup{justification=centering}
\subfloat[ $\tilde{G}=\{\tilde{V},\tilde{E},\tilde{\mathbf{W}}\}$ ]{\includegraphics[trim={0cm 0cm 0cm 0cm},clip,width=4.25cm]{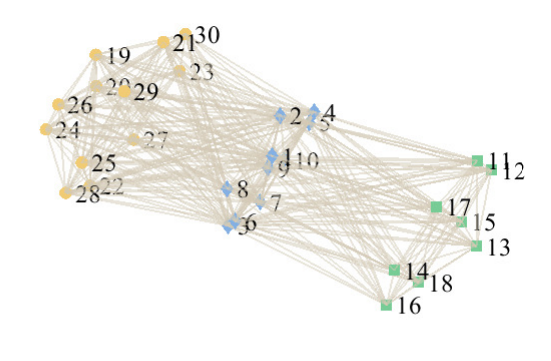}}
\subfloat[$\tilde{\mathbf{W}}\in\mathbb{R}^{\dimN\times\dimN}$]{\includegraphics[trim={0mm 0mm 0mm 0mm},clip,width=4cm]{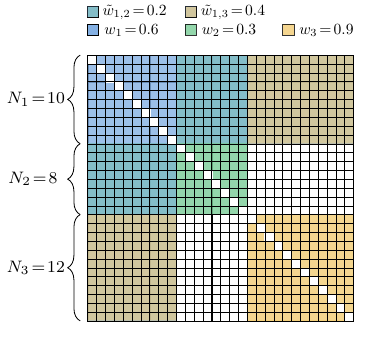}}\hspace{2mm}
\subfloat[$\tilde{\mathbf{L}}\in\mathbb{R}^{\dimN\times\dimN}$]{\includegraphics[trim={0mm 0mm 0mm 0mm},clip,width=4.5cm]{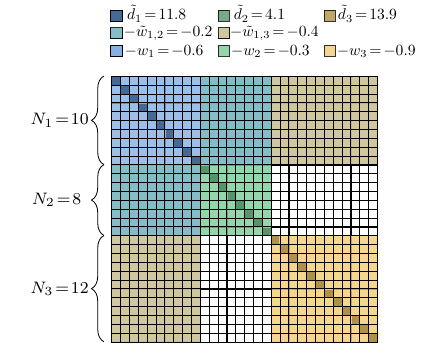}}
\subfloat[$\tilde{\boldsymbol{\lambda}}\in\mathbb{R}^{\dimN}$]{\includegraphics[trim={0mm 0mm 0mm 0mm},clip,width=4.75cm]{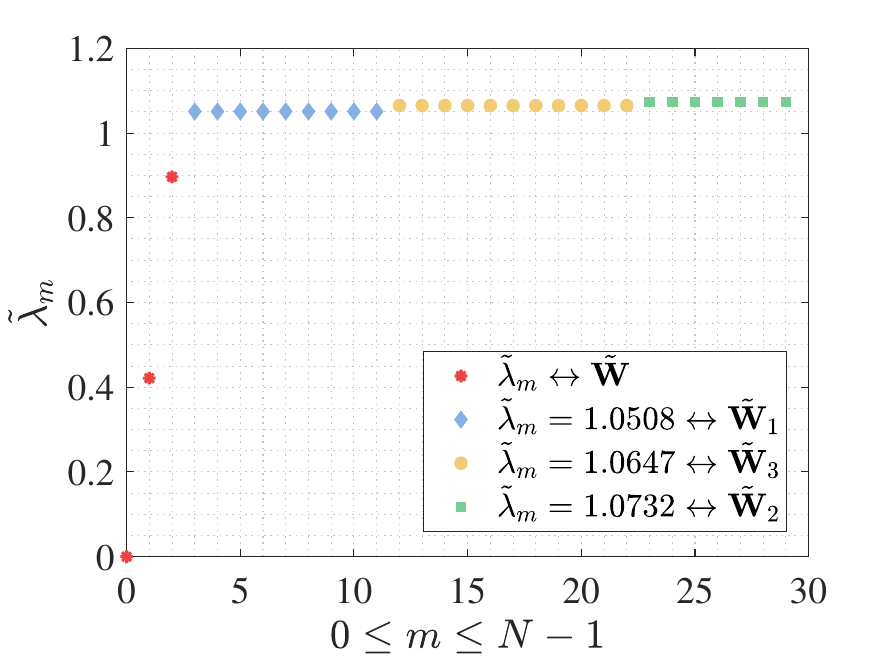}}
\\
\caption{Examplary plot of Theorem~2 ($\mathbf{n}=[10,8,12]^{\top}\in\mathbb{R}^{\blocknum}$, $\dimN=30$, $\blocknum=3$, $\indexi=1$).}
  \label{fig:ExamplaryPlotofCorollary4dot1iequal1}
\end{figure*}
\vspace{1mm}
\begin{figure*}[tbp!]
  \centering
  \captionsetup{justification=centering}
\subfloat[ $\tilde{G}=\{\tilde{V},\tilde{E},\tilde{\mathbf{W}}\}$ ]{\includegraphics[trim={0cm 0cm 0cm 0cm},clip,width=4.25cm]{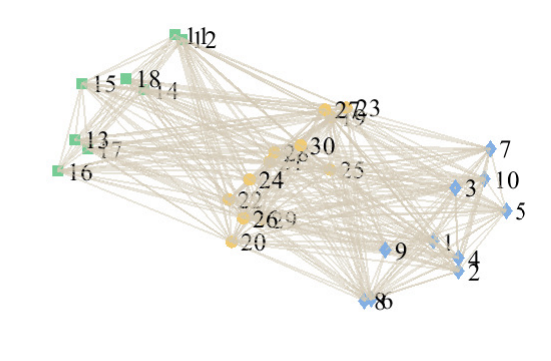}}
\subfloat[$\tilde{\mathbf{W}}\in\mathbb{R}^{\dimN\times\dimN}$]{\includegraphics[trim={0mm 0mm 0mm 0mm},clip,width=4.cm]{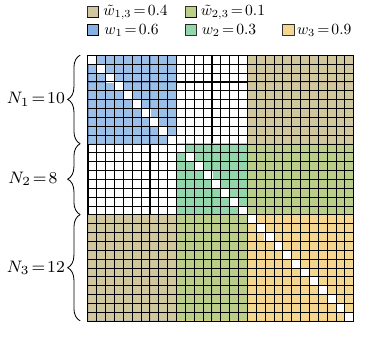}}\hspace{2mm}
\subfloat[$\tilde{\mathbf{L}}\in\mathbb{R}^{\dimN\times\dimN}$]{\includegraphics[trim={0mm 0mm 0mm 0mm},clip,width=4.5cm]{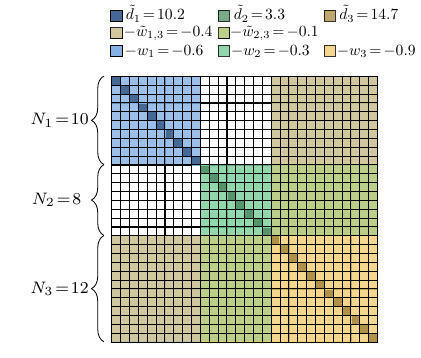}}
\subfloat[$\tilde{\boldsymbol{\lambda}}\in\mathbb{R}^{\dimN}$]{\includegraphics[trim={0mm 0mm 0mm 0mm},clip,width=4.75cm]{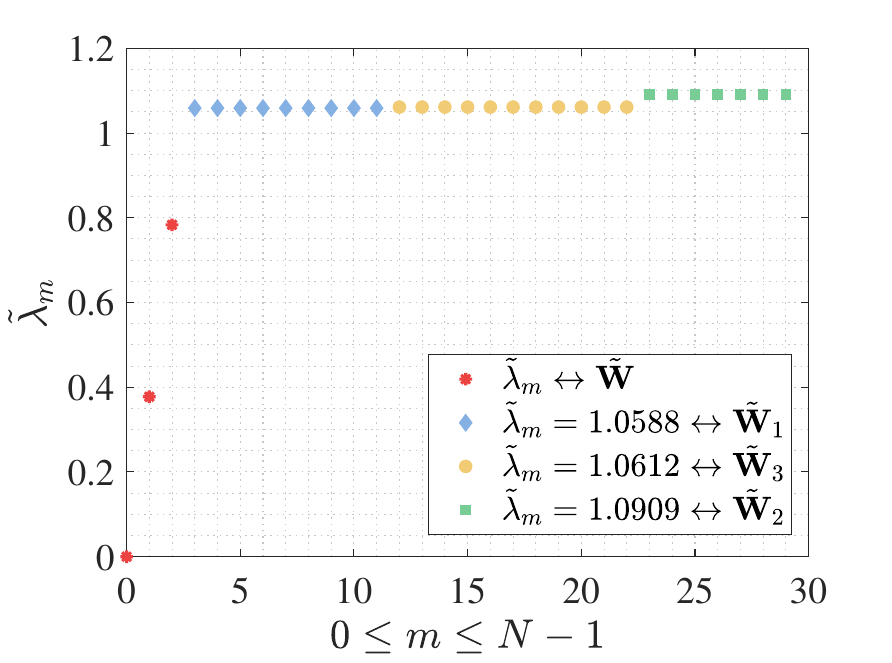}}
\\
\caption{Examplary plot of Theorem~2 ($\mathbf{n}=[10,8,12]^{\top}\in\mathbb{R}^{\blocknum}$, $\dimN=30$, $\blocknum=3$, $\indexi=\blocknum$).}
  \label{fig:ExamplaryPlotofCorollary4dot1iequalk}
\end{figure*}

\newpage
\section{Appendix B: The Standard Eigen-decomposition based Eigenvalue Analysis}
\subsection{B.1~Outlier Effects on Target Eigenvalues}
\begin{remark}\textbf{1.S.}\label{theorem:theoremtype2outliereffectEq1}\textit{ Let Let $\Tilde{\mathbf{W}}\hspace{-0.7mm}\in\hspace{-0.7mm}\mathbb{R}^{(\dimN+1)\times (\dimN+1)}$ define a symmetric affinity matrix, that is equal to $\mathbf{W}$, except for an additional Type~II outlier that shares similarity coefficients with $\blocknum$ blocks where $\Tilde{w}_{\TypeIIoutlier,\blocknum}\hspace{-0.7mm}>\hspace{-0.7mm}0$ denotes the similarity coefficient between the outlier $\mathrm{o}_{\TypeIIoutlier}$ and the $\blocknum$th block. Then, for the associated corrupted Laplacian matrix $\Tilde{\mathbf{L}}\hspace{-0.7mm}\in\hspace{-0.7mm}\mathbb{R}^{(\dimN+1)\times (\dimN+1)}$ with eigenvalues $\Tilde{\boldsymbol{\lambda}}\in\mathbb{R}^{\dimN+1}$,it holds that
\begin{align*}
\small
\begin{cases}
\begin{split}
    &\dimN_1-1\hspace{3mm}\mathrm{elements\hspace{1mm}of}\hspace{1mm}\tilde{\boldsymbol{\lambda}}\hspace{1mm}\mathrm{equal\hspace{1mm}to}\hspace{3mm}\dimN_1w_1+\Tilde{w}_{\TypeIIoutlier,1}\\
    &\dimN_2-1\hspace{3mm}\mathrm{elements\hspace{1mm}of}\hspace{1mm}\tilde{\boldsymbol{\lambda}}\hspace{1mm}\mathrm{equal\hspace{1mm}to}\hspace{3mm}\dimN_2w_2+\Tilde{w}_{\TypeIIoutlier,2}\\
    &\vdots\\
    &\dimN_{\blocknum}-1\hspace{3mm}\mathrm{elements\hspace{1mm}of}\hspace{1mm}\tilde{\boldsymbol{\lambda}}\hspace{1mm}\mathrm{equal\hspace{1mm}to}\hspace{3mm}\dimN_{\blocknum}w_{\blocknum}+\Tilde{w}_{\TypeIIoutlier,\blocknum}\\
    &\mathrm{the\hspace{1mm}smallest}\hspace{1mm}\mathrm{element\hspace{1mm}of}\hspace{1mm}\tilde{\boldsymbol{\lambda}}\hspace{1mm}\mathrm{equal\hspace{1mm}to}\hspace{1mm}\mathrm{zero}\\
     \end{split}
\end{cases}
\end{align*}
and the remaining $\blocknum$ eigenvalues are the roots of
\begin{align*}
\prod_{\indexj=1}^{\blocknum}(\Tilde{w}_{\TypeIIoutlier,\indexj}-\Tilde{\lambda})\Bigg(-\sum_{\indexj=1}^{\blocknum}\frac{\dimN_{\indexj}\Tilde{w}_{\TypeIIoutlier,\indexj}}{\Tilde{w}_{\TypeIIoutlier,\indexj}-\Tilde{\lambda}}-1 \Bigg)=0.
\end{align*}}
where $\Tilde{\lambda}\in\Tilde{\boldsymbol{\lambda}}$.
\end{remark}
\vspace{3mm}
\begin{proof}
Let $\tilde{\mathbf{W}}\in\mathbb{R}^{(\dimN+1)\times(\dimN+1)}$ and $\tilde{\mathbf{L}}\in\mathbb{R}^{(\dimN+1)\times(\dimN+1)}$, respectively, denote a block zero-diagonal symmetric affinity matrix and
associated Laplacian matrix for $\blocknum$ blocks with an additional Type II outlier that is correlated with all blocks, i.e.,\\\vspace{2mm}
\resizebox{\linewidth}{!}{
  \begin{minipage}{\linewidth}
\begin{align*}
\Tilde{\mathbf{W}}=
\begin{bmatrix}
\begin{smallmatrix}
0&\Tilde{w}_{\TypeIIoutlier,1}&\Tilde{w}_{\TypeIIoutlier,1} &\myydots&\Tilde{w}_{\TypeIIoutlier,1}&\Tilde{w}_{\TypeIIoutlier,2}&\Tilde{w}_{\TypeIIoutlier,2} &\myydots&\Tilde{w}_{\TypeIIoutlier,2} &\myydots& \Tilde{w}_{\TypeIIoutlier,\blocknum}& \Tilde{w}_{\TypeIIoutlier,\blocknum}&\myydots&\Tilde{w}_{\TypeIIoutlier,\blocknum}\\
\Tilde{w}_{\TypeIIoutlier,1}&0& w_1&\myydots&w_1&& & && && & &\\
\Tilde{w}_{\TypeIIoutlier,1}&w_1&0&\myydots&w_1& & & & && & & &\\
\vdots&\vdots&\vdots&\ddots&&& & && & & & &\\
\Tilde{w}_{\TypeIIoutlier,1}&w_1 &w_1 &\myydots & 0 & && &&& & & &\\
\Tilde{w}_{\TypeIIoutlier,2}&& & & &0& w_2&\myydots&w_2& & & & &\\
\Tilde{w}_{\TypeIIoutlier,2}& & & &  &w_2&0&\myydots&w_2& & & & &\\
\vdots&& & &  &\vdots&\vdots&\ddots&& & & & &\\
\Tilde{w}_{\TypeIIoutlier,2}& & & & &w_2 &w_2 &\myydots & 0 & & & & &\\
\vdots&& & & && & & &\ddots & &\vdots& & \vdots\\
\Tilde{w}_{\TypeIIoutlier,\blocknum}&&& & &&&&&&0&w_{\blocknum}&\myydots&w_{\blocknum}\\
\Tilde{w}_{\TypeIIoutlier,\blocknum}&& & & &&&&&&w_{\blocknum}&0& \myydots& w_{\blocknum} \\
\vdots&& & &&&&&&&\vdots&\vdots&\ddots\\
\Tilde{w}_{\TypeIIoutlier,\blocknum}&& & & &&&&&\myydots&w_{\blocknum}& w_{\blocknum} &\myydots&0\\
\end{smallmatrix}
\end{bmatrix}
\end{align*}
\end{minipage}}\\
and\\\vspace{2mm}
\resizebox{\linewidth}{!}{
  \begin{minipage}{\linewidth}
\begin{align*}
\Tilde{\mathbf{L}}=
\begin{bmatrix}
\begin{smallmatrix}
\Tilde{d}_{\TypeIIoutlier}&-\Tilde{w}_{\TypeIIoutlier,1}&-\Tilde{w}_{\TypeIIoutlier,1} &\myydots&-\Tilde{w}_{\TypeIIoutlier,1}&-\Tilde{w}_{\TypeIIoutlier,2}&-\Tilde{w}_{\TypeIIoutlier,2} &\myydots&-\Tilde{w}_{\TypeIIoutlier,2} &\myydots& -\Tilde{w}_{\TypeIIoutlier,\blocknum}& -\Tilde{w}_{\TypeIIoutlier,\blocknum}&\myydots&-\Tilde{w}_{\TypeIIoutlier,\blocknum}\\
-\Tilde{w}_{\TypeIIoutlier,1}&\Tilde{d}_1& -w_1&\myydots&-w_1&& & && && & &\\
-\Tilde{w}_{\TypeIIoutlier,1}&-w_1&\Tilde{d}_1&\myydots&-w_1& & & & && & & &\\
\vdots&\vdots&\vdots&\ddots&&& & && & & & &\\
-\Tilde{w}_{\TypeIIoutlier,1}&-w_1 &-w_1 &\myydots & \Tilde{d}_1 & && &&& & & &\\
-\Tilde{w}_{\TypeIIoutlier,2}&& & & &\Tilde{d}_2& -w_2&\myydots&-w_2& & & & &\\
-\Tilde{w}_{\TypeIIoutlier,2}& & & &  &-w_2&\Tilde{d}_2&\myydots&-w_2& & & & &\\
\vdots&& & &  &\vdots&\vdots&\ddots&& & & & &\\
-\Tilde{w}_{\TypeIIoutlier,2}& & & & &-w_2 &-w_2 &\myydots & \Tilde{d}_2 & & & & &\\
\vdots&& & & && & & &\ddots & &\vdots& & \vdots\\
-\Tilde{w}_{\TypeIIoutlier,\blocknum}&&& & &&&&&&\Tilde{d}_{\blocknum}&-w_{\blocknum}&\myydots&-w_{\blocknum}\\
-\Tilde{w}_{\TypeIIoutlier,\blocknum}&& & & &&&&&&-w_{\blocknum}&\Tilde{d}_{\blocknum}& \myydots& -w_{\blocknum} \\
\vdots&& & &&&&&&&\vdots&\vdots&\ddots\\
-\Tilde{w}_{\TypeIIoutlier,\blocknum}&& & & &&&&&\myydots&-w_{\blocknum}& -w_{\blocknum} &\myydots&\Tilde{d}_{\blocknum}\\
\end{smallmatrix}
\end{bmatrix}
\end{align*}
\end{minipage}}\\\vspace{1mm}
where $\Tilde{d}_{\TypeIIoutlier}=\sum_{\indexj=1}^{\blocknum}\dimN_{\indexj}\Tilde{w}_{\TypeIIoutlier,\indexj}$ and $\Tilde{d}_{\indexj}=(\dimN_{\indexj}-1)w_{\indexj}+\Tilde{w}_{\TypeIIoutlier,\indexj}$. To compute the eigenvalues of the Laplacian
matrix ${\Tilde{\mathbf{L}}}$, ${\mathrm{det}(\Tilde{\mathbf{L}}-\Tilde{\lambda}\mathbf{I})=0}$ is considered which can equivalently be written in matrix form as follows\\
\resizebox{\linewidth}{!}{
  \begin{minipage}{\linewidth}
\begin{align*}
\mathrm{det}(\Tilde{\mathbf{L}}-\Tilde{\lambda}\mathbf{I})=
\begin{vmatrix}
\begin{smallmatrix}
\Tilde{d}_{\TypeIIoutlier}-\Tilde{\lambda}&-\Tilde{w}_{\TypeIIoutlier,1}&-\Tilde{w}_{\TypeIIoutlier,1} &\myydots&-\Tilde{w}_{\TypeIIoutlier,1}&-\Tilde{w}_{\TypeIIoutlier,2}&-\Tilde{w}_{\TypeIIoutlier,2} &\myydots&-\Tilde{w}_{\TypeIIoutlier,2} &\myydots& -\Tilde{w}_{\TypeIIoutlier,\blocknum}& -\Tilde{w}_{\TypeIIoutlier,\blocknum}&\myydots&-\Tilde{w}_{\TypeIIoutlier,\blocknum}\\
-\Tilde{w}_{\TypeIIoutlier,1}&\Tilde{d}_1-\Tilde{\lambda}& -w_1&\myydots&-w_1&& & && && & &\\
-\Tilde{w}_{\TypeIIoutlier,1}&-w_1&\Tilde{d}_1-\Tilde{\lambda}&\myydots&-w_1& & & & && & & &\\
\vdots&\vdots&\vdots&\ddots&&& & && & & & &\\
-\Tilde{w}_{\TypeIIoutlier,1}&-w_1 &-w_1 &\myydots & \Tilde{d}_1-\Tilde{\lambda} & && &&& & & &\\
-\Tilde{w}_{\TypeIIoutlier,2}&& & & &\Tilde{d}_2-\Tilde{\lambda}& -w_2&\myydots&-w_2& & & & &\\
-\Tilde{w}_{\TypeIIoutlier,2}& & & &  &-w_2&\Tilde{d}_2-\Tilde{\lambda}&\myydots&-w_2& & & & &\\
\vdots&& & &  &\vdots&\vdots&\ddots&& & & & &\\
-\Tilde{w}_{\TypeIIoutlier,2}& & & & &-w_2 &-w_2 &\myydots & \Tilde{d}_2-\Tilde{\lambda}& & & & &\\
\vdots&& & & && & & &\ddots & &\vdots& & \vdots\\
-\Tilde{w}_{\TypeIIoutlier,\blocknum}&&& & &&&&&&\Tilde{d}_{\blocknum}-\Tilde{\lambda}&-w_{\blocknum}&\myydots&-w_{\blocknum}\\
-\Tilde{w}_{\TypeIIoutlier,\blocknum}&& & & &&&&&&-w_{\blocknum}&\Tilde{d}_{\blocknum}-\Tilde{\lambda}& \myydots& -w_{\blocknum} \\
\vdots&& & &&&&&&&\vdots&\vdots&\ddots\\
-\Tilde{w}_{\TypeIIoutlier,\blocknum}&& & & &&&&&\myydots&-w_{\blocknum}& -w_{\blocknum} &\myydots&\Tilde{d}_{\blocknum}-\Tilde{\lambda}\\
\end{smallmatrix}
\end{vmatrix}=0\\\vspace{2mm}
\end{align*}
\end{minipage}}\\
To simplify this determinant, the matrix determinant lemma \cite{matrixdeterminantlemma} can be generalized as follows\footnote{For a detailed information about generalization of the matrix determinant lemma, see Appendix~D.}
\begin{align*}
\mathrm{det}(\mathbf{H}+\mathbf{U}\mathbf{V}^{\top})=\mathrm{det}(\mathbf{H})\mathrm{det}(\mathbf{I}+\mathbf{V}^{\top}\mathbf{H}^{\dagger}\mathbf{U}) 
\end{align*}
where $\mathbf{H}\in\mathbb{R}^{(\dimN+1)\times(\dimN+1)}$ denotes an invertible matrix, $\mathbf{I}$ is the identity matrix and $\mathbf{U},\mathbf{V}\in\mathbb{R}^{(\dimN+1)\times(\dimN+1)}$. Then, for $\mathrm{det}(\tilde{\mathbf{L}}-\tilde{\lambda}\mathbf{I})=\mathrm{det}(\mathbf{H}+\mathbf{U}\mathbf{V}^{\top})=0$, it follows that\\
\resizebox{1.05\linewidth}{!}{
  \begin{minipage}{\linewidth}\vspace{5mm}
\begin{align*}
\begin{split}
\hspace{-0.7mm}0\hspace{-0.7mm}=&
\mathrm{det}\begin{pmatrix}
\begin{bmatrix}
\begin{smallmatrix}
z_{\TypeIIoutlier}-\Tilde{\lambda}&\myydots &&&& & & & & & &\myydots&\hspace{-2mm}0\\\vspace{-1mm}
&\hspace{-2mm}z_1-\Tilde{\lambda}& &&&& & & & & & &\hspace{-2mm}\vdots\\\vspace{-0.7mm}
&&\hspace{-5mm}\ddots&&&& & && & & & \hspace{-2mm}\\\vspace{-0.7mm}
&& & &\hspace{-5mm}z_1-\Tilde{\lambda} & & & & & & & & & \\\vspace{-1mm}
&&&& &\hspace{-5mm}z_2-\Tilde{\lambda}& & & & & & &\\\vspace{-0.6mm}
&&&& &&\hspace{-5mm}\ddots& && & & & \\\vspace{-0.6mm}
&& & & && & &\hspace{-5mm}z_2-\Tilde{\lambda}  & & & & & \\\vspace{-1mm}
&& & & && & &  &\hspace{-9mm}\ddots& & & & \\\vspace{-1mm}
&&&& && & & &\hspace{-1mm}z_{\blocknum}-\Tilde{\lambda}& & & \\\vspace{-1mm}
\vdots&&&& &&& & & &\hspace{-4mm}\ddots& & \\\vspace{-1mm}
0&\myydots && & && & && & & &\hspace{-6mm}z_{\blocknum}-\Tilde{\lambda}  &  \\\vspace{0.6mm}
\end{smallmatrix}
\end{bmatrix}+
\begin{bmatrix}
\begin{smallmatrix}
1&\hspace{-1mm}\bovermat{$\blocknum$}{0&\myydots& 
&}&&\hspace{-2mm}&&0\hspace{1mm}&\bovermat{$\dimN-\blocknum$}{0&\myydots&} &\myydots &0\\
0&\hspace{-1.3mm}1&0&\myydots 
&&&\hspace{-2mm}&&0\hspace{1mm}&0&\myydots& &\myydots &0\\
\vdots&\vdots&\vdots&&&&\hspace{-2mm}&&\vdots&\vdots&\ddots& &\iddots &\vdots\\
0&1&0& & & &\hspace{-2mm} &&&&& & &\\
&0&1& &&&&\hspace{-2mm}&&&& & &\\
&\vdots&\vdots & &&\hspace{-2mm}&  &&&&& & &\\
& &1 &&\hspace{-2mm}& & &&&&& & &\\
& &0 && &\hspace{-2mm} & &&0&&& & &\\
& &&&&\hspace{-2mm} & &&1&&& & &\\
&\vdots&\vdots & &&& \hspace{-2mm} &&\vdots&\vdots&\iddots& &\ddots &\vdots\\
0&0 &0 &\myydots& &\hspace{-2mm}&& & 1&0&\myydots& &\myydots &0\\
\end{smallmatrix}
\end{bmatrix}
\begin{bmatrix}
\begin{smallmatrix}
0&\bovermat{$\dimN_1$}{-\Tilde{w}_{\TypeIIoutlier,1}&\myydots &}-\Tilde{w}_{\TypeIIoutlier,1}&\bovermat{$\dimN_2$}{-\Tilde{w}_{\TypeIIoutlier,2}&\myydots&}-\Tilde{w}_{\TypeIIoutlier,2} &\myydots&\bovermat{$\dimN_{\blocknum}$}{-\Tilde{w}_{\TypeIIoutlier,\blocknum} &\myydots &}-\Tilde{w}_{\TypeIIoutlier,\blocknum}\\
-\Tilde{w}_{\TypeIIoutlier,1}&-w_1&\myydots&-w_1&0&\myydots && &&\myydots &0 \\
-\Tilde{w}_{\TypeIIoutlier,2}&0&\myydots&0&-w_2&\myydots &-w_2& &&\myydots &0 \\
\vdots&& & & & & & & & &\vdots &\\
-\Tilde{w}_{\TypeIIoutlier,\blocknum}&0&\myydots &&&& & &-w_{\blocknum}&\myydots &-w_{\blocknum} \\
0&\myydots&& &&& & & &\myydots &0\\
\vdots&\ddots&& &&&& &&\iddots&\vdots  \\\vspace{-2mm}
& & & && & && & & \\
\vdots&\iddots & && && & &  &\ddots&\vdots & \\
0&\myydots&&& && & & &\myydots&  0\\
\end{smallmatrix}
\end{bmatrix}
\end{pmatrix}\\\\
\hspace{-0.7mm}0\hspace{-0.7mm}=&\mathrm{det}\hspace{-1mm}\begin{pmatrix}\hspace{-1mm}
\mathbf{I}\hspace{-1mm}+\hspace{-2mm}
\begin{bmatrix}
\begin{smallmatrix}
0\hspace{-1.25mm}&-\Tilde{w}_{\TypeIIoutlier,1}(z_1-\Tilde{\lambda})^{-1}\hspace{-1.25mm}&\myydots \hspace{-1.25mm}&-\Tilde{w}_{\TypeIIoutlier,1}(z_1-\Tilde{\lambda})^{-1}\hspace{-1.25mm}&-\Tilde{w}_{\TypeIIoutlier,2}(z_2-\Tilde{\lambda})^{-1}\hspace{-1.25mm}&\myydots\hspace{-1.25mm}&-\Tilde{w}_{\TypeIIoutlier,2}(z_2-\Tilde{\lambda})^{-1} \hspace{-1.25mm}&\myydots\hspace{-1.25mm}&-\Tilde{w}_{\TypeIIoutlier,\blocknum}(z_{\blocknum}-\Tilde{\lambda})^{-1} \hspace{-1.25mm}&\myydots \hspace{-1.25mm}&-\Tilde{w}_{\TypeIIoutlier,\blocknum}(z_{\blocknum}-\Tilde{\lambda})^{-1}\\
-\Tilde{w}_{\TypeIIoutlier,1}(z_{\TypeIIoutlier}-\Tilde{\lambda})^{-1}\hspace{-1.25mm}&-w_1(z_1-\Tilde{\lambda})^{-1}\hspace{-1.25mm}&\myydots\hspace{-1.25mm}&-w_1(z_1-\Tilde{\lambda})^{-1}\hspace{-1.25mm}&0\hspace{-1.25mm}&\myydots \hspace{-1.25mm}&\hspace{-1.25mm}& \hspace{-1.25mm}&\hspace{-1.25mm}&\myydots \hspace{-1.25mm}&0 \\
-\Tilde{w}_{\TypeIIoutlier,2}(z_{\TypeIIoutlier}-\Tilde{\lambda})^{-1}\hspace{-1.25mm}&0\hspace{-1.25mm}&\myydots\hspace{-1.25mm}&0\hspace{-1.25mm}&-w_2(z_2-\Tilde{\lambda})^{-1}\hspace{-1.25mm}&\myydots \hspace{-1.25mm}&-w_2(z_2-\Tilde{\lambda})^{-1}\hspace{-1.25mm}& \hspace{-1.25mm}&\hspace{-1.25mm}&\myydots \hspace{-1.25mm}&0 \\
\vdots\hspace{-1.25mm}&\hspace{-1.25mm}& \hspace{-1.25mm}& \hspace{-1.25mm}& \hspace{-1.25mm}& \hspace{-1.25mm}& \hspace{-1.25mm}& \hspace{-1.25mm}& \hspace{-1.25mm}& \hspace{-1.25mm}&\vdots \hspace{-1.25mm}&\\
-\Tilde{w}_{\TypeIIoutlier,\blocknum}(z_{\TypeIIoutlier}-\Tilde{\lambda})^{-1}\hspace{-1.25mm}&0\hspace{-1.25mm}&\myydots \hspace{-1.25mm}&\hspace{-1.25mm}&\hspace{-1.25mm}&\hspace{-1.25mm}& \hspace{-1.25mm}& \hspace{-1.25mm}&-w_{\blocknum}(z_{\blocknum}-\Tilde{\lambda})^{-1}\hspace{-1.25mm}&\myydots \hspace{-1.25mm}&-w_{\blocknum}(z_{\blocknum}-\Tilde{\lambda})^{-1} \\
0\hspace{-1.25mm}&\myydots\hspace{-1.25mm}&\hspace{-1.25mm}& \hspace{-1.25mm}&\hspace{-1.25mm}&\hspace{-1.25mm}& \hspace{-1.25mm}& \hspace{-1.25mm}& \hspace{-1.25mm}&\myydots \hspace{-1.25mm}&0\\
\vdots\hspace{-1.25mm}&\ddots\hspace{-1.25mm}&\hspace{-1.25mm}& \hspace{-1.25mm}&\hspace{-1.25mm}&\hspace{-1.25mm}&\hspace{-1.25mm}& \hspace{-1.25mm}&\hspace{-1.25mm}&\iddots\hspace{-1.25mm}&\vdots  \\\vspace{-2mm}
\hspace{-1.25mm}& \hspace{-1.25mm}& \hspace{-1.25mm}& \hspace{-1.25mm}&\hspace{-1.25mm}& \hspace{-1.25mm}& \hspace{-1.25mm}&\hspace{-1.25mm}& \hspace{-1.25mm}& \hspace{-1.25mm}& \\
\vdots\hspace{-1.25mm}&\iddots \hspace{-1.25mm}& \hspace{-1.25mm}&\hspace{-1.25mm}& \hspace{-1.25mm}&\hspace{-1.25mm}& \hspace{-1.25mm}& \hspace{-1.25mm}&  \hspace{-1.25mm}&\ddots\hspace{-1.25mm}&\vdots \hspace{-1.25mm}& \\
0\hspace{-1.25mm}&\myydots\hspace{-1.25mm}&\hspace{-1.25mm}&\hspace{-1.25mm}& \hspace{-1.25mm}&\hspace{-1.25mm}& \hspace{-1.25mm}& \hspace{-1.25mm}& \hspace{-1.25mm}&\myydots\hspace{-1.25mm}&  0\\
\end{smallmatrix}
\end{bmatrix}\hspace{-2mm}
\begin{bmatrix}
\begin{smallmatrix}
1&\hspace{-1mm}0&\myydots& 
&&&\hspace{-2mm}&&0&0&\myydots&&\myydots &0\\\vspace{0.2mm}
0&\hspace{-1.3mm}1&0&\myydots 
&&&\hspace{-2mm}&\hspace{-2mm}&0&0&\myydots& &\myydots &0\\\vspace{0.2mm}
\vdots&\vdots&\vdots&&&&\hspace{-2mm}&&\vdots&\vdots&\ddots& &\iddots &\vdots\\\vspace{0.2mm}
0&1&0& & & &\hspace{-2mm} &&&&& & &\\\vspace{0.2mm}
&0&1& &&&&\hspace{-2mm}&&&& & &\\\vspace{0.2mm}
&\vdots&\vdots & &&\hspace{-2mm}&  &&&&& & &\\\vspace{0.2mm}
& &1 &&\hspace{-2mm}& & &&&&& & &\\\vspace{0.2mm}
& &0 && &\hspace{-2mm} & &&0&&& & &\\\vspace{0.2mm}
& &&&&\hspace{-2mm} & &&1&&& & &\\\vspace{0.2mm}
&\vdots&\vdots & &&& \hspace{-2mm} &&\vdots&\vdots&\iddots& &\ddots &\vdots\\\vspace{0.2mm}
0&0 &0 &\myydots& &\hspace{-2mm}&& & 1&0&\myydots& &\myydots &0\\
\end{smallmatrix}
\end{bmatrix}
\end{pmatrix}\mathrm{det}(\mathbf{H})\\\\
0=&
\begin{vmatrix}
\begin{smallmatrix}
1&-\dimN_1\Tilde{w}_{\TypeIIoutlier,1}(z_1-\Tilde{\lambda})^{-1}&-\dimN_2\Tilde{w}_{\TypeIIoutlier,2}(z_2-\Tilde{\lambda})^{-1} &\myydots&-\dimN_{\blocknum}\Tilde{w}_{\TypeIIoutlier,\blocknum}(z_{\blocknum}-\Tilde{\lambda})^{-1}&0&\myydots&\hspace{1cm}& &\myydots &0\\
-\Tilde{w}_{\TypeIIoutlier,1}(z_{\TypeIIoutlier}-\Tilde{\lambda})^{-1}&-\dimN_1w_1(z_1-\Tilde{\lambda})^{-1}+1&0&\myydots&0&\vdots&\ddots& &&\iddots &\vdots \\
-\Tilde{w}_{\TypeIIoutlier,2}(z_{\TypeIIoutlier}-\Tilde{\lambda})^{-1}&0&-\dimN_2w_2(z_2-\Tilde{\lambda})^{-1}+1&\myydots&0& && && & \\
\vdots&& & &\vdots &\vdots &\iddots& & &\ddots &\vdots &\\
-\Tilde{w}_{\TypeIIoutlier,\blocknum}(z_{\TypeIIoutlier}-\Tilde{\lambda})^{-1}&0&\myydots &&-\dimN_{\blocknum}w_{\blocknum}(z_{\blocknum}-\Tilde{\lambda})^{-1}+1&0&\myydots & &&\myydots &0 \\
0&\myydots&& &0&1 &\myydots & & &\myydots &0\\
\vdots&\ddots&&\iddots &\vdots&\vdots&\ddots& &&\iddots&\vdots  \\\vspace{-2mm}
& & & && & && & & \\
\vdots&\iddots & &\ddots&\vdots &\vdots&\iddots & &  &\ddots&\vdots & \\
0&\myydots&&\myydots&0 &0&\myydots & & &\myydots&  1\\
\end{smallmatrix}
\end{vmatrix}\mathrm{det}(\mathbf{H})
\end{split}
\end{align*}\vspace{3mm}
\end{minipage}}\\
where $z_{\TypeIIoutlier}=\sum_{\indexj=1}^{\blocknum}\dimN_{\indexj}\Tilde{w}_{\TypeIIoutlier,\indexj}$ and $z_{\indexj}=\dimN_{\indexj}w_{\indexj}+\Tilde{w}_{\TypeIIoutlier,\indexj}$ for $\indexj=1,\myydots,\blocknum$. . Using
determinant properties of block matrices \cite{detblockmatrices}, it holds that
\begin{align*}
    \begin{split}
       0=&
\begin{vmatrix}
\begin{smallmatrix}
1&-\dimN_1\Tilde{w}_{\TypeIIoutlier,1}(z_1-\Tilde{\lambda})^{-1}&-\dimN_2\Tilde{w}_{\TypeIIoutlier,2}(z_2-\Tilde{\lambda})^{-1} &\myydots&-\dimN_{\blocknum}\Tilde{w}_{\TypeIIoutlier,\blocknum}(z_{\blocknum}-\Tilde{\lambda})^{-1}\\
-\Tilde{w}_{\TypeIIoutlier,1}(z_{\TypeIIoutlier}-\Tilde{\lambda})^{-1}&-\dimN_1w_1(z_1-\Tilde{\lambda})^{-1}+1&0&\myydots&0\\
-\Tilde{w}_{\TypeIIoutlier,2}(z_{\TypeIIoutlier}-\Tilde{\lambda})^{-1}&0&-\dimN_2w_2(z_2-\Tilde{\lambda})^{-1}+1&\myydots&0\\
\vdots&& & &\vdots\\
-\Tilde{w}_{\TypeIIoutlier,\blocknum}(z_{\TypeIIoutlier}-\Tilde{\lambda})^{-1}&0&\myydots &&-\dimN_{\blocknum}w_{\blocknum}(z_{\blocknum}-\Tilde{\lambda})^{-1}+1
\end{smallmatrix}
\end{vmatrix}\mathrm{det}(\mathbf{H})
    \end{split}.
\end{align*}
\newpage
To simplify the determinant of the first matrix, it transformed into a lower diagonal matrix by applying the following
Gaussian elimination steps
\begin{align*}
\small
\begin{split}
    \frac{\dimN_1\Tilde{w}_{\TypeIIoutlier,1}(z_1-\Tilde{\lambda})^{-1}}{-\dimN_1w_1(z_1-\Tilde{\lambda})^{-1}+1}R_2+R_1 &\rightarrow R_1\\
    \frac{\dimN_2\Tilde{w}_{\TypeIIoutlier,2}(z_2-\Tilde{\lambda})^{-1}}{-\dimN_2w_2(z_2-\Tilde{\lambda})^{-1}+1}R_3+R_1 &\rightarrow R_1\\
    &\vdots\\
    \frac{\dimN_{\blocknum}\Tilde{w}_{\TypeIIoutlier,\blocknum}(z_{\blocknum}-\Tilde{\lambda})^{-1}}{-\dimN_{\blocknum}w_{\blocknum}(z_{\blocknum}-\Tilde{\lambda})^{-1}+1}R_{\blocknum+1}+R_1 &\rightarrow R_1
\end{split}
\end{align*}
where $R_{\blocknum}$ denotes $\blocknum$th row. Then, the simplified determinant yields\\
\resizebox{\linewidth}{!}{
  \begin{minipage}{\linewidth}
\begin{align*}
\begin{split}
    0=&c_{\TypeIIoutlier}(-\dimN_1w_1(z_1-\Tilde{\lambda})^{-1}+1)(-\dimN_2w_2(z_2-\Tilde{\lambda})^{-1}+1)\myydots(-\dimN_{\blocknum}w_{\blocknum}(z_{\blocknum}-\Tilde{\lambda})^{-1}+1)(z_{\TypeIIoutlier}-\Tilde{\lambda})(z_1-\Tilde{\lambda})^{\dimN_1}(z_2-\Tilde{\lambda})^{\dimN_2}\myydots(z_{\blocknum}-\Tilde{\lambda})^{\dimN_{\blocknum}}
\end{split}
\end{align*}
\end{minipage}}\\
where\\
\resizebox{\linewidth}{!}{
  \begin{minipage}{\linewidth}
\begin{align*}
\begin{split}
    c_{\TypeIIoutlier}=&\Bigg(1-\frac{\dimN_1\Tilde{w}_{\TypeIIoutlier,1}(z_1-\Tilde{\lambda})^{-1}\Tilde{w}_{\TypeIIoutlier,1}(z_{\TypeIIoutlier}-\Tilde{\lambda})^{-1}}{-\dimN_1w_1(z_1-\Tilde{\lambda})^{-1}+1}-\frac{\dimN_2\Tilde{w}_{\TypeIIoutlier,2}(z_2-\Tilde{\lambda})^{-1}\Tilde{w}_{\TypeIIoutlier,2}(z_{\TypeIIoutlier}-\Tilde{\lambda})^{-1}}{-\dimN_2w_2(z_2-\Tilde{\lambda})^{-1}+1}-\myydots-\frac{\dimN_{\blocknum}\Tilde{w}_{\TypeIIoutlier,\blocknum}(z_{\blocknum}-\Tilde{\lambda})^{-1}\Tilde{w}_{\TypeIIoutlier,\blocknum}(z_{\TypeIIoutlier}-\Tilde{\lambda})^{-1}}{-\dimN_{\blocknum}w_{\blocknum}(z_{\blocknum}-\Tilde{\lambda})^{-1}+1}   \Bigg).
\end{split}
\end{align*}
\end{minipage}}\\
For $z_{\TypeIIoutlier}=\sum_{\indexj=1}^{\blocknum}\dimN_{\indexj}\Tilde{w}_{\TypeIIoutlier,\indexj}$ and $z_{\indexj}=\dimN_{\indexj}w_{\indexj}+\Tilde{w}_{\TypeIIoutlier,\indexj}$ such that $\indexj=1,\myydots,\blocknum$ the determinant yields
\begin{align*}
    \begin{split}
      0=&(-\dimN_1w_1(z_1-\Tilde{\lambda})^{-1}+1)(-\dimN_2w_2(z_2-\Tilde{\lambda})^{-1}+1)\myydots(-\dimN_{\blocknum}w_{\blocknum}(z_{\blocknum}-\Tilde{\lambda})^{-1}+1)(z_{\TypeIIoutlier}-\Tilde{\lambda})(z_1-\Tilde{\lambda})^{\dimN_1}(z_2-\Tilde{\lambda})^{\dimN_2}\myydots(z_{\blocknum}-\Tilde{\lambda})^{\dimN_{\blocknum}}\\&(z_{\TypeIIoutlier}-\Tilde{\lambda})^{-1}
      \Bigg(z_{\TypeIIoutlier}-\Tilde{\lambda}-\frac{\dimN_1\Tilde{w}_{\TypeIIoutlier,1}(z_1-\Tilde{\lambda})^{-1}\Tilde{w}_{\TypeIIoutlier,1}}{-\dimN_1w_1(z_1-\Tilde{\lambda})^{-1}+1}-\frac{\dimN_2\Tilde{w}_{\TypeIIoutlier,2}(z_2-\Tilde{\lambda})^{-1}\Tilde{w}_{\TypeIIoutlier,2}}{-\dimN_2w_2(z_2-\Tilde{\lambda})^{-1}+1}-\myydots-\frac{\dimN_{\blocknum}\Tilde{w}_{\TypeIIoutlier,\blocknum}(z_{\blocknum}-\Tilde{\lambda})^{-1}\Tilde{w}_{\TypeIIoutlier,\blocknum}}{-\dimN_{\blocknum}w_{\blocknum}(z_{\blocknum}-\Tilde{\lambda})^{-1}+1}   \Bigg)\\
        0=&(\Tilde{w}_{\TypeIIoutlier,1}-\Tilde{\lambda})(\Tilde{w}_{\TypeIIoutlier,2}-\Tilde{\lambda})\myydots(\Tilde{w}_{\TypeIIoutlier,\blocknum}-\Tilde{\lambda})(\dimN_1w_1+\Tilde{w}_{\TypeIIoutlier,1}-\Tilde{\lambda})^{\dimN_1-1}(\dimN_2w_2+\Tilde{w}_{\TypeIIoutlier,2}-\Tilde{\lambda})^{\dimN_2-1}\myydots(\dimN_{\blocknum}w_{\blocknum}+\Tilde{w}_{\TypeIIoutlier,\blocknum}-\Tilde{\lambda})^{\dimN_{\blocknum}-1}\\
        &\Bigg( \sum_{\indexj=1}^{\blocknum}\dimN_{\indexj}\Tilde{w}_{\TypeIIoutlier,\indexj}-\Tilde{\lambda}-\frac{\dimN_1\Tilde{w}_{\TypeIIoutlier,1}^2}{\Tilde{w}_{\TypeIIoutlier,1}-\Tilde{\lambda}}-\frac{\dimN_2\Tilde{w}_{\TypeIIoutlier,2}^2}{\Tilde{w}_{\TypeIIoutlier,2}-\Tilde{\lambda}}\myydots-\frac{\dimN_{\blocknum}\Tilde{w}_{\TypeIIoutlier,\blocknum}^2}{\Tilde{w}_{\TypeIIoutlier,\blocknum}-\Tilde{\lambda}} \Bigg)\\
        0=&(\Tilde{w}_{\TypeIIoutlier,1}-\Tilde{\lambda})(\Tilde{w}_{\TypeIIoutlier,2}-\Tilde{\lambda})\myydots(\Tilde{w}_{\TypeIIoutlier,\blocknum}-\Tilde{\lambda})(\dimN_1w_1+\Tilde{w}_{\TypeIIoutlier,1}-\Tilde{\lambda})^{\dimN_1-1}(\dimN_2w_2+\Tilde{w}_{\TypeIIoutlier,2}-\Tilde{\lambda})^{\dimN_2-1}\myydots(\dimN_{\blocknum}w_{\blocknum}+\Tilde{w}_{\TypeIIoutlier,\blocknum}-\Tilde{\lambda})^{\dimN_{\blocknum}-1}\\
        &\Bigg( \dimN_1\Tilde{w}_{\TypeIIoutlier,1}-\frac{\dimN_1\Tilde{w}_{\TypeIIoutlier,1}^2}{\Tilde{w}_{\TypeIIoutlier,1}-\Tilde{\lambda}}+\dimN_2\Tilde{w}_{\TypeIIoutlier,2}-\frac{\dimN_2\Tilde{w}_{\TypeIIoutlier,2}^2}{\Tilde{w}_{\TypeIIoutlier,2}-\Tilde{\lambda}}\myydots+\dimN_{\blocknum}\Tilde{w}_{\TypeIIoutlier,\blocknum}-\frac{\dimN_{\blocknum}\Tilde{w}_{\TypeIIoutlier,\blocknum}^2}{\Tilde{w}_{\TypeIIoutlier,\blocknum}-\Tilde{\lambda}}-\Tilde{\lambda}\Bigg)\\
        0=&(\Tilde{w}_{\TypeIIoutlier,1}-\Tilde{\lambda})(\Tilde{w}_{\TypeIIoutlier,2}-\Tilde{\lambda})\myydots(\Tilde{w}_{\TypeIIoutlier,\blocknum}-\Tilde{\lambda})(\dimN_1w_1+\Tilde{w}_{\TypeIIoutlier,1}-\Tilde{\lambda})^{\dimN_1-1}(\dimN_2w_2+\Tilde{w}_{\TypeIIoutlier,2}-\Tilde{\lambda})^{\dimN_2-1}\myydots(\dimN_{\blocknum}w_{\blocknum}+\Tilde{w}_{\TypeIIoutlier,\blocknum}-\Tilde{\lambda})^{\dimN_{\blocknum}-1}\\
        &\Bigg( -\frac{\dimN_1\Tilde{w}_{\TypeIIoutlier,1}\Tilde{\lambda}}{\Tilde{w}_{\TypeIIoutlier,1}-\Tilde{\lambda}}-\frac{\dimN_2\Tilde{w}_{\TypeIIoutlier,2}\Tilde{\lambda}}{\Tilde{w}_{\TypeIIoutlier,2}-\Tilde{\lambda}}\myydots-\frac{\dimN_{\blocknum}\Tilde{w}_{\TypeIIoutlier,\blocknum}\Tilde{\lambda}}{\Tilde{w}_{\TypeIIoutlier,\blocknum}-\Tilde{\lambda}}-\Tilde{\lambda}\Bigg)\\
        0=&(\Tilde{w}_{\TypeIIoutlier,1}-\Tilde{\lambda})(\Tilde{w}_{\TypeIIoutlier,2}-\Tilde{\lambda})\myydots(\Tilde{w}_{\TypeIIoutlier,\blocknum}-\Tilde{\lambda})(\dimN_1w_1+\Tilde{w}_{\TypeIIoutlier,1}-\Tilde{\lambda})^{\dimN_1-1}(\dimN_2w_2+\Tilde{w}_{\TypeIIoutlier,2}-\Tilde{\lambda})^{\dimN_2-1}\myydots(\dimN_{\blocknum}w_{\blocknum}+\Tilde{w}_{\TypeIIoutlier,\blocknum}-\Tilde{\lambda})^{\dimN_{\blocknum}-1}\Tilde{\lambda}\\
        &\Bigg( -\frac{\dimN_1\Tilde{w}_{\TypeIIoutlier,1}}{\Tilde{w}_{\TypeIIoutlier,1}-\Tilde{\lambda}}-\frac{\dimN_2\Tilde{w}_{\TypeIIoutlier,2}}{\Tilde{w}_{\TypeIIoutlier,2}-\Tilde{\lambda}}\myydots-\frac{\dimN_{\blocknum}\Tilde{w}_{\TypeIIoutlier,\blocknum}}{\Tilde{w}_{\TypeIIoutlier,\blocknum}-\Tilde{\lambda}}-1\Bigg)\\
    \end{split}
\end{align*}
\vspace{2mm}
Now, $\dimN+1-\blocknum$ number of eigenvalues can be computed as
\begin{align*}
\begin{cases}
\begin{split}
    &\dimN_1-1\hspace{3mm}\mathrm{elements\hspace{1mm}of\hspace{1mm}\tilde{\boldsymbol{\lambda}}}\hspace{1mm}\mathrm{equal\hspace{1mm}to}\hspace{3mm} z_1=\dimN_1 w_1+\Tilde{w}_{\TypeIIoutlier,1}\\
    &\dimN_2-1\hspace{3mm}\mathrm{elements\hspace{1mm}of\hspace{1mm}\tilde{\boldsymbol{\lambda}}}\hspace{1mm}\mathrm{equal\hspace{1mm}to}\hspace{3mm} z_2=\dimN_2w_2+\Tilde{w}_{\TypeIIoutlier,2}\\
    &\vdots\\
    &\dimN_{\blocknum}-1\hspace{3mm}\mathrm{elements\hspace{1mm}of\hspace{1mm}\tilde{\boldsymbol{\lambda}}}\hspace{1mm}\mathrm{equal\hspace{1mm}to}\hspace{3mm}
    z_{\blocknum}=\dimN_{\blocknum}w_{\blocknum}+\Tilde{w}_{\TypeIIoutlier,\blocknum}\\
    &\mathrm{the\hspace{1mm}smallest}\hspace{1mm}\mathrm{element\hspace{1mm}of\hspace{1mm}\tilde{\boldsymbol{\lambda}}}\hspace{1mm}\mathrm{equal\hspace{1mm}to}\hspace{1mm}\mathrm{zero}\\
     \end{split}
\end{cases}
\end{align*}
and the remaining $\blocknum$ number of eigenvalues are roots of the equation
\begin{align*}
\begin{split}
0=&(\Tilde{w}_{\TypeIIoutlier,1}-\Tilde{\lambda})(\Tilde{w}_{\TypeIIoutlier,2}-\Tilde{\lambda})\myydots(\Tilde{w}_{\TypeIIoutlier,\blocknum}-\Tilde{\lambda})\Bigg( -\frac{\dimN_1\Tilde{w}_{\TypeIIoutlier,1}}{\Tilde{w}_{\TypeIIoutlier,1}-\Tilde{\lambda}}-\frac{\dimN_2\Tilde{w}_{\TypeIIoutlier,2}}{\Tilde{w}_{\TypeIIoutlier,2}-\Tilde{\lambda}}\myydots-\frac{\dimN_{\blocknum}\Tilde{w}_{\TypeIIoutlier,\blocknum}}{\Tilde{w}_{\TypeIIoutlier,\blocknum}-\Tilde{\lambda}}-1\Bigg)\\
0=&\prod_{\indexj=1}^{\blocknum}(\Tilde{w}_{\TypeIIoutlier,\indexj}-\Tilde{\lambda})\Bigg(-\sum_{\indexj=1}^{\blocknum}\frac{\dimN_{\indexj}\Tilde{w}_{\TypeIIoutlier,\indexj}}{\Tilde{w}_{\TypeIIoutlier,\indexj}-\Tilde{\lambda}}-1 \Bigg)
\end{split}
\end{align*}
\end{proof}

\begin{figure*}[tbp!]
\vspace{2mm}
  \centering
  \captionsetup{justification=centering}
\subfloat[ $\tilde{G}=\{\tilde{V},\tilde{E},\tilde{\mathbf{W}}\}$ ]{\includegraphics[trim={0cm 0cm 0cm 0cm},clip,width=4.25cm]{Images_Accompanying/GraphModel_oequal1_TypeIIOutliers.pdf}}
\subfloat[$\tilde{\mathbf{W}}\in\mathbb{R}^{(\dimN+1)\times (\dimN+1)}$]{\includegraphics[trim={0mm 0mm 0mm 0mm},clip,width=3.9cm]{Images_Accompanying/TypeIIOutlierEffect_oequal1_Affinity.pdf}}\hspace{1mm}
\subfloat[$\tilde{\mathbf{L}}\in\mathbb{R}^{(\dimN+1)\times (\dimN+1)}$]{\includegraphics[trim={0mm 0mm 0mm 0mm},clip,width=4.95cm]{Images_Accompanying/TypeIIOutlierEffect_oequal1_Laplacian.pdf}}
\subfloat[$\tilde{\boldsymbol{\lambda}}\in\mathbb{R}^{\dimN+1}$]{\includegraphics[trim={0mm 0mm 0mm 0mm},clip,width=4.75cm]{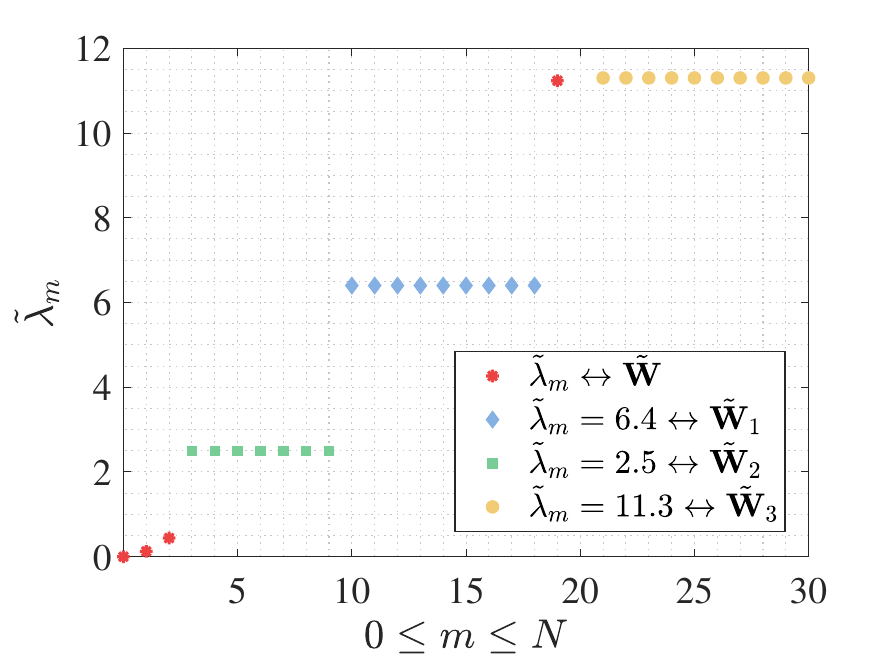}}
\\
\caption{Examplary plot of Theorem~1.S. ($\mathbf{n}=[1,10,8,12]^{\top}\in\mathbb{R}^{\blocknum+1}$, $\dimN+1=31$, $\blocknum=3$).}
  \label{fig:TypeIIstandardeigen-decomposition}
\end{figure*}

\newpage
\begin{remark}\textbf{2.S.} \label{theorem:kgroupcorrelationeffectonstandareigendecomposition} \textit{Let $\Tilde{\mathbf{W}}\in\mathbb{R}^{\dimN\times \dimN}$ define an affinity matrix, that is equal to $\mathbf{W}$, except that block $\indexi$ has similarity with the remaining $\blocknum\hspace{-0.2mm}-\hspace{-0.2mm}1$ blocks with $\tilde{w}_{\indexi,\indexj}\hspace{-0.7mm}=\hspace{-0.7mm}\Tilde{w}_{\indexj,\indexi}\hspace{-0.7mm}>\hspace{-0.7mm}0$ denoting the value around which the similarity coefficients between blocks $\indexi$ and $\indexj$ are concentrated for $\indexj=1,\myydots,\blocknum$ and $\indexi\neq\indexj$. Then, 
the eigenvalues 
$\Tilde{\boldsymbol{\lambda}}\in\mathbb{R}^{\dimN}$ of $\Tilde{\mathbf{L}}\in\mathbb{R}^{\dimN\times \dimN}$ are as follows:
\begin{align*}
\begin{cases}
\begin{split}
    &\dimN_{\indexi}-1\hspace{3mm}\mathrm{elements\hspace{1mm}of\hspace{1mm}\tilde{\boldsymbol{\lambda}}}\hspace{1mm}\mathrm{equal\hspace{1mm}to}\hspace{3mm}\dimN_{\indexi}w_{\indexi}+\sum\limits_{\substack{\indexj=1 \\ \indexj\neq\indexi}}^{\blocknum}\dimN_{\indexj}\tilde{w}_{\indexi,\indexj}\\
    &\dimN_{\indexj}-1\hspace{3mm}\mathrm{elements\hspace{1mm}of\hspace{1mm}\tilde{\boldsymbol{\lambda}}}\hspace{1mm}\mathrm{equal\hspace{1mm}to}\hspace{3mm}\dimN_{\indexj}w_{\indexj}+\dimN_{\indexi}\tilde{w}_{\indexi,\indexj}\\
    &\vdots\\
    &\dimN_{\blocknum}-1\hspace{3mm}\mathrm{elements\hspace{1mm}of\hspace{1mm}\tilde{\boldsymbol{\lambda}}}\hspace{1mm}\mathrm{equal\hspace{1mm}to}\hspace{3mm}\dimN_{\blocknum}w_{\blocknum}+\dimN_{\indexi}\tilde{w}_{\indexi,\blocknum}\\
    &\mathrm{the\hspace{1mm}smallest}\hspace{1mm}\mathrm{element\hspace{1mm}of\hspace{1mm}\tilde{\boldsymbol{\lambda}}}\hspace{1mm}\mathrm{equal\hspace{1mm}to}\hspace{1mm}\mathrm{zero}\\
     \end{split}
\end{cases}
\end{align*}
and the remaining $\blocknum\hspace{-0.45mm}-\hspace{-0.45mm}1$ eigenvalues in ${\Tilde{\boldsymbol{\lambda}}}$ are the roots of
\begin{align*}
\prod_{\substack{\indexj=1 \\ \indexj\neq\indexi}}^{\blocknum}(\dimN_{\indexi}\tilde{w}_{\indexi,\indexj}-\Tilde{\lambda})\Bigg(\sum\limits_{\substack{\indexj=1 \\ \indexj\neq\indexi}}^{\blocknum}-\frac{\dimN_{\indexj} \tilde{w}_{\indexi,\indexj}}{\dimN_{\indexi}\tilde{w}_{\indexi,\indexj}-\Tilde{\lambda}}-1\Bigg)=&0
\end{align*}}
where $\Tilde{\lambda}\in\Tilde{\boldsymbol{\lambda}}$.
\end{remark}
\vspace{2mm}
\begin{proof}
Let $\Tilde{\mathbf{W}}\in\mathbb{R}^{\dimN\times\dimN}$ and $\Tilde{\mathbf{L}}\in\mathbb{R}^{\dimN\times\dimN}$ denote $\blocknum$ block zero diagonal affinity matrix and associated Laplacian in which $\indexi$th block has similarity with remaining $\blocknum-1$ number of blocks. For simplicity, let $\indexi=1$, i.e.\\\vspace{2mm}
\resizebox{0.9\linewidth}{!}{
  \begin{minipage}{\linewidth}
\begin{align*}
\Tilde{\mathbf{W}}=
\begin{bmatrix}
\begin{smallmatrix}
0& w_1&\myydots&w_1&\tilde{w}_{1,2}&\myydots & &\tilde{w}_{1,2} &\myydots & \tilde{w}_{1,\blocknum}& \myydots& &\tilde{w}_{1,\blocknum}\\
w_1&0&\myydots&w_1&\tilde{w}_{1,2} & & & \tilde{w}_{1,2} &\myydots & \tilde{w}_{1,\blocknum}& \myydots& &\tilde{w}_{1,\blocknum}\\
\vdots&\vdots&\ddots&&\vdots& & \ddots& \vdots & & & &\ddots &\\
w_1 &w_1 &\myydots & 0 &\tilde{w}_{1,2} &\myydots & &\tilde{w}_{1,2}&\myydots & \tilde{w}_{1,\blocknum}& \myydots& &\tilde{w}_{1,\blocknum}\\
\tilde{w}_{1,2}&\myydots & &\tilde{w}_{1,2} &0& w_2&\myydots&w_2& & & & &\\
\tilde{w}_{1,2} & & & \tilde{w}_{1,2} &w_2&0&\myydots&w_2& & & & &\\
\vdots& & \ddots& \vdots &\vdots&\vdots&\ddots&& & & & &\\
\tilde{w}_{1,2} &\myydots & &\tilde{w}_{1,2} &w_2 &w_2 &\myydots & 0 & & & & &\\
\vdots& & &\vdots &\vdots& \vdots& & &\ddots & &\vdots& & \vdots\\
\tilde{w}_{1,\blocknum}&\myydots & &\tilde{w}_{1,\blocknum} &&&&&&0&w_{\blocknum}&\myydots&w_{\blocknum}\\
\tilde{w}_{1,\blocknum}& & &\tilde{w}_{1,\blocknum} &&&&&\myydots&w_{\blocknum}&0& \myydots& w_{\blocknum} \\
\vdots& & &\vdots&&&&&&\vdots&\vdots&\ddots\\
\tilde{w}_{1,\blocknum}& & &\tilde{w}_{1,\blocknum} &&&&&\myydots&w_{\blocknum}& w_{\blocknum} &\myydots&0\\
\end{smallmatrix}
\end{bmatrix}
\end{align*}
\end{minipage}}\\
and\\
\resizebox{0.9\linewidth}{!}{
  \begin{minipage}{\linewidth}
\begin{align*}
\Tilde{\mathbf{L}}=
\begin{bmatrix}
\begin{smallmatrix}
\Tilde{d}_1& -w_1&\myydots&-w_1&-\tilde{w}_{1,2}&\myydots & &-\tilde{w}_{1,2} &\myydots & -\tilde{w}_{1,\blocknum}& \myydots& &-\tilde{w}_{1,\blocknum}\\
-w_1&\Tilde{d}_1&\myydots&-w_1&-\tilde{w}_{1,2} & & & -\tilde{w}_{1,2} &\myydots & -\tilde{w}_{1,\blocknum}& \myydots& &-\tilde{w}_{1,\blocknum}\\
\vdots&\vdots&\ddots&&\vdots& & \ddots& \vdots & & & &\ddots &\\
-w_1 &-w_1 &\myydots & \Tilde{d}_1 &-\tilde{w}_{1,2} &\myydots & &-\tilde{w}_{1,2}&\myydots & -\tilde{w}_{1,\blocknum}& \myydots& &-\tilde{w}_{1,\blocknum}\\
-\tilde{w}_{1,2}&\myydots & &-\tilde{w}_{1,2} &\Tilde{d}_2& -w_2&\myydots&-w_2& & & & &\\
-\tilde{w}_{1,2} & & & -\tilde{w}_{1,2} &-w_2&\Tilde{d}_2&\myydots&-w_2& & & & &\\
\vdots& & \ddots& \vdots &\vdots&\vdots&\ddots&& & & & &\\
-\tilde{w}_{1,2} &\myydots & &-\tilde{w}_{1,2} &-w_2 &-w_2 &\myydots & \Tilde{d}_2 & & & & &\\
\vdots& & &\vdots &\vdots& \vdots& & &\ddots & &\vdots& & \vdots\\
-\tilde{w}_{1,\blocknum}&\myydots & &-\tilde{w}_{1,\blocknum} &&&&&&\Tilde{d}_{\blocknum}&-w_{\blocknum}&\myydots&-w_{\blocknum}\\
-\tilde{w}_{1,\blocknum}& & &-\tilde{w}_{1,\blocknum} &&&&&\myydots&-w_{\blocknum}&\Tilde{d}_{\blocknum}& \myydots&-w_{\blocknum} \\
\vdots& & &\vdots&&&&&&\vdots&\vdots&\ddots\\
-\tilde{w}_{1,\blocknum}& & &-\tilde{w}_{1,\blocknum} &&&&&\myydots&-w_{\blocknum}&-w_{\blocknum} &\myydots&\Tilde{d}_{\blocknum}\\
\end{smallmatrix}
\end{bmatrix}\\
\end{align*}
\end{minipage}}\\
where $\Tilde{d}_1=(\dimN_1-1)w_1+\sum_{\indexj=2}^{\blocknum}\dimN_{\indexj}\tilde{w}_{1,\indexj}$ and $\Tilde{d}_{\indexj}=(\dimN_{\indexj}-1)w_{\indexj}+\dimN_1\tilde{w}_{1,\indexj}, \indexj=1,\myydots,\blocknum$. To estimate the eigenvalues of the Laplacian
matrix $\Tilde{\mathbf{L}}$, ${\mathrm{det}(\Tilde{\mathbf{L}}-\Tilde{\lambda}\mathbf{I})=0}$ is considered which can equivalently be written in matrix form as follows\vspace{3mm}
\begin{align*}
\mathrm{det}(\Tilde{\mathbf{L}}-\Tilde{\lambda}\mathbf{I})=
\begin{vmatrix}
\begin{smallmatrix}
\Tilde{d}_1-\Tilde{\lambda}& -w_1&\myydots&-w_1&-\tilde{w}_{1,2}&\myydots & &-\tilde{w}_{1,2} &\myydots & -\tilde{w}_{1,\blocknum}& \myydots& &-\tilde{w}_{1,\blocknum}\\
-w_1&\Tilde{d}_1-\Tilde{\lambda}&\myydots&-w_1&-\tilde{w}_{1,2} & & & -\tilde{w}_{1,2} &\myydots & -\tilde{w}_{1,\blocknum}& \myydots& &-\tilde{w}_{1,\blocknum}\\
\vdots&\vdots&\ddots&&\vdots& & \ddots& \vdots & & & &\ddots &\\
-w_1 &-w_1 &\myydots & \Tilde{d}_1-\Tilde{\lambda} &-\tilde{w}_{1,2} &\myydots & &-\tilde{w}_{1,2}&\myydots & -\tilde{w}_{1,\blocknum}& \myydots& &-\tilde{w}_{1,\blocknum}\\
-\tilde{w}_{1,2}&\myydots & &-\tilde{w}_{1,2} &\Tilde{d}_2-\Tilde{\lambda}& -w_2&\myydots&-w_2& & & & &\\
-\tilde{w}_{1,2} & & & -\tilde{w}_{1,2} &-w_2&\Tilde{d}_2-\Tilde{\lambda}&\myydots&-w_2& & & & &\\
\vdots& & \ddots& \vdots &\vdots&\vdots&\ddots&& & & & &\\
-\tilde{w}_{1,2} &\myydots & &-\tilde{w}_{1,2} &-w_2 &-w_2 &\myydots & \Tilde{d}_2-\Tilde{\lambda} & & & & &\\
\vdots& & &\vdots &\vdots& \vdots& & &\ddots & &\vdots& & \vdots\\
-\tilde{w}_{1,\blocknum}&\myydots & &-\tilde{w}_{1,\blocknum} &&&&&&\Tilde{d}_{\blocknum}-\Tilde{\lambda}&-w_{\blocknum}&\myydots&-w_{\blocknum}\\
-\tilde{w}_{1,\blocknum}& & &-\tilde{w}_{1,\blocknum} &&&&&\myydots&-w_{\blocknum}&\Tilde{d}_{\blocknum}-\Tilde{\lambda}& \myydots&-w_{\blocknum} \\
\vdots& & &\vdots&&&&&&\vdots&\vdots&\ddots\\
-\tilde{w}_{1,\blocknum}& & &-\tilde{w}_{1,\blocknum} &&&&&\myydots&-w_{\blocknum}&-w_{\blocknum} &\myydots&\Tilde{d}_{\blocknum}-\Tilde{\lambda}\\
\end{smallmatrix}
\end{vmatrix}=0.
\end{align*}
To simplify this determinant, the matrix determinant lemma \cite{matrixdeterminantlemma} can be generalized as follows\footnote{For a detailed information about generalization of the matrix determinant lemma, see Appendix~D.}
\begin{align*}
\mathrm{det}(\mathbf{H}+\mathbf{U}\mathbf{V}^{\top})=\mathrm{det}(\mathbf{H})\mathrm{det}(\mathbf{I}+\mathbf{V}^{\top}\mathbf{H}^{\dagger}\mathbf{U}) 
\end{align*}
where $\mathbf{H}\in\mathbb{R}^{(\dimN+1)\times(\dimN+1)}$ denotes an invertible matrix, $\mathbf{I}$ is the identity matrix and $\mathbf{U},\mathbf{V}\in\mathbb{R}^{(\dimN+1)\times(\dimN+1)}$. Then, for $\mathrm{det}(\tilde{\mathbf{L}}-\tilde{\lambda}\mathbf{I})=\mathrm{det}(\mathbf{H}+\mathbf{U}\mathbf{V}^{\top})=0$, it follows that\vspace{2mm}\\
\resizebox{0.95\linewidth}{!}{
  \begin{minipage}{\linewidth}
\begin{align*}
\begin{split}
\hspace{-0.7mm}0\hspace{-0.7mm}&=
\mathrm{det}\begin{pmatrix}
\begin{bmatrix}
\begin{smallmatrix}
z_1-\Tilde{\lambda}&\myydots &&&& & & & & & \myydots&\hspace{-2mm}0\\\vspace{-1mm}
&\hspace{-5mm}\ddots&&&& & && & & & \hspace{-2mm}\vdots\\\vspace{-0.7mm}
& & &\hspace{-5mm}z_1-\Tilde{\lambda} & & & & & & & & & \\\vspace{-1mm}
&&& &\hspace{-5mm}z_2-\Tilde{\lambda}& & & & & & &\\\vspace{-0.6mm}
&&& &&\hspace{-5mm}\ddots& && & & & \\\vspace{-0.6mm}
& & & && & &\hspace{-5mm}z_2-\Tilde{\lambda}  & & & & & \\\vspace{-1mm}
& & & && & &  &\hspace{-9mm}\ddots& & & & \\\vspace{-1mm}
&&& && & & &\hspace{-1mm}z_{\blocknum}-\Tilde{\lambda}& & & \\\vspace{-1mm}
\vdots&&& &&& & & &\hspace{-4mm}\ddots& & \\\vspace{-1mm}
0&\myydots & & && & && & & &\hspace{-6mm}z_{\blocknum}-\Tilde{\lambda}  &  \\\vspace{0.6mm}
\end{smallmatrix}
\end{bmatrix}+
\begin{bmatrix}
\begin{smallmatrix}
\bovermat{$\blocknum$}{\hspace{-1.3mm}1&0&\myydots } &&&\hspace{-2mm}&&0\hspace{1mm}&\bovermat{$\dimN-\blocknum$}{0&\myydots&} &\myydots &0\\
\vdots&\vdots&&&&\hspace{-2mm}&&\vdots&\vdots&\ddots& &\iddots &\vdots\\
1&0& & & &\hspace{-2mm} &&&&& & &\\
0&1& &&&&\hspace{-2mm}&&&& & &\\
\vdots&\vdots & &&\hspace{-2mm}&  &&&&& & &\\
 &1 &&\hspace{-2mm}& & &&&&& & &\\
 &0 && &\hspace{-2mm} & &&0&&& & &\\
 &&&&\hspace{-2mm} & &&1&&& & &\\
\vdots&\vdots & &&& \hspace{-2mm} &&\vdots&\vdots&\iddots& &\ddots &\vdots\\
0 &0 &\myydots& &\hspace{-2mm}&& & 1&0&\myydots& &\myydots &0\\
\end{smallmatrix}
\end{bmatrix}
\begin{bmatrix}
\begin{smallmatrix}
\bovermat{$\dimN_1$}{-w_1&\myydots &}-w_1&\bovermat{$\dimN_2$}{-\tilde{w}_{1,2}&\myydots&}-\tilde{w}_{1,2} &\myydots&\bovermat{$\dimN_{\blocknum}$}{-\tilde{w}_{1,\blocknum} &\myydots &}-\tilde{w}_{1,\blocknum}\\
-\tilde{w}_{1,2}&\myydots&-\tilde{w}_{1,2}&-w_2&\myydots &-w_2& &&\myydots &0 \\
\vdots& & & & & & & & &\vdots &\\
-\tilde{w}_{1,\blocknum}&\myydots &-\tilde{w}_{1,\blocknum}&0&\myydots& & &-w_{\blocknum}&\myydots &-w_{\blocknum} \\
0&\myydots&& && & & &\myydots &0\\
\vdots&\ddots&& &&& &&\iddots&\vdots  \\\vspace{-2mm}
& & & && & && & & \\
\vdots&\iddots & & && & &  &\ddots&\vdots & \\
0&\myydots&& && & & &\myydots&  0\\
\end{smallmatrix}
\end{bmatrix}
\end{pmatrix}\\
\\
\hspace{-0.7mm}0\hspace{-0.7mm}&=\hspace{-0.7mm}\mathrm{det}(\mathbf{H})
\mathrm{det}\hspace{-1.1mm}\begin{pmatrix}
\hspace{-1mm}\mathbf{I}\hspace{-0.7mm}+\hspace{-1.7mm}
\begin{bmatrix}
\begin{smallmatrix}
\hspace{-1.2mm}-w_1(z_1-\Tilde{\lambda})^{-1}&\hspace{-1.2mm}\myydots &\hspace{-1.2mm}-w_1(z_1-\Tilde{\lambda})^{-1}&\hspace{-1.2mm}-\tilde{w}_{1,2}(z_2-\Tilde{\lambda})^{-1}&\hspace{-1.2mm}\myydots&\hspace{-1.2mm}-\tilde{w}_{1,2}(z_2-\Tilde{\lambda})^{-1} &\hspace{-1.2mm}\myydots&\hspace{-1.2mm}-\tilde{w}_{1,\blocknum}(z_{\blocknum}-\Tilde{\lambda})^{-1} &\hspace{-1.2mm}\myydots &\hspace{-1.2mm}-\tilde{w}_{1,\blocknum}(z_{\blocknum}-\Tilde{\lambda})^{-1}\\
\hspace{-1.2mm}-\tilde{w}_{1,2}(z_1-\Tilde{\lambda})^{-1}&\hspace{-1.2mm}\myydots&\hspace{-1.2mm}-\tilde{w}_{1,2}(z_1-\Tilde{\lambda})^{-1}&\hspace{-1.2mm}-w_2(z_2-\Tilde{\lambda})^{-1}&\hspace{-1.2mm}\myydots &\hspace{-1.2mm}-w_2(z_2-\Tilde{\lambda})^{-1}&\hspace{-1.2mm} &\hspace{-1.2mm}&\myydots &\hspace{-1.2mm}0 \\
\hspace{-1.2mm}\vdots&\hspace{-1.2mm} &\hspace{-1.2mm} &\hspace{-1.2mm} &\hspace{-1.2mm} &\hspace{-1.2mm} &\hspace{-1.2mm} &\hspace{-1.2mm} &\hspace{-1.2mm} &\hspace{-1.2mm}\vdots &\hspace{-1.2mm}\\
\hspace{-1.2mm}-\tilde{w}_{1,\blocknum}(z_1-\Tilde{\lambda})^{-1}&\hspace{-1.2mm}\myydots &\hspace{-1.2mm}-\tilde{w}_{1,\blocknum}(z_1-\Tilde{\lambda})^{-1}&\hspace{-1.2mm}0&\hspace{-1.2mm}\myydots&\hspace{-1.2mm} &\hspace{-1.2mm} &\hspace{-1.2mm}-w_{\blocknum}(z_{\blocknum}-\Tilde{\lambda})^{-1}&\hspace{-1.2mm}\myydots &\hspace{-1.2mm}-w_{\blocknum}(z_{\blocknum}-\Tilde{\lambda})^{-1} \\
\hspace{-1.2mm}0&\hspace{-1.2mm}\myydots&\hspace{-1.2mm}&\hspace{-1.2mm} &\hspace{-1.2mm}&\hspace{-1.2mm} &\hspace{-1.2mm} &\hspace{-1.2mm} &\hspace{-1.2mm}\myydots &\hspace{-1.2mm}0\\
\hspace{-1.2mm}\vdots&\hspace{-1.2mm}\ddots&\hspace{-1.2mm}&\hspace{-1.2mm} &\hspace{-1.2mm}&\hspace{-1.2mm}&\hspace{-1.2mm} &\hspace{-1.2mm}&\hspace{-1.2mm}\iddots &\hspace{-1.2mm}\vdots \\\vspace{-3mm}
\hspace{-1.2mm}&\hspace{-1.2mm} &\hspace{-1.2mm} &\hspace{-1.2mm} &\hspace{-1.2mm}&\hspace{-1.2mm} &\hspace{-1.2mm} &\hspace{-1.2mm}& \hspace{-1.2mm}&\hspace{-1.2mm} &\hspace{-1.2mm} \\
\hspace{-1.2mm}\vdots&\hspace{-1.2mm}\iddots &\hspace{-1.2mm} &\hspace{-1.2mm} &\hspace{-1.2mm}& \hspace{-1.2mm}& \hspace{-1.2mm}& \hspace{-1.2mm} &\hspace{-1.2mm}\ddots&\hspace{-1.2mm}\vdots &\hspace{-1.2mm} \\
\hspace{-1.2mm}0&\hspace{-1.2mm}\myydots&\hspace{-1.2mm}&\hspace{-1.2mm} &\hspace{-1.2mm}& \hspace{-1.2mm}&\hspace{-1.2mm} &\hspace{-1.2mm} &\hspace{-1.2mm}\myydots&\hspace{-1.2mm}0  \\
\end{smallmatrix}
\end{bmatrix}
\hspace{-1.5mm}
\begin{bmatrix}
\begin{smallmatrix}
1&0&\myydots&&&\hspace{-1mm}&&0&0&\myydots& &\myydots &0\\
\vdots&\vdots&&&&\hspace{-1mm}&&\vdots&\vdots&\ddots& &\iddots &\vdots\\
1&0& & & &\hspace{-1mm} &&&&& & &\\
0&1& &&&&\hspace{-1mm}&&&& & &\\
\vdots&\vdots & &&\hspace{-1mm}&  &&&&& & &\\
 &1 &&\hspace{-1mm}& & &&&&& & &\\
 &0 && &\hspace{-1mm} & &&0&&& & &\\
 &&&&\hspace{-1mm} & &&1&&& & &\\
\vdots&\vdots & &&& \hspace{-1mm} &&\vdots&\vdots&\iddots& &\ddots &\vdots\\
0 &0 &\myydots& &\hspace{-1mm}&& & 1&0&\myydots& &\myydots &0\\
\end{smallmatrix}
\end{bmatrix}
\hspace{-1mm}
\end{pmatrix}\\
\\
0&=\mathrm{det}(\mathbf{H})
\begin{vmatrix}
\begin{smallmatrix}
-\dimN_1w_1(z_1-\Tilde{\lambda})^{-1}+1 & -\dimN_2\tilde{w}_{1,2}(z_2-\Tilde{\lambda})^{-1} & -\dimN_3\tilde{w}_{1,3}(z_3-\Tilde{\lambda})^{-1} & \myydots & -\dimN_{\blocknum}\tilde{w}_{1,\blocknum}(z_{\blocknum}-\Tilde{\lambda})^{-1}& 0 &\myydots &\hspace{1.5cm} &\myydots & 0\\
-\dimN_1\tilde{w}_{1,2}(z_1-\Tilde{\lambda})^{-1} & -\dimN_2w_2(z_2-\Tilde{\lambda})^{-1}+1 & 0 & \myydots & 0& \vdots &\ddots &\hspace{1.5cm} &\iddots & \vdots\\
-\dimN_1\tilde{w}_{1,3}(z_1-\Tilde{\lambda})^{-1} & 0 &-\dimN_3w_3(z_3-\Tilde{\lambda})^{-1}+1 & \myydots & 0 &  &&\hspace{1.5cm} & & \\
\vdots & & & & \vdots&\vdots &\iddots &\hspace{1.5cm} &\ddots & \vdots\\
-\dimN_1\tilde{w}_{1,\blocknum}(z_1-\Tilde{\lambda})^{-1} &0 &\myydots & &-\dimN_{\blocknum}w_{\blocknum}(z_{\blocknum}-\Tilde{\lambda})^{-1}+1& 0 &\myydots &\hspace{1.5cm} &\myydots & 0\\
0 &\myydots & &\myydots &0& 1 & &\hspace{1.5cm} &\myydots & 0\\
\vdots &\ddots & &\iddots &\vdots&  & &\hspace{0.25cm}\ddots\hspace{1.25cm} & & \vdots\\
 & & & &&  & &\hspace{1.5cm} && \\
\vdots &\iddots & &\ddots &\vdots&  \vdots& &\hspace{1.5cm} &\hspace{-0.25cm}\ddots\hspace{0.25cm} & \vdots\\
0 &\myydots & &\myydots &0& 0 &\myydots&\hspace{1.5cm} & & 1\\
\end{smallmatrix}
\end{vmatrix}
\end{split}
\end{align*}
\end{minipage}}\\
where $z_1=\dimN_1w_1+\sum_{\indexj=2}^{\blocknum}\dimN_{\indexj}\tilde{w}_{1,\indexj}$ and $z_{\indexj}=\dimN_{\indexj}w_{\indexj}+\dimN_1\tilde{w}_{1,\indexj}$ for $\indexj=2,\myydots,\blocknum$.  Using determinant properties of block matrices \cite{detblockmatrices}, it holds that
\begin{align*}
    \begin{split}
        0&=\mathrm{det}(\mathbf{H})
\begin{vmatrix}
\begin{smallmatrix}
-\dimN_1w_1(z_1-\Tilde{\lambda})^{-1}+1 & -\dimN_2\tilde{w}_{1,2}(z_2-\Tilde{\lambda})^{-1} & -\dimN_3\tilde{w}_{1,3}(z_3-\Tilde{\lambda})^{-1} & \myydots & -\dimN_{\blocknum}\tilde{w}_{1,\blocknum}(z_{\blocknum}-\Tilde{\lambda})^{-1}\\
-\dimN_1\tilde{w}_{1,2}(z_1-\Tilde{\lambda})^{-1} & -\dimN_2w_2(z_2-\Tilde{\lambda})^{-1}+1 & 0 & \myydots & 0\\
-\dimN_1\tilde{w}_{1,3}(z_1-\Tilde{\lambda})^{-1} & 0 &-\dimN_3w_3(z_3-\Tilde{\lambda})^{-1}+1 & \myydots & 0 \\
\vdots & & & & \vdots\\
-\dimN_1\tilde{w}_{1,\blocknum}(z_1-\Tilde{\lambda})^{-1}&0 &\myydots & &-\dimN_{\blocknum}w_{\blocknum}(z_{\blocknum}-\Tilde{\lambda})^{-1}+1\\
\end{smallmatrix}
\end{vmatrix}
    \end{split}.
\end{align*}
To simplify the determinant of the first matrix, it transformed into a lower diagonal matrix by applying the following Gaussian elimination steps
\begin{align*}
\small
\begin{split}
    \frac{\dimN_2\tilde{w}_{1,2}(z_2-\Tilde{\lambda})^{-1}}{-\dimN_2w_2(z_2-\Tilde{\lambda})^{-1}+1}R_2+R_1 &\rightarrow R_1\\
    \frac{\dimN_3\tilde{w}_{1,3}(z_3-\Tilde{\lambda})^{-1}}{-\dimN_3w_3(z_3-\Tilde{\lambda})^{-1}+1}R_3+R_1 &\rightarrow R_1\\
    &\vdots\\
    \frac{\dimN_{\blocknum}\tilde{w}_{1,\blocknum}(z_{\blocknum}-\Tilde{\lambda})^{-1}}{-\dimN_{\blocknum}w_{\blocknum}(z_{\blocknum}-\Tilde{\lambda})^{-1}+1}R_{\blocknum}+R_1 &\rightarrow R_1
\end{split}
\end{align*}
where $R_{\blocknum}$ denotes $\blocknum$th row.
Then, the simplified determinant yields
\begin{align*}
    \begin{split}
        0&=\mathrm{det}(\mathbf{H})
\begin{vmatrix}
\begin{smallmatrix}
c_1 & 0 & 0 & \myydots & 0 \\
-\dimN_1\tilde{w}_{1,2}(z_1-\Tilde{\lambda})^{-1} & -\dimN_2w_2(z_2-\Tilde{\lambda})^{-1}+1 & 0 & \myydots & 0\\
-\dimN_1\tilde{w}_{1,3}(z_1-\Tilde{\lambda})^{-1} & 0 &-\dimN_3w_3(z_3-\Tilde{\lambda})^{-1}+1 & \myydots & 0 \\
\vdots & & & & \vdots\\
-\dimN_1\tilde{w}_{1,\blocknum}(z_1-\Tilde{\lambda})^{-1} &0 &\myydots & &-\dimN_{\blocknum}w_{\blocknum}(z_{\blocknum}-\Tilde{\lambda})^{-1}+1\\
\end{smallmatrix}
\end{vmatrix}
    \end{split}
\end{align*}
where $c_1$ equals to\\
\resizebox{0.95\linewidth}{!}{
  \begin{minipage}{\linewidth}
\begin{align*}
c_1=-\dimN_1w_1(z_1-\Tilde{\lambda})^{-1}+1- \frac{\dimN_2\tilde{w}_{1,2}(z_2-\Tilde{\lambda})^{-1}\dimN_1\tilde{w}_{1,2}(z_1-\Tilde{\lambda})^{-1}}{-\dimN_2w_2(z_2-\Tilde{\lambda})^{-1}+1}-\myydots-\frac{\dimN_{\blocknum}\tilde{w}_{1,\blocknum}(z_{\blocknum}-\Tilde{\lambda})^{-1}\dimN_1\tilde{w}_{1,\blocknum}(z_1-\Tilde{\lambda})^{-1}}{-\dimN_{\blocknum}w_{\blocknum}(z_{\blocknum}-\Tilde{\lambda})^{-1}+1}.
\end{align*}
\end{minipage}}\\
For $z_1=\dimN_1w_1+\sum_{\indexj=2}^{\blocknum}\dimN_{\indexj}\tilde{w}_{1,\indexj}$ and $z_{\indexj}=\dimN_{\indexj}w_{\indexj}+\dimN_1\tilde{w}_{1,\indexj}$ such that $\indexj=2,\myydots,\blocknum$, the determinant $\mathrm{det}(\Tilde{\mathbf{L}}-\Tilde{\lambda}\mathbf{I)=0}$ yields\\
\resizebox{\linewidth}{!}{
  \begin{minipage}{\linewidth}
\begin{align*}
\begin{split}
   0=&(z_1-\Tilde{\lambda})^{\dimN_1}(z_2-\Tilde{\lambda})^{\dimN_2}\myydots(z_{\blocknum}-\Tilde{\lambda})^{\dimN_{\blocknum}}(-\dimN_2w_2(z_2-\Tilde{\lambda})^{-1}+1)\myydots(-\dimN_{\blocknum}w_{\blocknum}(z_{\blocknum}-\Tilde{\lambda})^{-1}+1)c_1\\
   0=&(z_1-\Tilde{\lambda})^{\dimN_1}(-\dimN_2w_2(z_2-\Tilde{\lambda})^{\dimN_2-1}+(z_2-\Tilde{\lambda})^{\dimN_2})\myydots(-\dimN_{\blocknum}w_{\blocknum}(z_{\blocknum}-\Tilde{\lambda})^{\dimN_{\blocknum}-1}+(z_{\blocknum}-\Tilde{\lambda})^{\dimN_{\blocknum}})c_1\\
   0=&(z_1-\Tilde{\lambda})^{\dimN_1}(z_2-\Tilde{\lambda})^{\dimN_2-1}\myydots(z_{\blocknum}-\Tilde{\lambda})^{\dimN_{\blocknum}-1}(-\dimN_2w_2+z_2-\Tilde{\lambda})\myydots(-\dimN_{\blocknum}w_{\blocknum}+z_{\blocknum}-\Tilde{\lambda})c_1\\
   0=& (z_1-\Tilde{\lambda})^{\dimN_1}(z_2-\Tilde{\lambda})^{\dimN_2-1}\myydots(z_{\blocknum}-\Tilde{\lambda})^{\dimN_{\blocknum}-1}(-\dimN_2w_2+z_2-\Tilde{\lambda})\myydots(-\dimN_{\blocknum}w_{\blocknum}+z_{\blocknum}-\Tilde{\lambda})\\&(z_1-\Tilde{\lambda})^{-1}\Bigg(-\dimN_1w_1+z_1-\Tilde{\lambda}- \frac{\dimN_2\tilde{w}_{1,2}(z_2-\Tilde{\lambda})^{-1}\dimN_1\tilde{w}_{1,2}}{-\dimN_2w_2(z_2-\Tilde{\lambda})^{-1}+1}-\myydots-\frac{\dimN_{\blocknum}\tilde{w}_{1,\blocknum}(z_{\blocknum}-\Tilde{\lambda})^{-1}\dimN_1\tilde{w}_{1,\blocknum}}{-\dimN_{\blocknum}w_{\blocknum}(z_{\blocknum}-\Tilde{\lambda})^{-1}+1}\Bigg)\\
   0=& (z_1-\Tilde{\lambda})^{\dimN_1-1}(z_2-\Tilde{\lambda})^{\dimN_2-1}\myydots(z_{\blocknum}-\Tilde{\lambda})^{\dimN_{\blocknum}-1}(-\dimN_2w_2+z_2-\Tilde{\lambda})\myydots(-\dimN_{\blocknum}w_{\blocknum}+z_{\blocknum}-\Tilde{\lambda})\\&\Bigg(-\dimN_1w_1+z_1-\Tilde{\lambda}- \frac{\dimN_2\tilde{w}_{1,2}(z_2-\Tilde{\lambda})^{-1}\dimN_1\tilde{w}_{1,2}}{-\dimN_2w_2(z_2-\Tilde{\lambda})^{-1}+1}-\myydots-\frac{\dimN_{\blocknum}\tilde{w}_{1,\blocknum}(z_{\blocknum}-\Tilde{\lambda})^{-1}\dimN_1\tilde{w}_{1,\blocknum}}{-\dimN_{\blocknum}w_{\blocknum}(z_{\blocknum}-\Tilde{\lambda})^{-1}+1}\Bigg)\\
   0=&\Big(\dimN_1w_1+\sum_{\indexj=2}^{\blocknum}\dimN_{\indexj}\tilde{w}_{1,\indexj}-\Tilde{\lambda}\Big)^{\dimN_1-1}\Big(\dimN_2w_2+\dimN_1\tilde{w}_{1,2}-\Tilde{\lambda}\Big)^{\dimN_2-1}\myydots\Big(\dimN_{\blocknum}w_{\blocknum}+\dimN_1\tilde{w}_{1,\blocknum}-\Tilde{\lambda}\Big)^{\dimN_{\blocknum}-1}(\dimN_1\tilde{w}_{1,2}-\Tilde{\lambda})\myydots(\dimN_1\tilde{w}_{1,\blocknum}-\Tilde{\lambda})\\&\Bigg(\sum_{\indexj=2}^{\blocknum}\dimN_{\indexj}\tilde{w}_{1,\indexj}-\Tilde{\lambda}- \frac{\dimN_1\dimN_2\tilde{w}^2_{1,2}}{\dimN_1\tilde{w}_{1,2}-\Tilde{\lambda}}-\myydots-\frac{\dimN_1\dimN_{\blocknum}\tilde{w}^2_{1,\blocknum}}{\dimN_1\tilde{w}_{1,\blocknum}-\Tilde{\lambda}}\Bigg)\\
   0=&\Big(\dimN_1w_1+\sum_{\indexj=2}^{\blocknum}\dimN_{\indexj}\tilde{w}_{1,\indexj}-\Tilde{\lambda}\Big)^{\dimN_1-1}\Big(\dimN_2w_2+\dimN_1\tilde{w}_{1,2}-\Tilde{\lambda}\Big)^{\dimN_2-1}\myydots\Big(\dimN_{\blocknum}w_{\blocknum}+\dimN_1\tilde{w}_{1,\blocknum}-\Tilde{\lambda}\Big)^{\dimN_{\blocknum}-1}(\dimN_1\tilde{w}_{1,2}-\Tilde{\lambda})\myydots(\dimN_1\tilde{w}_{1,\blocknum}-\Tilde{\lambda})\\&\Bigg(-\frac{\dimN_2\tilde{w}_{1,2}\Tilde{\lambda}}{\dimN_1\tilde{w}_{1,2}-\Tilde{\lambda}}\myydots-\frac{\dimN_{\blocknum}\tilde{w}_{1,\blocknum}\Tilde{\lambda}}{\dimN_1\tilde{w}_{1,\blocknum}-\Tilde{\lambda}}-\Tilde{\lambda}\Bigg)\\
   0=&\Big(\dimN_1w_1+\sum_{\indexj=2}^{\blocknum}\dimN_{\indexj}\tilde{w}_{1,\indexj}-\Tilde{\lambda}\Big)^{\dimN_1-1}\Big(\dimN_2w_2+\dimN_1\tilde{w}_{1,2}-\Tilde{\lambda}\Big)^{\dimN_2-1}\myydots\Big(\dimN_{\blocknum}w_{\blocknum}+\dimN_1\tilde{w}_{1,\blocknum}-\Tilde{\lambda}\Big)^{\dimN_{\blocknum}-1}(\dimN_1\tilde{w}_{1,2}-\Tilde{\lambda})\myydots(\dimN_1\tilde{w}_{1,\blocknum}-\Tilde{\lambda})\Tilde{\lambda}\\&\Bigg(-\frac{\dimN_2\tilde{w}_{1,2}}{\dimN_1\tilde{w}_{1,2}-\Tilde{\lambda}}\myydots-\frac{\dimN_{\blocknum}\tilde{w}_{1,\blocknum}}{\dimN_1\tilde{w}_{1,\blocknum}-\Tilde{\lambda}}-1\Bigg)
\end{split}\vspace{1mm}
\end{align*}
\end{minipage}}\vspace{2mm}\\
Based on this, $\dimN+1-\blocknum$ number of eigenvalues are\vspace{-0.5mm}
\begin{align*}
\begin{cases}
\begin{split}
    &\dimN_1-1\hspace{3mm}\mathrm{elements\hspace{1mm}of\hspace{1mm}\tilde{\boldsymbol{\lambda}}}\hspace{1mm}\mathrm{equal\hspace{1mm}to}\hspace{3mm}\dimN_1w_1+\sum_{\indexj=2}^{\blocknum} \dimN_{\indexj}\tilde{w}_{1,\indexj}\\
    &\dimN_2-1\hspace{3mm}\mathrm{elements\hspace{1mm}of\hspace{1mm}\tilde{\boldsymbol{\lambda}}}\hspace{1mm}\mathrm{equal\hspace{1mm}to}\hspace{3mm}\dimN_2w_2+\dimN_1\tilde{w}_{1,2}\\
    &\vdots\\
    &\dimN_{\blocknum}-1\hspace{3mm}\mathrm{elements\hspace{1mm}of\hspace{1mm}\tilde{\boldsymbol{\lambda}}}\hspace{1mm}\mathrm{equal\hspace{1mm}to}\hspace{3mm}\dimN_{\blocknum}w_{\blocknum}+\dimN_1\tilde{w}_{1,\blocknum}\\
    &\mathrm{the\hspace{1mm}smallest}\hspace{1mm}\mathrm{element\hspace{1mm}of\hspace{1mm}\tilde{\boldsymbol{\lambda}}}\hspace{1mm}\mathrm{equal\hspace{1mm}to}\hspace{1mm}\mathrm{zero}
     \end{split}
\end{cases}
\end{align*}
and additionally $\blocknum-1$ number of eigenvalues are roots of the following equation\vspace{1mm}
\begin{align*}
(\dimN_1\tilde{w}_{1,2}-\Tilde{\lambda})\myydots(\dimN_1\tilde{w}_{1,\blocknum}-\Tilde{\lambda})\Bigg(-\frac{\dimN_2\tilde{w}_{1,2}}{\dimN_1\tilde{w}_{1,2}-\Tilde{\lambda}}\myydots-\frac{\dimN_{\blocknum}\tilde{w}_{1,\blocknum}}{\dimN_1\tilde{w}_{1,\blocknum}-\Tilde{\lambda}}-1\Bigg)=
\prod_{\indexj=2}^{\blocknum}(\dimN_1\tilde{w}_{1,\indexj}-\Tilde{\lambda})\Bigg(-\sum\limits_{\indexj=2}^{\blocknum}\frac{\dimN_{\indexj} \tilde{w}_{1,\indexj}}{\dimN_1\tilde{w}_{1,\indexj}-\Tilde{\lambda}}-1\Bigg)=0
\end{align*}
where $\indexi=1$.
\end{proof}
\begin{figure*}[tbp!]
  \centering
  \captionsetup{justification=centering}
\subfloat[ $\tilde{G}=\{\tilde{V},\tilde{E},\tilde{\mathbf{W}}\}$ ]{\includegraphics[trim={0cm 0cm 0cm 0cm},clip,width=4.25cm]{Images_Accompanying/GroupSimilarity_kgroup_iequal1_GraphModel.pdf}}
\subfloat[$\tilde{\mathbf{W}}\in\mathbb{R}^{\dimN\times\dimN}$]{\includegraphics[trim={0mm 0mm 0mm 0mm},clip,width=4cm]{Images_Accompanying/GroupSimilarity_kgroups_iequal1_Affinity.pdf}}\hspace{2mm}
\subfloat[$\tilde{\mathbf{L}}\in\mathbb{R}^{\dimN\times\dimN}$]{\includegraphics[trim={0mm 0mm 0mm 0mm},clip,width=4.5cm]{Images_Accompanying/GroupSimilarity_kgroups_iequal1_Laplacian.pdf}}
\subfloat[$\tilde{\boldsymbol{\lambda}}\in\mathbb{R}^{\dimN}$]{\includegraphics[trim={0mm 0mm 0mm 0mm},clip,width=4.75cm]{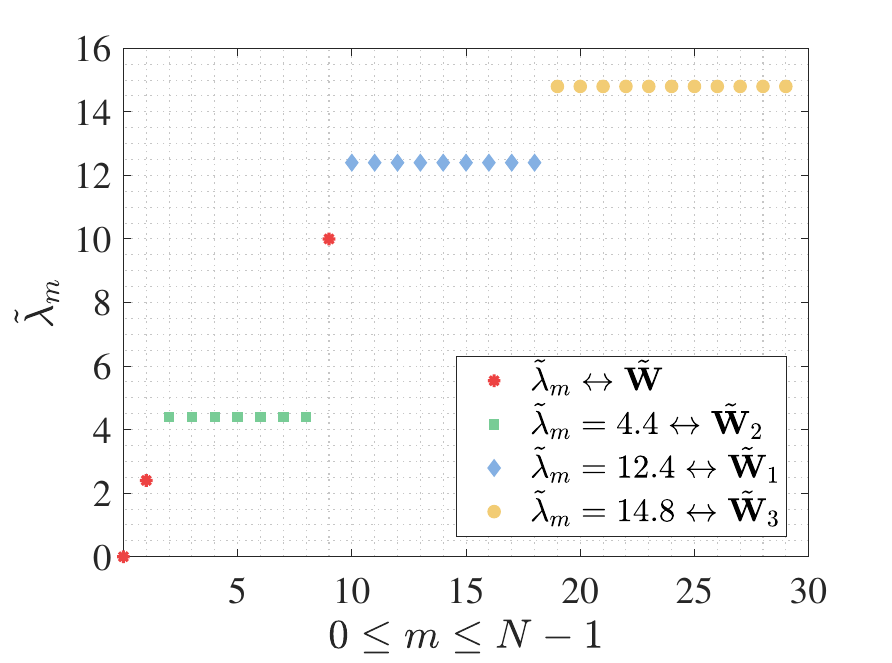}}
\\\vspace{-2mm}
\caption{Examplary plot of Theorem~2.S. ($\mathbf{n}=[10,8,12]^{\top}\in\mathbb{R}^{\blocknum}$, $\dimN=30$, $\blocknum=3$, $\indexi=1$).}
  \label{fig:ExamplaryPlotofCorollary41Soutlierindex1}
\end{figure*}
\begin{figure*}[tbp!]
  \centering
  \captionsetup{justification=centering}
\subfloat[ $\tilde{G}=\{\tilde{V},\tilde{E},\tilde{\mathbf{W}}\}$ ]{\includegraphics[trim={0cm 0cm 0cm 0cm},clip,width=4.25cm]{Images_Accompanying/GroupSimilarity_kgroup_iequalk_GraphModel.pdf}}
\subfloat[$\tilde{\mathbf{W}}\in\mathbb{R}^{\dimN\times\dimN}$]{\includegraphics[trim={0mm 0mm 0mm 0mm},clip,width=4cm]{Images_Accompanying/GroupSimilarity_kgroups_iequalk_Affinity.pdf}}
\subfloat[$\tilde{\mathbf{L}}\in\mathbb{R}^{\dimN\times\dimN}$]{\includegraphics[trim={0mm 0mm 0mm 0mm},clip,width=4.5cm]{Images_Accompanying/GroupSimilarity_kgroups_iequalk_Laplacian.pdf}}
\subfloat[$\tilde{\boldsymbol{\lambda}}\in\mathbb{R}^{\dimN}$]{\includegraphics[trim={0mm 0mm 0mm 0mm},clip,width=4.75cm]{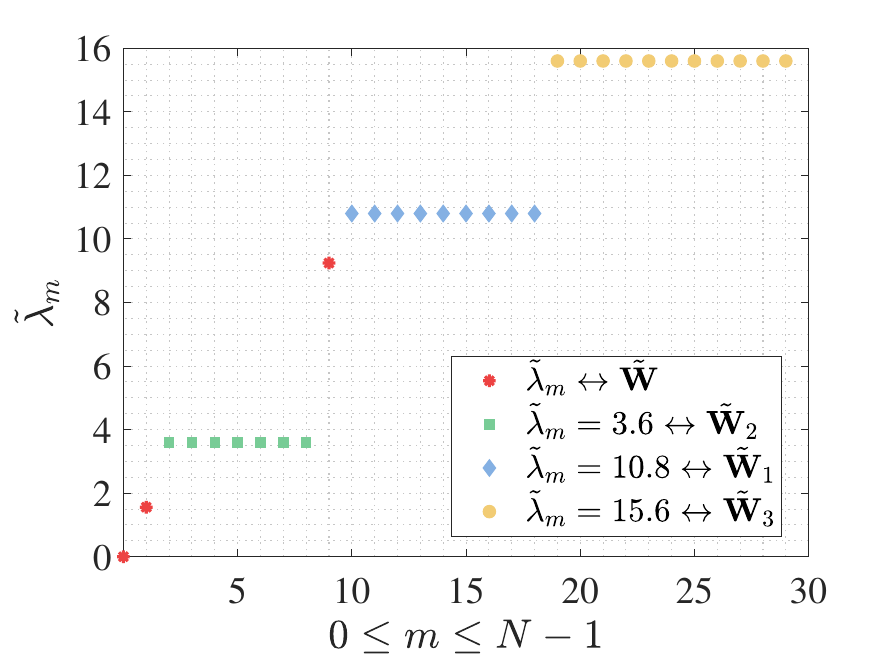}}
\\\vspace{-2mm}
\caption{Examplary plot of Theorem~2.S. ($\mathbf{n}=[10,8,12]^{\top}\in\mathbb{R}^{\blocknum}$, $\dimN=30$, $\blocknum=3$, $\indexi=\blocknum$).}
  \label{fig:ExamplaryPlotofCorollary41Soutlierindexk}
\end{figure*}
\newpage
\section{Appendix C: Simplified Laplacian Matrix Analysis and Outlier Effects}
\vspace{1mm}
\subsection{C.1~Outlier Effects on Target Vector $\mathbf{v}$: Proof of Theorem~3}
\vspace{1mm}
\begin{proof}[\unskip\nopunct]\vspace{-2mm}
To analyze different positions of Type~II outlier, it is shifted along the diagonal of corrupted Laplacian matrix $\Tilde{\mathbf{L}}\in\mathbb{R}^{(\dimN+1)\times (\dimN+1)}$ as follows\\
\resizebox{.85\linewidth}{!}{
  \begin{minipage}{\linewidth}
\begin{align*}
\Tilde{\mathbf{L}}_{(\indexm_\TypeIIoutlier=1)}=
\begin{bmatrix}
    \begin{smallmatrix}
\Tilde{d}_{\TypeIIoutlier}&-\Tilde{w}_{\TypeIIoutlier,1}& -\Tilde{w}_{\TypeIIoutlier,1}&\myydots&-\Tilde{w}_{\TypeIIoutlier,1}&-\Tilde{w}_{\TypeIIoutlier,2}&-\Tilde{w}_{\TypeIIoutlier,2}&\myydots&-\Tilde{w}_{\TypeIIoutlier,2}&\myydots &-\Tilde{w}_{\TypeIIoutlier,\blocknum} &-\Tilde{w}_{\TypeIIoutlier,\blocknum} &\myydots&-\Tilde{w}_{\TypeIIoutlier,\blocknum}\\
-\Tilde{w}_{\TypeIIoutlier,1}&\tilde{d}_1& -w_1&\myydots&-w_1&\myydots&&&& & & &&\\
-\Tilde{w}_{\TypeIIoutlier,1}&-w_1&\tilde{d}_1&\myydots&-w_1&\myydots&&&&\\
\vdots&\vdots&\vdots&\ddots&\vdots&&&&&& & & &\\
-\Tilde{w}_{\TypeIIoutlier,1}&-w_1 &-w_1 &\myydots & \tilde{d}_1 &\myydots&&&&&&&&\\
-\Tilde{w}_{\TypeIIoutlier,2}&\vdots&\vdots&&\vdots&\tilde{d}_2&-w_2&\myydots&-w_2&\myydots&&&&\\
-\Tilde{w}_{\TypeIIoutlier,2}&&&&&-w_2&\tilde{d}_2&\myydots&-w_2&\myydots \\
\vdots&&&&&\vdots&\vdots&\ddots&&& & &  &\\
-\Tilde{w}_{\TypeIIoutlier,2}&& &&&-w_2 &-w_2 &\myydots & \tilde{d}_2 &\myydots&& & &\\
\vdots&& & & &\vdots& \vdots& &\vdots &\ddots & &\vdots& & \vdots\\
-\Tilde{w}_{\TypeIIoutlier,\blocknum}&& & & &&&&&&\tilde{d}_{\blocknum}&-w_{\blocknum}&\myydots&-w_{\blocknum}\\
-\Tilde{w}_{\TypeIIoutlier,\blocknum}&& & & &&&&&\myydots&-w_{\blocknum}&\tilde{d}_{\blocknum}& \myydots& -w_{\blocknum} \\
\vdots&& & & &&&&&&\vdots&\vdots&\ddots\\
-\Tilde{w}_{\TypeIIoutlier,\blocknum}&& & & &&&&&\myydots&-w_{\blocknum}& -w_{\blocknum} &\myydots&\tilde{d}_{\blocknum}\
\end{smallmatrix}
    \end{bmatrix}\\\vspace{-1mm}
\Tilde{\mathbf{L}}_{(\indexm_{\TypeIIoutlier}=2)}
\underdot{
\begin{bmatrix}
    \begin{smallmatrix}
\tilde{d}_1&-\Tilde{w}_{\TypeIIoutlier,1}& -w_1&\myydots&-w_1&\myydots&&&& & & &&\\
-\Tilde{w}_{\TypeIIoutlier,1}&\Tilde{d}_{\TypeIIoutlier}& -\Tilde{w}_{\TypeIIoutlier,1}&\myydots&-\Tilde{w}_{\TypeIIoutlier,1}&-\Tilde{w}_{\TypeIIoutlier,2}&-\Tilde{w}_{\TypeIIoutlier,2}&\myydots&-\Tilde{w}_{\TypeIIoutlier,2}&\myydots &-\Tilde{w}_{\TypeIIoutlier,\blocknum} &-\Tilde{w}_{\TypeIIoutlier,\blocknum} &\myydots&-\Tilde{w}_{\TypeIIoutlier,\blocknum}\\
-w_1&-\Tilde{w}_{\TypeIIoutlier,1}&\tilde{d}_1&\myydots&-w_1&\myydots&&&&\\
\vdots&\vdots&\vdots&\ddots&\vdots&&&&&& & & &\\
-w_1 &-\Tilde{w}_{\TypeIIoutlier,1}&-w_1 &\myydots & \tilde{d}_1 &\myydots&&&&&&&&\\
\vdots&-\Tilde{w}_{\TypeIIoutlier,2}&\vdots&&\vdots&\tilde{d}_2&-w_2&\myydots&-w_2&\myydots&&&&\\
&-\Tilde{w}_{\TypeIIoutlier,2}&&&&-w_2&\tilde{d}_2&\myydots&-w_2&\myydots \\
&\vdots&&&&\vdots&\vdots&\ddots&&& & &  &\\
&-\Tilde{w}_{\TypeIIoutlier,2}& &&&-w_2 &-w_2 &\myydots & \tilde{d}_2 &\myydots&& & &\\
&\vdots& & & &\vdots& \vdots& &\vdots &\ddots & &\vdots& & \vdots\\
&-\Tilde{w}_{\TypeIIoutlier,\blocknum}& & & &&&&&&\tilde{d}_{\blocknum}&-w_{\blocknum}&\myydots&-w_{\blocknum}\\
&-\Tilde{w}_{\TypeIIoutlier,\blocknum}& & & &&&&&\myydots&-w_{\blocknum}&\tilde{d}_{\blocknum}& \myydots& -w_{\blocknum} \\
&\vdots& & & &&&&&&\vdots&\vdots&\ddots\\
&-\Tilde{w}_{\TypeIIoutlier,\blocknum}& & & &&&&&\myydots&-w_{\blocknum}& -w_{\blocknum} &\myydots&\tilde{d}_{\blocknum}
\end{smallmatrix}
\end{bmatrix}}\\\vspace{-5mm}
\Tilde{\mathbf{L}}_{(\indexm_{\TypeIIoutlier}=\ell_2-1)}=
\underdot{
\begin{bmatrix}
    \begin{smallmatrix}
\tilde{d}_1& -w_1&\myydots&-w_1&-\Tilde{w}_{\TypeIIoutlier,1}&\myydots&&&& & & &&\\
-w_1&\tilde{d}_1&\myydots&-w_1&-\Tilde{w}_{\TypeIIoutlier,1}&\myydots&&&&\\
\vdots&\vdots&\ddots&\vdots&\vdots&&&&&& & & &\\
-w_1 &-w_1 &\myydots & \tilde{d}_1 &-\Tilde{w}_{\TypeIIoutlier,1}&\myydots&&&&&&&&\\
-\Tilde{w}_{\TypeIIoutlier,1}& -\Tilde{w}_{\TypeIIoutlier,1}&\myydots&-\Tilde{w}_{\TypeIIoutlier,1}&\Tilde{d}_{\TypeIIoutlier}&-\Tilde{w}_{\TypeIIoutlier,2}&-\Tilde{w}_{\TypeIIoutlier,2}&\myydots&-\Tilde{w}_{\TypeIIoutlier,2}&\myydots &-\Tilde{w}_{\TypeIIoutlier,\blocknum} &-\Tilde{w}_{\TypeIIoutlier,\blocknum} &\myydots&-\Tilde{w}_{\TypeIIoutlier,\blocknum}\\
\vdots&\vdots&&\vdots&-\Tilde{w}_{\TypeIIoutlier,2}&\tilde{d}_2&-w_2&\myydots&-w_2&\myydots&&&&\\
&&&&-\Tilde{w}_{\TypeIIoutlier,2}&-w_2&\tilde{d}_2&\myydots&-w_2&\myydots \\
&&&&\vdots&\vdots&\vdots&\ddots&&& & &  &\\
& &&&-\Tilde{w}_{\TypeIIoutlier,2}&-w_2 &-w_2 &\myydots & \tilde{d}_2 &\myydots&& & &\\
& & & &\vdots&\vdots& \vdots& &\vdots &\ddots & &\vdots& & \vdots\\
& & & &-\Tilde{w}_{\TypeIIoutlier,\blocknum}&&&&&&\tilde{d}_{\blocknum}&-w_{\blocknum}&\myydots&-w_{\blocknum}\\
&& & & -\Tilde{w}_{\TypeIIoutlier,\blocknum}&&&&&\myydots&-w_{\blocknum}&\tilde{d}_{\blocknum}& \myydots& -w_{\blocknum} \\
&& & & \vdots&&&&&&\vdots&\vdots&\ddots\\
&& & &-\Tilde{w}_{\TypeIIoutlier,\blocknum} &&&&&\myydots&-w_{\blocknum}& -w_{\blocknum} &\myydots&\tilde{d}_{\blocknum}
\end{smallmatrix}
\end{bmatrix}}
\\\vspace{-5mm}
\Tilde{\mathbf{L}}_{(\indexm_{\TypeIIoutlier}=\dimN+1)}=
\begin{bmatrix}
    \begin{smallmatrix}
\tilde{d}_1& -w_1&\myydots&-w_1&\myydots&&&& & & &&&-\Tilde{w}_{\TypeIIoutlier,1}\\
-w_1&\tilde{d}_1&\myydots&-w_1&\myydots&&&&&&&&&-\Tilde{w}_{\TypeIIoutlier,1}\\
\vdots&\vdots&\ddots&\vdots&&&&&& & & &&\vdots\\
-w_1 &-w_1 &\myydots & \tilde{d}_1 &\myydots&&&&&&&&&-\Tilde{w}_{\TypeIIoutlier,1}\\
\vdots&\vdots&&\vdots&\tilde{d}_2&-w_2&\myydots&-w_2&\myydots&&&&&-\Tilde{w}_{\TypeIIoutlier,2}\\
&&&&-w_2&\tilde{d}_2&\myydots&-w_2&\myydots&&&&&-\Tilde{w}_{\TypeIIoutlier,2} \\
&&&&\vdots&\vdots&\ddots&&& & &  &&\vdots\\
& &&&-w_2 &-w_2 &\myydots & \tilde{d}_2 &\myydots&& & &&-\Tilde{w}_{\TypeIIoutlier,2}\\
& & & &\vdots&\vdots& &\vdots &\ddots &\vdots &&& & \vdots\\
& & & &&&&&&\tilde{d}_{\blocknum}&-w_{\blocknum}&\myydots&-w_{\blocknum}&-\Tilde{w}_{\TypeIIoutlier,\blocknum}\\
& & & &&&&&\myydots&-w_{\blocknum}&\tilde{d}_{\blocknum}& \myydots& -w_{\blocknum} &-\Tilde{w}_{\TypeIIoutlier,\blocknum}\\
& & & &&&&&&\vdots&\vdots&\ddots&\vdots\\
& & & &&&&&\myydots&-w_{\blocknum}& -w_{\blocknum} &\myydots&\tilde{d}_{\blocknum}&-\Tilde{w}_{\TypeIIoutlier,\blocknum}\\
-\Tilde{w}_{\TypeIIoutlier,1}& -\Tilde{w}_{\TypeIIoutlier,1}&\myydots&-\Tilde{w}_{\TypeIIoutlier,1}&-\Tilde{w}_{\TypeIIoutlier,2}&-\Tilde{w}_{\TypeIIoutlier,2}&\myydots&-\Tilde{w}_{\TypeIIoutlier,2}&\myydots &-\Tilde{w}_{\TypeIIoutlier,\blocknum} &-\Tilde{w}_{\TypeIIoutlier,\blocknum} &\myydots&-\Tilde{w}_{\TypeIIoutlier,\blocknum}&\Tilde{d}_{\TypeIIoutlier}\\
\end{smallmatrix}
\end{bmatrix}\vspace{-2mm}
\end{align*}
\end{minipage}}
\\
\newpage
Then, for each position of the outlier such that $0<m_{\TypeIIoutlier}\leq \dimN+1, m_{\TypeIIoutlier}\in\mathbb{Z}^{+}$, the vector $\tilde{\mathbf{v}}$ is computed as\\
\resizebox{\linewidth}{!}{
  \begin{minipage}{\linewidth}\vspace{3mm}
\begin{align*}
\begin{split}
 \Tilde{\mathbf{v}}_{(m_{\TypeIIoutlier}=1)}=&[\underbrace{0}_{\Tilde{v}_{\TypeIIoutlier}\in\mathbb{R}},\underbrace{\Tilde{w}_{\TypeIIoutlier,1},\Tilde{w}_{\TypeIIoutlier,1}+w_1,\myydots,\Tilde{w}_{\TypeIIoutlier,1}+(\dimN_1-1)w_1}_{\Tilde{\mathbf{v}}_1\in\mathbb{R}^{\dimN_1}},\underbrace{\Tilde{w}_{\TypeIIoutlier,2},\Tilde{w}_{\TypeIIoutlier,2}+w_2,\myydots,\Tilde{w}_{\TypeIIoutlier,2}+(\dimN_2-1)w_2}_{\Tilde{\mathbf{v}}_2\in\mathbb{R}^{\dimN_2}},\myydots,\underbrace{\Tilde{w}_{\TypeIIoutlier,\blocknum},\Tilde{w}_{\TypeIIoutlier,\blocknum}+w_{\blocknum},\myydots,\Tilde{w}_{\TypeIIoutlier,\blocknum}+(\dimN_{\blocknum}-1)w_{\blocknum}}_{\Tilde{\mathbf{v}}_{\blocknum}\in\mathbb{R}^{\dimN_{\blocknum}}}]\\
 \Tilde{\mathbf{v}}_{(m_{\TypeIIoutlier}=2)}=&[ \underbrace{0,
 \underbrace{\Tilde{w}_{\TypeIIoutlier,1}}_{\Tilde{v}_{\TypeIIoutlier}\in\mathbb{R}},\Tilde{w}_{\TypeIIoutlier,1}+w_1,\myydots,\Tilde{w}_{\TypeIIoutlier,1}+(\dimN_1-1)w_1}_{\Tilde{\mathbf{v}}_1\in\mathbb{R}^{\dimN_1}\cup\Tilde{v}_{\TypeIIoutlier}\in\mathbb{R}},\underbrace{\Tilde{w}_{\TypeIIoutlier,2},\Tilde{w}_{\TypeIIoutlier,2}+w_2,\myydots,\Tilde{w}_{\TypeIIoutlier,2}+(\dimN_2-1)w_2}_{\Tilde{\mathbf{v}}_2\in\mathbb{R}^{\dimN_2}},\myydots,\underbrace{\Tilde{w}_{\TypeIIoutlier,\blocknum},\Tilde{w}_{\TypeIIoutlier,\blocknum}+w_{\blocknum},\myydots,\Tilde{w}_{\TypeIIoutlier,\blocknum}+(\dimN_{\blocknum}-1)w_{\blocknum}}_{\Tilde{\mathbf{v}}_{\blocknum}\in\mathbb{R}^{\dimN_{\blocknum}}}]\\
 \vdots&\\
 \Tilde{\mathbf{v}}_{(m_{\TypeIIoutlier}=\ell_2-1)}=&[ \underbrace{0,w_1,\myydots,(\dimN_1-1)w_1}_{\Tilde{\mathbf{v}}_1\in\mathbb{R}^{\dimN_1}},\underbrace{\dimN_1\Tilde{w}_{\TypeIIoutlier,1}}_{\Tilde{v}_{\TypeIIoutlier}\in\mathbb{R}},\underbrace{\Tilde{w}_{\TypeIIoutlier,2},\Tilde{w}_{\TypeIIoutlier,2}+w_2,\myydots,\Tilde{w}_{\TypeIIoutlier,2}+(\dimN_2-1)w_2}_{\Tilde{\mathbf{v}}_2\in\mathbb{R}^{\dimN_2}},\myydots,\underbrace{\Tilde{w}_{\TypeIIoutlier,\blocknum},\Tilde{w}_{\TypeIIoutlier,\blocknum}+w_{\blocknum},\myydots,\Tilde{w}_{\TypeIIoutlier,\blocknum}+(\dimN_{\blocknum}-1)w_{\blocknum}}_{\Tilde{\mathbf{v}}_{\blocknum}\in\mathbb{R}^{\dimN_{\blocknum}}}]\\
 \vdots&\\
 \Tilde{\mathbf{v}}_{(m_{\TypeIIoutlier}=\dimN+1)}=&[ \underbrace{0,w_1,\myydots,(\dimN_1-1)w_1}_{\Tilde{\mathbf{v}}_1\in\mathbb{R}^{\dimN_1}},\underbrace{0,w_2,\myydots,(\dimN_2-1)w_2}_{\Tilde{\mathbf{v}}_2\in\mathbb{R}^{\dimN_2}},\myydots,\underbrace{0,w_{\blocknum},\myydots,(\dimN_{\blocknum}-1)w_{\blocknum}}_{\Tilde{\mathbf{v}}_{\blocknum}\in\mathbb{R}^{\dimN_{\blocknum}}},\underbrace{\sum\limits_{\indexj=1}^{\blocknum}\dimN_{\indexj}\Tilde{w}_{\TypeIIoutlier,\indexj}}_{\Tilde{v}_{\TypeIIoutlier}\in\mathbb{R}}]
\end{split}
\end{align*}
\end{minipage}}\vspace{3mm}\\
As it can be seen, the components, whose indexes are valued between the outlier
index and the largest index of the $\indexj$th block such that $m_{\TypeIIoutlier}<m\leq u_{\indexj}$, increase by $\Tilde{w}_{\TypeIIoutlier,\indexj}$ in the corrupted vector $\Tilde{\mathbf{v}}\in\mathbb{R}^{\dimN+1}$. Further, the component associated with the outlier is valued as follows
\begin{align*}
   \Tilde{v}_{\TypeIIoutlier}=\begin{cases}
   0\hspace{1.3cm}&\mathrm{if}\hspace{5mm}0<m_{\TypeIIoutlier}<\ell_1\\
   (m_{\TypeIIoutlier}-\ell_1)\Tilde{w}_{\TypeIIoutlier,1}&\mathrm{if}\hspace{5mm}\ell_1<m_{\TypeIIoutlier}<\ell_2\\
   &\vdots\\
   \sum\limits_{\indexj=1}^{\blocknum-1}\dimN_{\indexj}\Tilde{w}_{\TypeIIoutlier,\indexj}+(m_{\TypeIIoutlier}-\ell_{\blocknum})\Tilde{w}_{\TypeIIoutlier,\blocknum}&\mathrm{if}\hspace{5mm}\ell_{\blocknum}<m_{\TypeIIoutlier}\leq \dimN+1
   \end{cases},
\end{align*}
where $\ell_{\indexj}$ denotes the lowest index of the $\indexj$th block for $\indexj=1,\myydots,\blocknum$.
\end{proof}

\begin{figure*}[tbp!]
  \centering
  \captionsetup{justification=centering}
\subfloat[ $\tilde{G}=\{\tilde{V},\tilde{E},\tilde{\mathbf{W}}\}$ ]{\includegraphics[trim={0cm 0cm 0cm 0cm},clip,width=4.25cm]{Images_Accompanying/GraphModel_oequal1_TypeIIOutliers.pdf}}
\subfloat[$\tilde{\mathbf{W}}\in\mathbb{R}^{(\dimN+1)\times (\dimN+1)}$]{\includegraphics[trim={0mm 0mm 0mm 0mm},clip,width=3.9cm]{Images_Accompanying/TypeIIOutlierEffect_oequal1_Affinity.pdf}}\hspace{1mm}
\subfloat[$\tilde{\mathbf{L}}\in\mathbb{R}^{(\dimN+1)\times (\dimN+1)}$]{\includegraphics[trim={0mm 0mm 0mm 0mm},clip,width=4.95cm]{Images_Accompanying/TypeIIOutlierEffect_oequal1_Laplacian.pdf}}
\subfloat[$\tilde{\mathbf{v}}\in\mathbb{R}^{\dimN+1}$]{\includegraphics[trim={0mm 0mm 0mm 0mm},clip,width=4.75cm]{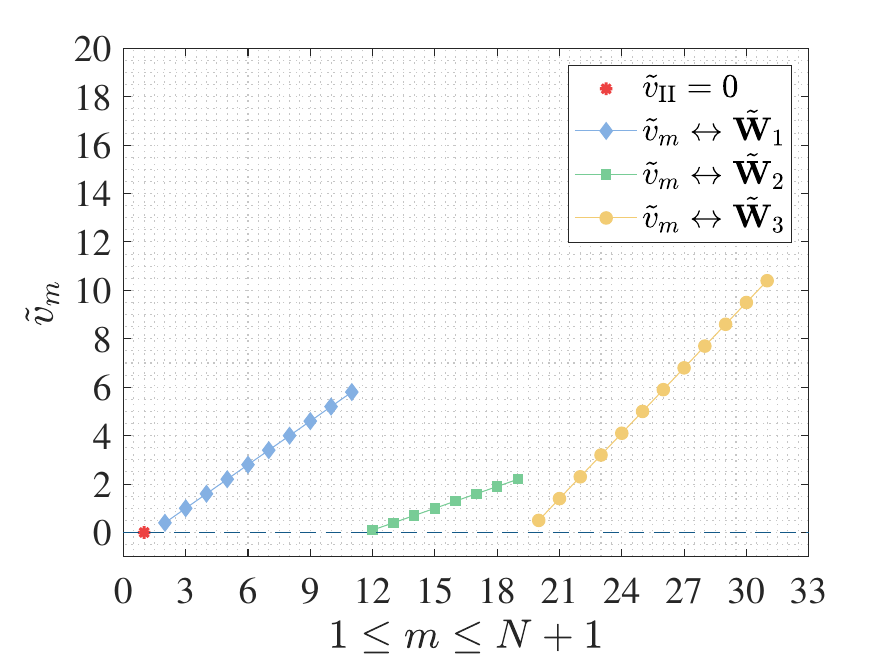}}
\\
\caption{Examplary plot of Theorem~3 ($\mathbf{n}=[1,10,8,12]^{\top}\in\mathbb{R}^{\blocknum+1}$, $\dimN+1=31$, $\blocknum=3$, $m_{\TypeIIoutlier}=1$).}
  \label{fig:TypeIIOutlierEffectoequal1vectorvanalysis}
\end{figure*}

\begin{figure*}[tbp!]
  \centering
  \captionsetup{justification=centering}
\subfloat[ $\tilde{G}=\{\tilde{V},\tilde{E},\tilde{\mathbf{W}}\}$ ]{\includegraphics[trim={0cm 0cm 0cm 0cm},clip,width=4.25cm]{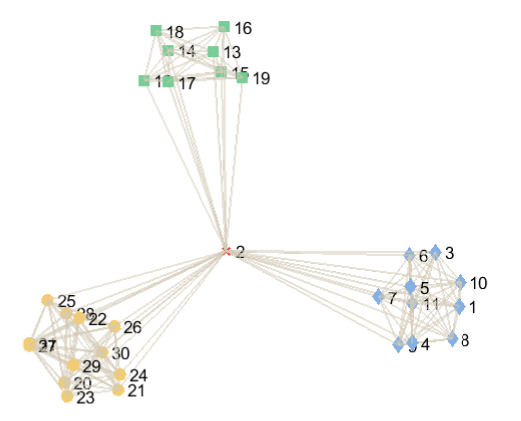}}
\subfloat[$\tilde{\mathbf{W}}\in\mathbb{R}^{(\dimN+1)\times (\dimN+1)}$]{\includegraphics[trim={0mm 0mm 0mm 0mm},clip,width=3.93cm]{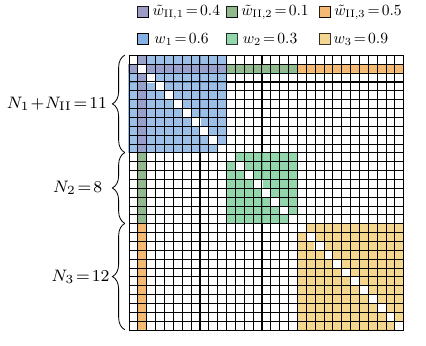}}\hspace{1mm}
\subfloat[$\tilde{\mathbf{L}}\in\mathbb{R}^{(\dimN+1)\times (\dimN+1)}$]{\includegraphics[trim={0mm 0mm 0mm 0mm},clip,width=4.92cm]{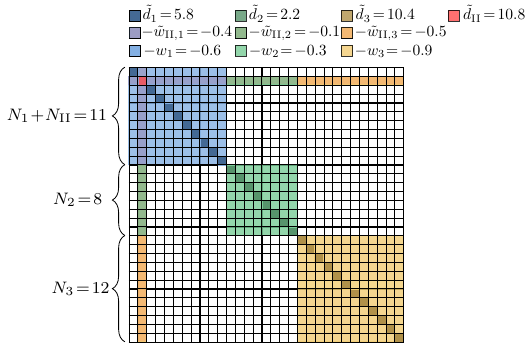}}
\subfloat[$\tilde{\mathbf{v}}\in\mathbb{R}^{\dimN+1}$]{\includegraphics[trim={0mm 0mm 0mm 0mm},clip,width=4.75cm]{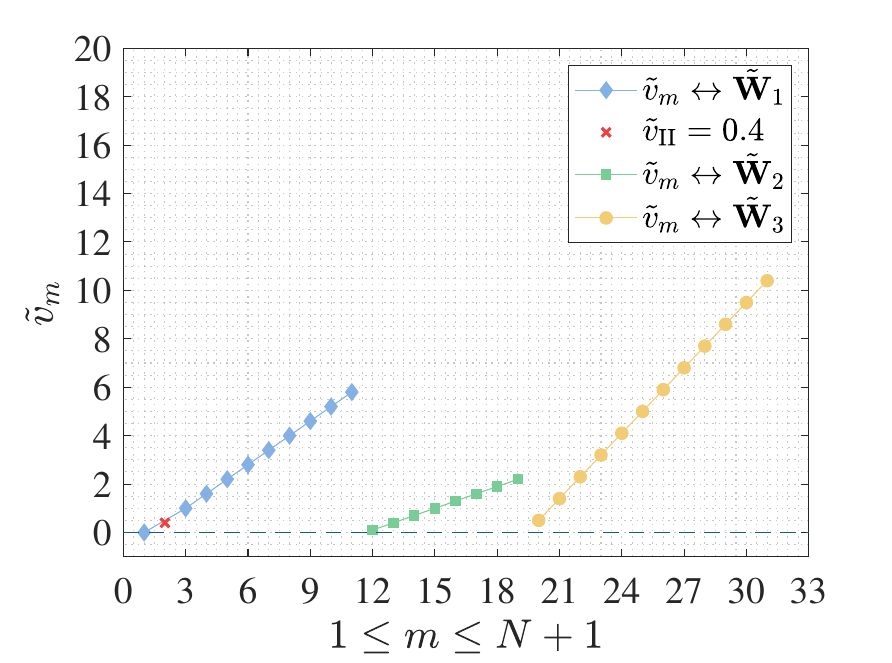}}
\\
\caption{Examplary plot of Theorem~3 ($\mathbf{n}=[1,1,9,8,12]^{\top}\in\mathbb{R}^{\blocknum+2}$, $\dimN+1=31$, $\blocknum=3$, $m_{\TypeIIoutlier}=2$).}
  \label{fig:TypeIIOutlierEffectoequal2vectorvanalysis}
\end{figure*}

\begin{figure*}[tbp!]
  \centering
  \captionsetup{justification=centering}
\subfloat[ $\tilde{G}=\{\tilde{V},\tilde{E},\tilde{\mathbf{W}}\}$ ]{\includegraphics[trim={0cm 0cm 0cm 0cm},clip,width=4.25cm]{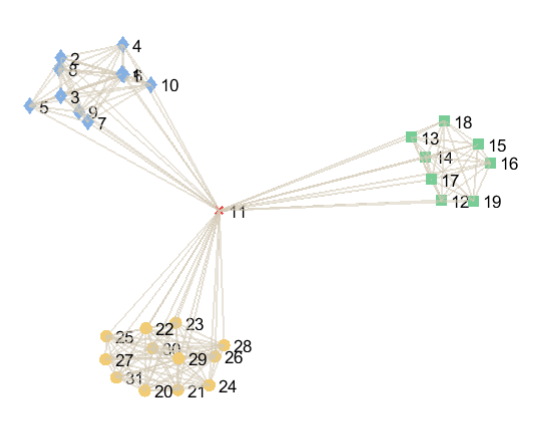}}
\subfloat[$\tilde{\mathbf{W}}\in\mathbb{R}^{(\dimN+1)\times (\dimN+1)}$]{\includegraphics[trim={0mm 0mm 0mm 0mm},clip,width=3.9cm]{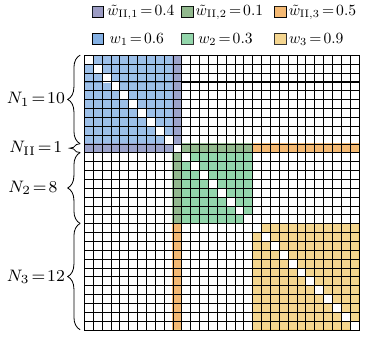}}\hspace{1mm}
\subfloat[$\tilde{\mathbf{L}}\in\mathbb{R}^{(\dimN+1)\times (\dimN+1)}$]{\includegraphics[trim={0mm 0mm 0mm 0mm},clip,width=4.95cm]{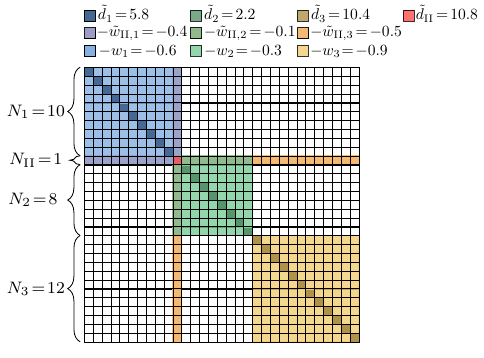}}
\subfloat[$\tilde{\mathbf{v}}\in\mathbb{R}^{\dimN+1}$]{\includegraphics[trim={0mm 0mm 0mm 0mm},clip,width=4.75cm]{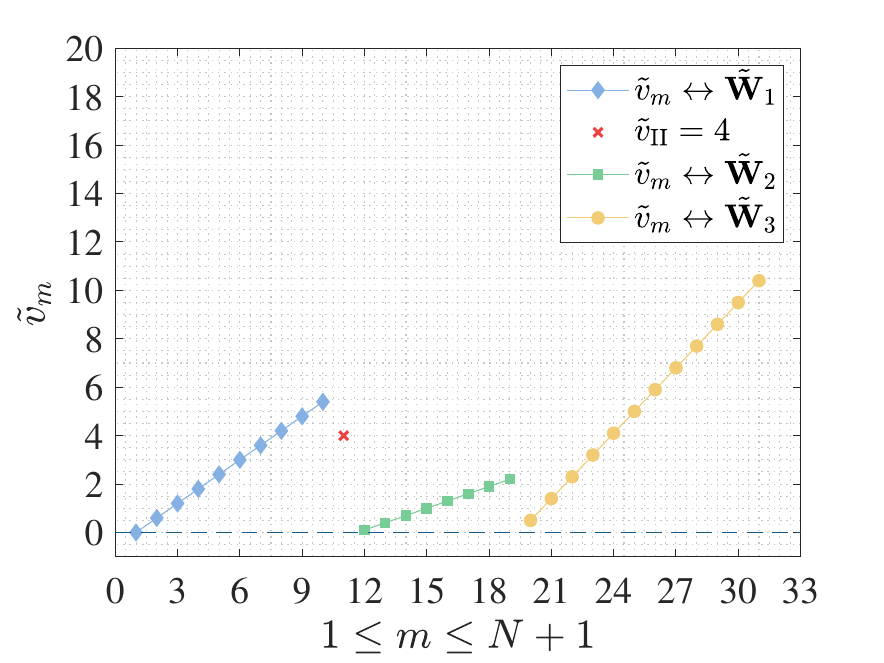}}
\\
\caption{Examplary plot of Theorem~3 ($\mathbf{n}=[10,1,8,12]^{\top}\in\mathbb{R}^{\blocknum+1}$, $\dimN+1=31$, $\blocknum=3$, $m_{\TypeIIoutlier}=\ell_2-1$).}
  \label{fig:TypeIIOutlierEffectoequal\dimN_1plus1vectorvanalysis}\vspace{-4mm}
\end{figure*}

\begin{figure*}[tbp!]\vspace{-3mm}
  \centering
  \captionsetup{justification=centering}
\subfloat[ $\tilde{G}=\{\tilde{V},\tilde{E},\tilde{\mathbf{W}}\}$ ]{\includegraphics[trim={0cm 0cm 0cm 0cm},clip,width=4.25cm]{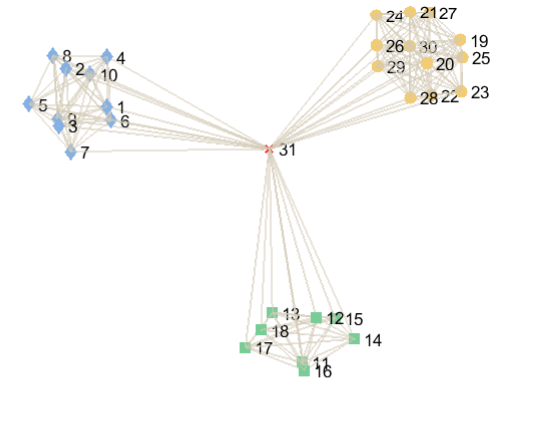}}
\subfloat[$\tilde{\mathbf{W}}\in\mathbb{R}^{(\dimN+1)\times (\dimN+1)}$]{\includegraphics[trim={0mm 0mm 0mm 0mm},clip,width=3.9cm]{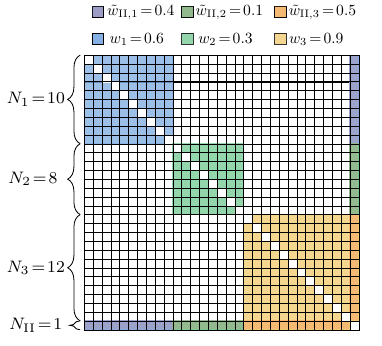}}\hspace{1mm}
\subfloat[$\tilde{\mathbf{L}}\in\mathbb{R}^{(\dimN+1)\times (\dimN+1)}$]{\includegraphics[trim={0mm 0mm 0mm 0mm},clip,width=4.95cm]{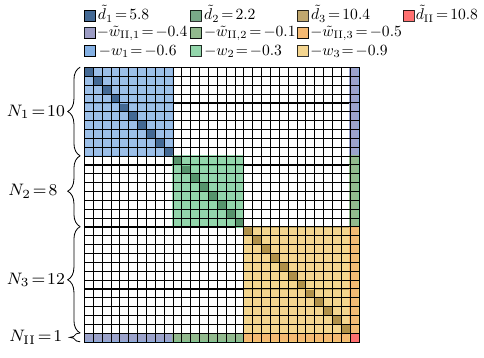}}
\subfloat[$\tilde{\mathbf{v}}\in\mathbb{R}^{\dimN+1}$]{\includegraphics[trim={0mm 0mm 0mm 0mm},clip,width=4.75cm]{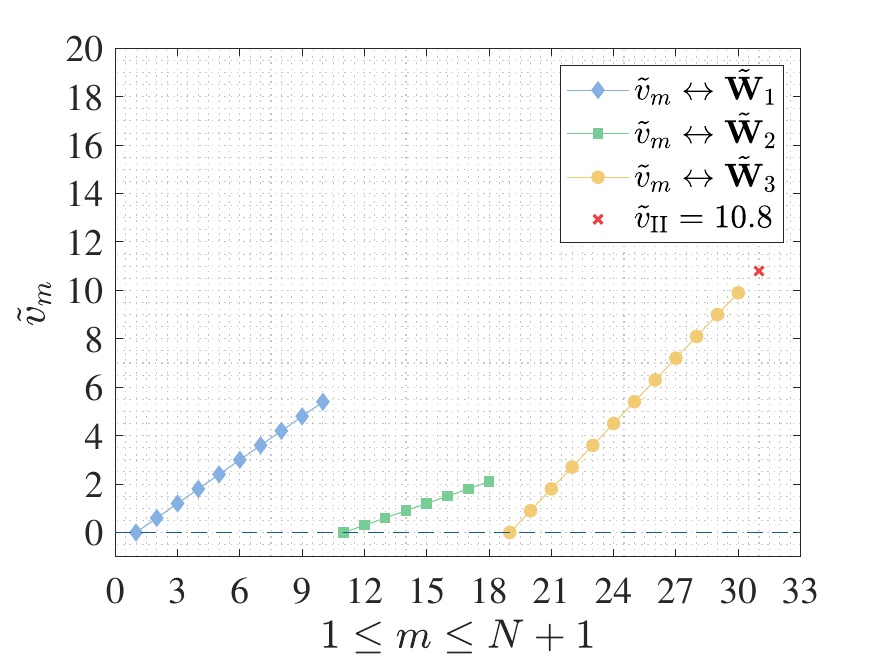}}
\\
\caption{Examplary plot of Theorem~3 ($\mathbf{n}=[10,8,12,1]^{\top}\in\mathbb{R}^{\blocknum+1}$, $\dimN+1=31$, $\blocknum=3$, $m_{\TypeIIoutlier}=\dimN+1$).}
  \label{fig:TypeIIOutlierEffectoequalnplus1vectorvanalysis}\vspace{-3mm}
\end{figure*}
\newpage

\subsection{C.2~Outlier Effects on Target Vector $\mathbf{v}$: Proof of Theorem~4 and Corollary~4.1}
\subsubsection{Proof of Theorem~4}
\begin{proof}[\unskip\nopunct]
To analyze different positions of group similarity on the target vector $\mathbf{v}$, the block $\indexi$ that has similarity with the remaining $\blocknum-1$ blocks is shifted along the diagonal of corrupted Laplacian matrix $\Tilde{\mathbf{L}}\in\mathbb{R}^{\dimN\times\dimN}$ as follows
\begin{align*}
\Tilde{\mathbf{L}}_{\indexi=1}=
\underdot{\begin{bmatrix}
    \begin{smallmatrix}
\tilde{d}_1& -w_1&\myydots&-w_1&-\tilde{w}_{1,2}&-\tilde{w}_{1,2} &\myydots &-\tilde{w}_{1,2} &\myydots &-\tilde{w}_{1,\blocknum} &-\tilde{w}_{1,\blocknum} &\myydots &-\tilde{w}_{1,\blocknum}\\
-w_1&\tilde{d}_1&\myydots&-w_1&-\tilde{w}_{1,2} &-\tilde{w}_{1,2} &\myydots &-\tilde{w}_{1,2}  &\myydots &-\tilde{w}_{1,\blocknum} &-\tilde{w}_{1,\blocknum} &\myydots &-\tilde{w}_{1,\blocknum}\\
\vdots&\vdots&\ddots&\vdots&\vdots&\vdots &\ddots &\vdots  &&\vdots &\vdots &\ddots &\vdots\\
-w_1 &-w_1 &\myydots & \tilde{d}_1 &-\tilde{w}_{1,2}&-\tilde{w}_{1,2}&\myydots &-\tilde{w}_{1,2} &\myydots &-\tilde{w}_{1,\blocknum} &-\tilde{w}_{1,\blocknum} &\myydots &-\tilde{w}_{1,\blocknum}\\
-\tilde{w}_{1,2}&-\tilde{w}_{1,2} &\myydots &-\tilde{w}_{1,2} &\tilde{d}_2& -w_2&\myydots&-w_2&\myydots& & & & \\
-\tilde{w}_{1,2}&-\tilde{w}_{1,2} &\myydots & -\tilde{w}_{1,2} &-w_2&\tilde{d}_2&\myydots&-w_2&\myydots \\
\vdots&\vdots &\ddots &\vdots &\vdots&\vdots&\ddots&&& & &  &\\
-\tilde{w}_{1,2} &-\tilde{w}_{1,2} &\myydots&-\tilde{w}_{1,2} &-w_2 &-w_2 &\myydots & \tilde{d}_2 &\myydots&& & &\\
\vdots&\vdots & &\vdots &\vdots& \vdots& & &\ddots & &\vdots& & \vdots\\
-\tilde{w}_{1,\blocknum}&-\tilde{w}_{1,\blocknum} &\myydots &-\tilde{w}_{1,\blocknum} &&&&&&\tilde{d}_{\blocknum}&-w_{\blocknum}&\myydots&-w_{\blocknum}\\
-\tilde{w}_{1,\blocknum}&-\tilde{w}_{1,\blocknum} &\myydots &-\tilde{w}_{1,\blocknum} &&&&&\myydots&-w_{\blocknum}&\tilde{d}_{\blocknum}& \myydots& -w_{\blocknum} \\
\vdots&\vdots &\ddots &\vdots &&&&&&\vdots&\vdots&\ddots\\
-\tilde{w}_{1,\blocknum}&-\tilde{w}_{1,\blocknum} &\myydots &-\tilde{w}_{1,\blocknum} &&&&&\myydots&-w_{\blocknum}& -w_{\blocknum} &\myydots&\tilde{d}_{\blocknum}\
\end{smallmatrix}
    \end{bmatrix}\vspace{5mm}}\\\vspace{5mm}
\Tilde{\mathbf{L}}_{\indexi=\blocknum}=
\begin{bmatrix}
    \begin{smallmatrix}
\tilde{d}_1& -w_1&\myydots&-w_1&\myydots& & &&\myydots &-\tilde{w}_{1,\blocknum} &-\tilde{w}_{1,\blocknum} &\myydots &-\tilde{w}_{1,\blocknum}\\
-w_1&\tilde{d}_1&\myydots&-w_1&\myydots & & &&\myydots &-\tilde{w}_{1,\blocknum} &-\tilde{w}_{1,\blocknum} &\myydots &-\tilde{w}_{1,\blocknum}\\
\vdots&\vdots&\ddots&\vdots&&&&&&\vdots &\vdots &\ddots &\vdots\\
-w_1 &-w_1 &\myydots & \tilde{d}_1 &\myydots&& &&\myydots &-\tilde{w}_{1,\blocknum} &-\tilde{w}_{1,\blocknum} &\myydots &-\tilde{w}_{1,\blocknum}\\
\vdots&\vdots & &\vdots &\tilde{d}_2& -w_2&\myydots&-w_2&\myydots&-\Tilde{w}_{2,\blocknum} &-\Tilde{w}_{2,\blocknum} &\myydots &-\Tilde{w}_{2,\blocknum} \\
&& & &-w_2&\tilde{d}_2&\myydots&-w_2&\myydots&-\Tilde{w}_{2,\blocknum} &-\Tilde{w}_{2,\blocknum} &\myydots &-\Tilde{w}_{2,\blocknum}\\
&&&&\vdots&\vdots&\ddots&&&\vdots &\vdots &\ddots &\vdots\\
 & && &-w_2 &-w_2 &\myydots & \tilde{d}_2 &\myydots&-\Tilde{w}_{2,\blocknum} &-\Tilde{w}_{2,\blocknum} &\myydots &-\Tilde{w}_{2,\blocknum}\\
\vdots&\vdots & &\vdots &\vdots& \vdots& & &\ddots & &\vdots& & \vdots\\
-\tilde{w}_{1,\blocknum}&-\tilde{w}_{1,\blocknum} &\myydots &-\tilde{w}_{1,\blocknum} &-\Tilde{w}_{2,\blocknum}&-\Tilde{w}_{2,\blocknum}&\myydots&-\Tilde{w}_{2,\blocknum}&\myydots&\tilde{d}_{\blocknum}&-w_{\blocknum}&\myydots&-w_{\blocknum}\\
-\tilde{w}_{1,\blocknum}&-\tilde{w}_{1,\blocknum} &\myydots &-\tilde{w}_{1,\blocknum} &-\Tilde{w}_{2,\blocknum}&-\Tilde{w}_{2,\blocknum}&\myydots&-\Tilde{w}_{2,\blocknum}&\myydots&-w_{\blocknum}&\tilde{d}_{\blocknum}& \myydots& -w_{\blocknum} \\
\vdots&\vdots &\ddots &\vdots &\vdots&\vdots&\ddots&\vdots&&\vdots&\vdots&\ddots\\
-\tilde{w}_{1,\blocknum}&-\tilde{w}_{1,\blocknum} &\myydots &-\tilde{w}_{1,\blocknum} &-\Tilde{w}_{2,\blocknum}&-\Tilde{w}_{2,\blocknum}&\myydots&-\Tilde{w}_{2,\blocknum}&\myydots&-w_{\blocknum}& -w_{\blocknum} &\myydots&\tilde{d}_{\blocknum}\
\end{smallmatrix}
    \end{bmatrix}.
\end{align*}
\newpage
Then, for each position of the block $\indexi$ such that $\indexi=1,\myydots,\blocknum$, the vector $\tilde{\mathbf{v}}\in\mathbb{R}^{\dimN}$ is computed as\\
\resizebox{\linewidth}{!}{
  \begin{minipage}{\linewidth}
\begin{align*}
\begin{split}
 \Tilde{\mathbf{v}}_{(\indexi=1)}=&[\underbrace{0,w_1,\myydots,(\dimN_1-1)w_1}_{\Tilde{\mathbf{v}}_1\in\mathbb{R}^{\dimN_1}},\underbrace{\dimN_1\tilde{w}_{1,2},w_2+\dimN_1\tilde{w}_{1,2},\myydots,(\dimN_2-1)w_2+\dimN_1\tilde{w}_{1,2}}_{\Tilde{\mathbf{v}}_2\in\mathbb{R}^{\dimN_2}},\myydots,\underbrace{\dimN_1\tilde{w}_{1,\blocknum},w_{\blocknum}+\dimN_1\tilde{w}_{1,\blocknum},\myydots,(\dimN_{\blocknum}-1)w_{\blocknum}+\dimN_1\tilde{w}_{1,\blocknum}}_{\Tilde{\mathbf{v}}_{\blocknum}\in\mathbb{R}^{\dimN_{\blocknum}}}]\\
 \vdots&\\
 \Tilde{\mathbf{v}}_{(\indexi=\blocknum)}=&[\underbrace{0,w_1,\myydots,(\dimN_1-1)w_1}_{\Tilde{\mathbf{v}}_1\in\mathbb{R}^{\dimN_1}},\underbrace{0,w_2,\myydots,(\dimN_2-1)w_2}_{\Tilde{\mathbf{v}}_2\in\mathbb{R}^{\dimN_2}},\myydots,\underbrace{\sum\limits_{\indexj=1}^{\blocknum-1}\dimN_{\indexj}\Tilde{w}_{\blocknum,\indexj},w_{\blocknum}+\sum\limits_{\indexj=1}^{\blocknum-1}\dimN_{\indexj}\Tilde{w}_{\blocknum,\indexj},\myydots,(\dimN_{\blocknum}-1)w_{\blocknum}+\sum\limits_{\indexj=1}^{\blocknum-1}\dimN_{\indexj}\Tilde{w}_{\blocknum,\indexj}}_{\Tilde{\mathbf{v}}_{\blocknum}\in\mathbb{R}^{\dimN_{\blocknum}}}].
 \end{split}
\end{align*}
\end{minipage}}\vspace{3mm}\\
Herein, the components associated with the blocks $\indexj>\indexi$ are increase by $\dimN_{\indexi}\tilde{w}_{\indexi,\indexj}$ while remain the same on the contrary. Further,
if $2\leq\indexi\leq\blocknum$ the components associated with the block $\indexi$ increase by $\sum_{\indexj=1}^{\indexi-1}\dimN_{\indexj}\tilde{w}_{\indexi,\indexj}$ and remain the same otherwise.\\
\vspace{-1mm}

In more details, starting from the $\indexi+1$th block undesired similarity blocks are located only on the lower triangular side. Therefore, summing upper triangular part of the Laplacian matrix results in an increase by $\dimN_{\indexi}\tilde{w}_{\indexi,\indexj}$ in these blocks. Additionally, for the $\indexi$th block $\indexi-1$ number of undesired similarity blocks are located on the lower triangular side which results in an increase by $\sum\limits_{\indexj=1}^{\indexi-1}\dimN_{\indexj}\tilde{w}_{\indexi,\indexj}$.

\end{proof}
\begin{figure*}[tbp!]
  \centering
  \captionsetup{justification=centering}
\subfloat[ $\tilde{G}=\{\tilde{V},\tilde{E},\tilde{\mathbf{W}}\}$ ]{\includegraphics[trim={0cm 0cm 0cm 0cm},clip,width=4.25cm]{Images_Accompanying/GroupSimilarity_kgroup_iequal1_GraphModel.pdf}}
\subfloat[$\tilde{\mathbf{W}}\in\mathbb{R}^{\dimN\times\dimN}$]{\includegraphics[trim={0mm 0mm 0mm 0mm},clip,width=4.25cm]{Images_Accompanying/GroupSimilarity_kgroups_iequal1_Affinity.pdf}}
\subfloat[$\tilde{\mathbf{L}}\in\mathbb{R}^{\dimN\times\dimN}$]{\includegraphics[trim={0mm 0mm 0mm 0mm},clip,width=4.75cm]{Images_Accompanying/GroupSimilarity_kgroups_iequal1_Laplacian.pdf}}
\subfloat[$\tilde{\mathbf{v}}\in\mathbb{R}^{\dimN}$]{\includegraphics[trim={0mm 0mm 0mm 0mm},clip,width=4.75cm]{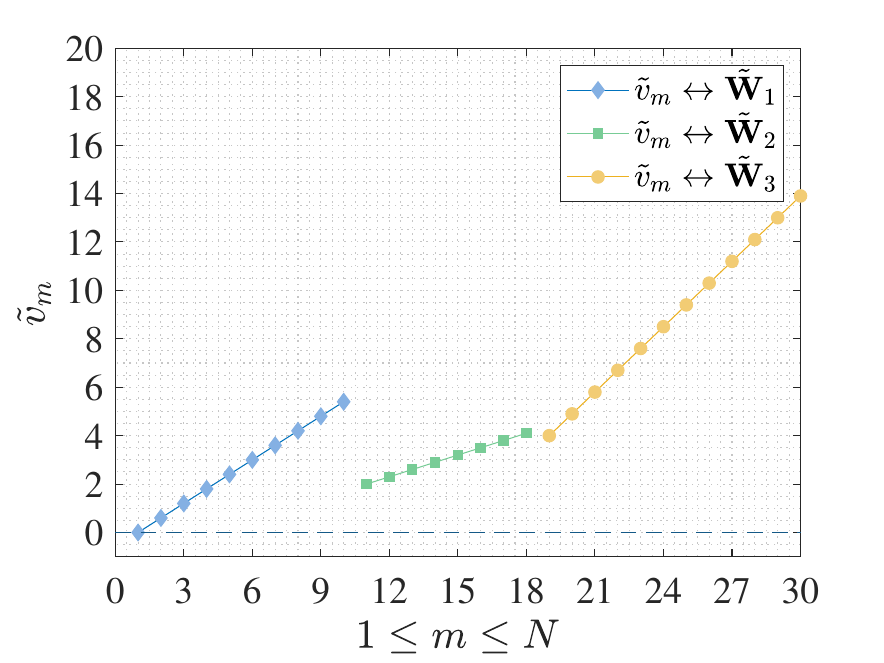}}
\\
\caption{Examplary plot of Theorem~4 ($\mathbf{n}=[10,8,12]^{\top}\in\mathbb{R}^{\blocknum}$, $\dimN=30$, $\blocknum=3$, $\indexi=1$).}
  \label{fig:Groupsimilarityeffectkgroupiequal1vectorv}
\end{figure*}
\begin{figure*}[tbp!]
  \centering
  \captionsetup{justification=centering}
\subfloat[ $\tilde{G}=\{\tilde{V},\tilde{E},\tilde{\mathbf{W}}\}$ ]{\includegraphics[trim={0cm 0cm 0cm 0cm},clip,width=4.25cm]{Images_Accompanying/GroupSimilarity_kgroup_iequalk_GraphModel.pdf}}
\subfloat[$\tilde{\mathbf{W}}\in\mathbb{R}^{\dimN\times\dimN}$]{\includegraphics[trim={0mm 0mm 0mm 0mm},clip,width=4.25cm]{Images_Accompanying/GroupSimilarity_kgroups_iequalk_Affinity.pdf}}
\subfloat[$\tilde{\mathbf{L}}\in\mathbb{R}^{\dimN\times\dimN}$]{\includegraphics[trim={0mm 0mm 0mm 0mm},clip,width=4.75cm]{Images_Accompanying/GroupSimilarity_kgroups_iequalk_Laplacian.pdf}}
\subfloat[$\tilde{\mathbf{v}}\in\mathbb{R}^{\dimN}$]{\includegraphics[trim={0mm 0mm 0mm 0mm},clip,width=4.75cm]{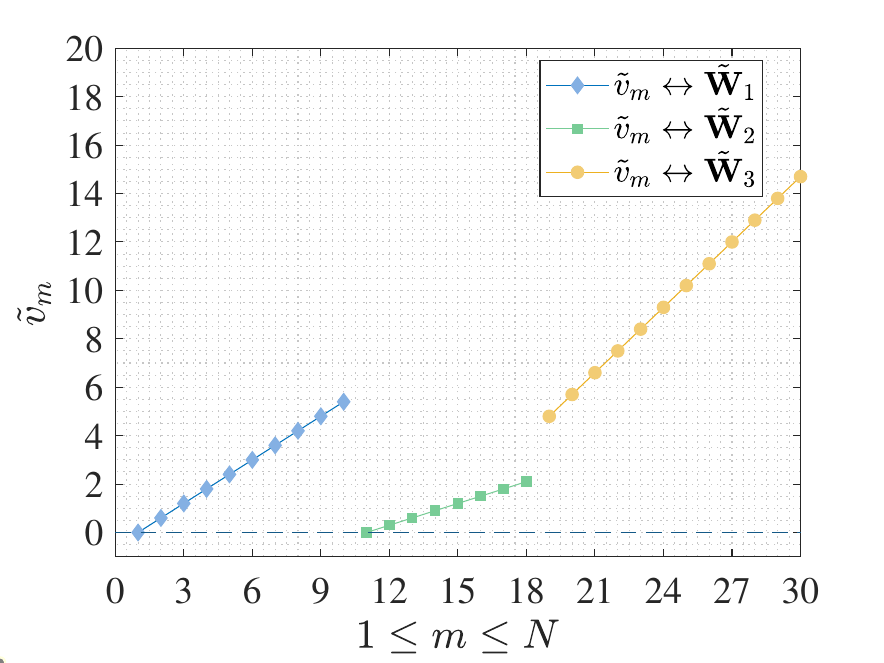}}
\\
\caption{Examplary plot of Theorem~4 ($\mathbf{n}=[10,8,12]^{\top}\in\mathbb{R}^{\blocknum}$, $\dimN=30$, $\blocknum=3$, $\indexi=\blocknum$).}
  \label{fig:Groupsimilarityeffectkgroupiequalkvectorv}
\end{figure*}
\newpage
\subsubsection{Proof of Corollary~4.1}
\begin{proof}[\unskip\nopunct]
Let $\Tilde{\mathbf{L}}\in\mathbb{R}^{\dimN\times\dimN}$ a corrupted Laplacian matrix, that is identical to $\mathbf{L}\in\mathbb{R}^{\dimN\times\dimN}$ except that each block ${\indexi=1,\myydots,\blocknum}$ has non-zero similarity coefficients with the remaining $\blocknum-1$ blocks, i.e.
\begin{align*}
\Tilde{\mathbf{L}}=
\begin{bmatrix}
    \begin{smallmatrix}
\tilde{d}_1& -w_1&\myydots&-w_1&-\tilde{w}_{1,2}&-\tilde{w}_{1,2} &\myydots &-\tilde{w}_{1,2} &\myydots &-\tilde{w}_{1,\blocknum} &-\tilde{w}_{1,\blocknum} &\myydots &-\tilde{w}_{1,\blocknum}\\
-w_1&\tilde{d}_1&\myydots&-w_1&-\tilde{w}_{1,2} &-\tilde{w}_{1,2} &\myydots &-\tilde{w}_{1,2}  &\myydots &-\tilde{w}_{1,\blocknum} &-\tilde{w}_{1,\blocknum} &\myydots &-\tilde{w}_{1,\blocknum}\\
\vdots&\vdots&\ddots&\vdots&\vdots&\vdots &\ddots &\vdots  &&\vdots &\vdots &\ddots &\vdots\\
-w_1 &-w_1 &\myydots & \tilde{d}_1 &-\tilde{w}_{1,2}&-\tilde{w}_{1,2}&\myydots &-\tilde{w}_{1,2} &\myydots &-\tilde{w}_{1,\blocknum} &-\tilde{w}_{1,\blocknum} &\myydots &-\tilde{w}_{1,\blocknum}\\
-\tilde{w}_{1,2}&-\tilde{w}_{1,2} &\myydots &-\tilde{w}_{1,2} &\tilde{d}_2& -w_2&\myydots&-w_2&\myydots&-\Tilde{w}_{2,\blocknum} &-\Tilde{w}_{2,\blocknum} &\myydots &-\Tilde{w}_{2,\blocknum} \\
-\tilde{w}_{1,2}&-\tilde{w}_{1,2} &\myydots & -\tilde{w}_{1,2} &-w_2&\tilde{d}_2&\myydots&-w_2&\myydots &-\Tilde{w}_{2,\blocknum} &-\Tilde{w}_{2,\blocknum} &\myydots &-\Tilde{w}_{2,\blocknum} \\
\vdots&\vdots &\ddots &\vdots &\vdots&\vdots&\ddots&&&\vdots &\vdots &\ddots  &\vdots\\
-\tilde{w}_{1,2} &-\tilde{w}_{1,2} &\myydots&-\tilde{w}_{1,2} &-w_2 &-w_2 &\myydots & \tilde{d}_2 &\myydots&-\Tilde{w}_{2,\blocknum} &-\Tilde{w}_{2,\blocknum} &\myydots &-\Tilde{w}_{2,\blocknum} \\
\vdots&\vdots & &\vdots &\vdots& \vdots& & &\ddots & &\vdots& & \vdots\\
-\tilde{w}_{1,\blocknum}&-\tilde{w}_{1,\blocknum} &\myydots &-\tilde{w}_{1,\blocknum} &-\Tilde{w}_{2,\blocknum}&-\Tilde{w}_{2,\blocknum}&\myydots&-\Tilde{w}_{2,\blocknum}&\myydots&\tilde{d}_{\blocknum}&-w_{\blocknum}&\myydots&-w_{\blocknum}\\
-\tilde{w}_{1,\blocknum}&-\tilde{w}_{1,\blocknum} &\myydots &-\tilde{w}_{1,\blocknum} &-\Tilde{w}_{2,\blocknum}&-\Tilde{w}_{2,\blocknum}&\myydots&-\Tilde{w}_{2,\blocknum}&\myydots&-w_{\blocknum}&\tilde{d}_{\blocknum}& \myydots& -w_{\blocknum} \\
\vdots&\vdots &\ddots &\vdots &\vdots&\vdots&\ddots&\vdots&&\vdots&\vdots&\ddots\\
-\tilde{w}_{1,\blocknum}&-\tilde{w}_{1,\blocknum} &\myydots &-\tilde{w}_{1,\blocknum} &-\Tilde{w}_{2,\blocknum}&-\Tilde{w}_{2,\blocknum}&\myydots&-\Tilde{w}_{2,\blocknum}&\myydots&-w_{\blocknum}& -w_{\blocknum} &\myydots&\tilde{d}_{\blocknum}\
\end{smallmatrix}
    \end{bmatrix}.
\end{align*}\\
Then, the vector $\Tilde{\mathbf{v}}$ associated with $\Tilde{\mathbf{L}}$ yields\\
\resizebox{\linewidth}{!}{
  \begin{minipage}{\linewidth}
\begin{align*}
 \hspace{-2mm}\Tilde{\mathbf{v}}\hspace{-0.7mm}=\hspace{-0.7mm}[\underbrace{0,w_1,\hspace{-0.5mm}\myydots,\hspace{-0.5mm}(\dimN_1-1)w_1}_{\Tilde{\mathbf{v}}_1\in\mathbb{R}^{\dimN_1}},\underbrace{\dimN_1\tilde{w}_{1,2},w_2\hspace{-0.5mm}+\hspace{-0.5mm}\dimN_1\tilde{w}_{1,2},\hspace{-0.5mm}\myydots,\hspace{-0.5mm}(\dimN_2-1)w_2\hspace{-0.5mm}+\hspace{-0.5mm}\dimN_1\tilde{w}_{1,2}}_{\Tilde{\mathbf{v}}_2\in\mathbb{R}^{\dimN_2}},\hspace{-0.5mm}\myydots,\hspace{-0.5mm}\underbrace{\hspace{-1mm}\sum\limits_{\indexi=1}^{\blocknum-1}\dimN_{\indexi}\tilde{w}_{\indexi,\blocknum},w_{\blocknum}\hspace{-0.5mm}+\hspace{-1mm}\sum\limits_{\indexi=1}^{\blocknum-1}\dimN_{\indexi}\tilde{w}_{\indexi,\blocknum},\hspace{-0.5mm}\myydots,\hspace{-0.5mm}(\dimN_{\blocknum}-1)w_{\blocknum}\hspace{-0.5mm}+\hspace{-1mm}\sum\limits_{\indexi=1}^{\blocknum-1}\dimN_{\indexi}\tilde{w}_{\indexi,\blocknum}}_{\Tilde{\mathbf{v}}_{\blocknum}\in\mathbb{R}^{\dimN_{\blocknum}}}]
\end{align*}
\end{minipage}}\vspace{3mm}\\
which concludes the proof that the vector $\tilde{\mathbf{v}}$ is piece-wise linear function in the following form
\begin{align*}
\small
\tilde{v}_\indexm\hspace{-0.7mm}=\hspace{-0.7mm}\tilde{f}(\indexm)\hspace{-0.7mm}=\hspace{-1mm}\begin{cases}(\indexm-\ell_1)w_1   &\hspace{-1.2mm}\mathit{if}\hspace{1.5mm}\ell_1\leq \indexm\leq u_1\\
(u_1-\ell_1+1)\tilde{w}_{1,2}\hspace{-0.7mm}+\hspace{-0.7mm}(\indexm-\ell_2)w_2 &\hspace{-1.2mm}\underdot{\mathit{if}}\hspace{1.5mm}\ell_2\leq \indexm\leq u_2
\\\vspace{2mm}
\sum\limits_{\indexi=1}^{\blocknum-1}(u_{\indexi}-\ell_{\indexi}+1)\tilde{w}_{\indexi,\blocknum}\hspace{-0.7mm}+\hspace{-0.7mm}(\indexm-\ell_{\blocknum})w_{\blocknum}&\hspace{-1.2mm}\mathit{if}\hspace{1.5mm}\ell_{\blocknum}\leq \indexm\leq u_{\blocknum}
\end{cases}
\end{align*}
where $\ell_1\hspace{-1mm}=\hspace{-1mm}1$, $u_1\hspace{-1mm}=\hspace{-1mm}\dimN_1$, $\ell_{\indexi}\hspace{-1mm}=\hspace{-1.52mm}\sum\limits_{\indexk=1}^{\indexi-1}\dimN_{\indexk}\hspace{-0.2mm}+\hspace{-0.35mm}1$ and ${u_{\indexi}\hspace{-1mm}=\hspace{-1.52mm}\sum\limits_{\indexk=1}^{\indexi}\dimN_{\indexk}}$ for $\indexi\hspace{-1mm}=\hspace{-1mm}2,\hspace{-0.3mm}\myydots,\hspace{-0.3mm}\blocknum$.
\end{proof}

\begin{figure*}[tbp!]
  \centering
  \captionsetup{justification=centering}
\subfloat[ $\tilde{G}=\{\tilde{V},\tilde{E},\tilde{\mathbf{W}}\}$ ]{\includegraphics[trim={0cm 0cm 0cm 0cm},clip,width=4.25cm]{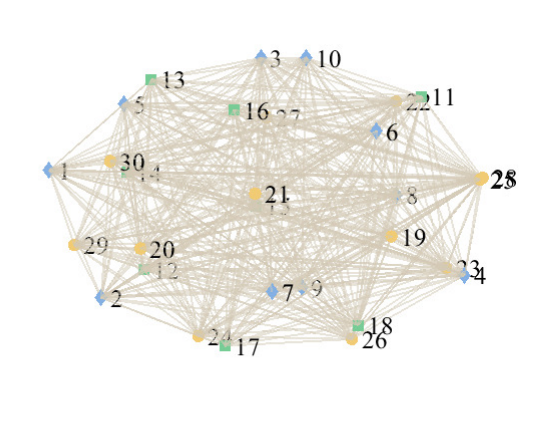}}
\subfloat[$\tilde{\mathbf{W}}\in\mathbb{R}^{\dimN\times\dimN}$]{\includegraphics[trim={0mm 0mm 0mm 0mm},clip,width=4.32cm]{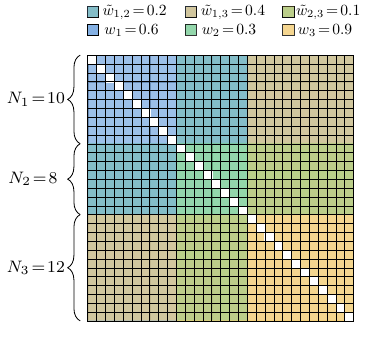}}
\subfloat[$\tilde{\mathbf{L}}\in\mathbb{R}^{\dimN\times\dimN}$]{\includegraphics[trim={0mm 0mm 0mm 0mm},clip,width=4.68cm]{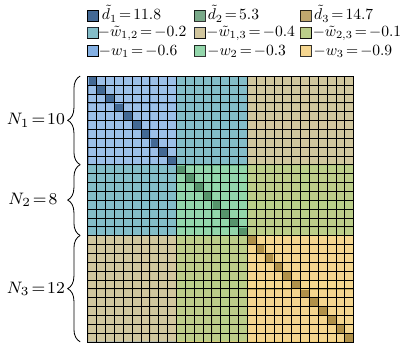}}
\subfloat[$\tilde{\mathbf{v}}\in\mathbb{R}^{\dimN}$]{\includegraphics[trim={0mm 0mm 0mm 0mm},clip,width=4.75cm]{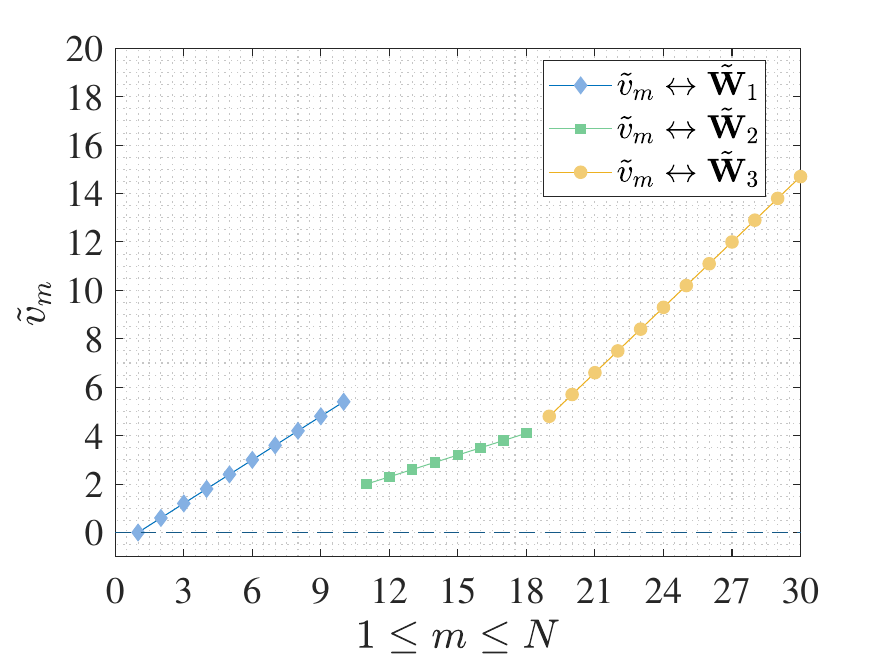}}
\caption{Examplary plot of Corollary~4.1 ($\mathbf{n}=[10,8,12]^{\top}\in\mathbb{R}^{\blocknum}$, $\dimN=30$, $\blocknum=3$).}
  \label{fig:ExamplaryPlotofCorollary82}
\end{figure*}

\newpage
\section{Appendix D: The Generalized Matrix Determinant Lemma}
Let $\mathbf{M}\in\mathbb{R}^{2\dimN \times 2\dimN}$ be a block matrix that can be shown as
\begin{align*}
    \begin{bmatrix}
    \mathbf{H} & \mathbf{U}\\
    \mathbf{-V^{\top}}& \mathbf{I}\\
    \end{bmatrix}
\end{align*}
where $\mathbf{I}\in\mathbb{R}^{\dimN\times\dimN}$ is the identity matrix and $\mathbf{H}, \mathbf{U}, \mathbf{V} \in\mathbb{R}^{\dimN \times \dimN}$. Using determinant properties of block matrices with non-commuting blocks \cite{detblockmatrices}, for $\mathbf{-V^{\top}}\mathbf{I}=\mathbf{I}(\mathbf{-V^{\top}})$ the determinant of $\mathbf{M}$ can be written as follows
\begin{align*}
    \mathrm{det}(\mathbf{M})= \mathrm{det}(\mathbf{H}\mathbf{I}-\mathbf{U}(-\mathbf{V}^{\top}))=\mathrm{det}(\mathbf{H}+\mathbf{U}\mathbf{V}^{\top}).
\end{align*}
Now, the next step is to simplify the determinant $\mathrm{det}(\mathbf{H}+\mathbf{U}\mathbf{V}^{\top})$ by computing a block diagonal matrix using Gaussian elimination. First, the entry under $\mathbf{H}$ is eliminated as follows
\begin{align*}
   \begin{bmatrix}
    \mathbf{I} & \mathbf{0}\\
    \mathbf{V^{\top}}\mathbf{H}^{\dagger}& \mathbf{I}\\
    \end{bmatrix}
    \begin{bmatrix}
    \mathbf{H} & \mathbf{U}\\
    \mathbf{-V^{\top}}& \mathbf{I}\\
    \end{bmatrix}=
    \begin{bmatrix}
    \mathbf{H} & \mathbf{U}\\
    \mathbf{0}& \mathbf{V^{\top}}\mathbf{H}^{\dagger}\mathbf{U}+\mathbf{I}\\
    \end{bmatrix}.
\end{align*}
Then, the entry above $\mathbf{I}$ is eliminated as 
\begin{align*}
    \begin{bmatrix}
    \mathbf{H} & \mathbf{U}\\
    \mathbf{-V^{\top}}& \mathbf{I}\\
    \end{bmatrix}
     \begin{bmatrix}
    \mathbf{I} & -\mathbf{H}^{\dagger}\mathbf{U}\\
    \mathbf{0}& \mathbf{I}\\
    \end{bmatrix}=
    \begin{bmatrix}
    \mathbf{H} & \mathbf{0}\\
    \mathbf{-V^{\top}}& \mathbf{V^{\top}}\mathbf{H}^{\dagger}\mathbf{U}+\mathbf{I}\\
    \end{bmatrix}.
\end{align*}
Combining these two operations leads to
\begin{align*}
     \begin{bmatrix}
    \mathbf{I} & \mathbf{0}\\
    \mathbf{V^{\top}}\mathbf{H}^{\dagger}& \mathbf{I}\\
    \end{bmatrix}
    \begin{bmatrix}
    \mathbf{H} & \mathbf{U}\\
    \mathbf{-V^{\top}}& \mathbf{I}\\
    \end{bmatrix}
     \begin{bmatrix}
    \mathbf{I} & -\mathbf{H}^{\dagger}\mathbf{U}\\
    \mathbf{0}& \mathbf{I}\\
    \end{bmatrix}=
    \begin{bmatrix}
    \mathbf{H} & \mathbf{0}\\
    \mathbf{0}& \mathbf{V^{\top}}\mathbf{H}^{\dagger}\mathbf{U}+\mathbf{I}\\
    \end{bmatrix}.
\end{align*}
Consequently, the determinant yields
\begin{align*}
\begin{split}
    \mathrm{det}\Bigg(
     \begin{bmatrix}
    \mathbf{I} & \mathbf{0}\\
    \mathbf{V^{\top}}\mathbf{H}^{\dagger}& \mathbf{I}\\
    \end{bmatrix}
    \begin{bmatrix}
    \mathbf{H} & \mathbf{U}\\
    \mathbf{-V^{\top}}& \mathbf{I}\\
    \end{bmatrix}
     \begin{bmatrix}
    \mathbf{I} & -\mathbf{H}^{\dagger}\mathbf{U}\\
    \mathbf{0}& \mathbf{I}\\
    \end{bmatrix}\Bigg)&=\mathrm{det}\Bigg(
    \begin{bmatrix}
    \mathbf{H} & \mathbf{0}\\
    \mathbf{0}& \mathbf{V^{\top}}\mathbf{H}^{\dagger}\mathbf{U}+\mathbf{I}\\
    \end{bmatrix}\Bigg)\\
    \mathrm{det}(\mathbf{H}+\mathbf{U}\mathbf{V}^{\top})&=\mathrm{det}(\mathbf{H})\mathrm{det}(\mathbf{I}+\mathbf{V^{\top}}\mathbf{H}^{\dagger}\mathbf{U})
    \end{split}.
\end{align*}

\newpage

\section{Appendix E: Real-world Data Examples}

\subsection{E.1~Type~I Outliers' Occurrence Analysis}
\begin{figure}[h!]\vspace{2mm}
\centering
\includegraphics[trim={0mm 0mm 0mm 0mm},clip,width=0.5\linewidth]{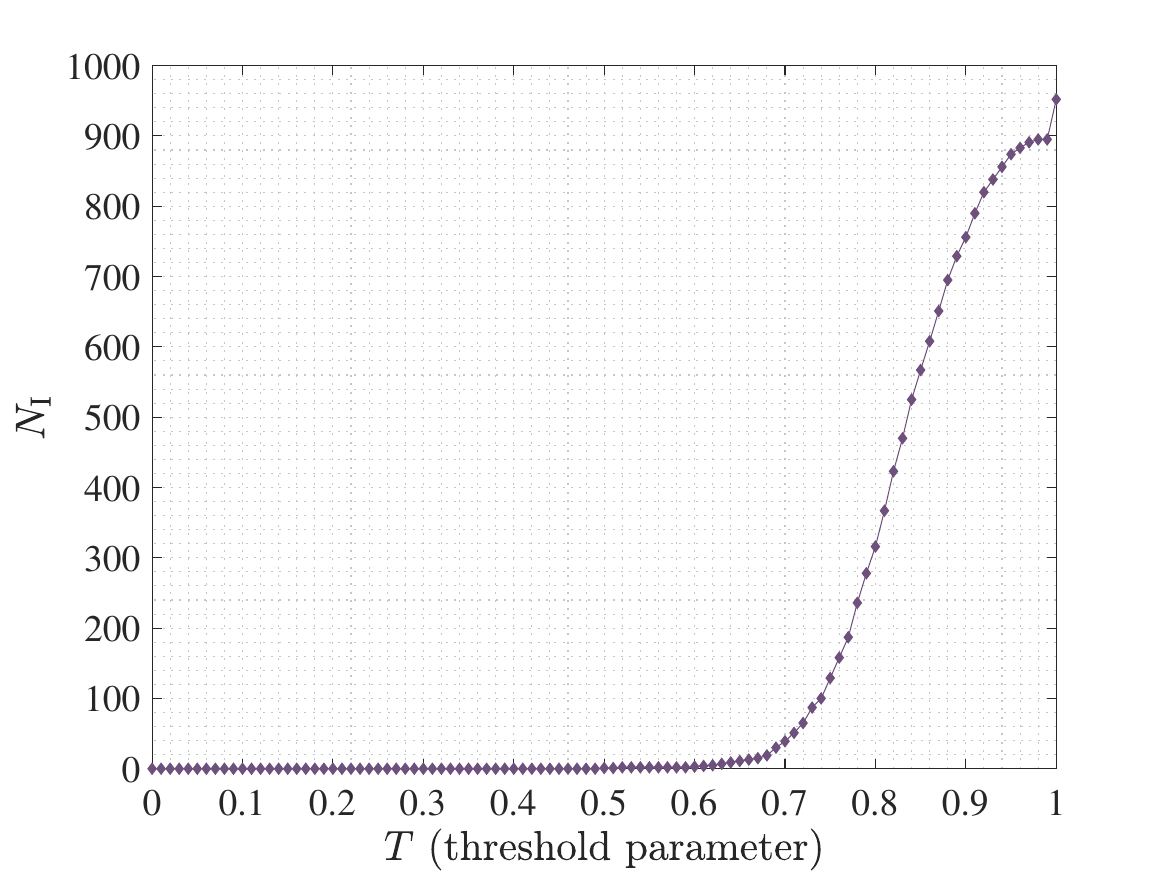}
\caption{Type~I outliers' occurrence in MNIST \cite{MNIST} data base for increasing sparsity. An initial affinity matrix is defined by $\mathbf{W}=\mathbf{X}^{\top}\mathbf{X}$ and the example graphs are obtained by removing the edges whose corresponding edge weight is smaller than the defined threshold value $T$.}
  \label{fig:OccurenceTypeIOutliers}\vspace{2mm}
\end{figure}

\begin{figure}[h!]
\centering
\includegraphics[trim={0mm 0mm 0mm 0mm},clip,width=0.5\linewidth]{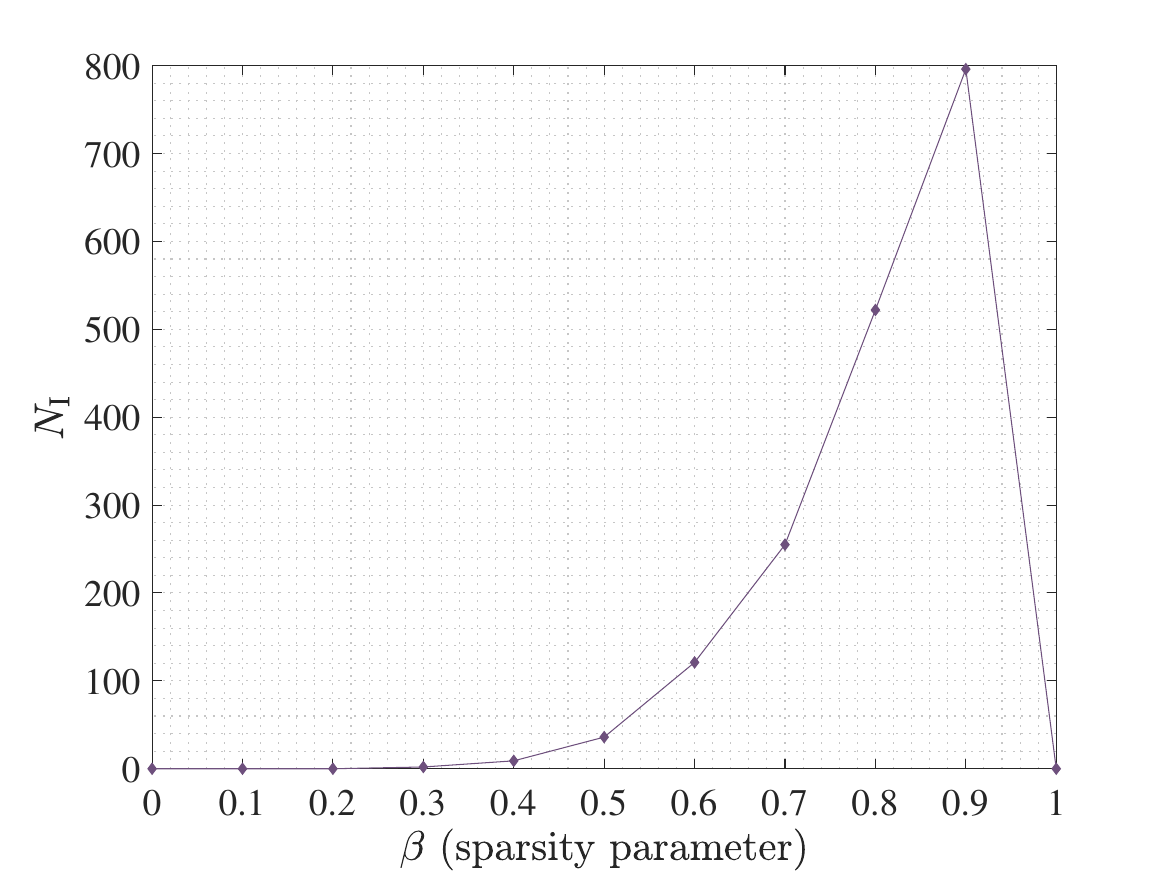}
\caption{Example of Type~I outliers' occurrence in the MNIST \cite{MNIST} data base for an increasing sparsity level. Sparse affinity matrices are computed based on elastic net similarity measure for an increasing value of the penalty parameter $\beta$.}
  \label{fig:OccurenceTypeIOutliersenet}
\end{figure}

\newpage

\subsection{E.2~Examplary Deviations from the Target Vector $\mathbf{v}$}
\begin{figure}[h!]
  \centering
\subfloat[MNIST]{\includegraphics[trim={0mm 0mm 0mm 0mm},clip,width=6cm]{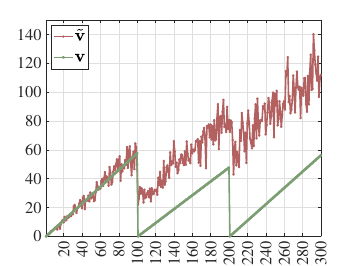}}
\subfloat[COIL20]{\includegraphics[trim={0mm 0mm 0mm 0mm},clip,width=6cm]{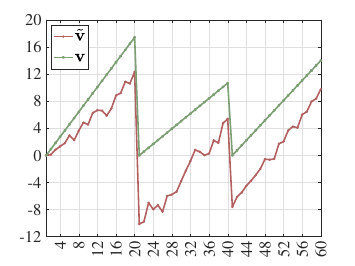}}
\subfloat[ORL]{\includegraphics[trim={0mm 0mm 0mm 0mm},clip,width=6cm]{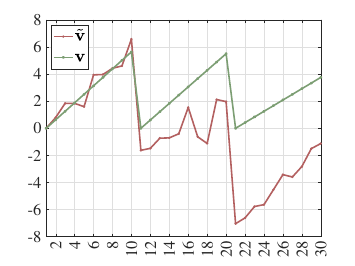}}\\ \vspace{3mm}
\subfloat[JAFFE]{\includegraphics[trim={0mm 0mm 0mm 0mm},clip,width=6cm]{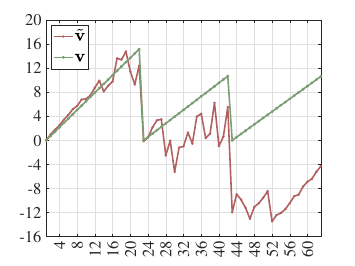}}
\subfloat[Ceramic]{\includegraphics[trim={0mm 0mm 0mm 0mm},clip,width=6cm]{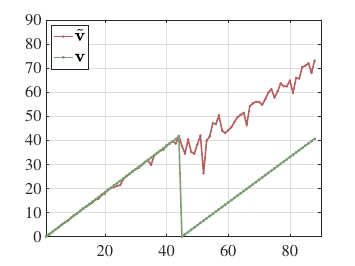}}
\subfloat[Iris]{\includegraphics[trim={0mm 0mm 0mm 0mm},clip,width=6cm]{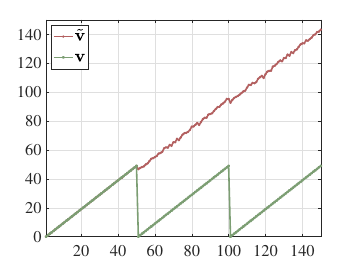}\label{fig:irisvectorvcomparison}}\\\vspace{3mm}
\subfloat[Person Identification]{\includegraphics[trim={0mm 0mm 0mm 0mm},clip,width=6cm]{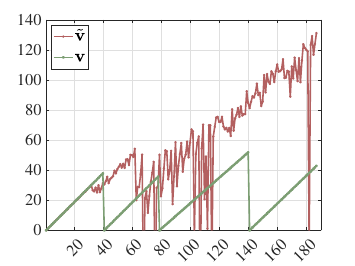}\label{fig:personidvectorvcomparison}}
\caption{Deviations from the ideal vector v based on the initial affinity matrix structure, i.e., $\mathbf{W}=\mathbf{X}^{\top}\mathbf{X}$ for different data sets. The observed outlier effects are consistent with the derived theory.}
\label{fig:vectorvcomparisonsdatasets}
\end{figure}

The deviations from an ``ideal'' piece-wise linear function is related to the structure of the affinity matrix, e.g. if it produces an internally dense cluster structure, and if it includes Type~I and Type~II outliers and/or group similarity. This means that the chosen similarity measure and the determined level of sparsity will have a direct impact on the deviations from an ``ideal'' piece-wise linear function. Therefore, it is difficult to make general statements and the following discussion is based on examples from our own practice.

\begin{figure}[h!]
  \centering
\subfloat[MNIST]{\includegraphics[trim={0mm 0mm 0mm 0mm},clip,width=6cm]{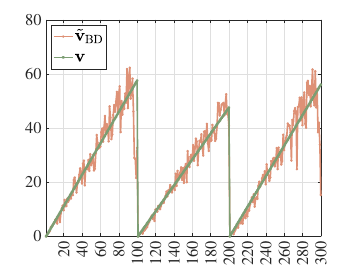}}
\subfloat[COIL20]{\includegraphics[trim={0mm 0mm 0mm 0mm},clip,width=6cm]{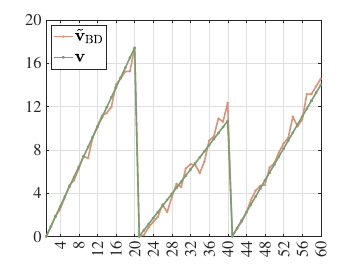}}
\subfloat[ORL]{\includegraphics[trim={0mm 0mm 0mm 0mm},clip,width=6cm]{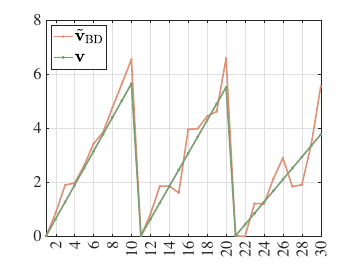}}\\ \vspace{3mm}
\subfloat[JAFFE]{\includegraphics[trim={0mm 0mm 0mm 0mm},clip,width=6cm]{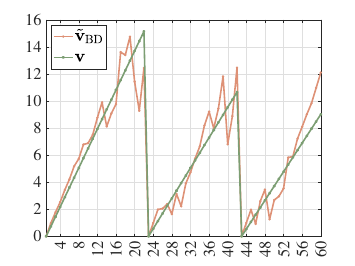}}
\subfloat[Ceramic]{\includegraphics[trim={0mm 0mm 0mm 0mm},clip,width=6cm]{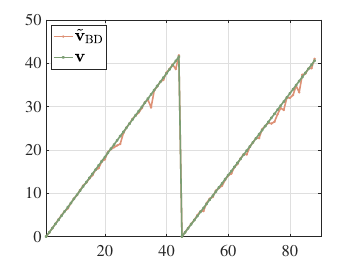}}
\subfloat[Iris]{\includegraphics[trim={0mm 0mm 0mm 0mm},clip,width=6cm]{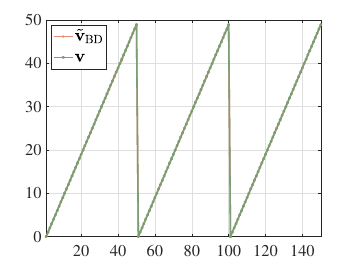}\label{fig:irisvectorvcomparison}}\\\vspace{3mm}
\subfloat[Person Identification]{\includegraphics[trim={0mm 0mm 0mm 0mm},clip,width=6cm]{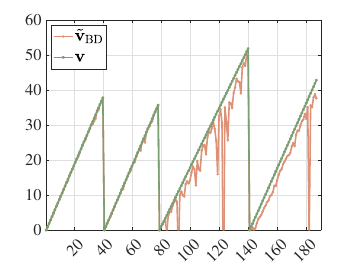}\label{fig:personidvectorvcomparison}}
\caption{Deviations from the ideal vector $\mathbf{v}$ based on block diagonally structured affinity matrices for different data sets. After removing all between-cluster edges, the remaining deviations are from noisy or heterogeneously connected within-cluster edges.}
\label{fig:vectorvcomparisonsdatasetsblock}
\end{figure}

To analyze the assumed affinity matrix structure in Corollary~4.1 in comparison to the corrupted initial affinity matrix structure in real-world data sets, for every examplary data set, we first compute the corrupted vector $\Tilde{\mathbf{v}}$ associated with the initial affinity matrix that is defined by $\mathbf{W}=\mathbf{X}^{\top}\mathbf{X}$. Similar to the theoretical analysis in Corollary~4.1, we consider deviations from the ``ideal'' piece-wise linear function $\mathbf{v}$ that is computed by assuming that the similarity coefficients within the blocks are concentrated around the mean value. The obtained corrupted $\Tilde{\mathbf{v}}$ and that of the ``ideal'' $\mathbf{v}$ structures are shown in Fig.~\ref{fig:vectorvcomparisonsdatasets} for different  real-world data sets. As can be seen, undesired similarity coefficients between different blocks result in shifts from the ``ideal'' piece-wise linear functions starting from the second linear pieces, consistent with our theory in Corollary~4.1. In particular, assumptions and findings of Corollary~4.1 hold in real-world data sets, especially, when the data sets include densely connected clusters of points, e.g. Ceramic \cite{Ceramic} and Iris \cite{Iris}.  Additionally, corrupted data sets, e.g. Person Identification \cite{PersonIdentification} whose corresponding affinity matrix is subject to Type~I outliers and group similarity results in large deviations from the ``ideal'' piece-wise linear function with group similarity shifts and small-valued $\Tilde{\mathbf{v}}$ components corresponding to Type~I outliers as it has been theoretically shown in Section~IV of the FRS-BDR paper. Even though highly corrupted data sets generate large deviations from the assumed models in real-world scenarios, an appropriate block diagonal representation suppresses these outlier effects by providing an optimal level of sparsity as it is exemplifed in Fig.~\ref{fig:vectorvcomparisonsdatasetsblock}.

\newpage
\begin{figure}[h!]
  \centering
\subfloat[MNIST]{\includegraphics[trim={0mm 0mm 0mm 0mm},clip,width=6cm]{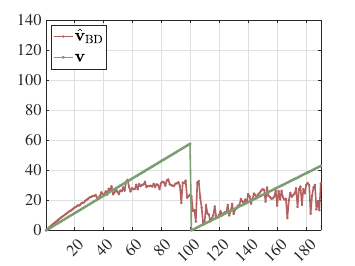}}
\subfloat[COIL20]{\includegraphics[trim={0mm 0mm 0mm 0mm},clip,width=6cm]{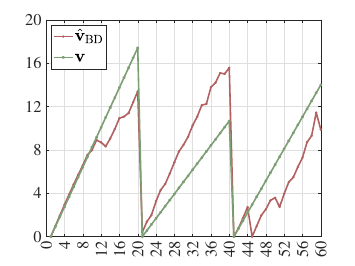}}
\subfloat[ORL]{\includegraphics[trim={0mm 0mm 0mm 0mm},clip,width=6cm]{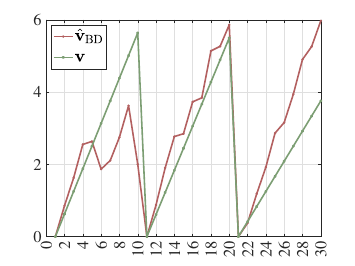}}\\
\subfloat[JAFFE]{\includegraphics[trim={0mm 0mm 0mm 0mm},clip,width=6cm]{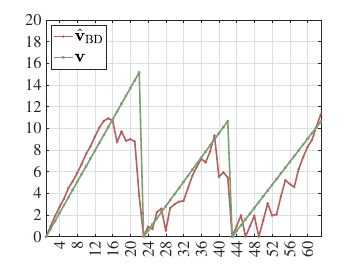}}
\subfloat[Ceramic]{\includegraphics[trim={0mm 0mm 0mm 0mm},clip,width=6cm]{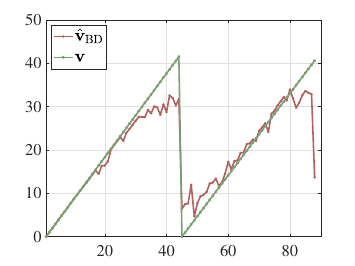}}
\subfloat[Iris]{\includegraphics[trim={0mm 0mm 0mm 0mm},clip,width=6cm]{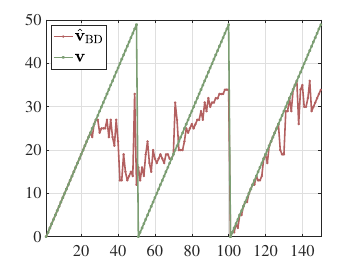}\label{fig:irisvectorvcomparison}}\\
\subfloat[Person Identification]{\includegraphics[trim={0mm 0mm 0mm 0mm},clip,width=6cm]{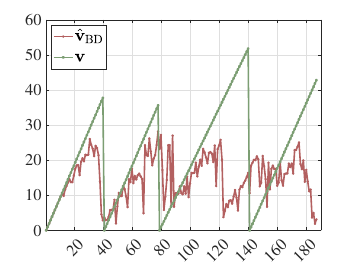}\label{fig:personidvectorvcomparison}}
\caption{Deviations from the ideal vector $\mathbf{v}$ based on the estimated block diagonally structured affinity matrix for different data sets.}
\label{fig:estimatedvectorvcomparisonsdatasets}
\end{figure}

A further set of analysis illustrating the deviations from ``ideal'' piece-wise linear functions, ``ideal'' $\mathbf{v}$ vectors and corresponding vector $\hat{\mathbf{v}}_{\mathrm{BD}}$ approximations, that represent the estimated FRS-BDR block diagonal affinity matrix structures, are provided in Fig.~\ref{fig:estimatedvectorvcomparisonsdatasets} for different data sets. As can be seen, heavily corrupted data sets, such as, Person Identification \cite{PersonIdentification} result in large deviations from the ``ideal'' piece-wise linear function, while densely connected more easily separable clusters of points such as in the Ceramic \cite{Ceramic} data set provide reasonably good approximations to the ``ideal'' structure. Although intra-cluster edges may result in large deviations from the determined ``ideal'' piece-wise linear functions in outlier-corrupted data sets, it is important to see that FRS-BDR algorithm removes the undesired edges between clusters in almost all cases, which is sufficient to obtain high clustering performance in these example. Another important point is that deviations from the ``ideal'' piece-wise linear function are considerably small for the FRS-BDR algorithm-based affinity matrix construction in comparison to the corrupted piece-wise linear function that has been shown in Fig.~\ref{fig:vectorvcomparisonsdatasets}.

\newpage
\subsection{E.3~Similarity Coefficients' Analysis}
\textcolor{white}{[tbp!]}\\\vspace{-3mm}
\\
\begin{figure}[!h]
  \centering
\subfloat[MNIST]{\includegraphics[trim={0mm 0mm 0mm 0mm},clip,width=7cm]{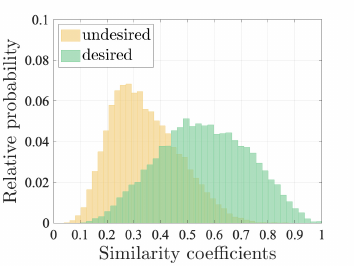}}\hspace{1cm}
\subfloat[USPS]{\includegraphics[trim={0mm 0mm 0mm 0mm},clip,width=7cm]{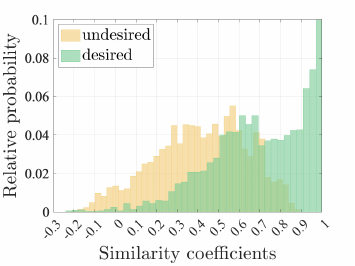}}\\ \vspace{3mm}
\subfloat[COIL20]{\includegraphics[trim={0mm 0mm 0mm 0mm},clip,width=7cm]{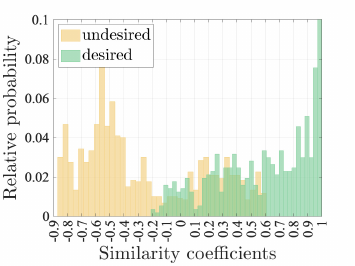}}\hspace{1cm}
\subfloat[ORL]{\includegraphics[trim={0mm 0mm 0mm 0mm},clip,width=7cm]{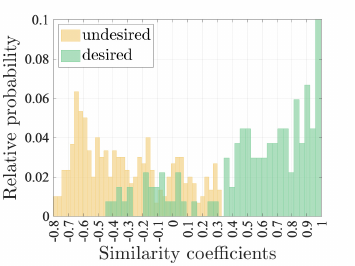}}\\ \vspace{3mm}
\subfloat[JAFFE]{\includegraphics[trim={0mm 0mm 0mm 0mm},clip,width=7cm]{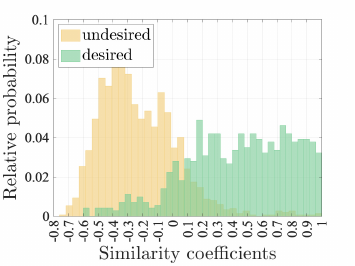}}\hspace{1cm}
\subfloat[YALE]{\includegraphics[trim={0mm 0mm 0mm 0mm},clip,width=7cm]{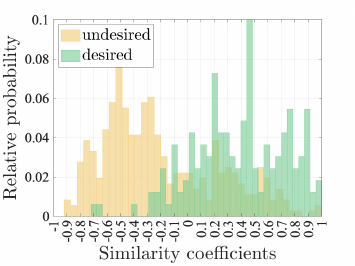}}
\caption{Relative probability of similarity coefficients for different data sets.}\vspace{2cm}
\end{figure}
\\
\newpage

\section{Appendix F: Additional Information for FRS-BDR}
\subsection{F.1~Visual Summary of FRS-BDR}

\begin{figure*}[h!]
  \centering
  \captionsetup{justification=centering}
\subfloat[Initialization]{\includegraphics[trim={0cm 0cm 0cm 0cm},clip,width=18cm]{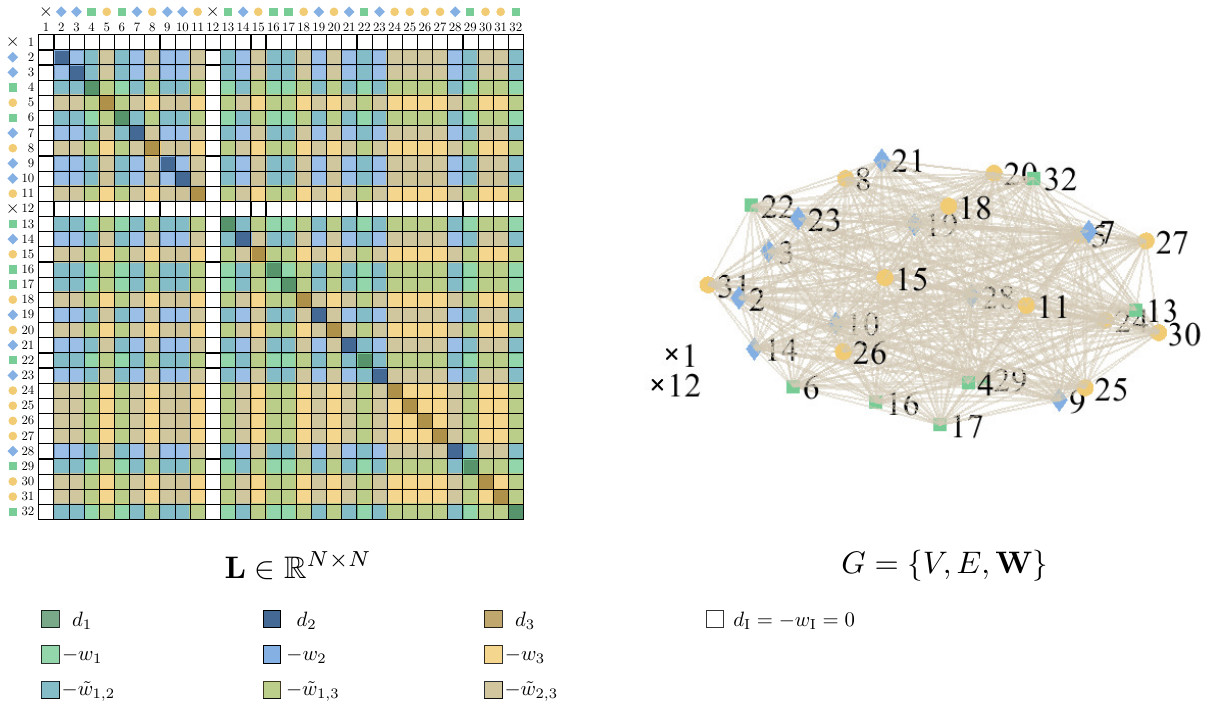}}\\
\subfloat[Step 1.1: Type~I Outlier Removal ]{\includegraphics[trim={0cm 0cm 0cm 0cm},clip,width=18cm]{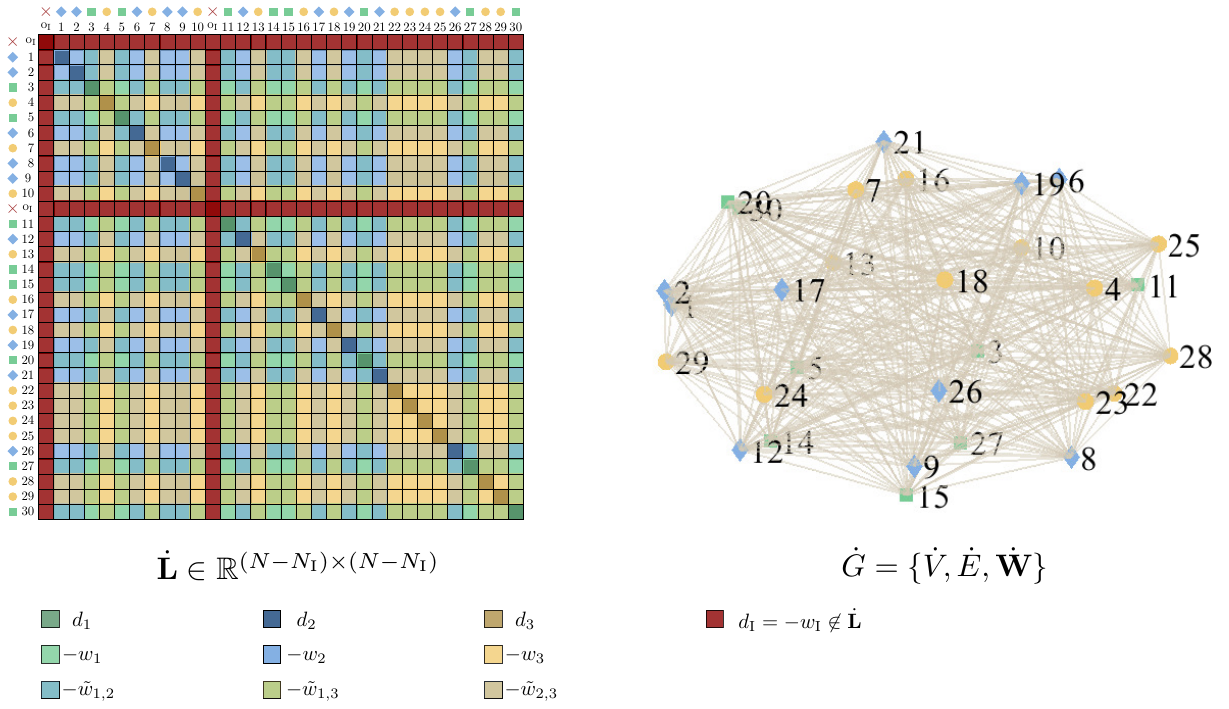}}
\vspace{-1cm}
\end{figure*}

\newpage
\begin{figure*}[h!]
\vspace{-3mm}
  \centering
  \captionsetup{justification=centering}
\setcounter{subfigure}{2}
\subfloat[Step 1.2: Similarity-based Block Diagonal Ordering (sBDO)]{\includegraphics[trim={0cm 0cm 0cm 0cm},clip,height=11.5cm]{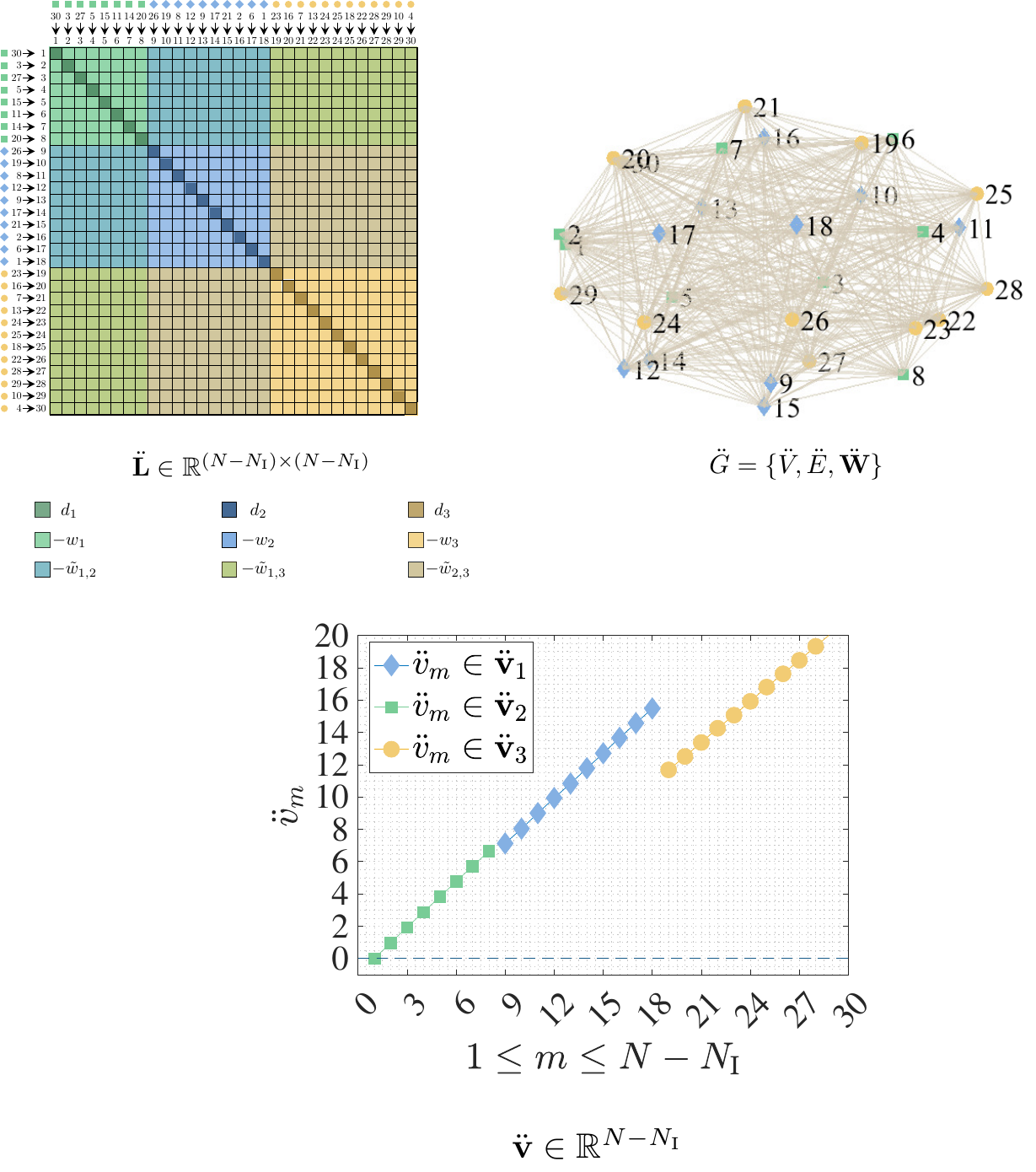}}\\
\subfloat[Step 1.3: Increase Sparsity for Excessive Group Similarity]{\includegraphics[trim={0cm 1.1cm 0cm 0cm},clip,height=11.5cm]{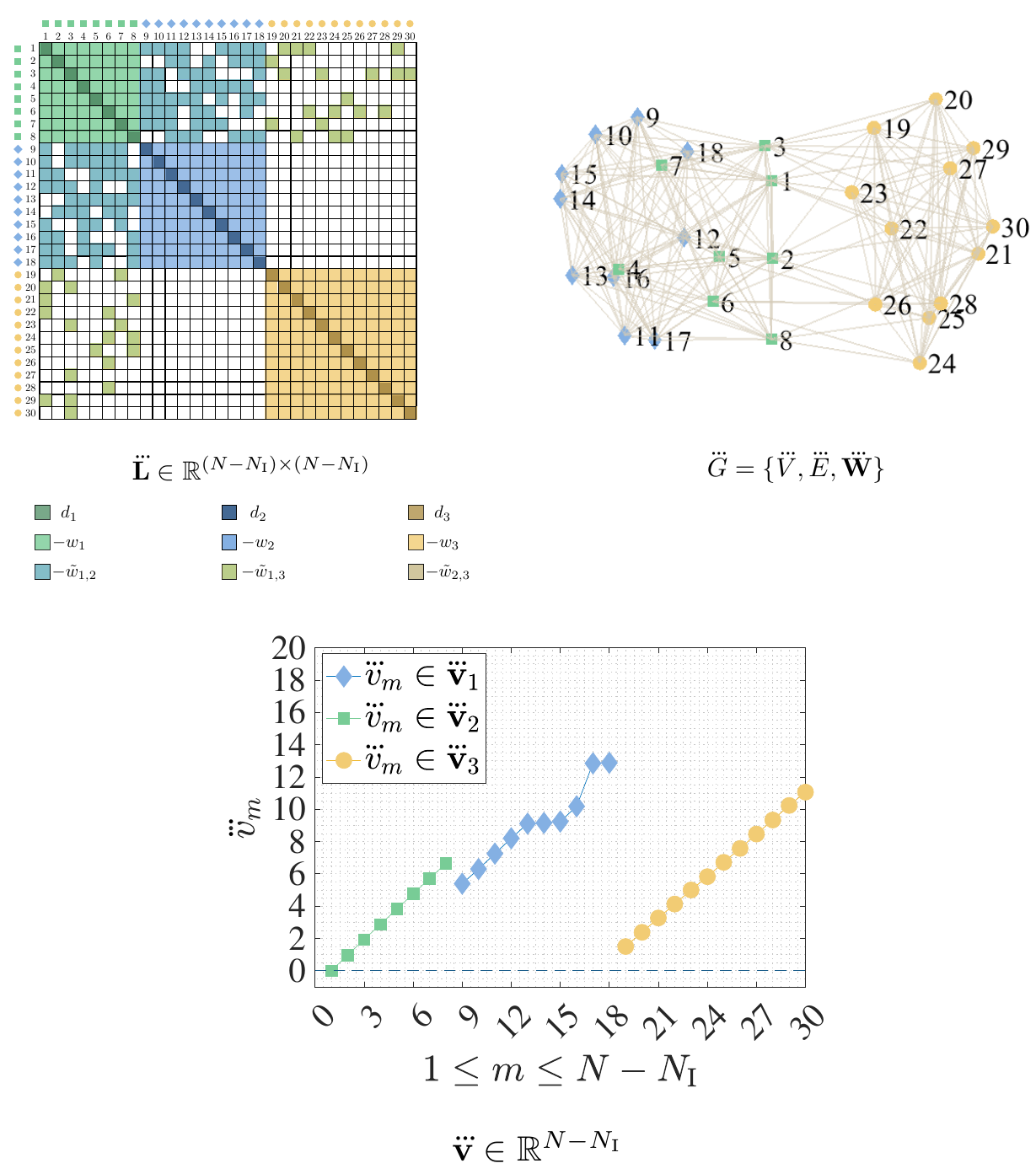}}
\vspace{-1cm}
\end{figure*}

\newpage
\begin{figure*}[h!]
\vspace{-3mm}
  \centering
  \captionsetup{justification=centering}
\setcounter{subfigure}{4}
\subfloat[Step 2.1: Compute Candidate Block Sizes]{\includegraphics[trim={0cm 0cm 0cm 0cm},clip,height=23cm]{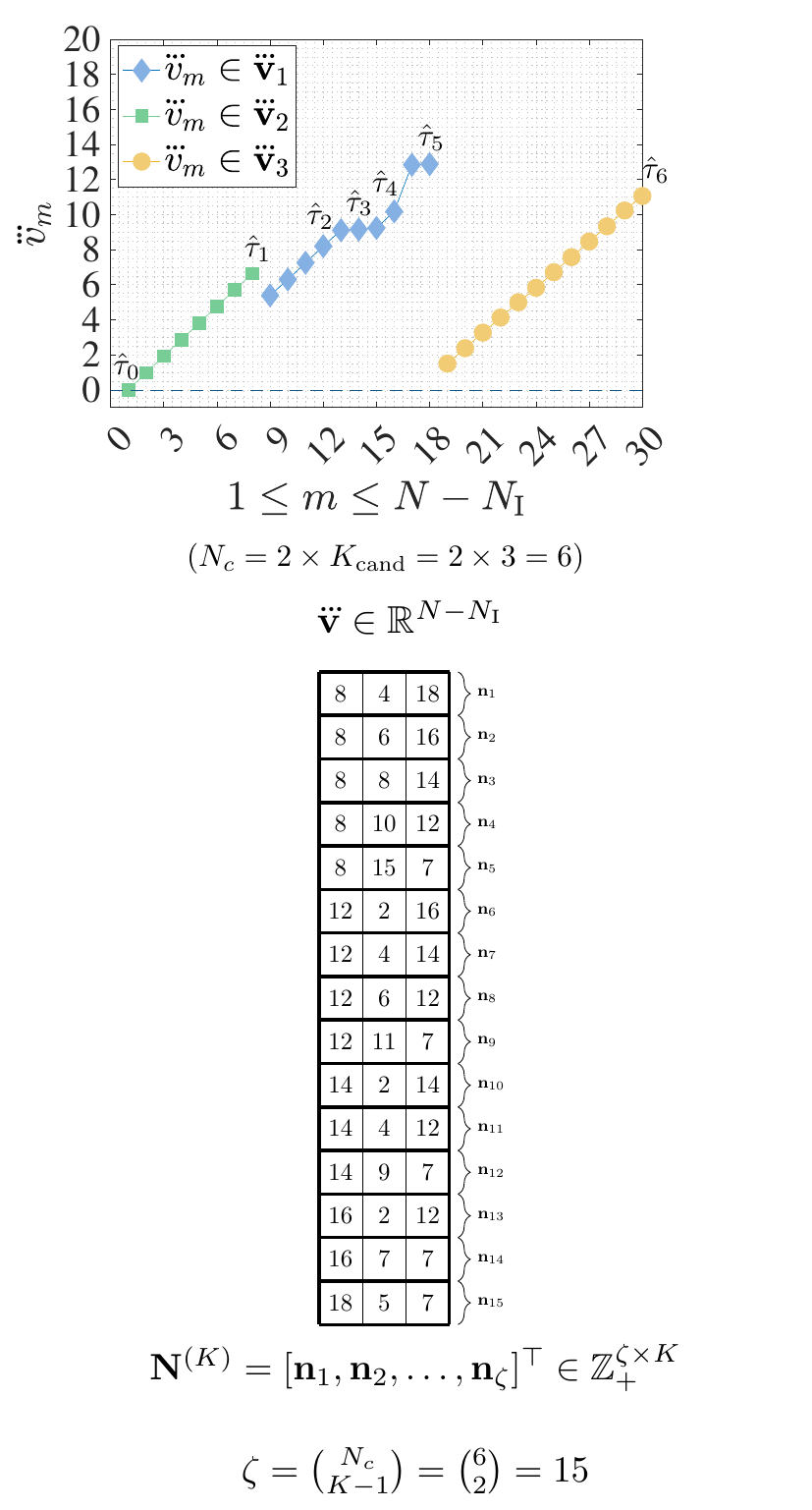}}\\
\vspace{-1cm}
\end{figure*}

\newpage
\begin{figure*}[h!]
\vspace{1cm}
  \centering
  \captionsetup{justification=centering}
\setcounter{subfigure}{5}
\subfloat[\vspace{5mm}Step 2.2.1:  Estimate Target Similarity Coefficients $\mathrm{diag}(\mathbf{W}_{\mathrm{sim}})$]{\hspace{-1.3cm}\includegraphics[trim={0cm 0cm 0cm 0cm},clip,height=18cm]{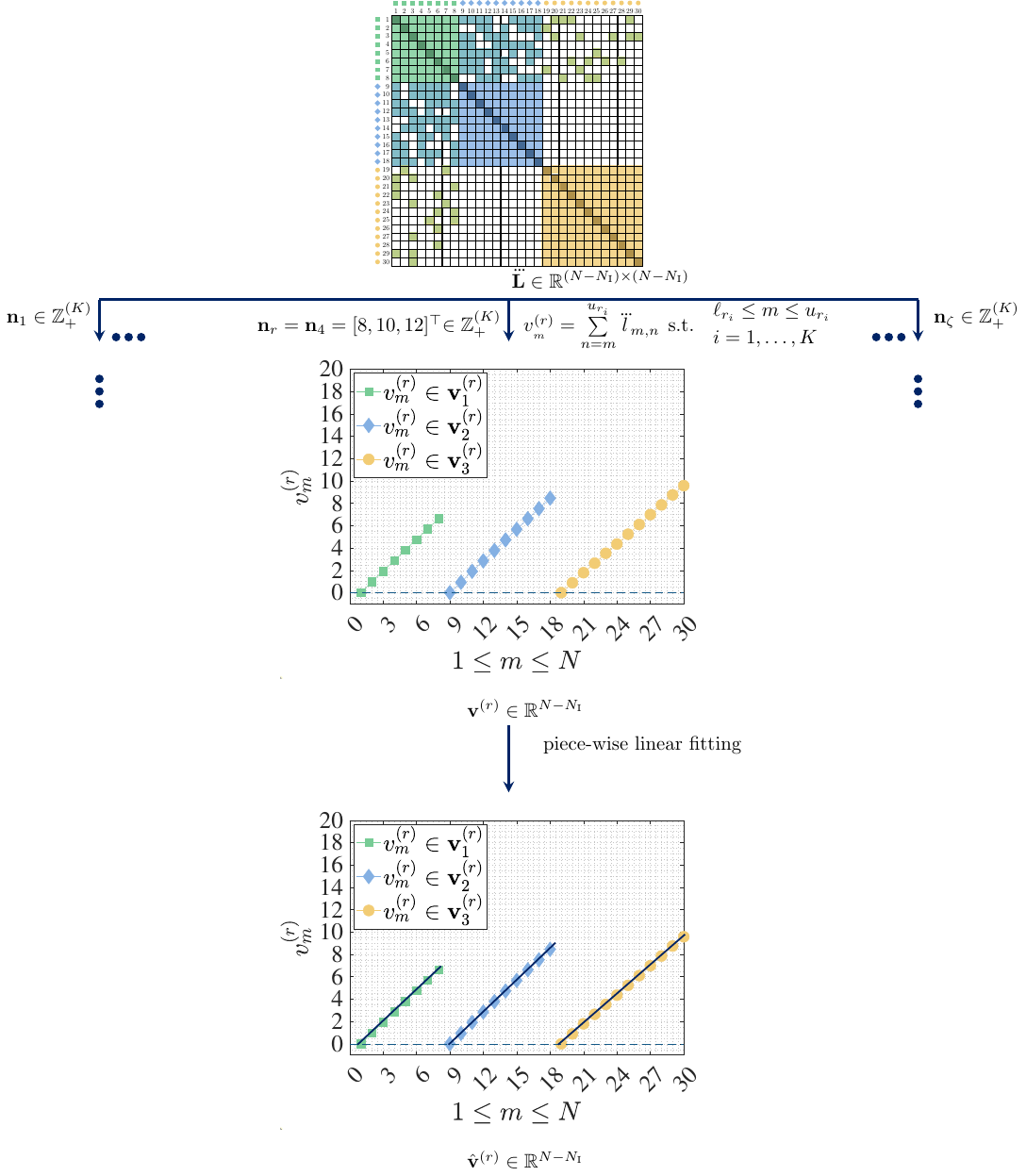}}
\end{figure*}

\newpage
\begin{figure*}[h!]
  \centering
  \captionsetup{justification=centering}
\setcounter{subfigure}{6}
\subfloat[Step 2.2.2: Estimate Undesired Similarity Coefficients ($\indexi=2,\indexj=1$)]{\includegraphics[trim={0cm 0cm 0cm 0cm},clip,height=24cm]{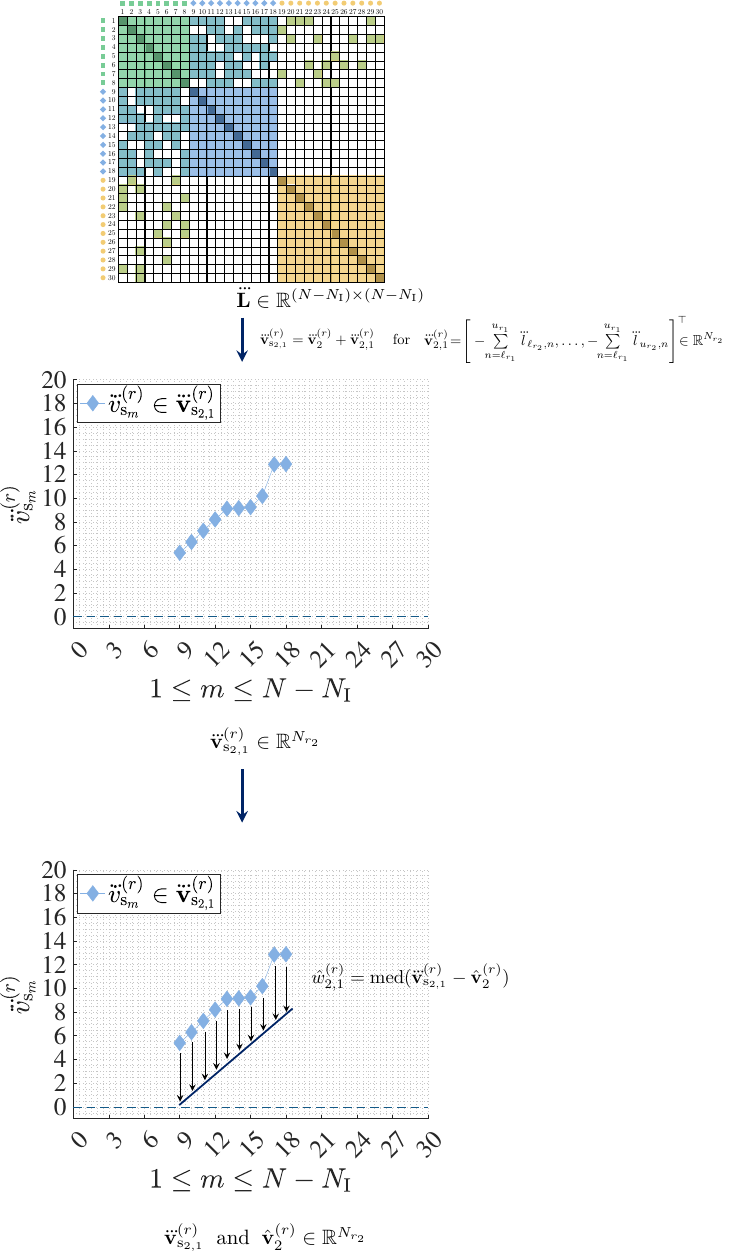}}
\vspace{-1cm}
\end{figure*}

\newpage
\begin{figure*}[h!]
  \centering
  \captionsetup{justification=centering}
\setcounter{subfigure}{7}
\subfloat[Step 2.2.2: Estimate Undesired Similarity Coefficients ($\indexi=3,\indexj=1$)]{\includegraphics[trim={0cm 0cm 0cm 0cm},clip,height=24cm]{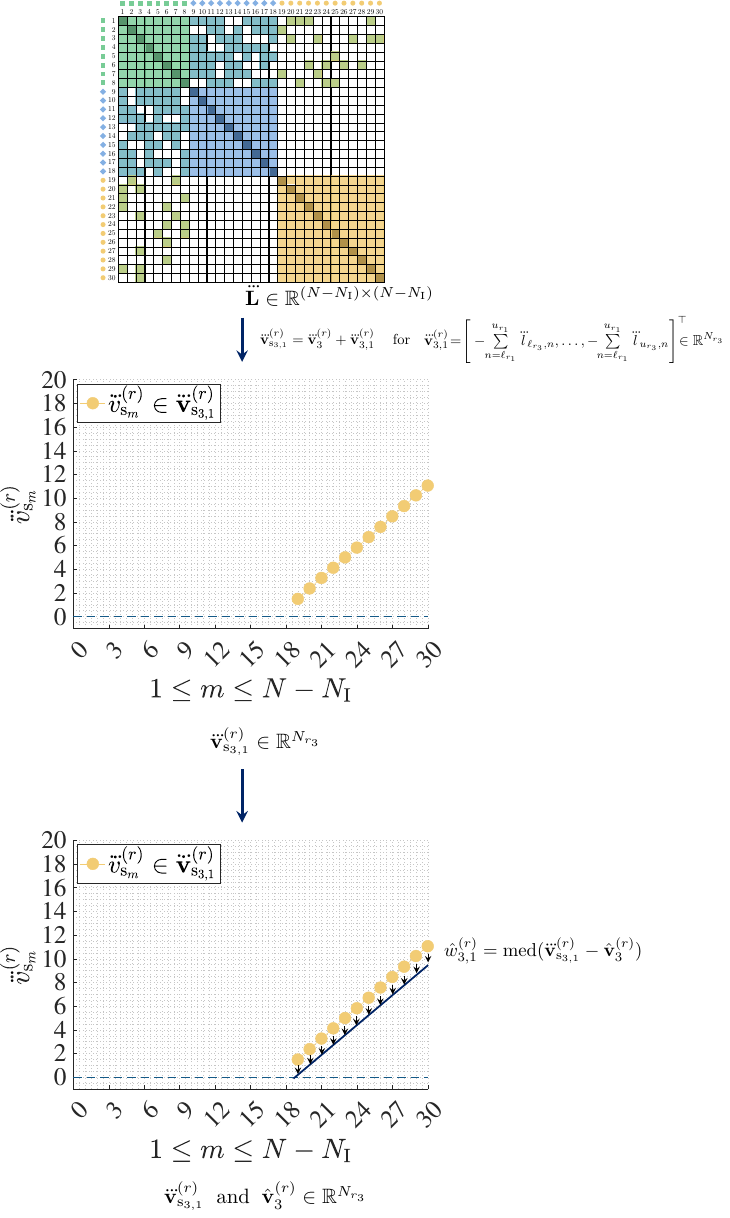}}
\vspace{-1cm}
\end{figure*}
\newpage
\begin{figure*}[h!]
  \centering
  \captionsetup{justification=centering}
\setcounter{subfigure}{8}
\subfloat[Step 2.2.2: Estimate Undesired Similarity Coefficients ($\indexi=3,\indexj=2$)]{\includegraphics[trim={0cm 0cm 0cm 0cm},clip,height=24cm]{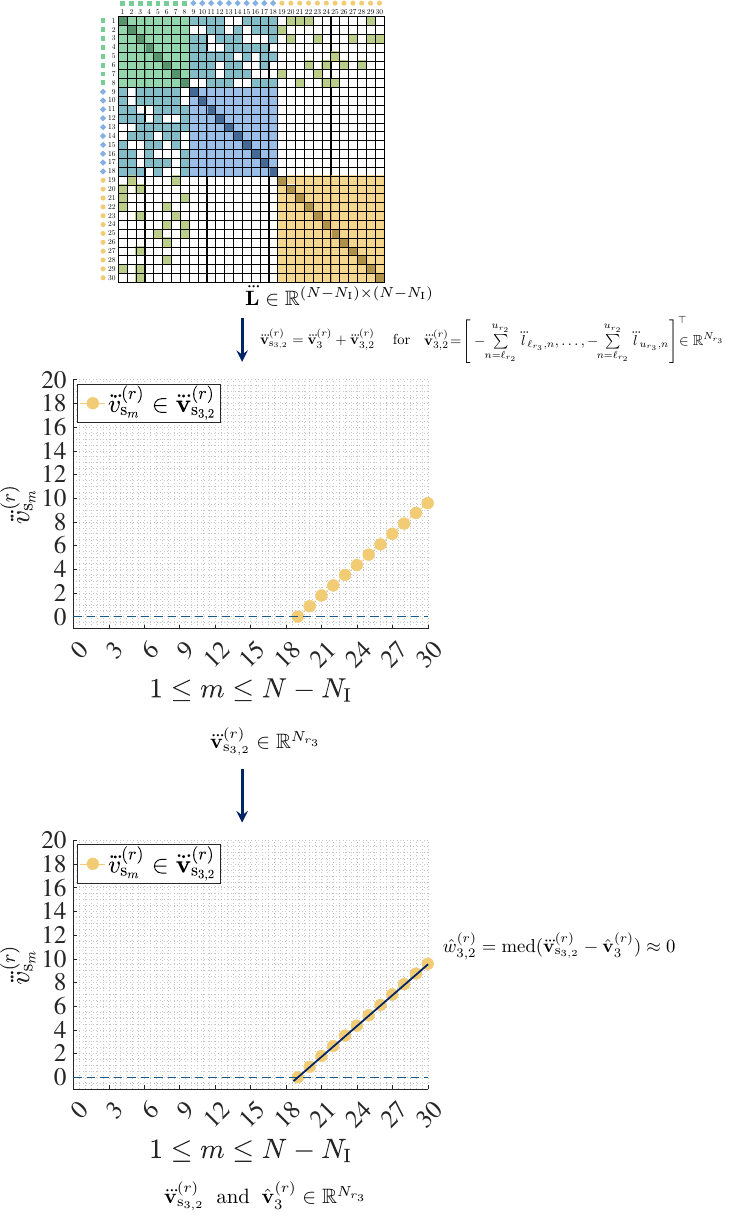}}
\vspace{-1cm}
\end{figure*}

\newpage
\begin{figure*}[h!]
  \centering\vspace{1cm}
  \captionsetup{justification=centering}
\setcounter{subfigure}{9}
\subfloat[Step 2.3: Estimate Vector $\dddot{\mathbf{v}}$]{\hspace{-1cm}\includegraphics[trim={0cm 0cm 0cm 0cm},clip,height=15.5cm]{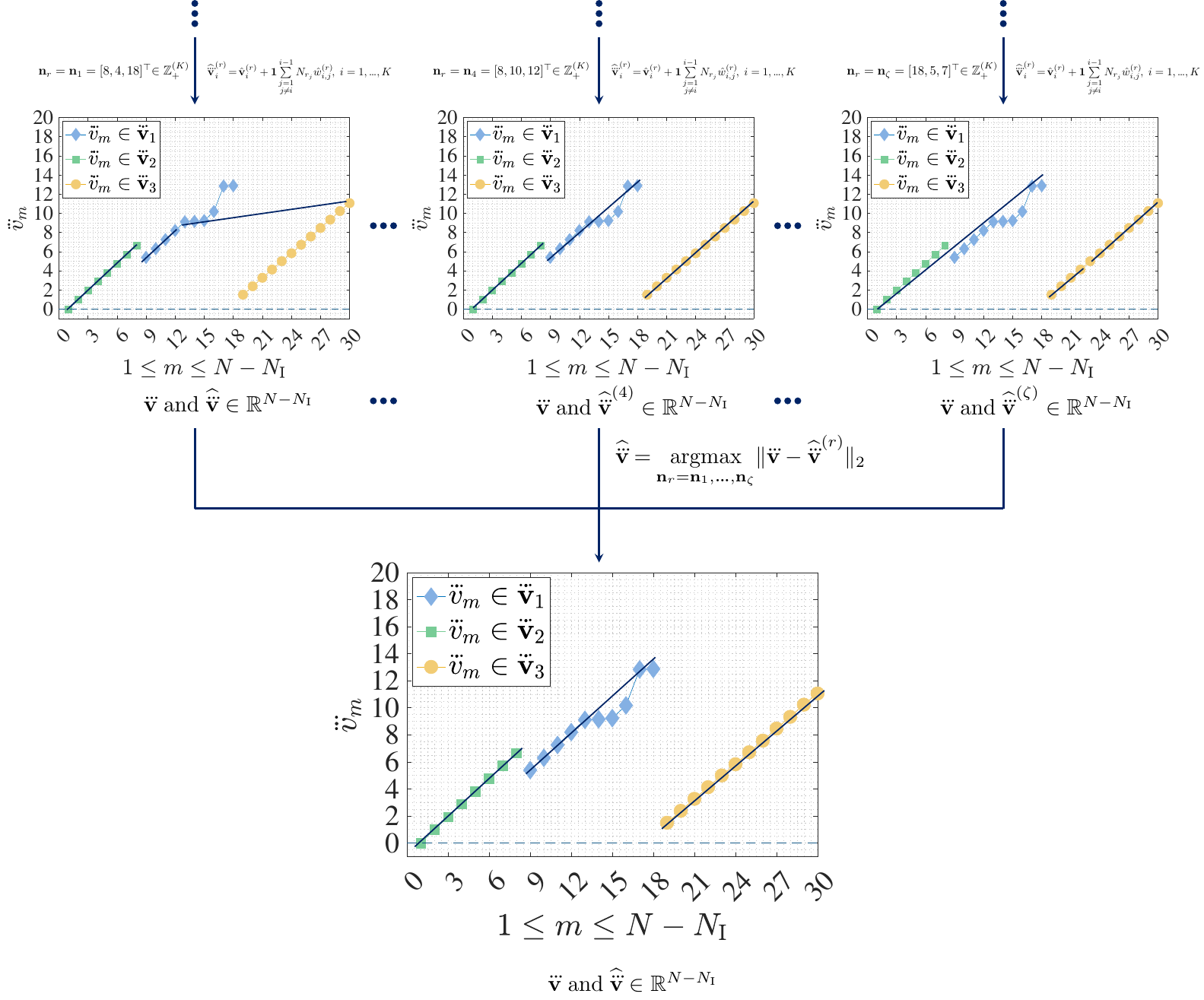}}
\caption{Visual summary of FRS-BDR}
\end{figure*}

\newpage
\subsection{F.2~Sparse Laplacian Matrix Analysis}
Step 1.3 of FRS-BDR determines a sparsity improved Laplacian matrix whose second smallest eigenvalue $\lambda_1$ closes to zero. Considering eigenvalues of Laplacian matrix, that is associated with target block zero-diagonal symmetric affinity matrix, $\lambda_1$ is definitely zero-valued. However, in real-world applications the distinct blocks may include negligibly small valued undesired similarity coefficients between different blocks. These coefficients result in an increase of $\lambda_1$ and affect definition of "close to zero". Therefore, this section provides set of experiments for determining a sparse Laplacian matrix.

\begin{figure*}[h!]\vspace{-5mm}
  \centering
  \captionsetup{justification=centering}
\includegraphics[trim={0cm 0cm 0cm 0cm},clip,width=16cm]{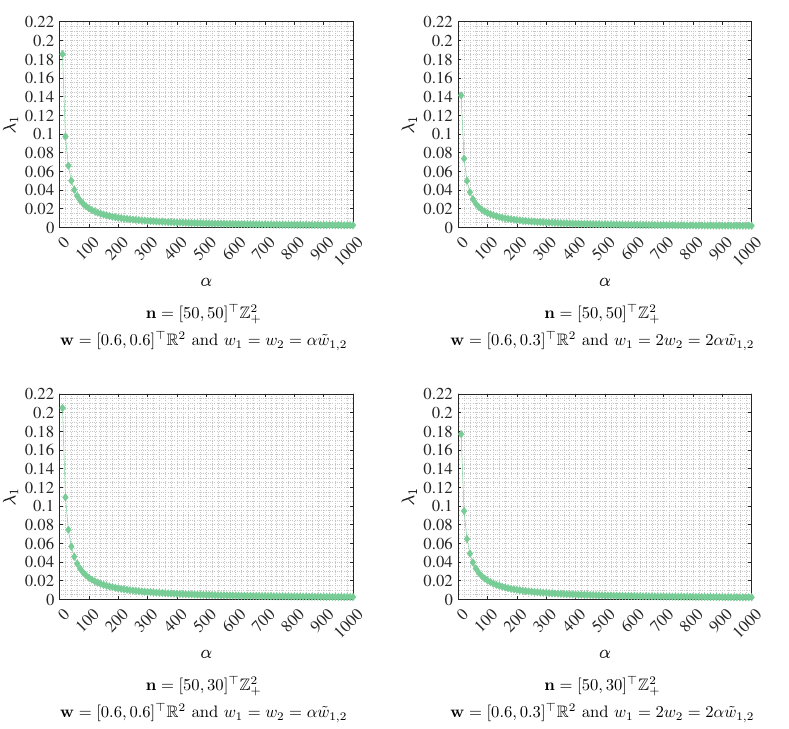}
\caption{$\lambda_1$ for increasing values of $\alpha$ associated with generalized eigen-decomposition}\label{fig:sparseLaplaciananalysisgen}\vspace{-5mm}
\end{figure*}

In Appendix~A.2 it has been shown that multiple group similarity results in additional increase in the vector of eigenvalues. Therefore, for simplicity, $\tilde{\mathbf{W}}\in\mathbb{R}^{\dimN\times\dimN}$ defining a $\blocknum=2$ block affinity matrix and associated Laplacian matrix $\tilde{\mathbf{L}}\in\mathbb{R}^{\dimN\times\dimN}$ are considered in the experiments. In $\tilde{\mathbf{W}}\in\mathbb{R}^{\dimN\times\dimN}$ each block $\mathbf{W}_{\indexi}, \indexi=1,2$ is associated to a number $\dimN_{\indexi}\in\mathbb{Z}_{+}>1$ of
feature vectors and concentrated around a similarity constant $w_{\indexi}\in\mathbb{R}_{+}, \indexi=1,2$ with negligibly small variations. Further, $\tilde{w}_{\indexi,\indexj}$ denotes a constant around which the similarity coefficients between blocks $\indexi$ and $\indexj$ are concentrated, i.e.
\begin{align*}
\Tilde{\mathbf{W}}=
\begin{bmatrix}
\begin{smallmatrix}
0& w_1&\myydots&w_1&\tilde{w}_{1,2}&\myydots & &\tilde{w}_{1,2}\\
w_1&0&\myydots&w_1&\tilde{w}_{1,2} & & & \tilde{w}_{1,2}\\
\vdots&\vdots&\ddots&&\vdots& & \ddots& \vdots\\
w_1 &w_1 &\myydots & 0 &\tilde{w}_{1,2} &\myydots & &\tilde{w}_{1,2}\\
\tilde{w}_{1,2}&\myydots & &\tilde{w}_{1,2} &0& w_2&\myydots&w_2\\
\tilde{w}_{1,2} & & & \tilde{w}_{1,2} &w_2&0&\myydots&w_2\\
\vdots& & \ddots& \vdots &\vdots&\vdots&\ddots&\vdots\\
\tilde{w}_{1,2} &\myydots & &\tilde{w}_{1,2} &w_2 &w_2 &\myydots & 0 \\
\end{smallmatrix}
\end{bmatrix}\hspace{3mm}\mathrm{and}\hspace{3mm}
\Tilde{\mathbf{L}}=
\begin{bmatrix}
\begin{smallmatrix}
\Tilde{d}_1& -w_1&\myydots&-w_1&-\tilde{w}_{1,2}&\myydots & &-\tilde{w}_{1,2}\\
-w_1&\Tilde{d}_1&\myydots&-w_1&-\tilde{w}_{1,2} & & & -\tilde{w}_{1,2}\\
\vdots&\vdots&\ddots&&\vdots& & \ddots& \vdots\\
-w_1 &-w_1 &\myydots & \Tilde{d}_1 &-\tilde{w}_{1,2} &\myydots & &-\tilde{w}_{1,2}\\
-\tilde{w}_{1,2}&\myydots & &-\tilde{w}_{1,2} &\Tilde{d}_2& -w_2&\myydots&-w_2\\
-\tilde{w}_{1,2} & & & -\tilde{w}_{1,2}&-w_2&\Tilde{d}_2&\myydots&-w_2\\
\vdots& & \ddots& \vdots &\vdots&\vdots&\ddots&\vdots\\
-\tilde{w}_{1,2} &\myydots & &-\tilde{w}_{1,2} &-w_2 &-w_2 &\myydots & \Tilde{d}_2\\
\end{smallmatrix}
\end{bmatrix},\vspace{3mm}
\end{align*}
where $\Tilde{d}_1=(\dimN_1-1)w_1+\dimN_2\tilde{w}_{1,2}$,  $\Tilde{d}_2=(\dimN_2-1)w_2+\dimN_1\tilde{w}_{1,2}$ and $w_{\indexi}>\tilde{w}_{1,2}, \indexi=1,2$. According to the generalized eigen-decomposition, the second smallest eigenvalue $\lambda_1$ of $\Tilde{\mathbf{L}}$ is
\begin{align*}
    \lambda_1=\frac{\Tilde{d}_1\dimN_{1}\tilde{w}_{1,2}+\Tilde{d}_2\dimN_{2}\tilde{w}_{1,2}}{\Tilde{d}_1\Tilde{d}_2}.
\end{align*}
Based on this, for $\tilde{w}_{1,2}\rightarrow 1$ and $w_{\indexi}>\tilde{w}_{1,2}, \indexi=1,2$, $\lambda_1$ reaches its maximum value
\begin{align*}
    \lambda_{1_\mathrm{max}}=\frac{\dimN_1+\dimN_2}{\dimN_1+\dimN_2-1}
\end{align*}
which tends to $1$ for $\dimN_1+\dimN_2\gg1$. Even though the maximum value of $\lambda_1$ is explicit, its minimum value depends on different variables $\dimN_{\indexi}$, $w_{\indexi}$ and $\tilde{w}_{i,j}$ for $\indexi=1,2$, $\indexj=1,2$ and $\indexi\neq\indexj$. Therefore, $\lambda_1$ is analyzed as a function of $\alpha$ for different block size values in Fig.~\ref{fig:sparseLaplaciananalysisgen}. Here, $\alpha$ denotes the ratio between smallest target similarity coefficient $w_{\mathrm{min}}$ and that of undesired $\tilde{w}_{i,j}$ for $\indexi=1,2$, $\indexj=1,2$ and $\indexi\neq\indexj$. As can be seen, the second smallest eigenvalue $\lambda_1$ decreases to zero for $w_1\gg\tilde{w}_{1,2}$ and $w_2\gg\tilde{w}_{1,2}$.

\begin{figure*}[h!]
  \centering
  \captionsetup{justification=centering}
\includegraphics[trim={0cm 0cm 0cm 0cm},clip,width=16cm]{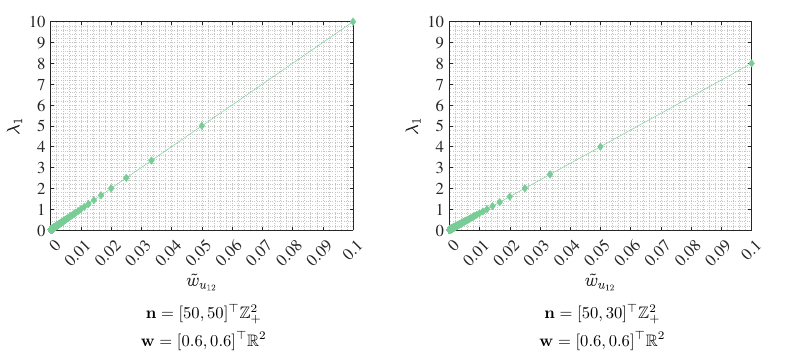}
\caption{$\lambda_1$ for increasing values of $\tilde{w}_{1,2}$ associated with standard eigen-decomposition}\label{fig:sparseLaplaciananalysisstand}
\end{figure*}

In contrast to generalized eigen-decomposition, the second smallest eigenvalue $\lambda_1$ of $\Tilde{\mathbf{L}}$ associated to the standard eigen-decomposition does not affected by target similarity coefficients, i.e.
\begin{align*}
    \lambda_1=\tilde{w}_{1,2}(\dimN_1+\dimN_2).
\end{align*}
As $\dimN_{\indexi}>1, \indexi=1,2$, $\lambda_1$ can be reduced to zero for considerably small-valued undesired similarity coefficients. Therefore, $\lambda_1$ is analyzed as a function of $\tilde{w}_{1,2}$ for different block size values in Fig.~\ref{fig:sparseLaplaciananalysisstand}. The figure implies that the value of undesired similarity coefficients is directly linked to closeness to zero. Therefore, the closeness to zero can be determined according to total sample size and desired value of undesired similarity.

To summarize, the eigenvalues based on generalized eigen-decomposition are less sensitive to block sizes which makes definition of close easier. As a result, based on the assumption that target similarity coefficients considerably larger valued than that of undesired coefficients, a selected $\lambda_1$ value smaller than $0.05$ might be sufficient to obtain sparse Laplacian matrices. In the proposed default setting, a Laplacian matrix is assumed to be sparse if its second smallest eigenvalue is valued by $0\leq\lambda_1<10^{-3}$.

\newpage

\newpage
\subsection{F.3~Additional Algorithms}
In this section, the design of sparse Laplacian matrix $\dddot{\mathbf{L}}$ is detailed using different methods. In both algorithms, if the method does not provide any sparse Laplacian matrix $\dddot{\mathbf{L}}$, the sparsest nonnegative definite Laplacian matrix $\dddot{\mathbf{L}}_{\mathrm{alt}}$ is selected. The example algorithms have the same operational principle which can be further adapted to different graph construction methods.\\

\begin{figure}[htb]\vspace{-7mm}
\scalebox{0.9}{
\begin{minipage}{\linewidth}
\begin{algorithm}[H]
\setstretch{1}
\DontPrintSemicolon
\KwIn{a non-sparse affinity matrix $\ddot{\mathbf{W}}\in \mathbb{R}^{(\dimN-\dimN_{\mathrm{I}})\times(\dimN-\dimN_{\mathrm{I}})}$, initial threshold $T_{\mathrm{ini}}$ (optional, default is $T_{\mathrm{ini}}=0.5$), increasement constant $T_{\mathrm{inc}}$ (optional, default is $T_{\mathrm{inc}}=10^{-3}$)}\vspace{-0.5mm}
Set $T=T_{\mathrm{ini}}$\\
\While{$T<1$}{\SetInd{0.5em}{0.25em}
Compute affinity matrix $\ddot{\mathbf{W}}^{(T)}$ which is equal to $\ddot{\mathbf{W}}$ except that the similarity coefficients smaller than $T$ in\\ $\ddot{\mathbf{W}}^{(T)}$ are zero, i.e.
$\forall \ddot{w}^{(T)}_{\indexm\indexn}<T \in \ddot{\mathbf{W}}^{(T)}, \ddot{w}^{(T)}_{\indexm\indexn}=0$ where $\indexm=1,\myydots,\dimN-\dimN_{\mathrm{I}}$, $\indexn=1,\myydots,\dimN-\dimN_{\mathrm{I}}$ and $\indexm\neq \indexn$\\
Based on obtained $\ddot{\mathbf{W}}^{(T)}$, compute $\ddot{\mathbf{D}}^{(T)}$ and $\ddot{\mathbf{L}}^{(T)}$ of dimension $\mathbb{R}^{(\dimN-\dimN_{\mathrm{I}})\times(\dimN-\dimN_{\mathrm{I}})}$\\
Compute $\ddot{\boldsymbol{\lambda}}^{(T)}\hspace{-0.8mm}=\hspace{-0.7mm}[\ddot{\lambda}^{(T)}_0\hspace{-0.3mm},\ddot{\lambda}^{(T)}_1\hspace{-0.3mm},\myydots\hspace{-0.3mm},\ddot{\lambda}^{(T)}_{\dimN\hspace{-0.1mm}-\hspace{-0.1mm}\dimN_{\mathrm{I}}\hspace{-0.1mm}-\hspace{-0.1mm}1}]\hspace{-0.8mm}\in\hspace{-0.7mm}\mathbb{R}^{\dimN\hspace{-0.3mm}-\hspace{-0.3mm}\dimN_{\mathrm{I}}}$ in ascending order using Eq.~(1) or Eq.~(2)$\hspace{-4cm}$\\
\begin{algorithmic}
\IF {$\ddot{\lambda}^{(T)}_{1}\cong0$ (For detailed analysis, see Appendix~F.2.) and $\forall\ddot{\lambda}^{(T)}_{\indexm}\in\ddot{\boldsymbol{\lambda}}^{(T)}, \ddot{\lambda}^{(T)}_{\indexm}\geq 0$ for $m=1,\myydots,\dimN-\dimN_{\mathrm{I}}-1$} 
  \STATE {$\dddot{\mathbf{W}}=\ddot{\mathbf{W}}^{(T)}$, $\dddot{\mathbf{D}}=\ddot{\mathbf{D}}^{(T)}$ and $\dddot{\mathbf{L}}=\ddot{\mathbf{L}}^{(T)}$}
  \STATE {$\mathbf{break}$;}
\ELSIF {$\ddot{\lambda}^{(T)}_{1}\neq0$ and $\forall\ddot{\lambda}^{(T)}_{\indexm}\in\ddot{\boldsymbol{\lambda}}^{(T)}, \ddot{\lambda}^{(T)}_{\indexm}\geq 0$ for $m=1,\myydots,\dimN-\dimN_{\mathrm{I}}-1$}
  \STATE {$\dddot{\mathbf{W}}_{\mathrm{alt}}=\ddot{\mathbf{W}}^{(T)}$, $\dddot{\mathbf{D}}_{\mathrm{alt}}=\ddot{\mathbf{D}}^{(T)}$ and $\dddot{\mathbf{L}}_{\mathrm{alt}}=\ddot{\mathbf{L}}^{(T)}$}
  \STATE {$T\leftarrow T+T_{\mathrm{inc}}$}
\ELSE\vspace{-0.5mm}
\STATE {$T\leftarrow T+T_{\mathrm{inc}}$}
\ENDIF\vspace{-0.5mm}
\end{algorithmic}}
\textbf{if} $\dddot{\mathbf{W}}$ does not exist \textbf{then}\\
\begin{algorithmic}
  \IF {$\dddot{\mathbf{W}}_{\mathrm{alt}}$ exists}
  \STATE {$\dddot{\mathbf{W}}=\dddot{\mathbf{W}}_{\mathrm{alt}}$}
  \ELSE\vspace{-0.7mm}
  \STATE{\textbf{error :} Please start with a smaller threshold}
  \ENDIF\vspace{-0.7mm}
\end{algorithmic}
\textbf{end if}\vspace{-0.5mm}
\caption{Sparse Laplacian Matrix Design using Adaptive Thresholding}
\label{alg:SparseLaplacianDesignAdapThr}
\KwOut{Estimated sparse matrices $\dddot{\mathbf{W}}$, $\dddot{\mathbf{D}}$ and $\dddot{\mathbf{L}}$}
\end{algorithm}
\end{minipage}}\vspace{-5mm}
\end{figure}\vspace{-2mm}
\begin{figure}[htb]
\scalebox{0.9}{
\begin{minipage}{\linewidth}
\begin{algorithm}[H]
\setstretch{1}
\DontPrintSemicolon
\KwIn{a non-sparse affinity matrix $\ddot{\mathbf{W}}\in \mathbb{R}^{(\dimN-\dimN_{\mathrm{I}})\times(\dimN-\dimN_{\mathrm{I}})}$, initial number of neighbors value $p_{\mathrm{ini}}$ (optional, default is $p_{\mathrm{ini}}=\dimN-2$), decreasement constant $p_{\mathrm{dec}}$ (optional, default is $p_{\mathrm{dec}}=1$), minimum number of nodes in per block $N_{\mathrm{min}}$ (optional, default is $N_{\mathrm{min}}\approx\frac{N}{\blocknum_{\mathrm{max}}}\in\mathbb{Z}_{+}$)
}\vspace{-0.5mm}
Set $p=p_{\mathrm{ini}}$\\
\While{$p>N_{\mathrm{min}}$}{\SetInd{0.5em}{0.25em}\vspace{-0.5mm}
Construct affinity matrix $\ddot{\mathbf{W}}^{(p)}$ using $p$ nearest neighbors as in\cite{RLPI}\\
Based on obtained $\ddot{\mathbf{W}}^{(p)}$, compute $\ddot{\mathbf{D}}^{(p)}$ and $\ddot{\mathbf{L}}^{(p)}$ of dimension $\mathbb{R}^{(\dimN-\dimN_{\mathrm{I}})\times(\dimN-\dimN_{\mathrm{I}})}$\\
Compute $\ddot{\boldsymbol{\lambda}}^{(p)}\hspace{-0.7mm}=\hspace{-0.7mm}[\ddot{\lambda}^{(p)}_0,\ddot{\lambda}^{(p)}_1,\myydots,\ddot{\lambda}^{(p)}_{\dimN\hspace{-0.1mm}-\hspace{-0.1mm}\dimN_{\mathrm{I}}\hspace{-0.1mm}-\hspace{-0.1mm}1}]\hspace{-0.7mm}\in\hspace{-0.7mm}\mathbb{R}^{\dimN\hspace{-0.1mm}-\hspace{-0.1mm}\dimN_{\mathrm{I}}}$ in ascending order using Eq.~(1) or Eq.~(2) $\hspace{-3cm}$\\
\begin{algorithmic}
\IF {$\ddot{\lambda}^{(p)}_{1}\cong0$ (For detailed analysis, see Appendix~F.2.) and $\forall\ddot{\lambda}^{(p)}_{\indexm}\in\ddot{\boldsymbol{\lambda}}^{(p)}, \ddot{\lambda}^{(p)}_{\indexm}\geq 0$ for $m=1,\myydots,\dimN-\dimN_{\mathrm{I}}-1$} 
  \STATE {$\dddot{\mathbf{W}}=\ddot{\mathbf{W}}^{(p)}$, $\dddot{\mathbf{D}}=\ddot{\mathbf{D}}^{(p)}$ and $\dddot{\mathbf{L}}=\ddot{\mathbf{L}}^{(p)}$}\vspace{-0.5mm}
  \STATE {$\mathbf{break}$;}
\ELSIF {$\ddot{\lambda}^{(p)}_{1}\neq0$ and
$\forall\ddot{\lambda}^{(p)}_{\indexm}\in\ddot{\boldsymbol{\lambda}}^{(p)}, \ddot{\lambda}^{(p)}_{\indexm}\geq 0$ for $m=1,\myydots,\dimN-\dimN_{\mathrm{I}}-1$}
  \STATE {$\dddot{\mathbf{W}}_{\mathrm{alt}}=\ddot{\mathbf{W}}^{(p)}$, $\dddot{\mathbf{D}}_{\mathrm{alt}}=\ddot{\mathbf{D}}^{(p)}$ and $\dddot{\mathbf{L}}_{\mathrm{alt}}=\ddot{\mathbf{L}}^{(p)}$}
  \STATE {$p\leftarrow p-p_{\mathrm{dec}}$}
\ELSE\vspace{-0.5mm}
\STATE {$p\leftarrow p-p_{\mathrm{dec}}$}
\ENDIF\vspace{-0.5mm}
\end{algorithmic}\vspace{-0.7mm}}
\textbf{if} $\dddot{\mathbf{W}}$ does not exist \textbf{then}\\\vspace{-0.5mm}
\begin{algorithmic}
  \IF {$\dddot{\mathbf{W}}_{\mathrm{alt}}$ exists}
  \STATE {$\dddot{\mathbf{W}}=\dddot{\mathbf{W}}_{\mathrm{alt}}$}
  \ELSE\vspace{-0.7mm}
  \STATE{\textbf{error :} Please start with a greater $p_{\mathrm{ini}}$}
  \ENDIF\vspace{-0.7mm}
\end{algorithmic}
\textbf{end if}\vspace{-0.5mm}
\caption{Sparse Laplacian Matrix Design using $p$-nearest Graph}
\label{alg:SparseLaplacianDesignEpsilonNearest}
\KwOut{Estimated sparse matrices $\dddot{\mathbf{W}}$, $\dddot{\mathbf{D}}$ and $\dddot{\mathbf{L}}$}
\end{algorithm}
\end{minipage}}
\end{figure}
\newpage
\subsection{F.4~Additional Experimental Results}
\subsubsection{F.4.1~Handwritten Digit Clustering}
\setlength{\parindent}{0pt}\paragraph{F.4.1.1~MNIST Data Set}
\textcolor{white}{[tbp!]}\\\vspace{-3mm}
\\
\begin{table*}[h!]\vspace{-2mm}
\centering
\resizebox{\linewidth}{!}{%
\begin{tabular}{p{4cm}M{1.5cm}M{1.5cm}M{1.5cm}M{1.5cm}M{1.5cm}M{1.5cm}M{1.5cm}M{1.5cm}M{1.5cm}M{1.5cm}M{1.5cm}M{1.5cm}M{1.5cm}M{1.8cm}}
\hline\hline
\\[-3mm]
 & \multicolumn{14}{c}{Subspace Clustering Performance for Different Block Diagonal Representation Methods}\\\\[-3mm]
\cline{2-15}\\[-3mm]
& &\multicolumn{11}{c}{Minimum-Maximum Clustering Accuracy $(c_\mathrm{accmin}-c_\mathrm{accmax})$ for Different Regularization Parameters} &  &\\\\[-3mm]
\cline{3-13}\\[-3mm]
MNIST Data Set & $\mathbf{W}$ & SSC & BD-SSC & LRR & BD-LRR & LSR & BDR-B & BDR-Z & RKLRR & IBDLR & FRPCAG & RSC & EBDR & \textbf{FRS-BDR} \\
\midrule
2 subjects & 87.1 & 50.5-82.6 & 50.5-90.3 & 51.0-89.2 & 51.1-89.9 & 51.5-86.8 & 50.9-91.8 & 51.0-89.8 & 50.5-91.6 & 50.5-92.2 &91.2-95.1 & 53.7-60.1 &85.5 & 89.7\\
3 subjects & 72.0 & 33.9-37.5 & 33.8-76.1 & 34.2-74.8 & 34.3-72.5 & 34.9-71.8 & 34.4-67.6 & 34.4-67.7 & 33.9-70.0 & 33.9-79.0 &62.3-79.7 &39.7-45.1 & 68.6 & 79.6\\
5 subjects & 60.9 & 20.6-25.1 & 20.4-63.7 & 21.0-62.4 & 21.1-62.5 & 20.6-60.8 & 23.0-48.6 & 20.5-54.4 & 20.6-59.2 & 20.6-65.4 &52.5-61.7 &28.8-31.7 & 52.3 & 67.5\\
8 subjects&53.4& 13.1-18.0 & 12.9-56.0 & 13.7-50.6 & 13.7-52.4 & 13.8-53.7 & 13.9-37.6 & 13.3-46.0 & 13.1-50.1 & 13.1-57.7 &50.0-56.5 &20.6-23.2 & 42.3 & 59.3\\
10 subjects&51.2&10.7-16.9 & 10.4-52.7 & 11.2-45.6 & 11.2-50.8 & 11.6-50.1 & 10.9-33.5 & 10.8-45.4 & 10.6-44.1 & 10.5-53.4 &48.5-57.2 &16.8-24.2 &38.9 & 57.6\\
\hline\\[-3mm]
Average & 64.9 & 25.8-36.0 & 25.6-67.8 & 26.2-64.5 & 26.3-65.6 & 26.5-64.6 & 26.6-55.8 & 26.0-60.7 & 25.7-63.0 & 25.7-69.5 &  60.9-70.0&31.9-36.8 & 57.5 & 70.7\\
\hline\hline
\end{tabular}}
\caption{Subspace clustering performance of different block diagonal representation approaches on MNIST data set. The results are summarized for the similarity measure $\mathbf{W}=\mathbf{X}^{\top}\mathbf{X}$.}
\end{table*}
\begin{table*}[h!]
\centering
\resizebox{9cm}{!}{%
\begin{tabular}{p{4cm}M{1.5cm}M{1.5cm}M{1.5cm}M{1.5cm}M{1.5cm}}
\hline\hline
\\[-3mm]
 & \multicolumn{5}{c}{Detailed Computation time $(t)$ for FRS-BDR Method}\\\\[-3mm]
\cline{2-6}\\[-3mm]
MNIST Data Set & Step 1.1 & Step 1.2 & Step 1.3 & Step 2\\
\midrule
2 subjects &  0.007 & 0.064  & 0.194 & 0.004  \\
3 subjects &  0.011 & 0.171  &  0.602& 0.007 \\
5 subjects &  0.033 & 1.106  & 3.262 & 0.018 \\
8 subjects &  0.096 &  5.045 & 15.797 & 0.018 \\
10 subjects &  0.164 & 10.053 &35.965 & 0.017 \\
\hline\hline
\end{tabular}}
\caption{Computation time performance of FRS-BDR method on MNIST data set. The results are summarized for the similarity measure $\mathbf{W}=\mathbf{X}^{\top}\mathbf{X}$.}
\end{table*}

\begin{table*}[h!]
\centering
\resizebox{\linewidth}{!}{%
\begin{tabular}{p{4cm}M{1.5cm}M{1.5cm}M{1.5cm}M{1.5cm}M{1.5cm}M{1.5cm}M{1.5cm}M{1.5cm}M{1.5cm}M{1.5cm}M{1.5cm}M{1.5cm}M{1.5cm}M{1.8cm}}
\hline\hline
\\[-3mm]
 & \multicolumn{14}{c}{Computation time $(t)$ for Different Block Diagonal Representation Methods}\\\\[-3mm]
\cline{2-15}\\[-3mm]
& &\multicolumn{11}{c}{Computation Time $(t)$ for Optimally Tuned Regularization Parameters} &  &\\\\[-3mm]
\cline{3-13}\\[-3mm]
MNIST Data Set & $\mathbf{W}$ & SSC & BD-SSC & LRR & BD-LRR & LSR & BDR-B & BDR-Z & RKLRR & IBDLR & FRPCAG & RSC &EBDR &\textbf{FRS-BDR} \\
\midrule
2 subjects & 0.002 & 0.971 & 1.081 & 3.093 & 3.224 & 0.004 & 0.624 & 0.959 & 4.774 & 13.743 &21.963&0.168& 0.009 & 0.075 \\
3 subjects & 0.003 & 0.666 & 1.681 & 5.168 & 5.517 & 0.007 & 1.580 & 2.166 & 11.782 & 1.565 &42.297&0.858& 0.017 & 0.189 \\
5 subjects & 0.006 & 1.574 & 4.315 & 14.679 & 14.725 & 0.016 & 7.783 & 7.776 & 244.397 & 8.169 &149.955& 14.508& 0.043  &1.157\\
8 subjects & 0.013 & 3.766 & 10.936 & 26.520 & 29.069 & 0.040 & 25.090 & 24.991 & 572.827 & 21.823 &209.952&46.374& 0.116 & 5.159 \\
10 subjects & 0.018 & 6.068 & 16.493 & 34.692 & 64.749 & 0.063 & 44.883 & 55.796 & 748.296 & 35.881 &247.014& 70.863& 0.208 & 10.235 \\
\hline\\[-3mm]
Average  & 0.008 & 2.609 & 6.901 & 16.830 & 23.457 & 0.026 &  15.992 & 18.338 & 316.415 & 16.236 &134.236&26.554& 0.079 & 3.363 \\
\hline\hline
\end{tabular}}
\caption{Computation time performance of different block diagonal representation approaches on MNIST data set. The results are summarized for the similarity measure $\mathbf{W}=\mathbf{X}^{\top}\mathbf{X}$ and sparsity assumed to be known for all sparse representation methods which means that computation time of FRS-BDR is detailed for Steps 1.1, 1.2 and 2.}
\end{table*}

\begin{table*}[h!]
\centering
\resizebox{\linewidth}{!}{%
\begin{tabular}{p{4cm}M{2cm}M{2cm}M{2cm}M{2cm}M{2cm}M{2cm}M{2cm}M{2cm}M{2cm}M{2cm}M{2cm}M{2cm}}
\hline\hline
\\[-3mm]
 & \multicolumn{12}{c}{Modularity Performance for Different Block Diagonal Representation Methods}\\\\[-3mm]
\cline{2-13}\\[-3mm]
& &\multicolumn{9}{c}{Modularity Performance corresponding to the Maximum Clustering Accuracy of Different Block Diagonal Representation Methods} &  &\\\\[-3mm]
\cline{3-11}\\[-3mm]
MNIST Data Set & $\mathbf{W}$ & SSC & BD-SSC & LRR & BD-LRR & LSR & BDR-B & BDR-Z & RKLRR & IBDLR & EBDR & \textbf{FRS-BDR} \\
\midrule
2 subjects & 0.0787	& 0.0328 &	0.2672 &	0	& 0	&0.3110&	0.3547&	0.3583 &0.1570 &0.2016	&0.3412 &	0.3334\\
3 subjects & 0.0730 & 0.0272 & 0.2830 &0 &0	&0.4486 & 0.3053 & 0.4300	& 0.1538 &0.2498 &0.4141 & 0.4102\\
5 subjects & 0.0565	& x	& 0.3147 &	0 &0 &0.4117 &0.3524 &0.5190 &0.1288	&0.2143 & 0.4696 &0.4580\\
8 subjects& 0.0403	&0.0668 & 0.2629 &0	&0	&0.5548	&0.3443	&0.5324	&0.0929	&0.1552	&0.4638	&0.4780\\
10 subjects&0.0346 &0.0549	&0.2259	&0	&0	&0.3287 & 0.2738 &0.4721 &0.0846	&0.1323	&0.4634	&0.4783\\
\hline\hline
\end{tabular}}
\caption{Modularity performance of different block diagonal representation approaches on the MNIST data set. The results are summarized for the similarity measure $\mathbf{W}=\mathbf{X}^{\top}\mathbf{X}$. All benchmark methods are 'oracle'-tuned, i.e., their tuning parameter is chosen such that maximal performance is achieved. `x' denotes the failed results due to the generated unconnected graphs.}\label{tab:mnistmaxmod}

\end{table*}

\begin{table*}[h!]
\centering
\resizebox{\linewidth}{!}{%
\begin{tabular}{p{4cm}M{2cm}M{2cm}M{2cm}M{2cm}M{2cm}M{2cm}M{2cm}M{2cm}M{2cm}M{2cm}M{2cm}M{2cm}}
\hline\hline
\\[-3mm]
 & \multicolumn{12}{c}{Modularity Performance for Different Block Diagonal Representation Methods}\\\\[-3mm]
\cline{2-13}\\[-3mm]
& &\multicolumn{9}{c}{Modularity Performance corresponding to the Minimum Clustering Accuracy of Different Block Diagonal Representation Methods} &  &\\\\[-3mm]
\cline{3-11}\\[-3mm]
MNIST Data Set & $\mathbf{W}$ & SSC & BD-SSC & LRR & BD-LRR & LSR & BDR-B & BDR-Z & RKLRR & IBDLR & EBDR & \textbf{FRS-BDR} \\
\midrule
2 subjects & 0.0787	&x &0.0354 &0 &0 &0.1466 &0.1302 &0.01966 &x &	x &	0.3412 &0.3334\\
3 subjects & 0.0730	&x	&0	&0	&0	&0.0985	&0.0764	&0.0273&x&x	&0.4141	&0.4102\\
5 subjects & 0.0565	&x & 0	&0	&0	&0.0203	&0.0535	&0.0186&x&	x &	0.4696&	0.4580\\
8 subjects& 0.0403	&x	&-0.0003	&0	&0	&0.0737&	0.0117	&0.0203	&x	&x	&0.4638	&0.4780\\
10 subjects&0.0346 &x &0 &0	&0 &0.1659&	0.0027	&0.0212	&x	&x &	0.4634 &0.4783\\
\hline\hline
\end{tabular}}
\caption{Modularity performance of different block diagonal representation approaches on the MNIST data set. The results are summarized for the similarity measure $\mathbf{W}=\mathbf{X}^{\top}\mathbf{X}$. All benchmark methods are 'oracle'-tuned, i.e., their tuning parameter is chosen such that minimal performance is obtained. `x' denotes the failed results due to the generated unconnected graphs.}\label{tab:mnistminmod}
\vspace{-2mm}
\end{table*}

\begin{table*}[h!]
\centering
\resizebox{\linewidth}{!}{%
\begin{tabular}{p{4cm}M{2cm}M{2cm}M{2cm}M{2cm}M{2cm}M{2cm}M{2cm}M{2cm}M{2cm}M{2cm}M{2cm}M{2cm}}
\hline\hline
\\[-3mm]
 & \multicolumn{12}{c}{Conductance Performance for Different Block Diagonal Representation Methods}\\\\[-3mm]
\cline{2-13}\\[-3mm]
& &\multicolumn{9}{c}{Conductance Performance corresponding to the Maximum Clustering Accuracy of Different Block Diagonal Representation Methods} &  &\\\\[-3mm]
\cline{3-11}\\[-3mm]
MNIST Data Set & $\mathbf{W}$ & SSC & BD-SSC & LRR & BD-LRR & LSR & BDR-B & BDR-Z & RKLRR & IBDLR & EBDR & \textbf{FRS-BDR} \\
\midrule
2 subjects & 0.4120	&0.0113	&0.0725	&0.3331	&0.1644	&0.1796	&0.0067&	0.0656&	0.3284	&0.2854	&0.0832	&0.1569\\
3 subjects &0.5647	&0.0054	&0.0868&	0.4609	&0.2571	&0.1862	&0.1222	&0.0890	&0.4111	&0.3844	&0.1325	&0.2387\\
5 subjects & 0.7200	&x	&0.4666	&0.7301	&0.3747	&0.3691 & 0.1728	&0.1453&	0.5902	&0.5450	&0.1831	&0.3236\\
8 subjects& 0.8169	&0.0440	&0.5985	&0.8417	&0.5438	&0.3041	&0.2686	&0.1735	&0.6645	&0.7051	&0.2722	&0.3848\\
10 subjects&0.8509	&0.0078&	0.6613	&0.8753	&0.5919	&0.5595&	0.2232	&0.1662	&0.7063	&0.7569	&0.2960&0.4119\\
\hline\hline
\end{tabular}}
\caption{Conductance performance of different block diagonal representation approaches on the MNIST data set. The results are summarized for the similarity measure $\mathbf{W}=\mathbf{X}^{\top}\mathbf{X}$. All benchmark methods are 'oracle'-tuned, i.e., their tuning parameter is chosen such that maximal performance is achieved. `x' denotes the failed results due to the generated unconnected graphs.}\label{tab:mnistmaxcond}
\end{table*}

\begin{table*}[h!]
\centering
\resizebox{\linewidth}{!}{%
\begin{tabular}{p{4cm}M{2cm}M{2cm}M{2cm}M{2cm}M{2cm}M{2cm}M{2cm}M{2cm}M{2cm}M{2cm}M{2cm}M{2cm}}
\hline\hline
\\[-3mm]
 & \multicolumn{12}{c}{Conductance Performance for Different Block Diagonal Representation Methods}\\\\[-3mm]
\cline{2-13}\\[-3mm]
& &\multicolumn{9}{c}{Conductance Performance corresponding to the Minimum Clustering Accuracy of Different Block Diagonal Representation Methods} &  &\\\\[-3mm]
\cline{3-11}\\[-3mm]
MNIST Data Set & $\mathbf{W}$ & SSC & BD-SSC & LRR & BD-LRR & LSR & BDR-B & BDR-Z & RKLRR & IBDLR & EBDR & \textbf{FRS-BDR} \\
\midrule
2 subjects & 0.4120	& x	&0.0061	&0 &0 &0 &0.0316 &0	 &x	 &x	&0.0832 &0.1569\\
3 subjects &0.5647 & x &0.0133	&0&	0&0 & 0.1030 &0	&x &x &0.1325 &	0.2387\\
5 subjects & 0.7200	&x	&0.0160& 0 & 0	&0	&0.0779	&0	&x	&x &	0.1831	&0.3236\\
8 subjects& 0.8169 & x	&0.0783 & 0& 0	&0.0438	&0.0030	&0	&x	&x&	0.2722	&0.3848\\
10 subjects&0.8509	&x&	0.0179	&0&	0&	0.1213&	0&	0&x&x &0.2960&	0.4119\\
\hline\hline
\end{tabular}}
\caption{Conductance performance of different block diagonal representation approaches on MNIST data set. The results are summarized for the similarity measure $\mathbf{W}=\mathbf{X}^{\top}\mathbf{X}$. All benchmark methods are 'oracle'-tuned, i.e., their tuning parameter is chosen such that minimal performance is obtained. `x' denotes the failed results due to the generated unconnected graphs.}\label{tab:mnistmincond}
\end{table*}

\newpage
\setlength{\parindent}{0pt}\paragraph{F.4.1.2~USPS Data Set}
\textcolor{white}{[tbp!]}\\\vspace{-3mm}
\\
\begin{table*}[h!]
\centering
\resizebox{\linewidth}{!}{%
\begin{tabular}{p{4cm}M{1.5cm}M{1.5cm}M{1.5cm}M{1.5cm}M{1.5cm}M{1.5cm}M{1.5cm}M{1.5cm}M{1.5cm}M{1.5cm}M{1.5cm}M{1.5cm}M{1.5cm}M{1.8cm}}
\hline\hline
\\[-3mm]
 & \multicolumn{14}{c}{Subspace Clustering Performance for Different Block Diagonal Representation Methods}\\\\[-3mm]
\cline{2-15}\\[-3mm]
& &\multicolumn{11}{c}{Minimum-Maximum Clustering Accuracy $(c_\mathrm{accmin}-c_\mathrm{accmax})$ for Different Regularization Parameters} &  &\\\\[-3mm]
\cline{3-13}\\[-3mm]
USPS Data Set & $\mathbf{W}$ & SSC & BD-SSC & LRR & BD-LRR & LSR & BDR-B & BDR-Z & RKLRR & IBDLR &FRPCAG & RSC & EBDR & \textbf{FRS-BDR} \\
\midrule
2 subjects & 87.6 & 51.0-69.5 & 54.3-90.6 & 52.1-74.3 & 52.2-85.9 & 56.0-88.6 & 51.1-93.0 & 51.1-93.1 & 51.0-92.9 & 51.0-93.2&77.3-88.9&51.3-65.3 & 87.0 & 92.6\\
3 subjects & 70.4 & 34.4-55.0 & 36.8-77.4 & 35.5-68.7 & 35.5-69.4 & 38.3-71.9 & 36.5-85.1 & 34.7-85.6 & 34.4-80.7 & 34.4-86.6 &65.0-78.6&37.7-48.8&77.2 & 87.5\\
5 subjects & 61.1 & 21.2-44.4 & 24.7-66.9 & 21.6-45.7 & 22.1-54.6 & 23.1-63.9 & 22.4-74.5 & 21.2-76.0 & 21.2-65.0 & 21.2-76.8 &67.1-73.3&28.1-33.7& 63.3 & 77.3\\
8 subjects& 50.7 & 13.7-35.7 & 18.1-59.4 & 14.4-44.2 & 14.5-46.1 & 15.3-58.7 & 13.8-69.2 & 14.7-68.4 & 13.7-60.1 & 13.7-70.7 &66.0-70.2&21.2-24.2& 52.4 & 65.2\\
10 subjects& 49.4 & 11.0-41.4 & 10.6-56.0 & 11.8-46.2 & 12.2-39.8 & 13.8-52.0 & 12.2-68.6 & 14.8-68.2 & 11.0-57.8 & 11.2-69.8 &56.2-70.4&17.1-23.4& 47.4 & 59.8\\
\hline\\[-3mm]
Average & 63.9 & 26.3-49.2 & 28.9-70.1 & 27.1-55.8 & 27.3-59.2 & 29.3-67.0 & 27.2-78.1 & 27.3-78.2 & 26.3-71.3 & 26.3-79.4 &66.3-76.3&31.1-39.9& 65.5 & 76.5\\
\hline\hline
\end{tabular}}
\caption{Subspace clustering performance of different block diagonal representation approaches on USPS data set. The results are summarized for the similarity measure $\mathbf{W}=\mathbf{X}^{\top}\mathbf{X}$.}
\end{table*}
\begin{table*}[h!]
\centering
\resizebox{9cm}{!}{%
\begin{tabular}{p{4cm}M{1.5cm}M{1.5cm}M{1.5cm}M{1.5cm}M{1.5cm}}
\hline\hline
\\[-3mm]
 & \multicolumn{5}{c}{Detailed Computation time $(t)$ for FRS-BDR Method}\\\\[-3mm]
\cline{2-6}\\[-3mm]
USPS Data Set & Step 1.1 & Step 1.2 & Step 1.3 & Step 2\\
\midrule
2 subjects & 0.003 & 0.015 & 0.049& 0.002  \\
3 subjects & 0.004 & 0.034 & 0.154 & 0.003  \\
5 subjects & 0.007 & 0.108 & 0.401 & 0.005  \\
8 subjects & 0.019 & 0.464 & 1.689 & 0.004  \\
10 subjects & 0.032 & 1.110 & 3.444& 0.119  \\
\hline\hline
\end{tabular}}
\caption{Computation time performance of FRS-BDR method on USPS data set. The results are summarized for the similarity measure $\mathbf{W}=\mathbf{X}^{\top}\mathbf{X}$.}
\end{table*}

\begin{table*}[h!]
\centering
\resizebox{\linewidth}{!}{%
\begin{tabular}{p{4cm}M{1.5cm}M{1.5cm}M{1.5cm}M{1.5cm}M{1.5cm}M{1.5cm}M{1.5cm}M{1.5cm}M{1.5cm}M{1.5cm}M{1.5cm}M{1.5cm}M{1.5cm}M{1.8cm}}
\hline\hline
\\[-3mm]
 & \multicolumn{14}{c}{Computation time $(t)$ for Different Block Diagonal Representation Methods}\\\\[-3mm]
\cline{2-15}\\[-3mm]
& &\multicolumn{11}{c}{Computation Time $(t)$ for Optimally Tuned Regularization Parameters} &  &\\\\[-3mm]
\cline{3-13}\\[-3mm]
USPS Data Set & $\mathbf{W}$ & SSC & BD-SSC & LRR & BD-LRR & LSR & BDR-B & BDR-Z & RKLRR & IBDLR &FRPCAG & RSC& EBDR & \textbf{FRS-BDR} \\
\midrule
2 subjects & 0.001 & 0.096 & 0.108 & 0.561 & 0.587 & 0.001 & 0.260 & 0.256 & 0.561 & 3.971 &1.461&0.060& 0.006 & 0.020\\
3 subjects & 0.001 & 0.228 & 0.252 & 1.239 & 1.115 & 0.001 & 0.607 & 0.606 & 2.358 & 0.952 &5.346&0.231& 0.008 & 0.041\\
5 subjects & 0.001 & 0.631 & 0.632 & 2.639 & 2.509 & 0.002 & 1.757 & 1.781 & 5.302 & 1.962 &8.373&1.542& 0.011 & 0.120\\
8 subjects & 0.003 & 1.429 & 2.011 & 4.106 & 4.286 & 0.006 & 4.837 & 5.095 & 11.177 & 4.579 &19.560&9.288& 0.028 & 0.487\\
10 subjects & 0.003 & 2.260 & 2.929 & 4.257 & 4.713 & 0.008 & 7.127 & 7.664 & 14.232 & 8.551 &15.804&22.698& 0.050 & 1.261\\
\hline\\[-3mm]
Average  & 0.002 & 0.929 & 1.186 & 2.561 & 2.642 & 0.004 & 2.918 & 3.080 & 6.726 & 4.003 & 10.109&6.764&0.020 & 0.386\\
\hline\hline
\end{tabular}}
\caption{Computation time performance of different block diagonal representation approaches on USPS data set. The results are summarized for the similarity measure $\mathbf{W}=\mathbf{X}^{\top}\mathbf{X}$ and sparsity assumed to be known for all sparse representation methods which means that computation time of FRS-BDR is detailed for Steps 1.1, 1.2 and 2.}\vspace{-3mm}
\end{table*}
\begin{figure}[!h]
  \centering
\includegraphics[trim={0mm 0mm 0mm 0mm},clip,width=\linewidth]{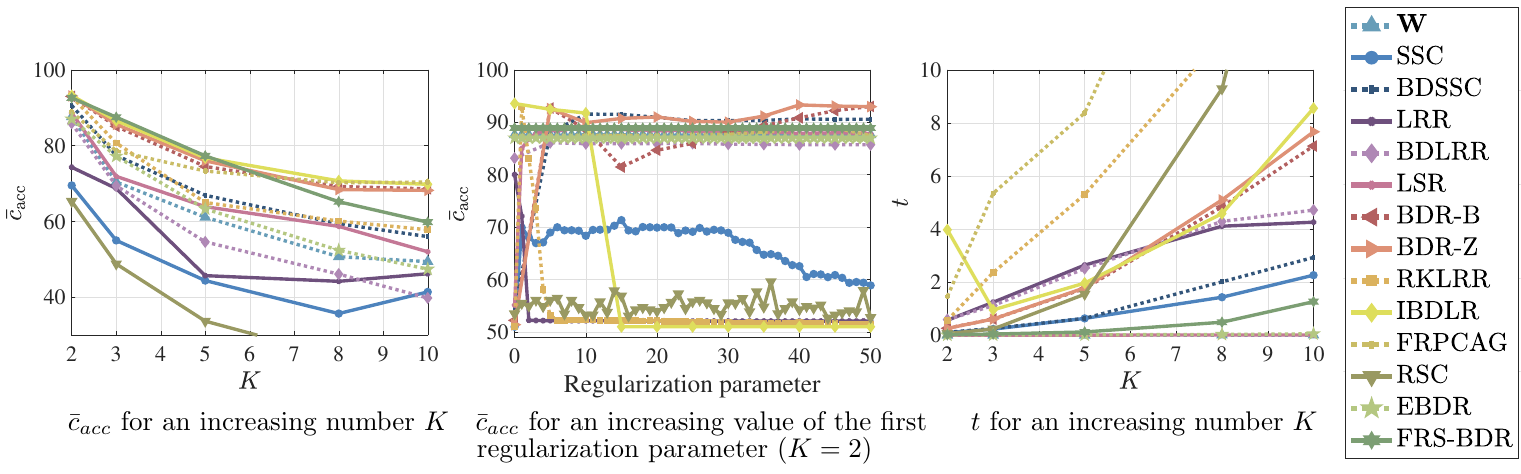}
\caption{Numerical results for USPS data set.}
\end{figure}

\newpage

\subsubsection{F.4.2~Object Clustering}
\setlength{\parindent}{0pt}\paragraph{F.4.2.1~COIL20 Data Set}
\textcolor{white}{[tbp!]}\\\vspace{-3mm}
\\
\begin{figure}[!h]
  \centering
\includegraphics[trim={0mm 0mm 0mm 0mm},clip,width=11.6cm]{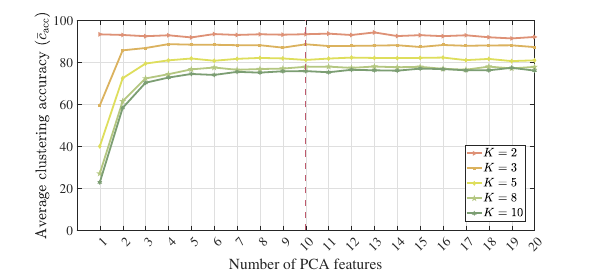}
\caption{Average clustering accuracy ($\bar{c}_\mathrm{acc}$) of FRS-BDR for increasing number of PCA features.}
\end{figure}


\begin{table*}[h!]
\centering
\resizebox{\linewidth}{!}{%
\begin{tabular}{p{4cm}M{1.5cm}M{1.5cm}M{1.5cm}M{1.5cm}M{1.5cm}M{1.5cm}M{1.5cm}M{1.5cm}M{1.5cm}M{1.5cm}M{1.5cm}M{1.5cm}M{1.5cm}M{1.8cm}}
\hline\hline
\\[-3mm]
 & \multicolumn{14}{c}{Subspace Clustering Performance for Different Block Diagonal Representation Methods}\\\\[-3mm]
\cline{2-15}\\[-3mm]
& &\multicolumn{11}{c}{Minimum-Maximum Clustering Accuracy $(c_\mathrm{accmin}-c_\mathrm{accmax})$ for Different Regularization Parameters} &  &\\\\[-3mm]
\cline{3-13}\\[-3mm]
COIL20 Data Set & $\mathbf{W}$ & SSC & BD-SSC & LRR & BD-LRR & LSR & BDR-B & BDR-Z & RKLRR & IBDLR &FRPCAG & RSC & EBDR & \textbf{FRS-BDR} \\
\midrule
2 subjects & 68.9 & 52.5-86.6 & 54.5-65.9 & 57.1-64.9 & 60.8-63.6 & 58.0-61.8 & 54.3-95.9 & 53.3-95.6 & 52.5-72.9 & 52.5-69.7 &90.6-91.6&55.6-60.7& 95.8 & 93.1\\
3 subjects & 42.3 & 36.0-83.3 & 39.3-54.9 & 66.7-68.1& 58.2-67.1 & 44.1-47.3 & 38.2-89.1 & 37.8-88.8 & 36.2-66.0 & 35.9-72.1 & 83.9-87.8&42.3-45.8&90.1 & 88.1 \\
5 subjects & 26.4 & 23.0-80.6 & 26.2-57.6 & 64.0-76.5 & 65.1-76.4 & 34.3-37.0 & 25.3-83.3 & 27.1-83.1 & 23.1-73.6 & 22.8-75.4 &72.5-79.7&31.1-33.8& 80.9 & 82.5\\
8 subjects & 17.3 & 15.6-75.6 & 18.2-63.3 & 56.3-73.4 & 64.0-73.4 & 25.2-28.0 & 18.0-75.6 & 21.9-75.6 & 15.5-71.3 & 15.5-74.2 &67.1-73.4&24.7-26.1& 72.4 & 77.3\\
10 subjects & 14.6 & 13.1-72.5 & 15.6-65.3 & 54.2-72.9 & 63.7-73.0 & 20.6-24.9 & 15.3-74.5 & 20.0-74.2 & 13.1-72.0 & 13.0-73.4 &65.6-71.3&22.2-24.2&69.1 & 75.9\\
\hline\\[-3mm]
Average & 33.9 & 28.1-79.7 & 30.8-61.4 & 59.7-71.1 & 62.4-70.7 & 36.4-39.8 & 30.2-83.7 & 32.0-83.5 & 28.1-71.2 & 27.9-73.0 &75.9-80.7&35.2-38.1& 81.7 & 83.4\\
\hline\hline
\end{tabular}}
\caption{Subspace clustering performance of different block diagonal representation approaches on COIL20 data set. The results are summarized for the similarity measure $\mathbf{W}=\mathbf{X}^{\top}\mathbf{X}$.}
\end{table*}

\begin{table*}[h!]
\centering
\resizebox{9cm}{!}{%
\begin{tabular}{p{4cm}M{1.5cm}M{1.5cm}M{1.5cm}M{1.5cm}M{1.5cm}}
\hline\hline
\\[-3mm]
 & \multicolumn{5}{c}{Detailed Computation time $(t)$ for FRS-BDR Method}\\\\[-3mm]
\cline{2-6}\\[-3mm]
COIL20 Data Set & Step 1.1 & Step 1.2 & Step 1.3 & Step 2\\
\midrule
2 subjects &   0.002 &   0.004  &   0.014 & 0.002 \\
3 subjects &   0.003 &   0.007  &   0.029  & 0.003 \\
5 subjects &  0.006  &   0.014  &   0.071  & 0.007 \\
8 subjects &  0.019  &   0.039  &   0.192  & 0.098 \\
10 subjects & 0.032  &   0.064  &   0.319  &  0.206\\
\hline\hline
\end{tabular}}
\caption{Computation time performance of FRS-BDR method on COIL20 data set. The results are summarized for the similarity measure $\mathbf{W}=\mathbf{X}^{\top}\mathbf{X}$.}
\end{table*}

\begin{table*}[h!]
\centering
\resizebox{\linewidth}{!}{%
\begin{tabular}{p{4cm}M{1.5cm}M{1.5cm}M{1.5cm}M{1.5cm}M{1.5cm}M{1.5cm}M{1.5cm}M{1.5cm}M{1.5cm}M{1.5cm}M{1.5cm}M{1.5cm}M{1.5cm}M{1.8cm}}
\hline\hline
\\[-3mm]
 & \multicolumn{14}{c}{Computation time $(t)$ for Different Block Diagonal Representation Methods}\\\\[-3mm]
\cline{2-15}\\[-3mm]
& &\multicolumn{11}{c}{Computation Time $(t)$ for Optimally Tuned Regularization Parameters} &  &\\\\[-3mm]
\cline{3-13}\\[-3mm]
COIL20 Data Set & $\mathbf{W}$ & SSC & BD-SSC & LRR & BD-LRR & LSR & BDR-B & BDR-Z & RKLRR & IBDLR &FRPCAG & RSC& EBDR & \textbf{FRS-BDR} \\
\midrule
2 subjects & $3\times 10^{-4}$  & 0.017 & 0.031 & 0.012 & 0.021 & $3\times 10^{-4}$ & 0.043 & 0.040 & 0.001 & 1.120 &3.738&0.030& 0.003 & 0.007 \\
3 subjects & $3\times 10^{-4}$  & 0.041 & 0.049 & 0.013 & 0.017 & $2\times 10^{-4}$  & 0.098 & 0.119 & 0.002 & 0.275 &5.199&0.066 &0.005 & 0.013 \\
5 subjects & $4\times 10^{-4}$  & 0.092 & 0.104 & 0.038 & 0.054 & $3\times 10^{-4}$  & 0.244 & 0.249 & 0.004 & 3.688 &9.657&0.294& 0.005 & 0.027 \\
8 subjects & 0.001 & 0.190 & 0.184 & 0.050 & 0.080 & 0.001 & 0.686 & 0.702 & 0.006 & 1.075 &16.737&1.206& 0.008 & 0.156 \\
10 subjects & 0.001 & 0.250 & 0.254 & 0.058 & 0.108 & 0.001 & 0.908 & 0.878 & 0.011 & 1.058 &22.143&1.764& 0.041 & 0.302 \\
\hline\\[-3mm]
Average  & $4\times 10^{-4}$  & 0.118 & 0.124 & 0.034 & 0.056 & $5\times 10^{-4}$  & 0.396 & 0.397 & 0.005 & 1.443 &11.495&0.672& 0.012 &  0.101 \\
\hline\hline
\end{tabular}}
\caption{Computation time performance of different block diagonal representation approaches on COIL20 data set. The results are summarized for the similarity measure $\mathbf{W}=\mathbf{X}^{\top}\mathbf{X}$ and sparsity assumed to be known for all sparse representation methods which means that computation time of FRS-BDR is detailed for Steps 1.1, 1.2 and 2.}\vspace{2mm}
\end{table*}

\begin{table*}[h!]
\centering
\resizebox{\linewidth}{!}{%
\begin{tabular}{p{4cm}M{2cm}M{2cm}M{2cm}M{2cm}M{2cm}M{2cm}M{2cm}M{2cm}M{2cm}M{2cm}M{2cm}M{2cm}}
\hline\hline
\\[-3mm]
 & \multicolumn{12}{c}{Modularity Performance for Different Block Diagonal Representation Methods}\\\\[-3mm]
\cline{2-13}\\[-3mm]
& &\multicolumn{9}{c}{Modularity Performance corresponding to the Maximum Clustering Accuracy of Different Block Diagonal Representation Methods} &  &\\\\[-3mm]
\cline{3-11}\\[-3mm]
COIL20 Data Set & $\mathbf{W}$ & SSC & BD-SSC & LRR & BD-LRR & LSR & BDR-B & BDR-Z & RKLRR & IBDLR & EBDR & \textbf{FRS-BDR} \\
\midrule
2 subjects & 0	&0.3830	&-1	&0	&0	&1	&0.4818	&0.4611	&0.1638	&0.1906	&0.4728	&0.4116\\
3 subjects & -0.69&	0.5196	&0.4726	&0	&0	&1	&0.6243&0.6106&	0.1984	&0.1773	&0.6193	&0.5720\\
5 subjects & -1	&0.6611&0.6648	&0	&0	&1	&0.7672	&0.7482	&0.2505	&0.3196	&0.7320	&0.6638\\
8 subjects& -0.13	&0.7223	&0.7556	&0	&0	&1	&0.8403	&0.8261	&0.2311	&0.2333	&0.7903	&0.6829\\
10 subjects&0.1496&	0.7650	&0.7611	&0	&0	&1	&0.8738	&0.8592	&0.2068	&0.1888	&0.8053	&0.6756\\
\hline\hline
\end{tabular}}
\caption{Modularity performance of different block diagonal representation approaches on the COIL20 data set. The results are summarized for the similarity measure $\mathbf{W}=\mathbf{X}^{\top}\mathbf{X}$. All benchmark methods are 'oracle'-tuned, i.e., their tuning parameter is chosen such that maximal performance is achieved.}\label{tab:coil20maxmod}
\vspace{3mm}
\end{table*}

\begin{table*}[h!]
\centering
\resizebox{\linewidth}{!}{%
\begin{tabular}{p{4cm}M{2cm}M{2cm}M{2cm}M{2cm}M{2cm}M{2cm}M{2cm}M{2cm}M{2cm}M{2cm}M{2cm}M{2cm}}
\hline\hline
\\[-3mm]
 & \multicolumn{12}{c}{Modularity Performance for Different Block Diagonal Representation Methods}\\\\[-3mm]
\cline{2-13}\\[-3mm]
& &\multicolumn{9}{c}{Modularity Performance corresponding to the Minimum Clustering Accuracy of Different Block Diagonal Representation Methods} &  &\\\\[-3mm]
\cline{3-11}\\[-3mm]
COIL20 Data Set & $\mathbf{W}$ & SSC & BD-SSC & LRR & BD-LRR & LSR & BDR-B & BDR-Z & RKLRR & IBDLR & EBDR & \textbf{FRS-BDR} \\
\midrule
2 subjects &0	&x	&-1	&0	&0	&1	&0.0299	&0.0100	&x	&x	&0.4728&	0.4116\\
3 subjects & -0.69&	x	&-1	&0	&0	&1	&0.0281&0.0153&	x&x	&0.6193	&0.5720\\
5 subjects & -1	&x	&-0.0463 &0	&0	&1	&0.0473	&0.0513	&x	&x	&0.7320	&0.6638\\
8 subjects& -0.13	&x	&0.3647	&0	&0	&1	&0.0543	&0.0740	&x	&x	&0.7903	&0.6829\\
10 subjects&0.1496	&x	&0.1165	&0	&0	&1	&0.0492	&0.1039	&x	&x	&0.8053	&0.6756\\
\hline\hline
\end{tabular}}
\caption{Modularity performance of different block diagonal representation approaches on the COIL20 data set. The results are summarized for the similarity measure $\mathbf{W}=\mathbf{X}^{\top}\mathbf{X}$. All benchmark methods are 'oracle'-tuned, i.e., their tuning parameter is chosen such that minimal performance is obtained.  `x' denotes the failed results due to the generated unconnected graphs.}\label{tab:coil20minmod}
\vspace{3mm}
\end{table*}

\begin{table*}[h!]
\centering
\resizebox{\linewidth}{!}{%
\begin{tabular}{p{4cm}M{2cm}M{2cm}M{2cm}M{2cm}M{2cm}M{2cm}M{2cm}M{2cm}M{2cm}M{2cm}M{2cm}M{2cm}}
\hline\hline
\\[-3mm]
 & \multicolumn{12}{c}{Conductance Performance for Different Block Diagonal Representation Methods}\\\\[-3mm]
\cline{2-13}\\[-3mm]
& &\multicolumn{9}{c}{Conductance Performance corresponding to the Maximum Clustering Accuracy of Different Block Diagonal Representation Methods} &  &\\\\[-3mm]
\cline{3-11}\\[-3mm]
COIL20 Data Set & $\mathbf{W}$ & SSC & BD-SSC & LRR & BD-LRR & LSR & BDR-B & BDR-Z & RKLRR & IBDLR & EBDR & \textbf{FRS-BDR} \\
\midrule
2 subjects & 0	&0.0481	&0	&0.3024	&0.0072	&0	&0.0115	&0.0202	&0.2550	&0.1892	&0.0048	&0.0054\\
3 subjects &0.66	&0.0809	&0	&0.0277	&0.0276	&0	&0	&0.0158	&0.3802	&0.4372	&0.0104	&0.03\\
5 subjects & 0.7300	&0.0961	&0.0127	&0.0502	&0.0501	&0	&0	&0.0198	&0.5176	&0.4411	&0.0214	&0.0845\\
8 subjects& 0.5524	&0.1113	&0.0581	&0.0935	&0.0941	&0	&0.0076	&0.0157	&0.6253	&0.6255	&0.0292	&0.1626\\
10 subjects&0.6483	&0.0953	&0.1036	&0.1274	&0.1276	&0	&0.0037	&0.0170	&0.6797	&0.6674	&0.0485	&0.1967\\
\hline\hline
\end{tabular}}
\caption{Conductance performance of different block diagonal representation approaches on the COIL20 data set. The results are summarized for the similarity measure $\mathbf{W}=\mathbf{X}^{\top}\mathbf{X}$. All benchmark methods are 'oracle'-tuned, i.e., their tuning parameter is chosen such that maximal performance is achieved.}\label{tab:coil20maxcond}
\vspace{3mm}
\end{table*}

\begin{table*}[h!]
\centering
\resizebox{\linewidth}{!}{%
\begin{tabular}{p{4cm}M{2cm}M{2cm}M{2cm}M{2cm}M{2cm}M{2cm}M{2cm}M{2cm}M{2cm}M{2cm}M{2cm}M{2cm}}
\hline\hline
\\[-3mm]
 & \multicolumn{12}{c}{Conductance Performance for Different Block Diagonal Representation Methods}\\\\[-3mm]
\cline{2-13}\\[-3mm]
& &\multicolumn{9}{c}{Conductance Performance corresponding to the Minimum Clustering Accuracy of Different Block Diagonal Representation Methods} &  &\\\\[-3mm]
\cline{3-11}\\[-3mm]
COIL20 Data Set & $\mathbf{W}$ & SSC & BD-SSC & LRR & BD-LRR & LSR & BDR-B & BDR-Z & RKLRR & IBDLR & EBDR & \textbf{FRS-BDR} \\
\midrule
2 subjects & 0	&x	&0.3984	&0.0072	&0.0751	&0	&0	&0.0005	&x	&x	&0.0048	&0.0054\\
3 subjects &0.66	&x	&0.0958	&0.0278&	0.0699	&0	&0.1810	&0.0011	&x	&x	&0.0104	&0.0300\\
5 subjects & 0.7300	&x	&0	&0.4180	&0.1409 & 0	&0.0079	&0.0245	&x	&x	&0.0214	&0.0845\\
8 subjects& 0.5524	&x	&0	&0.5218	&0.2410	&0	&0.0085	&0.0560	&x	&x	&0.0292	&0.1626\\
10 subjects&0.6483	&x	&0	&0.5600	&0.2986	&0	&0.0059	&0.1228	&x	&x	&0.0485	&0.1967\\
\hline\hline
\end{tabular}}
\caption{Conductance performance of different block diagonal representation approaches on COIL20 data set. The results are summarized for the similarity measure $\mathbf{W}=\mathbf{X}^{\top}\mathbf{X}$. All benchmark methods are 'oracle'-tuned, i.e., their tuning parameter is chosen such that minimal performance is obtained. `x' denotes the failed results due to the generated unconnected graphs.}\label{tab:coil20mincond}
\vspace{1cm}
\end{table*}

\newpage
\subsubsection{F.4.3~Face Clustering}
\setlength{\parindent}{0pt}\paragraph{F.4.3.1~ORL Data Set}
\textcolor{white}{[tbp!]}\\\vspace{-3mm}
\\
\begin{figure}[!h]\vspace{-3mm}
  \centering
\includegraphics[trim={0mm 0mm 0mm 0mm},clip,width=11.6cm]{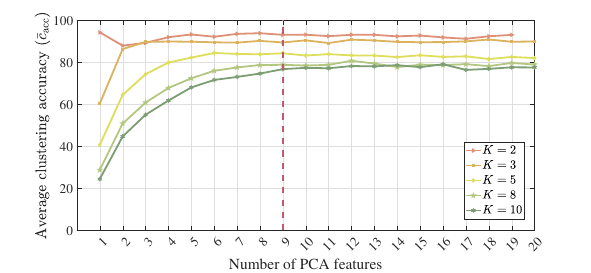}
\caption{Average clustering accuracy ($\bar{c}_\mathrm{acc}$) of FRS-BDR for increasing number of PCA features.}
\end{figure}

\begin{table*}[h!]
\centering
\resizebox{\linewidth}{!}{%
\begin{tabular}{p{4cm}M{1.5cm}M{1.5cm}M{1.5cm}M{1.5cm}M{1.5cm}M{1.5cm}M{1.5cm}M{1.5cm}M{1.5cm}M{1.5cm}M{1.5cm}M{1.5cm}M{1.5cm}M{1.8cm}}
\hline\hline
\\[-3mm]
 & \multicolumn{14}{c}{Subspace Clustering Performance for Different Block Diagonal Representation Methods}\\\\[-3mm]
\cline{2-15}\\[-3mm]
& &\multicolumn{11}{c}{Minimum-Maximum Clustering Accuracy $(c_\mathrm{accmin}-c_\mathrm{accmax})$ for Different Regularization Parameters} &  &\\\\[-3mm]
\cline{3-13}\\[-3mm]
ORL Data Set & $\mathbf{W}$ & SSC & BD-SSC & LRR & BD-LRR & LSR & BDR-B & BDR-Z & RKLRR & IBDLR &FRPCAG & RSC & EBDR & \textbf{FRS-BDR} \\
\midrule
2 subjects & 56.8 & 55.0-81.9 & 55.1-64.0 & 57.6-69.1 & 59.1-69.1 & 57.7-60.5 & 55.0-95.6 & 55.3-95.4 & 55.0-73.7 & x &90.1-91.8&58.7-61.5& 95.2 & 93.1\\
3 subjects & 41.5 & 38.8-80.8 & 43.4-50.7 & 58.6-63.4 & 58.0-59.7 & 45.1-46.6 & 38.9-90.5 & 40.0-90.5 & 39.0-62.9 & 38.8-69.0 &84.1-85.8&45.4-47.4& 90.5 & 90.1 \\
5 subjects &28.4 & 26.0-77.1 & 31.5-41.0 & 51.3-57.7& 57.0-66.7 & 36.3-38.9 & 26.2-80.1 & 30.8-80.1 & 26.1-60.6 & 25.7-70.8 &77.3-82.6&34.9-36.6& 83.0 & 84.2 \\
8 subjects &21.8& 18.6-74.4 & 25.3-35.8 & 45.0-72.6& 56.7-68.5 & 26.8-28.5 & 18.9-79.1 & 34.6-78.5 & 18.6-75.4 & 18.5-75.1 &74.6-82.2&28.7-30.0& 74.1 & 79.4\\
10 subjects & 18.7 & 16.3-74.2 & 22.7-36.1 & 41.8-72.6 & 62.7-72.8 & 24.0-25.3 & 16.4-78.4 & 31.2-78.9 & 16.5-72.5 & 16.1-72.4 &70.6-81.3&25.6-27.8& 69.9 & 77.2 \\
\hline\\[-3mm]
Average & 33.5 & 30.9-77.7 & 35.6-45.5 & 50.9-67.1 & 58.7-67.3 & 38.0-40.0 & 31.1-84.7 & 38.4-84.7 & 31.0-69.0 & 24.8-71.8 &79.3-84.7&38.7-40.7& 82.5 & 84.8\\
\hline\hline
\end{tabular}}
\caption{Subspace clustering performance of different block diagonal representation approaches on ORL data set. The results are summarized for the similarity measure $\mathbf{W}=\mathbf{X}^{\top}\mathbf{X}$. `x' denotes the failed results.}
\end{table*}

\begin{table*}[h!]
\centering
\resizebox{10cm}{!}{%
\begin{tabular}{p{4cm}M{1.5cm}M{1.5cm}M{1.5cm}M{1.5cm}M{1.5cm}}
\hline\hline
\\[-3mm]
 & \multicolumn{5}{c}{Detailed Computation time $(t)$ for FRS-BDR Method}\\\\[-3mm]
\cline{2-6}\\[-3mm]
ORL Data Set & Step 1.1 & Step 1.2 & Step 1.3 & Step 2\\
\midrule
2 subjects & 0.001 & 0.002 & 0.005 & 0.002 \\
3 subjects & 0.001 & 0.002 & 0.009 & 0.002 \\
5 subjects & 0.002 & 0.005 & 0.018 & 0.010 \\
8 subjects & 0.004 & 0.010 & 0.044 & 0.119 \\
10 subjects & 0.007 & 0.014 & 0.079 & 1.810 \\
\hline\hline
\end{tabular}}
\caption{Computation time performance of FRS-BDR method on ORL data set. The results are summarized for the similarity measure $\mathbf{W}=\mathbf{X}^{\top}\mathbf{X}$.}
\end{table*}

\begin{table*}[h!]
\centering
\resizebox{\linewidth}{!}{%
\begin{tabular}{p{4cm}M{1.5cm}M{1.5cm}M{1.5cm}M{1.5cm}M{1.5cm}M{1.5cm}M{1.5cm}M{1.5cm}M{1.5cm}M{1.5cm}M{1.5cm}M{1.5cm}M{1.5cm}M{1.8cm}}
\hline\hline
\\[-3mm]
 & \multicolumn{14}{c}{Computation time $(t)$ for Different Block Diagonal Representation Methods}\\\\[-3mm]
\cline{2-13}\\[-3mm]
& &\multicolumn{11}{c}{Computation Time $(t)$ for Optimally Tuned Regularization Parameters} &  &\\\\[-3mm]
\cline{3-11}\\[-3mm]
ORL Data Set & $\mathbf{W}$ & SSC & BD-SSC & LRR & BD-LRR & LSR & BDR-B & BDR-Z & RKLRR & IBDLR & FRPCAG & RSC & EBDR & \textbf{FRS-BDR}\\
\midrule
2 subjects & $3\times 10^{-4}$ & 0.016 & 0.023 & 0.011 & 0.012 & $4\times 10^{-5}$ & 0.009 & 0.009 & $2\times 10^{-4}$ & x &1.791&0.027& 0.003 & 0.005\\
3 subjects & $3\times 10^{-4}$ & 0.018 & 0.025 & 0.012 & 0.014 & $5\times 10^{-5}$ & 0.027 & 0.026 & $4\times 10^{-4}$ & 0.133 &1.959&0.042& 0.003 & 0.006 \\
5 subjects & $3\times 10^{-4}$ & 0.034 & 0.047 & 0.013 & 0.017 & $8\times 10^{-5}$ & 0.053 & 0.054 & 0.001 & 0.171 &1.503&0.090& 0.017 \\
8 subjects &$4\times 10^{-4}$& 0.114 & 0.078 & 0.032 & 0.051 & $2\times 10^{-4}$ & 0.063 & 0.065 & 0.003 & 2.089 &2.487&0.240& 0.011 & 0.133\\
10 subjects & $5\times 10^{-4}$ & 0.111 &0.108  &0.036  &0.051  &$3\times 10^{-4}$ & 0.116 & 0.085 & 0.004 & 3.274 &3.099&0.246&0.232 & 1.831 \\
\hline\\[-3mm]
Average & $4\times 10^{-4}$ & 0.058 & 0.056 & 0.021 & 0.029 & $2\times 10^{-4}$ & 0.054  & 0.048 & 0.002 & 1.417 &2.168&0.129 &0.050  & 0.398 \\
\hline\hline
\end{tabular}}
\caption{Computation time performance of different block diagonal representation approaches on ORL data set. The results are summarized for the similarity measure $\mathbf{W}=\mathbf{X}^{\top}\mathbf{X}$ and sparsity assumed to be known for all sparse representation methods which means that computation time of FRS-BDR is detailed for Steps 1.1, 1.2 and 2.}\vspace{2.5mm}
\end{table*}

\begin{table*}[h!]
\centering
\resizebox{\linewidth}{!}{%
\begin{tabular}{p{4cm}M{2cm}M{2cm}M{2cm}M{2cm}M{2cm}M{2cm}M{2cm}M{2cm}M{2cm}M{2cm}M{2cm}M{2cm}}
\hline\hline
\\[-3mm]
 & \multicolumn{12}{c}{Modularity Performance for Different Block Diagonal Representation Methods}\\\\[-3mm]
\cline{2-13}\\[-3mm]
& &\multicolumn{9}{c}{Modularity Performance corresponding to the Maximum Clustering Accuracy of Different Block Diagonal Representation Methods} &  &\\\\[-3mm]
\cline{3-11}\\[-3mm]
ORL Data Set & $\mathbf{W}$ & SSC & BD-SSC & LRR & BD-LRR & LSR & BDR-B & BDR-Z & RKLRR & IBDLR & EBDR & \textbf{FRS-BDR} \\
\midrule
2 subjects &-0.0305	&0.2731	&-1	&0	&0	&1	&0.4742	&0.4664	&0.1690	&x	&0.4827	&0.4287\\
3 subjects & -0.0566	&0.4634	&-1	&0	&0	&1	&0.6276	&0.6183	&0.2015	&0.1427	&0.5913	&0.5616\\
5 subjects & -0.0235	&0.5125	&-1	&0	&0	&1	&0.7381	&0.7205	&0.2043	&0.1891	&0.6659	&0.6104\\
8 subjects& -0.1393	&0.6923	&-1	&0	&0	&1	&0.2905	&0.2918	&0.1985	&0.2649	&0.6905	&0.5909\\
10 subjects&-0.7352	&0.7460	&-1	&0	&0	&1	&0.2743	&0.2762	&0.1736	&0.2584	&0.6766	&0.5743\\
\hline\hline
\end{tabular}}
\caption{Modularity performance of different block diagonal representation approaches on the ORL data set. The results are summarized for the similarity measure $\mathbf{W}=\mathbf{X}^{\top}\mathbf{X}$. All benchmark methods are 'oracle'-tuned, i.e., their tuning parameter is chosen such that maximal performance is achieved. `x' denotes the failed results due to the generated unconnected graphs.}\label{tab:ORLmaxmod}
\vspace{1cm}
\end{table*}

\begin{table*}[h!]
\centering
\resizebox{\linewidth}{!}{%
\begin{tabular}{p{4cm}M{2cm}M{2cm}M{2cm}M{2cm}M{2cm}M{2cm}M{2cm}M{2cm}M{2cm}M{2cm}M{2cm}M{2cm}}
\hline\hline
\\[-3mm]
 & \multicolumn{12}{c}{Modularity Performance for Different Block Diagonal Representation Methods}\\\\[-3mm]
\cline{2-13}\\[-3mm]
& &\multicolumn{9}{c}{Modularity Performance corresponding to the Minimum Clustering Accuracy of Different Block Diagonal Representation Methods} &  &\\\\[-3mm]
\cline{3-11}\\[-3mm]
ORL Data Set & $\mathbf{W}$ & SSC & BD-SSC & LRR & BD-LRR & LSR & BDR-B & BDR-Z & RKLRR & IBDLR & EBDR & \textbf{FRS-BDR} \\
\midrule
2 subjects &-0.0305	&x	&0.1398	&0	&0	&1	&0.0923	&0.0174	&x	&x	&0.4827	&0.4287\\
3 subjects & -0.0566 &x	&-1	&0	&0	&1	&0	&0.0074	&x	&x	&0.5913	&0.5616\\
5 subjects &-0.0235	&x	&-1	&0	&0	&1	&0	&0.0410	&x	&x	&0.6659	&0.6104\\
8 subjects& -0.1393	&x	&-1	&0	&0	&1	&0	&0.1382	&x	&x	&0.6905	&0.5909\\
10 subjects&-0.7352	&x	&-1	&0	&0	&1	&0	&0.1483	&x	&x	&0.6766	&0.5743\\
\hline\hline
\end{tabular}}
\caption{Modularity performance of different block diagonal representation approaches on the ORL data set. The results are summarized for the similarity measure $\mathbf{W}=\mathbf{X}^{\top}\mathbf{X}$. All benchmark methods are 'oracle'-tuned, i.e., their tuning parameter is chosen such that minimal performance is obtained. `x' denotes the failed results due to the generated unconnected graphs.}\label{tab:ORLminmod}
\vspace{1cm}
\end{table*}

\begin{table*}[h!]
\centering
\resizebox{\linewidth}{!}{%
\begin{tabular}{p{4cm}M{2cm}M{2cm}M{2cm}M{2cm}M{2cm}M{2cm}M{2cm}M{2cm}M{2cm}M{2cm}M{2cm}M{2cm}}
\hline\hline
\\[-3mm]
 & \multicolumn{12}{c}{Conductance Performance for Different Block Diagonal Representation Methods}\\\\[-3mm]
\cline{2-13}\\[-3mm]
& &\multicolumn{9}{c}{Conductance Performance corresponding to the Maximum Clustering Accuracy of Different Block Diagonal Representation Methods} &  &\\\\[-3mm]
\cline{3-11}\\[-3mm]
ORL Data Set & $\mathbf{W}$ & SSC & BD-SSC & LRR & BD-LRR & LSR & BDR-B & BDR-Z & RKLRR & IBDLR & EBDR & \textbf{FRS-BDR} \\
\midrule
2 subjects & 0.2303	&0.1195	&1	&0.0316	&0.0315	&0	&0	&0.0055	&0.2539	&x &	0.0127	&0.0099\\
3 subjects &0.4539	&0.1390	&1	&0.0638	&0.0985	&0	&0	&0.0103	&0.3884	&0	&0.0425	&0.0579\\
5 subjects & 0.5634	&0.2447	&1	&0.1103	&0.1417	&0	&0.1742	&0.0201	&0.5294	&0.5862	&0.0804	&0.1553\\
8 subjects& 0.9308	&0.1586	&1	&0.1781	&0.1779	&0	&0.5656	&0.5650	&0.6577	&0.5854	&0.1291	&0.2559\\
10 subjects&0.7982	&0.1347	&1	&0.2453	&0.2459	&0	&0.6159	&0.6145	&0.7146	&0.6306	&0.1674	&0.3090\\
\hline\hline
\end{tabular}}
\caption{Conductance performance of different block diagonal representation approaches on the ORL data set. The results are summarized for the similarity measure $\mathbf{W}=\mathbf{X}^{\top}\mathbf{X}$. All benchmark methods are 'oracle'-tuned, i.e., their tuning parameter is chosen such that maximal performance is achieved. `x' denotes the failed results due to the generated unconnected graphs.}\label{tab:ORLmaxcond}
\vspace{1cm}
\end{table*}

\begin{table*}[h!]
\centering
\resizebox{\linewidth}{!}{%
\begin{tabular}{p{4cm}M{2cm}M{2cm}M{2cm}M{2cm}M{2cm}M{2cm}M{2cm}M{2cm}M{2cm}M{2cm}M{2cm}M{2cm}}
\hline\hline
\\[-3mm]
 & \multicolumn{12}{c}{Conductance Performance for Different Block Diagonal Representation Methods}\\\\[-3mm]
\cline{2-13}\\[-3mm]
& &\multicolumn{9}{c}{Conductance Performance corresponding to the Minimum Clustering Accuracy of Different Block Diagonal Representation Methods} &  &\\\\[-3mm]
\cline{3-11}\\[-3mm]
ORL Data Set & $\mathbf{W}$ & SSC & BD-SSC & LRR & BD-LRR & LSR & BDR-B & BDR-Z & RKLRR & IBDLR & EBDR & \textbf{FRS-BDR} \\
\midrule
2 subjects & 0.2303	&x	&0.1221	&0.0316	&0.0673	&0	&0	&0	&x	&x	&0.0127	&0.0099\\
3 subjects &0.4539	&x	&0.2866	&0.0642	&0.1183	&0	&0	&0	&x	&x	&0.0425	&0.0579\\
5 subjects & 0.5634	&x	&0	&0.1106	&0.1109	&0	&0	&0.0944	&x	&x	&0.0804	&0.1553\\
8 subjects&0.9308	&x	&0.6875	&0.5875	&0.2976	&0	&0	&0.3606	&x	&x	&0.1291	&0.2559\\
10 subjects&0.7982	&x	&0	&0.6190	&0.3866	&0	&0	&0.3708	&x	&x	&0.1674	&0.3090\\
\hline\hline
\end{tabular}}
\caption{Conductance performance of different block diagonal representation approaches on ORL data set. The results are summarized for the similarity measure $\mathbf{W}=\mathbf{X}^{\top}\mathbf{X}$. All benchmark methods are 'oracle'-tuned, i.e., their tuning parameter is chosen such that minimal performance is obtained. `x' denotes the failed results due to the generated unconnected graphs.}\label{tab:ORLmincond}
\vspace{3.5cm}
\end{table*}

\newpage
\setlength{\parindent}{0pt}\paragraph{F.4.3.2~JAFFE Data Set}
\textcolor{white}{[tbp!]}\\\vspace{-3mm}
\\
\begin{figure}[!h]
  \centering
\includegraphics[trim={0mm 0mm 0mm 0mm},clip,width=11.6cm]{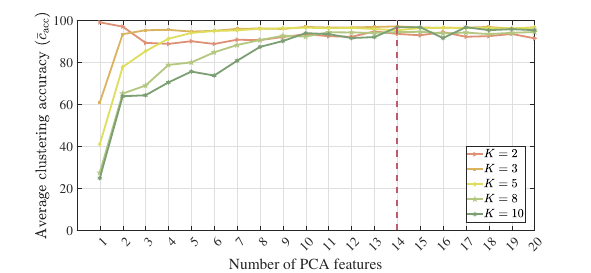}
\caption{Average clustering accuracy ($\bar{c}_\mathrm{acc}$) of FRS-BDR for increasing number of PCA features.}
\end{figure}

\begin{table*}[h!]
\centering
\resizebox{\linewidth}{!}{%
\begin{tabular}{p{4cm}M{1.5cm}M{1.5cm}M{1.5cm}M{1.5cm}M{1.5cm}M{1.5cm}M{1.5cm}M{1.5cm}M{1.5cm}M{1.5cm}M{1.5cm}M{1.5cm}M{1.5cm}M{1.8cm}}
\hline\hline
\\[-3mm]
 & \multicolumn{14}{c}{Subspace Clustering Performance for Different Block Diagonal Representation Methods}\\\\[-3mm]
\cline{2-15}\\[-3mm]
& &\multicolumn{11}{c}{Minimum-Maximum Clustering Accuracy $(c_\mathrm{accmin}-c_\mathrm{accmax})$ for Different Regularization Parameters} &  &\\\\[-3mm]
\cline{3-13}\\[-3mm]
JAFFE Data Set & $\mathbf{W}$ & SSC & BD-SSC & LRR & BD-LRR & LSR & BDR-B & BDR-Z & RKLRR & IBDLR &FRPCAG & RSC& EBDR & \textbf{FRS-BDR} \\
\midrule
2 subjects & 56.1 & 52.1-93.3 & 52.8-70.4 & 56.4-60.7 & 57.1-59.0 & 53.4-57.1 & 52.3-99.7 & 52.4-99.7 & 52.4-82.0 & 52.0-79.6 &98.8-99.8&56.3-59.1& 94.9 & 93.4\\
3 subjects & 39.7 & 36.2-95.2 & 37.1-59.7 & 49.1-78.6 & 49.0-65.3& 39.8-45.1 & 36.1-98.0& 36.5-98.0 & 36.2-76.7 & 35.9-75.5 &96.7-98.9&43.1-44.7& 92.4 & 96.9\\
5 subjects & 26.4 & 23.3-95.3 & 26.4-76.5 & 59.9-65.7 & 60.0-85.0 & 33.1-37.8 & 23.2-94.2 & 28.3-94.2 & 23.3-69.0 & 23.1-88.0 &91.4-95.3&31.7-33.0& 90.7 & 96.4\\
8 subjects & 19.1 & 15.8-93.6 & 18.6-91.9 & 56.0-85.5 & 80.3-89.6 & 24.2-38.1 & 15.8-94.3 & 34.2-94.3 & 15.8-87.4 & 15.7-94.2 &89.3-94.4&24.4-25.5& 82.4 & 94.4\\
10 subjects & x & 13.1-93.0 & 16.4-93.9 & 54.9-92.0 & 85.9-92.0 & 11.7-55.9 & 12.2-91.5 & 25.8-91.5 & 13.1-92.0 & 12.7-94.8 &84.5-97.7&x& 77.9  & 96.7\\
\hline\\[-3mm]
Average & 35.3 & 28.1-94.1 & 30.3-78.5 & 55.3-76.5 & 66.5-78.2 & 32.4-46.8 & 27.9-95.6 & 35.4-95.6 & 28.2-81.4 & 27.9-86.4 &92.2-97.2&38.9-40.5& 87.7 & 95.5\\
\hline\hline
\end{tabular}}
\caption{Subspace clustering performance of different block diagonal representation approaches on JAFFE data set. The results are summarized for the similarity measure $\mathbf{W}=\mathbf{X}^{\top}\mathbf{X}$. `x' denotes the failed results.}
\end{table*}

\begin{table*}[h!]
\centering
\resizebox{9cm}{!}{%
\begin{tabular}{p{4cm}M{1.5cm}M{1.5cm}M{1.5cm}M{1.5cm}M{1.5cm}}
\hline\hline
\\[-3mm]
 & \multicolumn{5}{c}{Detailed Computation time $(t)$ for FRS-BDR Method}\\\\[-3mm]
\cline{2-6}\\[-3mm]
JAFFE Data Set & Step 1.1 & Step 1.2 & Step 1.3 & Step 2\\
\midrule
2 subjects &   0.002    &  0.004     & 0.015      & 0.002 \\
3 subjects &  0.003     &  0.008     &  0.034     &  0.003\\
5 subjects &  0.007     &  0.016     &  0.087     &  0.013 \\
8 subjects &  0.022     &  0.047     &  0.245     & 0.298 \\
10 subjects &  0.043     &  0.076     &  0.469     &  0.777\\
\hline\hline
\end{tabular}}
\caption{Computation time performance of FRS-BDR method on JAFFE data set. The results are summarized for the similarity measure $\mathbf{W}=\mathbf{X}^{\top}\mathbf{X}$.}
\end{table*}

\begin{table*}[h!]
\centering
\resizebox{\linewidth}{!}{%
\begin{tabular}{p{4cm}M{1.5cm}M{1.5cm}M{1.5cm}M{1.5cm}M{1.5cm}M{1.5cm}M{1.5cm}M{1.5cm}M{1.5cm}M{1.5cm}M{1.5cm}M{1.5cm}M{1.5cm}M{1.8cm}}
\hline\hline
\\[-3mm]
 & \multicolumn{14}{c}{Computation time $(t)$ for Different Block Diagonal Representation Methods}\\\\[-3mm]
\cline{2-15}\\[-3mm]
& &\multicolumn{11}{c}{Computation Time $(t)$ for Optimally Tuned Regularization Parameters} &  &\\\\[-3mm]
\cline{3-13}\\[-3mm]
JAFFE Data Set & $\mathbf{W}$ & SSC & BD-SSC & LRR & BD-LRR & LSR & BDR-B & BDR-Z & RKLRR & IBDLR &FRPCAG & RSC& EBDR & \textbf{FRS-BDR} \\
\midrule
2 subjects & $3\times 10^{-4}$  & 0.022 & 0.032 & 0.015 & 0.018 & $8\times 10^{-5}$  & 0.060 & 0.060 & 0.006 & 0.505 &1.563&0.033& 0.004 &  0.008\\
3 subjects & $4\times 10^{-4}$  & 0.053 & 0.044 & 0.050 & 0.056 & $3\times 10^{-4}$  & 0.106 & 0.108 & 0.002 & 1.794 &5.691&0.108& 0.004 & 0.015 \\
5 subjects & $4\times 10^{-4}$  & 0.109 & 0.109 & 0.066 & 0.087 & $4\times 10^{-4}$  & 0.246 & 0.247 & 0.004 & 0.563 &11.703&0.333& 0.005 & 0.037 \\
8 subjects & $5\times 10^{-4}$  & 0.215 & 0.216 & 0.068 & 0.129 & 0.001 & 0.561 & 0.555  & 0.010 & 0.972 & 31.371&1.404&0.036 & 0.367 \\
10 subjects & $6\times 10^{-4}$  & 0.291 & 0.235 & 0.063 & 0.120 & 0.001 & 0.625 & 0.635 & 0.013 & 1.044 & 8.136&x&0.083 &  0.896\\
\hline\\[-3mm]
Average  &$4\times 10^{-4}$   & 0.138 & 0.127  & 0.052 & 0.082 & $4\times 10^{-4}$ & 0.320 & 0.321 & 0.007 & 0.976 &11.693&0.470& 0.026 & 0.264 \\
\hline\hline
\end{tabular}}
\caption{Computation time performance of different block diagonal representation approaches on JAFFE data set. The results are summarized for the similarity measure $\mathbf{W}=\mathbf{X}^{\top}\mathbf{X}$ and sparsity assumed to be known for all sparse representation methods which means that computation time of FRS-BDR is detailed for Steps 1.1, 1.2 and 2.}
\end{table*}

\newpage
\setlength{\parindent}{0pt}\paragraph{F.4.3.3~Yale Data Set}
\textcolor{white}{[tbp!]}\\\vspace{-3mm}
\\

\begin{figure}[!h]
  \centering
\includegraphics[trim={0mm 0mm 0mm 0mm},clip,width=11.6cm]{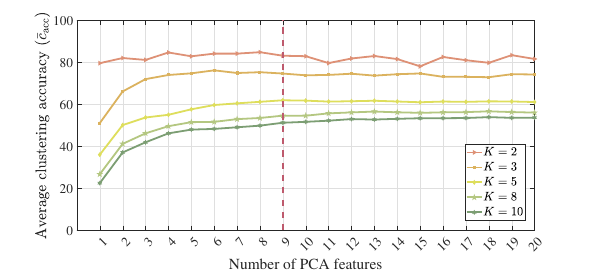}
\caption{Average clustering accuracy ($\bar{c}_\mathrm{acc}$) of FRS-BDR for increasing number of PCA features.}
\end{figure}

\begin{table*}[h!]
\centering
\resizebox{\linewidth}{!}{%
\begin{tabular}{p{4cm}M{1.5cm}M{1.5cm}M{1.5cm}M{1.5cm}M{1.5cm}M{1.5cm}M{1.5cm}M{1.5cm}M{1.5cm}M{1.5cm}M{1.5cm}M{1.5cm}M{1.5cm}M{1.8cm}}
\hline\hline
\\[-3mm]
 & \multicolumn{14}{c}{Subspace Clustering Performance for Different Block Diagonal Representation Methods}\\\\[-3mm]
\cline{2-15}\\[-3mm]
& &\multicolumn{9}{c}{Minimum-Maximum Clustering Accuracy $(c_\mathrm{accmin}-c_\mathrm{accmax})$ for Different Regularization Parameters} &  &\\\\[-3mm]
\cline{3-13}\\[-3mm]
Yale Data Set & $\mathbf{W}$ & SSC & BD-SSC & LRR & BD-LRR & LSR & BDR-B & BDR-Z & RKLRR & IBDLR &FRPCAG & RSC & EBDR & \textbf{FRS-BDR} \\
\midrule
2 subjects & 55.3 & 54.5-75.6 & 54.7-58.8 & 55.7-60.9 & 57.5-60.9 & 54.5-57.3 & 53.6-84.2 & 54.5-83.9 & 54.5-60.4 & x &66.1-90.4&55.5-62.2& 84.1 & 83.1\\
3 subjects & 41.9 & 38.4-67.2 & 43.0-46.7 & 51.0-51.4 & 50.9-53.6 & 42.1-43.8 & 38.6-68.5 & 39.0-68.6 & 38.6-52.4 & 38.2-55.1 &61.1-69.2&43.3-49.0& 71.6 & 74.5 \\
5 subjects &29.8 & 25.5-57.9 & 30.5-37.4 & 43.6-49.8 & 49.2-54.4 & 31.1-33.2 & 25.7-62.3 & 29.9-62.2 & 25.4-49.9 & 25.2-52.9 &52.9-62.7&33.8-36.9& 60.2 & 61.7 \\
8 subjects &21.6& 18.1-52.6 & 22.3-29.7 & 37.4-47.9 & 47.3-50.7 & 23.6-25.4 & 18.3-51.9 & 28.5-51.5 & 18.3-48.6 & 17.9-49.6 &44.3-57.1&27.5-30.3& 53.8 & 54.8\\
10 subjects & 18.7 & 15.6-50.0 & 20.2-30.2 & 34.5-47.5 & 46.2-47.9 & 20.8-22.0 & 15.8-51.6 & 24.5-51.6 & 15.9-46.3 & 15.4-46.3 &40.3-55.0&25.3-27.3& 49.2 & 51.2 \\
\hline\\[-3mm]
Average & 33.5 & 30.4-60.7 & 34.1-40.6 & 44.4-51.5 & 50.2-53.5 & 34.4-36.3 & 30.4-63.7 & 35.3-63.6 & 30.5-51.5 & 24.2-51.0 &52.9-66.9&37.1-41.2& 63.8 & 65.1\\
\hline\hline
\end{tabular}}
\caption{Subspace clustering performance of different block diagonal representation approaches on Yale data set. The results are summarized for the similarity measure $\mathbf{W}=\mathbf{X}^{\top}\mathbf{X}$. `x' denotes the failed results.}
\end{table*}

\begin{table*}[h!]
\centering
\resizebox{10cm}{!}{%
\begin{tabular}{p{4cm}M{1.5cm}M{1.5cm}M{1.5cm}M{1.5cm}M{1.5cm}}
\hline\hline
\\[-3mm]
 & \multicolumn{5}{c}{Detailed Computation time $(t)$ for FRS-BDR Method}\\\\[-3mm]
\cline{2-6}\\[-3mm]
Yale Data Set & Step 1.1 & Step 1.2 & Step 1.3 & Step 2\\
\midrule
2 subjects & 0.001 & 0.002 & 0.007 & 0.002 \\
3 subjects & 0.001 & 0.003 & 0.013 & 0.003 \\
5 subjects & 0.002 & 0.006 & 0.025 & 0.013 \\
8 subjects & 0.005 & 0.012 & 0.056 & 0.046 \\
10 subjects & 0.008 & 0.018 & 0.099 & 0.160 \\
\hline\hline
\end{tabular}}
\caption{Computation time performance of FRS-BDR method on Yale data set. The results are summarized for the similarity measure $\mathbf{W}=\mathbf{X}^{\top}\mathbf{X}$.}
\end{table*}

\begin{table*}[h!]
\centering
\resizebox{\linewidth}{!}{%
\begin{tabular}{p{4cm}M{1.5cm}M{1.5cm}M{1.5cm}M{1.5cm}M{1.5cm}M{1.5cm}M{1.5cm}M{1.5cm}M{1.5cm}M{1.5cm}M{1.5cm}M{1.5cm}M{1.5cm}M{1.8cm}}
\hline\hline
\\[-3mm]
 & \multicolumn{14}{c}{Computation time $(t)$ for Different Block Diagonal Representation Methods}\\\\[-3mm]
\cline{2-15}\\[-3mm]
& &\multicolumn{11}{c}{Computation Time $(t)$ for Optimally Tuned Regularization Parameters} &  &\\\\[-3mm]
\cline{3-13}\\[-3mm]
Yale Data Set & $\mathbf{W}$ & SSC & BD-SSC & LRR & BD-LRR & LSR & BDR-B & BDR-Z & RKLRR & IBDLR &FRPCAG & RSC&EBDR & \textbf{FRS-BDR} \\
\midrule
2 subjects & $3\times 10^{-4}$ & 0.011 & 0.018 & 0.013 & 0.011 & $5\times 10^{-5}$ & 0.010 & 0.010 & $2\times 10^{-4}$ & x & 2.661& 0.027& 0.003 & 0.005\\
3 subjects & $4\times 10^{-4}$ & 0.022 & 0.023 & 0.010 & 0.011 & $6\times 10^{-5}$ & 0.039 & 0.049 & $4\times 10^{-4}$ & 0.059 & 3.942& 0.045& 0.003 & 0.007 \\
5 subjects & $3\times 10^{-4}$ & 0.039 & 0.040 & 0.013 & 0.017 & $1\times 10^{-4}$ & 0.043 & 0.038 & 0.002 & 0.276 & 2.847& 0.105& 0.005 & 0.021 \\
8 subjects &$4\times 10^{-4}$& 0.093 & 0.090 & 0.031 & 0.053 & $3\times 10^{-4}$ &  0.055 & 0.053 & 0.003 & 0.705 & 3.609& 0.192& 0.006 & 0.063\\
10 subjects & $4\times 10^{-4}$ & 0.138 & 0.126 & 0.036 & 0.062 & $4\times 10^{-4}$ & 0.061 & 0.069 & 0.004 & 1.077 & 3.246& 0.270&0.022  & 0.186 \\
\hline\\[-3mm]
Average & $4\times 10^{-4}$ & 0.061 & 0.059 & 0.021 & 0.031 &  $2\times 10^{-4}$&  0.042& 0.044 & 0.002 & 0.529 & 3.261& 0.128&0.008  & 0.056 \\
\hline\hline
\end{tabular}}
\caption{Computation time performance of different block diagonal representation approaches on Yale data set. The results are summarized for the similarity measure $\mathbf{W}=\mathbf{X}^{\top}\mathbf{X}$ and sparsity assumed to be known for all sparse representation methods which means that computation time of FRS-BDR is detailed for Steps 1.1, 1.2 and 2.}
\end{table*}

\begin{figure}[!h]
  \centering
\includegraphics[trim={0mm 0mm 0mm 0mm},clip,width=\linewidth]{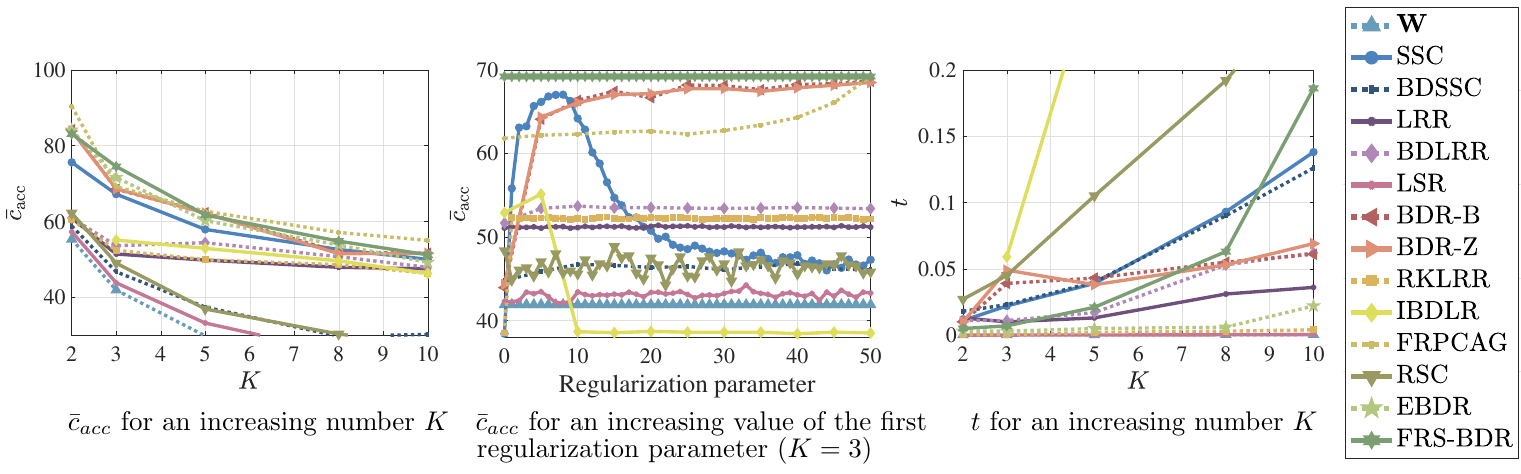}
\caption{Numerical results for Yale data set.}\vspace{-3mm}
\end{figure}


\newpage
\subsubsection{F.4.4~Additional Clustering Data Sets}

\setlength{\parindent}{0pt}\paragraph{F.4.4.1~Comparisons with Popular Block Diagonal Representation Approaches}
\textcolor{white}{[tbp!]}
\begin{table*}[h!]
\centering
\resizebox{\linewidth}{!}{%
\begin{tabular}{p{4cm}M{0.65cm}M{0.5cm}M{0.5cm}M{0.8cm}M{3.5cm}M{3.5cm}M{1.2cm}M{1.2cm}M{1.2cm}M{1.2cm}M{1.4cm}}
\hline\hline
\\[-3mm]
 & \multicolumn{11}{c}{Estimated Parameters of FRS-BDR for Different Clustering Data Sets}\\\\[-3mm]
\cline{2-12}\\[-3mm]
& & & & & & &\multicolumn{4}{c}{Detailed Computation Time $(t)$ }&Estimation Error\\\\[-3mm]
\cline{8-11}\\[-3mm]
Data Set & $\dimN$ & $\blocknum$ & $\hat{\blocknum}$ & $\bar{c}_\mathrm{acc}$ &$\mathbf{n}=[n_1,n_2,\myydots,n_{\blocknum}]^{\top}$& $\hat{\mathbf{n}}=[\hat{n}_1,\hat{n}_2,\myydots,\hat{n}_{\hat{\blocknum}}]^{\top}$ & Step 1.1 & Step 1.2 & Step 1.3 & Step 2  &  $\|\mathbf{v}-\hat{\mathbf{v}}\|$ \\
\midrule
Breast Cancer\cite{BreastCancer},&569 & 2 & 2 & 90.128 & $[212,357]^{\top}$ & $[173,396]^{\top}$ & 0.038 & 1.584 & 4.674 & 0.021 & 604.387 \\
Ceramic \cite{Ceramic},& 88 & 2 & 2 & 98.864 & $[44,44]^{\top}$ & $[44,44]^{\top}$ & 0.002 & 0.010 & 0.031 & 0.007 & 24.271  \\
Vertebral Column \cite{VertebralColumn},& 310 & 2,3 & 2 & 75.784 & $[100,210]^{\top}$ & $[120,190]^{\top}$& 0.009 & 0.183 & 0.543 & 0.011 & 140.272  \\
Iris \cite{Iris},& 150 & 3 & 3 & 96.667 & $[50,50,50]^{\top}$ & $[51,49,50]^{\top}$ & 0.003 & 0.033 & 0.098 & 0.018 & 48.121 \\
Human Gait \cite{humangait},& 800 & 5 & 5 & 77.125 & $[160,160,160,160,160]^{\top}$ & $[100,128,166,201,205]^{\top}$ & 0.087 & 5.011 & 15.327 & 1.139 & 466.761  \\
Ovarian Cancer \cite{OvarianCancer},& 216 & 2 & 2 & 77.315 & $[95,121]^{\top}$ & $[65,151]^{\top}$ & 0.012 & 0.075 & 0.208 & 0.011 & 131.006 \\
Person Identification \cite{PersonIdentification},& 187 & 4 & 4 & 96.791 & $[38,40,47,62]^{\top}$ & $[34,36,52,65]^{\top}$ & 0.027 & 0.053 & 0.190 & 0.108 & 70.456 \\
Parkinson \cite{Parkinson},& 240 & 2 & 2 & 58.208 & $[120,120]^{\top}$ & $[78,162]^{\top}$ & 0.006 & 0.095 & 0.266 & 0.013 & 111.626 \\
\hline\hline
\end{tabular}}
\caption{Estimation performance of FRS-BDR on well-known clustering data sets. The results are summarized for the similarity measure $\mathbf{W}=\mathbf{X}^{\top}\mathbf{X}$.}\vspace{-2mm}
\end{table*}
\begin{table*}[h!]
\centering
\resizebox{\linewidth}{!}{%
\begin{tabular}{p{4cm}M{1.5cm}M{1.5cm}M{1.5cm}M{1.5cm}M{1.5cm}M{1.5cm}M{1.5cm}M{1.5cm}M{1.5cm}M{1.5cm}M{1.5cm}M{1.5cm}M{1.5cm}M{1.8cm}}
\hline\hline
\\[-3mm]
 & \multicolumn{14}{c}{Subspace Clustering Performance for Different Block Diagonal Representation Methods}\\\\[-3mm]
\cline{2-15}\\[-3mm]
& &\multicolumn{11}{c}{Minimum-Maximum Clustering Accuracy $(c_\mathrm{accmin}-c_\mathrm{accmax})$ for Different Regularization Parameters} &  &\\\\[-3mm]
\cline{3-13}\\[-3mm]
Data Set & $\mathbf{W}$ & SSC & BD-SSC & LRR & BD-LRR & LSR & BDR-B & BDR-Z & RKLRR & IBDLR &FRPCAG & RSC & EBDR & \textbf{FRS-BDR} \\
\midrule
Breast Cancer\cite{BreastCancer},& 88.2 & 51.0-74.7 & 50.3-88.2 & 54.3-90.3 & 88.0-90.0 & 73.5-88.2 & 62.4-90.0 & 52.9-90.2 & 62.6-91.7 & 60.3-90.0 &60.5-88.2&50.1-58.5& 85.2 & 90.1\\
Ceramic \cite{Ceramic},&98.9 & 51.1-98.9 & 51.1-100 & 95.5-98.9 & 95.5-98.9 & 54.5-98.9 & 51.1-100 & 51.1-98.9 & 51.1-95.5 & 51.1-98.9 &50.0-100&50.0-69.3& 98.9 & 98.9 \\
Vertebral Column \cite{VertebralColumn},&73.2 & 50.0-77.7 & 50.3-74.8 & 53.9-72.6 & 72.6-72.6 & 62.6-75.8 & 67.4-76.8 & 71.9-76.8 & 67.4-71.3 & 67.4-76.1 &51.0-75.8&50.0-69.7& 74.8 & 75.8 \\
Iris \cite{Iris},&78.0& 34.7-82.7 & 34.0-83.3 & 38.7-80.7 & 80.0-98.0 & 78.0-82.7 & 34.0-96.7 & 65.3-96.7 & 34.0-80.0 & 34.7-84.0 &34.0-89.3&35.3-50.0& 98.0 & 96.7\\
Human Gait \cite{humangait},& 77.3 & 20.3-77.4 & 20.1-77.5 & 26.1-83.9 & 78.9-83.5 & 55.4-75.9 & 20.3-84.8 & 26.4-84.5 & 20.5-85.5 & 20.4-81.6 &50.8-77.8&22.1-26.0& 81.1 & 77.1 \\
Ovarian Cancer \cite{OvarianCancer},& 61.7 & 51.4-73.6 & 50.9-71.3 & 52.3-76.4 & 54.2-76.4 & 51.9-66.2 & 53.7-75.9 & 51.9-74.1 & 55.6-88.4 & 55.6-75.5 &77.8-89.3&50.0-69.4& 77.8 & 77.3\\
Person Identification \cite{PersonIdentification},& x & 33.7-96.8 & 31.6-95.7 & 49.7-94.7 & 71.1-94.7 & 33.2-64.2 & 31.6-96.3 & 59.4-95.7 & 34.2-94.1 & 33.7-95.7 &29.4-92.5&28.9-41.2& 97.3 & 96.8\\
Parkinson \cite{Parkinson},& 61.3 & 50.4-58.8 & 50.0-61.3 & 50.4-54.2 & 50.4-61.3 & 57.9-61.3 & 50.4-61.3 & 50.0-61.3 & 50.4-61.7 & 50.4-61.3 &50.0-67.5&50.4-72.1& 56.7 & 58.2\\
\hline\\[-3mm]
Average & 76.9 & 42.8-80.1 & 42.3-81.5 & 52.6-81.4 & 73.8-84.4 & 58.4-76.6 & 46.4-85.2 & 53.6-84.8 & 47.0-83.5 & 46.7-82.9 &50.4-85.1&42.1-57.0& 83.7 & 83.9\\
\hline\hline
\end{tabular}}
\caption{Subspace clustering performance of different block diagonal representation approaches on well-known clustering data sets. The results are summarized for the similarity measure $\mathbf{W}=\mathbf{X}^{\top}\mathbf{X}$. `x' denotes the failed results due to the complex-valued eigenvectors.}\vspace{-2mm}
\label{tab:subspaceclusteringadditionaldatasets}
\end{table*}

\begin{table*}[h!]
\centering
\resizebox{\linewidth}{!}{%
\begin{tabular}{p{4cm}M{1.5cm}M{1.5cm}M{1.5cm}M{1.5cm}M{1.5cm}M{1.5cm}M{1.5cm}M{1.5cm}M{1.5cm}M{1.5cm}M{1.5cm}M{1.5cm}M{1.5cm}M{1.8cm}}
\hline\hline
\\[-3mm]
 & \multicolumn{14}{c}{Computation time $(t)$ for Different Block Diagonal Representation Methods}\\\\[-3mm]
\cline{2-15}\\[-3mm]
& &\multicolumn{11}{c}{Computation Time $(t)$ for Optimally Tuned Regularization Parameters} &  &\\\\[-3mm]
\cline{3-13}\\[-3mm]
Data Set & $\mathbf{W}$ & SSC & BD-SSC & LRR & BD-LRR & LSR & BDR-B & BDR-Z & RKLRR & IBDLR &FRPCAG & RSC& EBDR & \textbf{FRS-BDR} \\
\midrule
Breast Cancer\cite{BreastCancer},& 0.003 & 1.123 & 2.967 & 0.456 & 1.034 & 0.009 & 4.510 & 4.502 & 2.793 & 10.782 &0.078&2.197& 0.116 & 1.643\\
Ceramic \cite{Ceramic},& $4\times 10^{-4}$ & 0.074 & 0.076 & 0.074 & 0.086 & $3\times 10^{-4}$ & 0.139 & 0.137 & 0.174 & 0.918 &0.511&0.052& 0.007 & 0.019 \\
Vertebral Column \cite{VertebralColumn},&0.001 & 0.518 & 0.490 & 0.122 & 0.230 & 0.002 & 0.897 & 0.902 & 0.031 & 2.018 &0.012&0.351& 0.038 & 0.203 \\
Iris \cite{Iris},&$5\times 10^{-4}$& 0.161 & 0.166 & 0.070 & 0.066 & 0.001 & 0.030 & 0.025 & 0.024 & 9.330 & 0.056&0.226&0.012 & 0.054\\
Human Gait \cite{humangait},& 0.006 & 6.465 & 6.642 & 1.214 & 2.877 & 0.018 & 1.377 & 1.212 & 7.392 & 865.055 &0.153&24.499& 0.583 & 6.237 \\
Ovarian Cancer \cite{OvarianCancer},& 0.008 & 3.036 & 3.883 & 8.316 & 8.387 & 0.016 & 0.503 & 0.506 & 9.192 & 17.590 &3.297&0.182& 0.026 & 0.098\\
Person Identification \cite{PersonIdentification},& $5\times 10^{-4}$ & 0.234 & 0.236 & 0.039 & 0.125 & 0.001 & 0.905 & 0.485 & 0.012 & 0.873 &0.008&0.578& 0.017 & 0.188\\
Parkinson \cite{Parkinson},& 0.001 & 0.350 & 0.354 & 0.321 & 0.399 & 0.001 & 0.066 & 0.065 & 0.170 & 2.397 & 1.806&0.224&0.025 & 0.113\\
\hline\\[-3mm]
Average & 0.002 & 1.495 & 1.851 & 1.327 & 1.651 & 0.006 & 1.053 & 0.979 & 2.473 & 113.620 &0.740&3.539& 0.103 & 1.069\\
\hline\hline
\end{tabular}}
\caption{Computation time performance of different block diagonal representation approaches on well-known clustering data sets. The results are summarized for the similarity measure $\mathbf{W}=\mathbf{X}^{\top}\mathbf{X}$ and sparsity assumed to be known for all sparse representation methods which means that computation time of FRS-BDR is detailed for Steps 1.1, 1.2 and 2.}\vspace{-1cm}
\label{tab:subspaceclusteringadditionaldatasets}
\end{table*}

\newpage
\setlength{\parindent}{0pt}\paragraph{F.4.4.2~Comparisons with Popular Community Detection Approaches}
\textcolor{white}{[tbp!]}
\begin{table*}[h!]
\centering
\small
\begin{tabular}{p{4.3cm}M{1cm}M{1cm}M{1cm}M{1cm}M{1cm}M{1cm}M{1.1cm}M{1.5cm}M{0.5cm}}
\hline\hline
\\[-3mm]
 & \multicolumn{8}{c}{$\hat{K}$ for Different Cluster Enumeration Methods} &  \\\\[-3mm]
\cline{2-9}\\[-3mm]
Data Set & Louvain & Martelot & BNMF & DenPeak & Combo & MAP & Sparcode & \textbf{FRS-BDR} & $K$\\
\midrule
Breast Cancer\cite{BreastCancer},&2 & 1 & 1 & 3 & 2 & 1 & 2 & 2 & 2 \\
Ceramic \cite{Ceramic},&2 & 2 & 1 & 3 & 2 & 1 & 2 & 2 & 2 \\
Vertebral Column \cite{VertebralColumn},&2 & 2 & 1 & 3 & 2 & 1 & 2 & 2 & 2,3\\
Iris \cite{Iris},&2& 2 & 1 & 3 & 2 & 1 & 1 & 3 &3 \\
Human Gait \cite{humangait},& 3 & 2 & 1 & 2 & 3 & 1 & 2 & 5 &5 \\
Ovarian Cancer \cite{OvarianCancer},& 2 & 2 & 1& 3 & 2& 1 & 4 & 2& 2\\
Person Identification \cite{PersonIdentification},& 3 & 3 & 1& 47&2 & 2 & 3&4&4\\
Parkinson \cite{Parkinson},& 1 & 1 &1 & 3 & 1 &1 &2 & 2&2\\
\hline\hline
\end{tabular}
\caption{Performance of different cluster enumeration approaches on well-known clustering data sets. The results are summarized for the similarity measure $\mathbf{W}=\mathbf{X}^{\top}\mathbf{X}$.}\label{tab:table-name}\vspace{-3mm}
\end{table*}
\vspace{-3mm}
\begin{table*}[h!]
\centering
\small
\begin{tabular}{p{4.3cm}M{1.1cm}M{1.1cm}M{1.1cm}M{1.1cm}M{1.1cm}M{1cm}M{1.1cm}M{1.5cm}}
\hline\hline
\\[-3mm]
 & \multicolumn{7}{c}{Modularity $(\mathrm{mod})$ for Different Cluster Enumeration Methods} &  \\\\[-3mm]
\cline{2-9}\\[-3mm]
Data Set & Louvain & Martelot & BNMF & DenPeak & Combo & MAP & Sparcode & \textbf{FRS-BDR}\\
\midrule
Breast Cancer\cite{BreastCancer},&0.001 & 0.000 & 0.000 & 0.000 & 0.001 & 0.000 & 0.022 & 0.345 \\
Ceramic \cite{Ceramic},&0.055 & 0.055 & 0.000 & 0.040 & 0.055 & 0.000 & 0.440 & 0.441 \\
Vertebral Column \cite{VertebralColumn},&0.015 & 0.015 & 0.000 & 0.000 & 0.015 & 0.000 & 0.330 & 0.418 \\
Iris \cite{Iris},&0.016& 0.016 & 0.000 & 0.016 & 0.016 & 0.000 & 0.000 & 0.472\\
Human Gait \cite{humangait},& 0.017 & 0.014 & 0.000 & 0.000 & 0.017 & 0.000 & 0.375 & 0.641 \\
Ovarian Cancer \cite{OvarianCancer},& 0.006 & 0.006 & 0.000 & 0.005 & 0.006& 0.000 & 0.019 & 0.334\\
Person Identification \cite{PersonIdentification},& 0.128 & 0.124 & 0.000& 0.051&0.128 & 0.000 & 0.515&0.694\\
Parkinson \cite{Parkinson},& 0.000 & 0.000 &0.000 & 0.000 & 0.000 &0.000 &0.179 & 0.342\\
\hline\\[-3mm]
Average & 0.030 & 0.029 &0.000 & 0.014 & 0.030 & 0.000 & 0.235 & 0.461\\
\hline\hline
\end{tabular}
\caption{Partitioning performance of different cluster enumeration approaches on well-known clustering data sets. The results summarized for modularity $(\mathrm{mod})$
using the similarity measure $\mathbf{W}=\mathbf{X}^{\top}\mathbf{X}$.}
\label{tab:clusterenumerationmodularity}\vspace{-3mm}
\end{table*}
\vspace{-3mm}
\begin{table*}[h!]
\centering
\small
\begin{tabular}{p{4.3cm}M{1.1cm}M{1.1cm}M{1.1cm}M{1.1cm}M{1.1cm}M{1cm}M{1.1cm}M{1.5cm}}
\hline\hline
\\[-3mm]
 & \multicolumn{7}{c}{Conductance $(\mathrm{cond})$ for Different Cluster Enumeration Methods} &  \\\\[-3mm]
\cline{2-9}\\[-3mm]
Data Set & Louvain & Martelot & BNMF & DenPeak & Combo & MAP & Sparcode & \textbf{FRS-BDR}\\
\midrule
Breast Cancer\cite{BreastCancer},&0.448 & 0.000 & 0.000 & 0.007 & 0.448 & 0.000 & 0.079 & 0.098 \\
Ceramic \cite{Ceramic},&0.445 & 0.445 & 0.000 & 0.449 & 0.445 & 0.000 & 0.058 & 0.059 \\
Vertebral Column \cite{VertebralColumn},&0.484 & 0.484 & 0.000 & 0.022 & 0.484 & 0.000 & 0.119 & 0.079 \\
Iris \cite{Iris},&0.424& 0.429 & 0.000 & 0.433 & 0.424 & 0.000 & 0.000 & 0.183\\
Human Gait \cite{humangait},& 0.615 & 0.479 & 0.000 & 0.003 & 0.615 & 0.000 & 0.121 & 0.138 \\
Ovarian Cancer \cite{OvarianCancer},& 0.451 & 0.476 & 0.000& 0.487 & 0.451& 0.000 & 0.047 & 0.140\\
Person Identification \cite{PersonIdentification},& 0.296 & 0.294 & 0.000 & 0.660 &0.296 & -0.035 & 0.095 &0.047\\
Parkinson \cite{Parkinson},& 0.000 & 0.000 & 0.000 & 0.017 & 0.000 & 0.000 & 0.312 & 0.157\\
\hline\\[-3mm]
Average & 0.395 & 0.326 &0.000 & 0.260 & 0.395 & -0.004 & 0.104 & 0.113\\
\hline\hline
\end{tabular}
\caption{Partitioning performance of different cluster enumeration approaches on well-known clustering data sets. The results summarized for conductance $(\mathrm{cond})$
using the similarity measure $\mathbf{W}=\mathbf{X}^{\top}\mathbf{X}$.}\label{tab:clusterenumerationconductance}\vspace{-3mm}
\end{table*}
\vspace{-3mm}
\begin{table*}[h!]
\centering
\small
\begin{tabular}{p{4.3cm}M{1.1cm}M{1.1cm}M{1.1cm}M{1.1cm}M{1.1cm}M{1cm}M{1.1cm}M{1.5cm}}
\hline\hline
\\[-3mm]
 & \multicolumn{7}{c}{Computation Time $(t)$ for Different Cluster Enumeration Methods} &  \\\\[-3mm]
\cline{2-9}\\[-3mm]
Data Set & Louvain & Martelot & BNMF & DenPeak & Combo & MAP & Sparcode & \textbf{FRS-BDR}\\
\midrule
Breast Cancer\cite{BreastCancer},&0.270 & 0.547 & 22.839 & 0.080 & 3.965 & 2.016 & 3.644 & 5.334 \\
Ceramic \cite{Ceramic},&0.008 & 0.004 & 0.273 & 0.009 & 0.083 & 0.219 & 0.103 & 0.059 \\
Vertebral Column \cite{VertebralColumn},&0.059 & 0.066 & 4.154 & 0.036 & 1.040 & 0.719 & 0.820 & 0.818 \\
Iris \cite{Iris},&0.012& 0.010 & 0.752 & 0.017 & 0.241 & 0.375 & 0.323 & 0.178\\
Human Gait \cite{humangait},& 0.629 & 1.654 & 48.883 & 0.118 & 3.706 & 2.766 & 5.723 & 21.321 \\
Ovarian Cancer \cite{OvarianCancer},& 0.026 & 0.023 & 1.769& 0.024 & 0.546& 0.484 & 0.509 & 0.337\\
Person Identification \cite{PersonIdentification},& 0.027 & 0.016 & 1.363& 0.047&0.366 & 0.344 & 0.420 &0.366\\
Parkinson \cite{Parkinson},& 0.029 & 0.032 &2.307 & 0.027 & 0.003 & 0.578 & 0.540 & 0.425\\
\hline\\[-3mm]
Average & 0.132 & 0.294 &10.293 & 0.045 & 1.243 & 0.938 & 1.510 & 3.605\\
\hline\hline
\end{tabular}
\caption{Computation performance of different cluster enumeration approaches on well-known clustering data sets. The results are summarized for the similarity measure $\mathbf{W}=\mathbf{X}^{\top}\mathbf{X}$ and  FRS-BDR is detailed for all steps.}\label{tab:table-name}
\end{table*}

\newpage

\newpage

\subsubsection{F.4.5~Similarity-based Block Diagonal Ordering (sBDO) Performance Analysis}

\setlength{\parindent}{0pt}\paragraph{F.4.5.1~Subspace Clustering Performance Comparisons with Reverse Cuthill-McKee (RCM) Algorithm}
\textcolor{white}{[tbp!]}
\begin{table*}[h!]

\centering
\resizebox{\linewidth}{!}{%
\begin{tabular}{p{4cm}M{2cm}M{2cm}M{2cm}M{2cm}M{2cm}M{2cm}}
\hline\hline
\\[-3mm]
 & \multicolumn{6}{c}{Average Clustering Accuracy $(\bar{c}_\mathrm{acc})$ for Different Data Sets using RCM Algorithm}\\\\[-3mm]
\cline{2-7}\\[-3mm]
Number of subjects & MNIST & USPS & COIL20 & ORL & JAFFE & YALE \\
\midrule
2  & 89.5 & 92.7 & 87.9 & 81.5 & 89 & 73.8 \\
3  & 79.7 & 87.5 & 82.3 & 74.9 & 93 & 73.6 \\
5  & 67.3 & 77 & 81.7 & 82.7 & 96 & 60.9 \\
8  & 59 & 65.5 & 77.6 & 79.8 & 94.8 &  54.8\\
10  & 54.7 & 62 & 75.9 & 76.2 &  92.9 & 51.1 \\
\hline\\[-3mm]
Average  & 70.0 & 76.9 & 81.0 & 79.0 & 93.1  & 62.8 \\
\hline\hline
\end{tabular}}
\caption{Subspace clustering performance for different data sets using RCM algorithm in Step 1.2 of the FRS-BDR method. The results are summarized for the similarity measure $\mathbf{W}=\mathbf{X}^{\top}\mathbf{X}$.}
\end{table*}


\begin{table*}[h!]

\centering
\resizebox{\linewidth}{!}{%
\begin{tabular}{p{4cm}M{2cm}M{2cm}M{2cm}M{2cm}M{2cm}M{2cm}}
\hline\hline
\\[-3mm]
 & \multicolumn{6}{c}{Average Clustering Accuracy $(\bar{c}_\mathrm{acc})$ for Different Data Sets using sBDO Algorithm}\\\\[-3mm]
\cline{2-7}\\[-3mm]
Number of subjects & MNIST & USPS & COIL20 & ORL & JAFFE & YALE \\
\midrule
2  & 89.7 & 92.6 & 93.1 & 93.1 & 93.4 & 83.1 \\
3  & 79.6 & 87.5 & 88.1 & 90.1 & 96.9 & 74.5 \\
5  & 67.5 & 77.3 & 82.5 & 84.2 & 96.4 & 61.7 \\
8  & 59.3 & 65.2 & 77.3 & 79.4 & 96.4 &  54.8\\
10  & 57.6 & 59.8 & 75.9 & 77.2 &  96.7 & 51.2 \\
\hline\\[-3mm]
Average  & 70.7 & 76.5 & 83.4 & 84.8 & 96.0  & 65.1 \\
\hline\hline
\end{tabular}}
\caption{Subspace clustering performance for different data sets using sBDO algorithm in Step 1.2 of the FRS-BDR method. The results are summarized for the similarity measure $\mathbf{W}=\mathbf{X}^{\top}\mathbf{X}$.}
\end{table*}


\begin{table*}[h!]

\centering
\resizebox{0.7\linewidth}{!}{%
\begin{tabular}{p{4cm}M{2cm}M{2cm}M{2cm}M{2cm}}
\hline\hline
\\[-3mm]
Data Set & $\dimN$ & $\blocknum$ & $\hat{\blocknum}$ & $\bar{c}_\mathrm{acc}$\\
\midrule
Breast Cancer\cite{BreastCancer},&569 & 2 & x & x\\
Ceramic \cite{Ceramic},& 88 & 2 & x & x\\
Vertebral Column \cite{VertebralColumn},& 310 & 2,3 & x & x\\
Iris \cite{Iris},& 150 & 3 & 2 & 66.0 \\
Human Gait \cite{humangait},& 800 & 5 & 2 & 20.0 \\
Ovarian Cancer \cite{OvarianCancer},& 216 & 2 & x & x\\
Person Identification \cite{PersonIdentification},& 187 & 4 & 2 & 26.6 \\
Parkinson \cite{Parkinson},& 240 & 2 & x & x \\
\hline\hline
\end{tabular}}
\caption{Estimation performance of FRS-BDR on well-known clustering data sets. The results are summarized for the similarity measure $\mathbf{W}=\mathbf{X}^{\top}\mathbf{X}$. FRS-BDR algorithm is performed using RCM in Step 1.2.  `x' denotes the failed results.}
\end{table*}

\begin{table*}[h!]

\centering
\resizebox{0.7\linewidth}{!}{%
\begin{tabular}{p{4cm}M{2cm}M{2cm}M{2cm}M{2cm}}
\hline\hline
\\[-3mm]
Data Set & $\dimN$ & $\blocknum$ & $\hat{\blocknum}$ & $\bar{c}_\mathrm{acc}$\\
\midrule
Breast Cancer\cite{BreastCancer},&569 & 2 & 2 & 90.1\\
Ceramic \cite{Ceramic},& 88 & 2 & 2 & 98.9\\
Vertebral Column \cite{VertebralColumn},& 310 & 2,3 & 2 & 75.8\\
Iris \cite{Iris},& 150 & 3 & 3 & 96.7 \\
Human Gait \cite{humangait},& 800 & 5 & 5 & 77.1 \\
Ovarian Cancer \cite{OvarianCancer},& 216 & 2 & 2 & 77.3\\
Person Identification \cite{PersonIdentification},& 187 & 4 & 4 & 96.8 \\
Parkinson \cite{Parkinson},& 240 & 2 & 2 & 58.2 \\
\hline\hline
\end{tabular}}
\caption{Estimation performance of FRS-BDR on well-known clustering data sets. The results are summarized for the similarity measure $\mathbf{W}=\mathbf{X}^{\top}\mathbf{X}$. FRS-BDR algorithm is performed using sBDO in Step 1.2.}
\end{table*}
\newpage

\setlength{\parindent}{0pt}\paragraph{F.4.5.2~Block Diagonal Structure Enhancement Comparisons with Reverse Cuthill-McKee (RCM) Algorithm}
\textcolor{white}{[tbp!]}\\\vspace{-3mm}
\\
\begin{figure}[!h]
\vspace{-3mm}
  \centering
\subfloat[RCM]{\includegraphics[trim={0mm 0mm 0mm 0mm},clip,width=8cm]{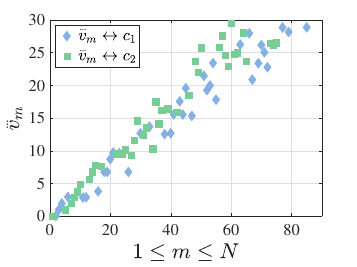}}\hspace{1cm}
\subfloat[sBDO]{\includegraphics[trim={0mm 0mm 0mm 0mm},clip,width=8cm]{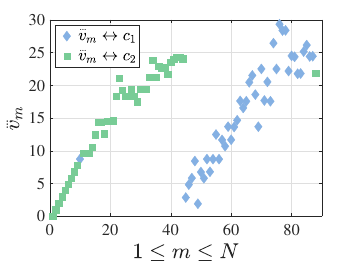}}
\caption{Examplary block diagonal enhancement results for Ceramic \cite{Ceramic} data set using different block diagonal ordering algorithms in Step 1.2. Components of the obtained vector $\ddot{\mathbf{v}}$ are highlighted in two different colors based on their associations to $K=2$ clusters.}
\end{figure}

\begin{figure}[!h]
  \centering
\subfloat[RCM]{\includegraphics[trim={0mm 0mm 0mm 0mm},clip,width=8cm]{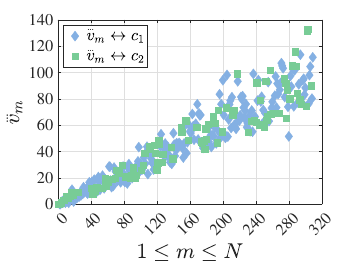}}\hspace{1cm}
\subfloat[sBDO]{\includegraphics[trim={0mm 0mm 0mm 0mm},clip,width=8cm]{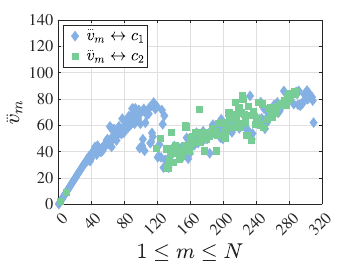}\label{fig:sBDOchallengingscenario}}
\caption{Examplary block diagonal enhancement results for Vertebral Column \cite{VertebralColumn} data set using different block diagonal ordering algorithms in Step 1.2. Components of the obtained vector $\ddot{\mathbf{v}}$ are highlighted in two different colors based on their associations to $K=2$ clusters.}
\end{figure}

To better understand robustness of the sBDO algorithm, let consider a challenging scenario such that a heavy tailed noise that connects two groups/clusters with a large weight is selected as a starting point in the proposed sBDO method. Since the similarity coefficients are valued in range $[-1,1]$, a single similarity coefficient would not lead to an unbounded effect in proposed ordering algorithm. In such cases, the algorithm starts with adding the most similar neighbor from one of the clusters. Then, the algorithm updates the set and finds the most similar neighbors based on their connectedness to the available set. This means that after the sBDO algorithm selects the first few neighbors from the same cluster, similarity of these vertices will quickly dominate the similarity of the noisy vertex. This scenario about selecting a Type~II outlier in the first few iterations is visible in Fig.~\ref{fig:sBDOchallengingscenario}. The real-world data example confirms the above information that selecting samples from within a cluster quickly dominates the effect of Type II outliers after a few iterations while RCM algorithm loses the cluster structure completely. 

\begin{figure}[tbp!]\vspace{2cm}
  \centering
\subfloat[RCM]{\includegraphics[trim={0mm 0mm 0mm 0mm},clip,width=8cm]{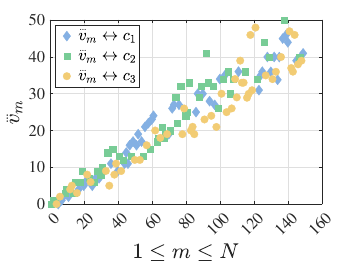}}\hspace{1cm}
\subfloat[sBDO]{\includegraphics[trim={0mm 0mm 0mm 0mm},clip,width=8cm]{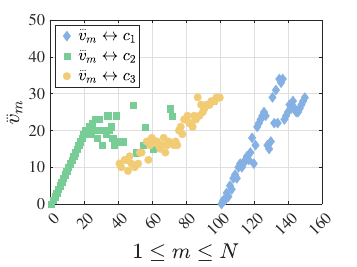}}
\caption{Examplary block diagonal enhancement results for Iris \cite{Iris} data set using different block diagonal ordering algorithms in Step 1.2. Components of the obtained vector $\ddot{\mathbf{v}}$ are highlighted in three different colors based on their associations to $K=3$ clusters.}
\vspace{3cm}
\end{figure}

\begin{figure}[tbp!]
  \centering
\subfloat[RCM]{\includegraphics[trim={0mm 0mm 0mm 0mm},clip,width=8cm]{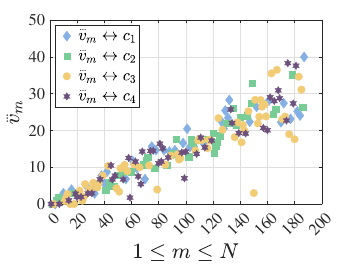}}\hspace{1cm}
\subfloat[sBDO]{\includegraphics[trim={0mm 0mm 0mm 0mm},clip,width=8cm]{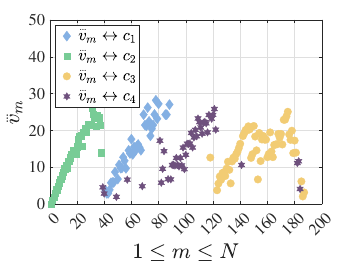}}
\caption{Examplary block diagonal enhancement results for Person Identification \cite{PersonIdentification} data set using different block diagonal ordering algorithms in Step 1.2. Components of the obtained vector $\ddot{\mathbf{v}}$ are highlighted in four different colors based on their associations to $K=4$ clusters.\color{white}data set using different block diagonal ordering algorithms in Step 1.2. Components of the obtained vector $\ddot{\mathbf{v}}$ are highlighted in four different colors based on their associations to $K=4$ clusters.data set using different block diagonal ordering algorithms in Step 1.2. Components of the obtained vector $\ddot{\mathbf{v}}$ are highlighted in four different colors based on their associations to $K=4$ clusters.data set using different block diagonal ordering algorithms in Step 1.2. Components of the obtained vector $\ddot{\mathbf{v}}$ are highlighted in four different colors based on their associations to $K=4$ clusters.\color{black}}
\end{figure}

\newpage
\subsubsection{F.4.6~Robustness Analysis}
\setlength{\parindent}{0pt}\paragraph{\color{white}F.4.6.1~Robustness Analysis for MNIST Data Set}

\begin{table*}[h!]
\centering
\resizebox{\linewidth}{!}{%
\begin{tabular}{p{1.5cm}M{1.5cm}M{1.5cm}M{1.5cm}M{1.5cm}M{1.5cm}M{1.5cm}M{1.5cm}M{1.5cm}M{1.5cm}M{1.5cm}M{1.5cm}M{1.5cm}M{1.5cm}M{2cm}}
\hline\hline
\\[-3mm]
 & \multicolumn{14}{c}{Subspace Clustering Performance for Different Block Diagonal Representation Methods}\\\\[-3mm]
\cline{2-15}\\[-3mm]
& &\multicolumn{9}{c}{Minimum-Maximum Clustering Accuracy $(c_\mathrm{accmin}-c_\mathrm{accmax})$ for Different Regularization Parameters} &  &\\\\[-3mm]
\cline{3-13}\\[-3mm]
Noise density & $\mathbf{W}$ & SSC & BD-SSC & LRR & BD-LRR & LSR & BDR-B & BDR-Z & RKLRR & IBDLR & FRPCAG  & RSC & EBDR & \textbf{FRS-BDR} \\
\midrule
0 & 72.0 & 33.9-37.5 & 33.8-76.1 & 34.2-74.8 & 34.3-72.5 & 34.9-71.8 & 34.4-67.6 & 34.4-67.7 & 33.9-70.0 & 33.9-79.0 &62.3-79.7&39.7-45.1& 68.6 & 79.6\\
0.1 & 72.9 & 33.9-75.2 & 33.8-90.7 & 34.7-65.7 & 34.3-70.6 & 66.1-74.7 & 33.7-76.8& 40.5-75.8 & 33.8-78.7 & 33.8-81.9 &66.0-75.0&39.8-45.5& 64.2 & 80.7\\
0.2 & 73.8 & 33.8-77.9 & 33.8-81.2 & 36.0-61.0 & 35.5-76.2 & 69.5-76.9&33.7-76.3 & 48.9-74.2 & 33.8-75.2 & 33.8-75.5 &68.6-81.8&39.9-44.4& 60.9 & 79.3\\
0.3& 74.1 & 33.8-75.5 & 33.8-84.2 & 36.1-57.6 & 36.0-72.3& 69.8-73.7&33.7-84.1 & 67.3-78.9 & 33.8-64.1 & 33.8-74.8 &72.1-80.5 &40.2-44.7& 58.8 & 77.2\\
0.4 & 74.0 & 33.8-72.5 & 33.7-80.7 & 35.7-39.3 & 41.6-77.8& 71.9-76.4&33.7-75.9 & 66.0-78.8 & 33.8-63.0 & 33.8-76.5 &46.4-77.5 &39.8-45.6&57.2 & 74.8\\
0.5 & 72.8 & 33.9-72.5 & 33.9-74.1 & 35.7-42.8 & 61.3-71.3& 73.4-77.9&33.7-66.1 &63.7-75.0  & 33.8-57.1 & 33.8-71.6 &34.1-74.9 &38.7-45.7& 54.7 & 72.4\\
\hline\hline
\end{tabular}}
\caption{Subspace clustering performance of different block diagonal representation approaches on corrupted MNIST data set. MNIST data set is computed for $K=3$ clusters and the noise density, which affects approximately the defined value of pixels based on salt and pepper noise, is increased. The results are summarized for the similarity measure $\mathbf{W}=\mathbf{X}^{\top}\mathbf{X}$.}\label{tab:robustnessMNISTSaltandPepper}
\end{table*}

\begin{table*}[h!]
\centering
\resizebox{\linewidth}{!}{%
\begin{tabular}{p{1.5cm}M{1.5cm}M{1.5cm}M{1.5cm}M{1.5cm}M{1.5cm}M{1.5cm}M{1.5cm}M{1.5cm}M{1.5cm}M{1.5cm}M{1.5cm}M{1.5cm}M{1.5cm}M{2cm}}
\hline\hline
\\[-3mm]
 & \multicolumn{14}{c}{Subspace Clustering Performance for Different Block Diagonal Representation Methods}\\\\[-3mm]
\cline{2-15}\\[-3mm]
& &\multicolumn{11}{c}{Minimum-Maximum Clustering Accuracy $(c_\mathrm{accmin}-c_\mathrm{accmax})$ for Different Regularization Parameters} &  &\\\\[-3mm]
\cline{3-13}\\[-3mm]
Noise type & $\mathbf{W}$ & SSC & BD-SSC & LRR & BD-LRR & LSR & BDR-B & BDR-Z & RKLRR & IBDLR & FRPCAG  & RSC & EBDR & \textbf{FRS-BDR} \\
\midrule
Poisson & 72.3 & 33.9-37.0 & 33.7-74.3 & 34.4-66.9 & 34.3-78.9& 36.3-75.4 & 34.4-71.4 & 34.4-68.8 & 33.8-72.6 & 33.8-79.6 & 70.5-80.2 &39.5-46.1& 65.6 & 81.2\\
\hline\hline
\end{tabular}}
\caption{Subspace clustering performance of different block diagonal representation approaches on corrupted MNIST data set. MNIST data set is computed for $K=3$ clusters and it is corrupted by interpreting input pixel values  as means of Poisson distributions that are scaled up by $10^{12}$. The results are summarized for the similarity measure $\mathbf{W}=\mathbf{X}^{\top}\mathbf{X}$.}\label{tab:robustnessMNISTpoisson}
\end{table*}

\setlength{\parindent}{0pt}\paragraph{\color{white}F.4.6.2~Robustness Analysis for USPS Data Set}
\begin{table*}[h!]
\centering
\resizebox{\linewidth}{!}{%
\begin{tabular}{p{1.5cm}M{1.5cm}M{1.5cm}M{1.5cm}M{1.5cm}M{1.5cm}M{1.5cm}M{1.5cm}M{1.5cm}M{1.5cm}M{1.5cm}M{1.5cm}M{1.5cm}M{1.5cm}M{2cm}}
\hline\hline
\\[-3mm]
 & \multicolumn{14}{c}{Subspace Clustering Performance for Different Block Diagonal Representation Methods}\\\\[-3mm]
\cline{2-13}\\[-3mm]
& &\multicolumn{11}{c}{Minimum-Maximum Clustering Accuracy $(c_\mathrm{accmin}-c_\mathrm{accmax})$ for Different Regularization Parameters} &  &\\\\[-3mm]
\cline{3-13}\\[-3mm]
Noise density & $\mathbf{W}$ & SSC & BD-SSC & LRR & BD-LRR & LSR & BDR-B & BDR-Z & RKLRR & IBDLR & FRPCAG  & RSC& EBDR & \textbf{FRS-BDR} \\
\midrule
0 & 70.4 & 34.4-55.0 & 36.8-77.4 & 35.5-68.7& 35.5-69.4 & 38.3-71.9 & 36.5-85.1 & 34.7-85.6 & 34.4-80.7 & 34.4-86.6 &  65.0-78.6 &37.7-48.8& 77.2 & 87.5\\
0.1 & 78.2& 34.4-59.8 & 35.0-77.7 & 36.6-54.7 & 36.7-80.3 & 35.8-74.5 & 34.2-74.5 & 35.7-76.9 & 34.3-63.0 & 34.3-77.1 &40.7-77.3 & 36.9-47.8& 68.3 & 77.9\\
0.2 & 73.3& 34.5-64.8 & 34.2-75.6 & 36.8-41.2 & 38.1-77.3 & 37.1-72.4 & 34.2-70.5 &  35.7-71.3& 34.3-53.5 & 34.3-64.5 & 39.1-75.6 & 38.1-46.9& 62.3 & 73.2\\
0.3& 67.5& 34.5-63.2& 34.1-67.4 & 37.5-41.4 & 38.4-71.6 & 40.7-69.9 & 34.2-66.9 & 38.8-68.2 & 34.3-47.6 & 34.3-58.2& 39.1-78.1  &36.9-47.5&  56.7 & 67\\
0.4 & 59.9 & 34.4-55.1& 34.2-64.6 & 37.2-39.6 & 37.9-66.2 & 40.5-61.5 & 34.1-59.4& 38.1-56.8 & 34.4-44.3 & 34.3-49.8& 40.6-74.6  &38.0-47.7&  50.3 & 57\\
0.5 & 50.0 & 34.4-49.2& 34.1-54.2 & 36.2-39.3 & 37.5-57.9 & 39.5-53.9 & 34.2-52.1 &  39.2-53.8& 34.3-41.1 & 34.3-43.3 &   39.1-72.8  &36.9-47.1& 44.4 & 48.6\\
\hline\hline
\end{tabular}}
\caption{Subspace clustering performance of different block diagonal representation approaches on corrupted USPS data set. USPS data set is computed for $K=3$ clusters and the noise density, which affects approximately the defined value of pixels based on salt and pepper noise, is increased. The results are summarized for the similarity measure $\mathbf{W}=\mathbf{X}^{\top}\mathbf{X}$.}\label{tab:robustnessUSPSSaltandPepper}
\end{table*}

\begin{table*}[h!]
\centering
\resizebox{\linewidth}{!}{%
\begin{tabular}{p{1.5cm}M{1.5cm}M{1.5cm}M{1.5cm}M{1.5cm}M{1.5cm}M{1.5cm}M{1.5cm}M{1.5cm}M{1.5cm}M{1.5cm}M{1.5cm}M{1.5cm}M{1.5cm}M{2cm}}
\hline\hline
\\[-3mm]
 & \multicolumn{14}{c}{Subspace Clustering Performance for Different Block Diagonal Representation Methods}\\\\[-3mm]
\cline{2-15}\\[-3mm]
& &\multicolumn{11}{c}{Minimum-Maximum Clustering Accuracy $(c_\mathrm{accmin}-c_\mathrm{accmax})$ for Different Regularization Parameters} &  &\\\\[-3mm]
\cline{3-13}\\[-3mm]
Noise type & $\mathbf{W}$ & SSC & BD-SSC & LRR & BD-LRR & LSR & BDR-B & BDR-Z & RKLRR & IBDLR & FRPCAG  & RSC & EBDR & \textbf{FRS-BDR} \\
\midrule
Poisson & 80.6 & 34.3-56.8 & 35.2-64.8 & 35.2-72.5 & 35.5-82.4& 34.5-71.6 & 34.3-55.6 & 34.9-67.5 & 34.4-84.3 & 34.3-67.6 &42.3-78.3 &37.5-50.3&  71.2 & 83.9\\
\hline\hline
\end{tabular}}
\caption{Subspace clustering performance of different block diagonal representation approaches on corrupted USPS data set. USPS data set is computed for $K=3$ clusters and it is corrupted by interpreting input pixel values  as means of Poisson distributions that are scaled up by $10^{12}$. The results are summarized for the similarity measure $\mathbf{W}=\mathbf{X}^{\top}\mathbf{X}$.}\label{tab:robustnessUSPSpoisson}
\end{table*}


\newpage
